\documentclass[runningheads]{llncs}
\usepackage{latexsym}
\usepackage{amssymb}
\usepackage{amsmath}
\usepackage{booktabs}
\usepackage{enumitem}
\usepackage{graphicx}
\usepackage{color}
\usepackage{epsfig}
\usepackage{amssymb}
\usepackage{pifont}
\usepackage{bbding}
\usepackage{array}
\usepackage{enumitem}
\usepackage{multirow}
\usepackage{colortbl}
\usepackage{xspace}
\usepackage{bbm}
\usepackage{import}
\usepackage{times}
\usepackage{soul}
\usepackage{url}
\usepackage[hidelinks]{hyperref}
\usepackage[utf8]{inputenc}
\usepackage{graphicx}
\usepackage{amsmath}
\usepackage{booktabs}
\usepackage{algorithm}
\usepackage{algorithmic}
\usepackage[switch]{lineno}

\usepackage{epsfig}
\usepackage{amssymb}
\usepackage{pifont}
\usepackage{bbding}
\usepackage{array}
\usepackage{enumitem}
\usepackage{multirow}
\usepackage{colortbl}
\usepackage{xspace}
\usepackage{bbm}
\usepackage{import}

\newcommand{\BibTeX}{B\kern-.05em{\sc i\kern-.025em b}\kern-.08em\TeX}
\renewcommand{\algorithmiccomment}[1]{{\color{green!50!black}\bgroup\hfill #1\egroup}}

\usepackage{color}
\definecolor{darkgreen}{rgb}{0,0.5,0}
\definecolor{eyecancerpink}{rgb}{1,0,1}
\definecolor{purple}{rgb}{1,0,1}

\newcommand{\projabbrv}{\textbf{Eidos}\xspace}

\makeatletter
\@namedef{ver@everyshi.sty}{}
\makeatother

\usepackage[dvipsnames,svgnames,x11names]{xcolor}
\usepackage{tikz}
\usetikzlibrary{arrows.meta,shapes,calc,matrix,fit,positioning,backgrounds,decorations.markings,fadings}
\usepackage{pgfplots}
\usepackage{pgfplotstable}
\pgfplotsset{compat=1.9}
\usepackage{xstring}

\usepgfplotslibrary{external}
\newcommand{\extfig}[2]{\tikzsetnextfilename{#1}{#2}}

\IfBeginWith*{\jobname}{fig/extern/}{\finalcopy}{}

\tikzstyle{every picture}+=[
	remember picture,
	every text node part/.style={align=center},
	every matrix/.append style={ampersand replacement=\&},
]
\tikzstyle{tight} = [inner sep=0pt,outer sep=0pt]
\tikzstyle{node}  = [draw,circle,tight,minimum size=12pt,anchor=center]
\tikzstyle{op}    = [draw,circle,tight]
\tikzstyle{dot}   = [fill,draw,circle,inner sep=1pt,outer sep=0]
\tikzstyle{pt}    = [fill,draw,circle,inner sep=1.5pt,outer sep=.2pt]
\tikzstyle{box}   = [draw,rectangle,inner sep=3pt]
\tikzstyle{high}  = [black!60]
\tikzstyle{group} = [high,box,opacity=.5]
\tikzstyle{dim1}  = [fill opacity=.3,text opacity=1]
\tikzstyle{dim2}  = [fill opacity=.5,text opacity=1]
\tikzstyle{dim3}  = [fill opacity=.7,text opacity=1]
\tikzstyle{rectc} = [tight,transform shape]
\tikzstyle{rect}  = [rectc,anchor=south west]

\newcommand{\leg}[1]{\addlegendentry{#1}}

\tikzset{every mark/.append style={solid}}
\pgfplotsset{
	grid=both, width=\columnwidth, try min ticks=5,
	every axis/.append style={font=\small},
	every axis plot/.append style={thick,mark=none,mark size=1.8,tension=0.18},
	legend cell align=left, legend style={fill opacity=0.8},
	xticklabel={\pgfmathprintnumber[assume math mode=true]{\tick}},
	yticklabel={\pgfmathprintnumber[assume math mode=true]{\tick}},
	nodes near coords math/.style={
		nodes near coords={\pgfmathprintnumber[assume math mode=true]{\pgfplotspointmeta}},
	},
}

\pgfplotsset{
	dash/.style={mark=o,dashed,opacity=0.6},
	dott/.style={mark=o,dotted,opacity=0.6},
	nolim/.style={enlargelimits=false},
	plain/.style={every axis plot/.append style={},nolim,grid=none},
}

\pgfdeclarelayer{bg4}
\pgfdeclarelayer{bg3}
\pgfdeclarelayer{bg2}
\pgfdeclarelayer{bg1}
\pgfdeclarelayer{fg1}
\pgfdeclarelayer{fg2}
\pgfdeclarelayer{fg3}
\pgfdeclarelayer{fg4}
\pgfsetlayers{bg4,bg3,bg2,bg1,main,fg1,fg2,fg3,fg4}

\pgfkeys{/tikz/.cd, aspect/.store in=\aspect, aspect=1}
\pgfkeys{/tikz/.cd, depth/.store in=\depth, depth=.5}
\pgfkeys{/tikz/.cd, stepx/.store in=\step, stepx=1}

\tikzstyle{geom} = [line join=bevel,aspect=1,depth=.5,z={(\depth*\aspect,\depth)}]
\tikzstyle{wire} = [geom,draw,thick]

\def\cx[#1,#2,#3]{#1}
\def\cy[#1,#2,#3]{#2}
\def\cz[#1,#2,#3]{#3}
\def\ex[#1,#2,#3]{#1,0,0}
\def\ey[#1,#2,#3]{0,#2,0}
\def\ez[#1,#2,#3]{0,0,#3}

\newcommand{\citep}[1]{\cite{#1}}

\newcommand{\geoa}{\textbf{GeoA-}${{ \textbf 3}}$\xspace}
\newcommand{\siadv}{\textbf{SI-Adv}\xspace}
\newcommand{\gsda}{\textbf{GSDA}\xspace}

\newcommand{\srs}{SRS\xspace}
\newcommand{\sor}{SOR\xspace}

\newcommand{\simba}{\textbf{Simba}\xspace}
\newcommand{\simbapp}{\textbf{Simba++}\xspace}

\newcommand{\asr}{$P_{suc}$\xspace}

\newcommand{\lt}{$L_2$\xspace}
\newcommand{\cd}{CD\xspace}
\newcommand{\hd}{HD\xspace}
\newcommand{\curv}{Curv\xspace}
\newcommand{\smooth}{Smooth\xspace}

\newcommand{\our}{\textbf{Eidos}\xspace}

\newcommand{\figt}[2][1]{\includegraphics[trim={50mm 15mm 50mm 15mm},clip,width=#1\linewidth]{figures/#2}}

\def \IterMax{K}

\newcommand{\func}[1]{\mathsf{#1}}

\newcommand{\stt}{\textrm{subject to}\ }

\newcommand{\defn}{\mathrel{:=}}

\newcommand{\Th}[1]{\textsc{#1}}

\newcommand{\mc}[2]{\multicolumn{#1}{c}{#2}}
\newcommand{\tb}[1]{\textbf{#1}}

\newcommand{\red}[1]{{\textcolor{red}{#1}}}
\newcommand{\blue}[1]{{\textcolor{blue}{#1}}}

\definecolor{LightCyan}{rgb}{0.88,1,1}

\newcommand{\citeme}[1]{\red{[XX]}}
\newcommand{\refme}[1]{\red{(XX)}}

\newcommand{\cmark}{\ding{51}}%
\newcommand{\xmark}{\ding{55}}%

\newcommand{\cD}{\mathcal{D}}

\newcommand{\cN}{\mathcal{N}}

\newcommand{\cS}{\mathcal{S}}

\newcommand{\cX}{\mathcal{X}}
\newcommand{\cY}{\mathcal{Y}}

\newcommand{\vd}{\mathbf{d}}

\newcommand{\vg}{\mathbf{g}}

\newcommand{\vn}{\mathbf{n}}

\newcommand{\vq}{\mathbf{q}}

\newcommand{\vv}{\mathbf{v}}

\newcommand{\vx}{\mathbf{x}}
\newcommand{\vy}{\mathbf{y}}

\makeatletter
\newcommand*\bdot{\mathpalette\bdot@{.7}}
\newcommand*\bdot@[2]{\mathbin{\vcenter{\hbox{\scalebox{#2}{$\m@th#1\bullet$}}}}}
\makeatother

\makeatletter
\DeclareRobustCommand\onedot{\futurelet\@let@token\@onedot}
\def\@onedot{\ifx\@let@token.\else.\null\fi\xspace}

\def\eg{\emph{e.g}\onedot} 
\def\ie{\emph{i.e}\onedot} 
\def\cf{\emph{cf}\onedot} 
 \def\vs{\emph{vs}\onedot}
\def\wrt{w.r.t\onedot}  
\def\etal{\emph{et al}\onedot}
\makeatother

\begin{document}
\title{\projabbrv: Efficient, Imperceptible Adversarial\\3D Point Clouds}

\author{Hanwei Zhang\inst{1,2} 
	\and
	Luo Cheng\inst{3,4} 
	\and
	Qisong He\inst{3,5} 
	\and
	Wei Huang\inst{6} 
	\and
	Renjue Li\inst{3,4} 
	\and
	Ronan Sicre\inst{7} 
	\and
	Xiaowei Huang\inst{5} 
	\and
	Holger Hermanns\inst{2} 
	\and
	Lijun Zhang\inst{3} 
}
\authorrunning{Hanwei Zhang et al.}

\institute{
Institute of Intelligent Software, Guangzhou \and
Universität des Saarlandes \and
Institute of Software, Chinese Academy of Sciences\and
University of Chinese Academy of Sciences \and
University of Liverpool \and
Purple Mountain Laboratories \and
LIS, Centrale Méditerranée Marseille
}
\maketitle              
\begin{abstract}
Classification of 3D point clouds is a challenging machine learning (ML) task with important real-world applications in a spectrum from autonomous driving and robot-assisted surgery to earth observation from low orbit. As with other ML tasks, classification models are notoriously brittle in the presence of adversarial attacks. These are rooted in imperceptible changes to inputs with the effect that a seemingly well-trained model ends up misclassifying the input. 
This paper adds to the understanding of adversarial attacks by  presenting \projabbrv, a framework providing \textbf{E}fficient \textbf{I}mperceptible a\textbf{D}versarial attacks on 3D p\textbf{O}int cloud\textbf{S}. 
\projabbrv\ supports a diverse set of imperceptibility metrics. It employs an iterative, two-step procedure to identify optimal adversarial examples, thereby enabling a runtime-imperceptibility trade-off. 
We provide empirical evidence relative to several popular 3D point cloud classification models and several established 3D attack methods, showing  \projabbrv' superiority with respect to efficiency as well as imperceptibility. \projabbrv is an open-source project, and its code is available on GitHub at \url{https://github.com/Uzukidd/eidos}.

\end{abstract}
\section{Introduction}

{3D point clouds are a crucial format for representing  shapes of 3D objects. Among others, they are the output produced by LiDAR sensors and thus of critical importannce in robotics  and autonomous vehicle applications.} They capture the surface geometry of the object by means of a discrete set of data points in 3D. The points within a point cloud are generally unordered, and this has been proven challenging when trying to apply (convolutional) Neural Network (NN) technology for 3D object recognition~\citep{singh20193d,maturana2015voxnet}.
Recent advances, including PointNet~\citep{pointnet}, PointNet++~\citep{pointnet++}, and other works~\citep{8953340,Densepoint,yang2019modeling}, address this challenge by capturing fine local structural information from the neighborhood of each point, leading to better performance on classification and segmentation tasks. With further advances in this direction, it can be expected that applications across ``high-risk'' sectors \citep{AIAct} are becoming in reach, including tasks like autonomous navigation and robot-assisted surgery, where any failure may have serious consequences.

\begin{figure*}[t]
\centering

\newcommand{\teaser}[5]{\extfig{intro-#1}{\tikz{
	\node[tight](a){\includegraphics[trim={10mm 1mm 10mm 1mm},clip,width=0.20\linewidth]{figures/firstFigure/#1}};
	\node[tight,draw=gray,fit=(a)](b){};
        \draw[dashed, red] (-0.6,0.3) rectangle (0.5,0.7);
        \node at(-0.6,-0.6) {\scriptsize #2};
        \node at(0.8,-0.6) {\scriptsize #3};
        \node at(-0.1,-0.6) {\tiny #4};
        \node at(-0.6, -0.6) {\scriptsize #5}
}}}
\setlength{\tabcolsep}{1pt}
\begin{tabular}{ccccc}
\teaser{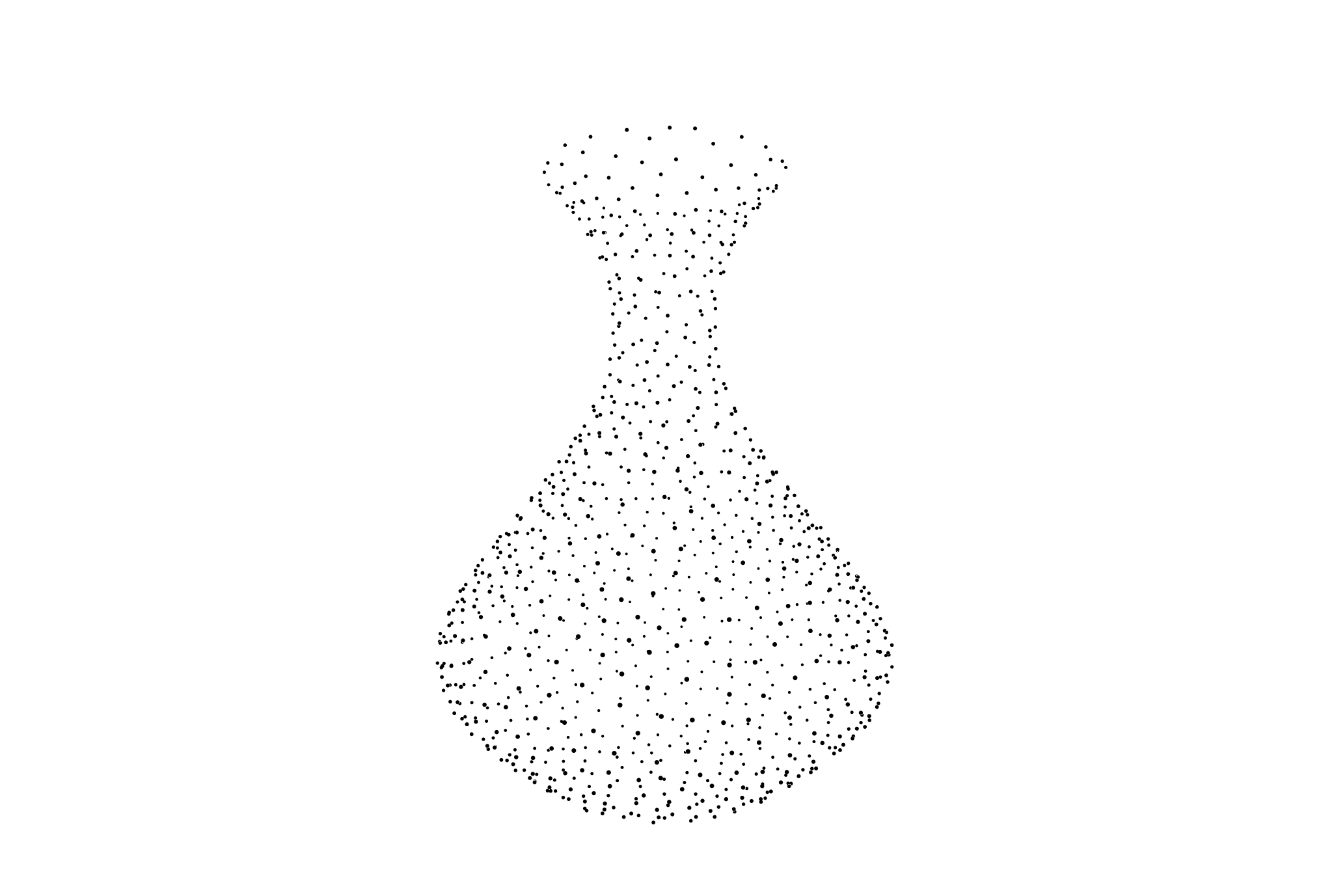}{\textbf{Clean}}{0}{}{} &
\teaser{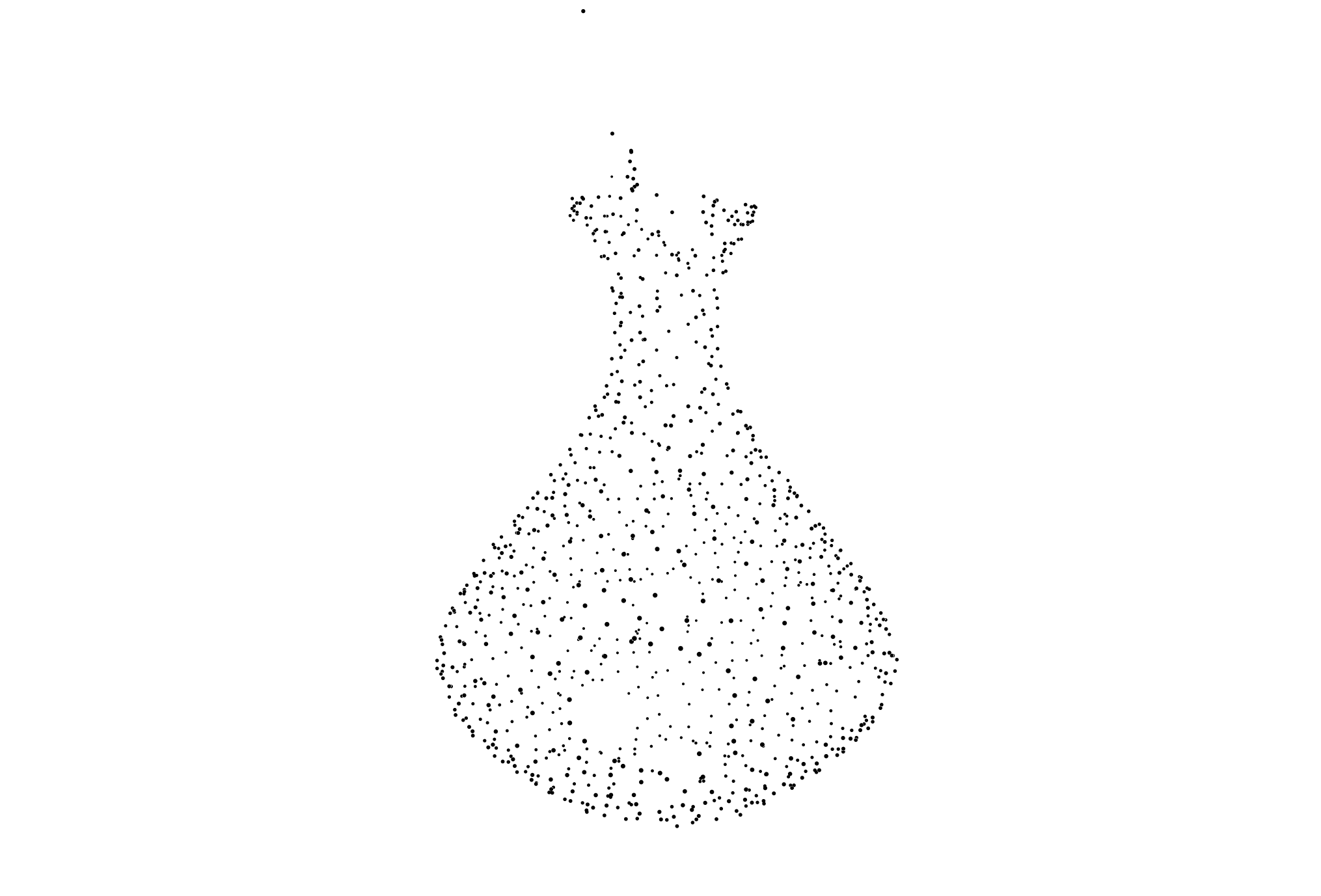}{\siadv}{0.65}{}{} & 
\teaser{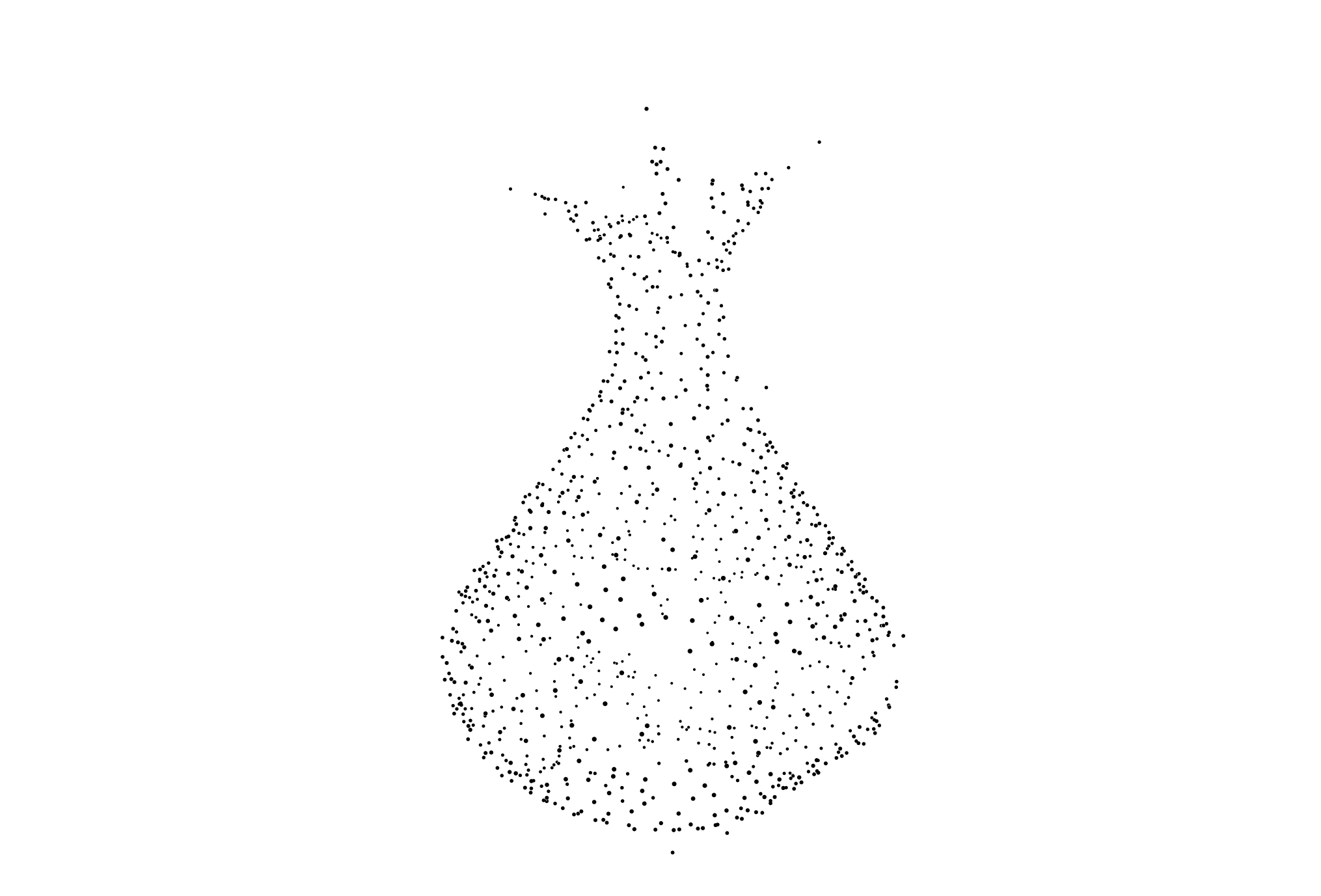}{\geoa}{92.61}{}{}  & 
\teaser{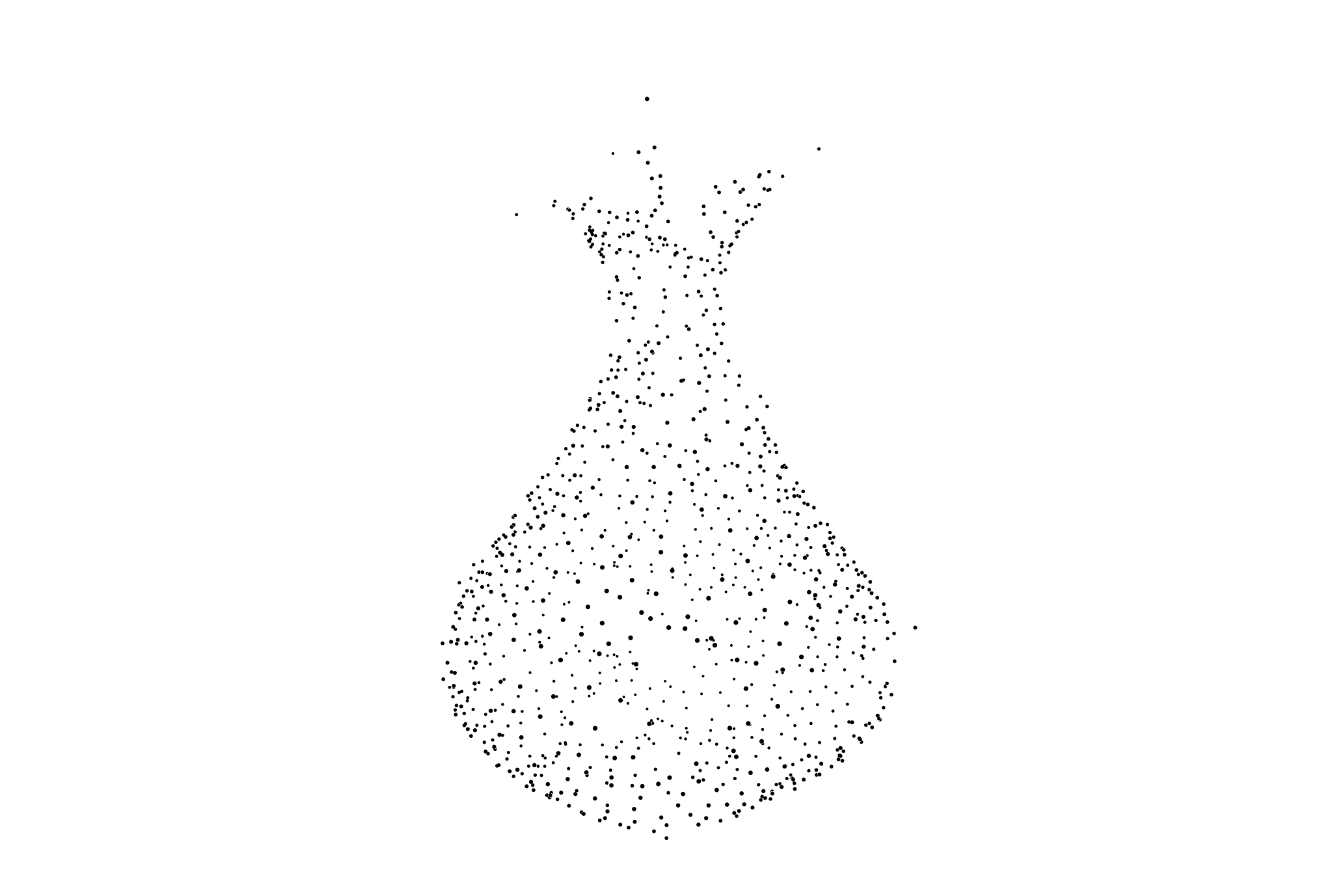}{\gsda}{93.59}{}{} & \teaser{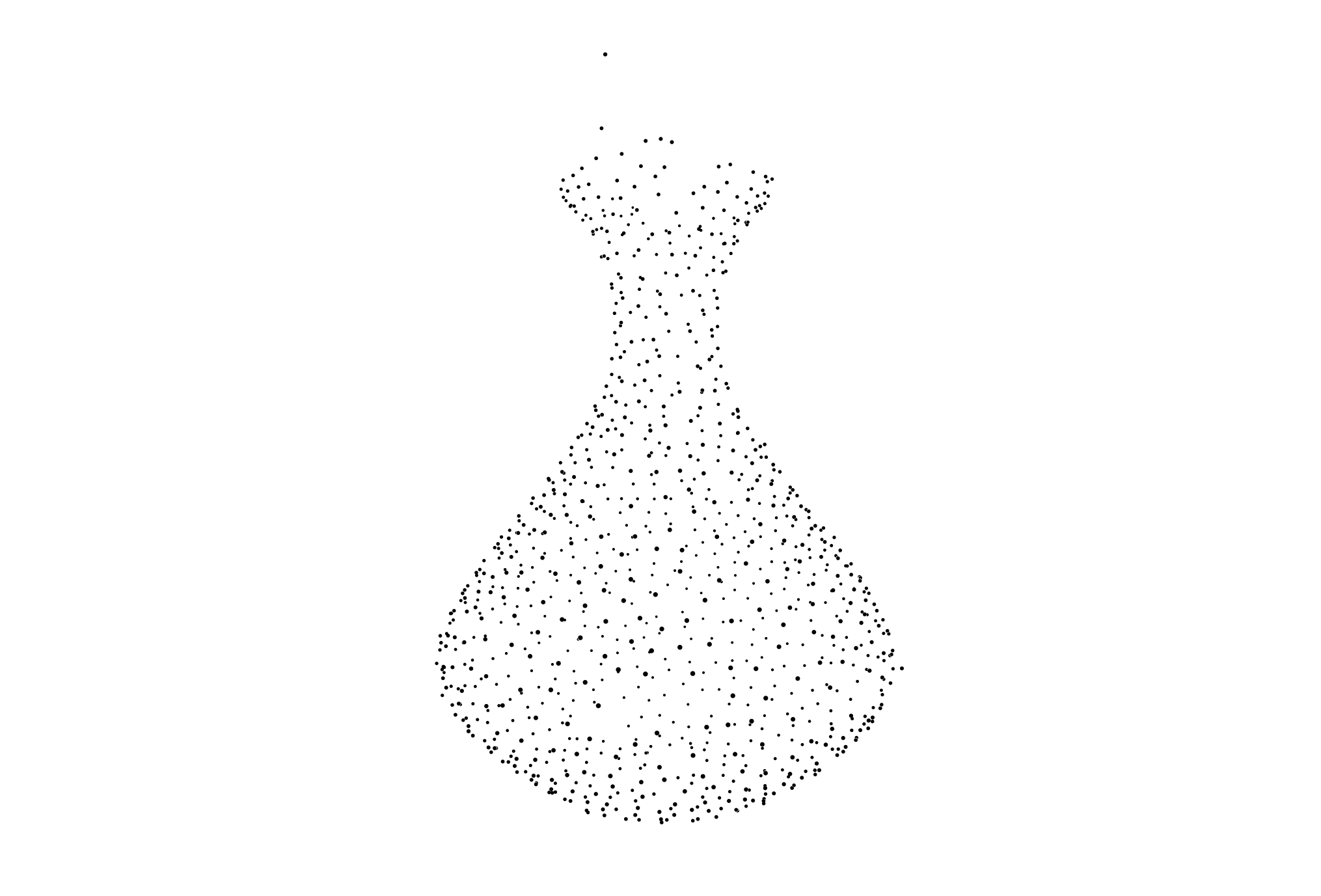}{\projabbrv}{0.07}{-\scalebox{0.8}{$D_{L_2}$}}{}\\
\end{tabular}
\caption{Visualization of adversarial point clouds. It shows the original sample and adversarial distortions generated by different attack methods. \projabbrv here is used with $D_{L_2}$ as imperceptibility regularization term. The number displayed in the bottom right denotes the mean $L_2$ norm of distortions, and it is clear that \our results in better imperceptibility than \siadv, \geoa, and \gsda.}
\label{fig:bp-teaser}
\end{figure*}


A prominent shortcoming across many advanced ML techniques -- especially those based on NN technology -- is their susceptibility to input distortions, meaning that small distortions of the input may induce a misclassification by the NN. 
Techniques to identify such issues are often devised in an \emph{adversarial} setting, where an adversary intentionally distorts the input slightly to induce a misclassification. 
For effective real-world applications, adversarial distortions need to be \emph{imperceptible} and should be computable \emph{efficiently}, i.e.,~the modifications should be subtle enough to avoid human detection and intervention, while also being computationally feasible.
However, recent attempts \citep{liu2019extending,geoa3,zhou2020lg} { fail to efficiently achieve imperceptibility
. The existing definitions of imperceptibility focus on  different aspects, and when optimizations are based on just one definition, the results often do not align with what is imperceptible to the human eye.}
As depicted in Figure~\ref{fig:bp-teaser}, attacks computed by  \siadv~\citep{si_adv_pc}, \geoa~\citep{geoa3}, and also \gsda~\citep{gsda}, all produce adversarial point clouds that fail to be imperceptibly different from the original (\cf the region inside the dash-dotted boxes). 
Incorporating multiple definitions therefore appears as a better approach to capture the true essence of imperceptibility, but this idea so far lacks computational efficiency \citep{geoa3}.
To address these limitations, we present a novel adversarial attack framework that computes imperceptible adversarial distortion of 3D point clouds in an efficient manner. We name our method \projabbrv\ after Plato's term ``eidos'', which means the permanent reality that makes a thing what it is (as opposed to the particulars that are finite and subject to change). Translated to our practice, \projabbrv\ makes optimal adversarial 3D point cloud generation a reality. 

\projabbrv\ is distinguished by its ability to work with a diverse set of imperceptibility regularization terms and to consider them altogether.
For this, we formalize the adversarial attack as a constrained optimization problem, with the goal of minimizing the imperceptibility of adversarial distortion with the additional constraint of enforcing misclassification.
With this formalization, we can study the relations between different regularization terms to guide the search for imperceptible adversarial distortions. 
Our work examines the power of the following imperceptibility regularizations: L2 norm, Chamfer Distance (CD), Hausdorff Distance (HD), and Consistency of Local Curvature (Curv), each of which echoeing distinct imperceptibility traits of adversarial distortions. As shown in Figure~\ref{fig:bp-teaser}, \projabbrv\ finds an adversarial example with extremely small distortion when only employing the L2 norm as an imperceptibility regularizer.

Two competing objectives make the optimization difficult. It has been observed that naively optimizing over the classification loss and imperceptibility together fails to generate adversarial distortions. Existing attacks, \eg \geoa~\citep{geoa3} use a hyper-parameter $\lambda$ to control the balance between imperceptibility and misclassification. However, this may lead to failing attacks or oscillations on the boundary, and thus comes with a waste of computational resources. In contrast, \our tackles the efficiency problem by decomposing the optimization into two phases, inspired by boundary projection (BP) attacks~\citep{zhang2020walking} originally designed for 2D images. The \textsc{IN} phase aims at minimizing the classification loss quickly to identify adversarial examples.
During the \textsc{OUT} phase, the search direction is determined by enforcing two conditions: minimizing imperceptibility and maintaining orthogonality to the gradient direction, thus preventing oscillations.
Therefore, this two-phase optimization can
find adversarial examples while minimizing the chosen imperceptibility in an efficient way. As we see in Figure~\ref{fig:bp-teaser}, \projabbrv\ can generate more imperceptible adversarial distortions than the state of the art.

We report on an extensive empirical evaluation of our approach, demonstrating its sensitivity and effectiveness. We assess effectiveness using success probability and imperceptibility metrics, while efficiency is measured as computation time. For sensitivity, we investigate different parameters of the algorithm, including step size, the number of iterations, and the number of imperceptibility regularization terms. 

\paragraph{Contributions} Our contributions are as follows. 
\begin{itemize}[itemsep=1pt,topsep=1pt,parsep=1pt]
    \item {We propose an efficient framework, which facilitates the incorporation of a diverse set of imperceptibility metrics. We further explore the relationship between them in depicting the imperceptibility traits of point clouds while existing works do not discuss this in detail.
    }
    \item {Our approach \emph{efficiently and effectively} handles adversarial optimization with several imperceptibility regularizations, avoids the competition between classification loss and imperceptibility, and provides a better trade-off than existing works.
    }
    \item Our attack achieves decent performance against models with defense and easily adapts to black-box settings.
    \item We provide a comprehensive and fair evaluation of \projabbrv\ and the state-of-the-art. 
\end{itemize}

\projabbrv is an open-source project, and its code is available on GitHub at \url{https://github.com/Uzukidd/eidos}.

\section{Problem, Background, and Related Work}
\label{sec:related}

\subsection{Problem Formulation}

\paragraph{Preliminaries}
Let $\cX \defn \{\vx_0, \cdots, \vx_n\}$ denote a set of $n$ 3D points represented by their 3D coordinates $\vx_i \defn [x_{i,x},x_{i,y},x_{i,z}]^\intercal \in \mathbb{R}^3$. A classifier $f: \mathbb{R}^{n \times 3} \rightarrow \mathbb{R}^c$ maps a point cloud $\cX$ to a vector $f(\cX)$ representing probabilities of $c$ classes. The classifier prediction $\pi : \mathbb{R}^{n \times 3} \rightarrow [c] \defn \{1, \cdots, c\}$ maps $\cX$ to the class label with maximum probability:
\begin{equation}
    \pi(\cX) \defn \arg \max_{k\in [c]} f(\cX)_k.
\end{equation}
The prediction is correct if $\pi(\cX) = t$ where $t \in [c]$ is the ground truth label.

\paragraph{Problem}
Let $\cX \in \mathbb{R}^{n \times 3}$ be a given 3D point cloud with known ground truth label $t$. An adversarial example $\cY \defn \cX + \Delta \in \mathbb{R}^{n \times 3}$ is a 3D point cloud such that:
\begin{itemize}
    \item[i)] probability of the ground truth class is small enough to result in \emph{misclassification}, \ie $f(\cY)_t < f(\cY)_{k\in [c], k \neq t}$;
    \item[ii)] the distortion $\Delta$ is \emph{imperceptible}.
\end{itemize}

\paragraph{Imperceptibility}
\label{sec:imper}
We list well-known metrics of distortion:
\begin{enumerate}[itemsep=1pt,topsep=1pt,parsep=1pt]
    \item \emph{$L_p$ Norm}
    \begin{equation}
        D_{L_p}(\cX,\cY) \defn (\sum_i(\vx_i - \vy_i)^p)^{\frac{1}{p}},
    \label{eq:dis-lp}
    \end{equation}
    where $\vy_i$ is the corresponding point of $\vx_i$ in set $\cY$, {assuming} {
    $|\cX| = |\cY|$}. Following the assumption made in~\citep{origin} that the modification $\Delta$ is small enough, norm $L_p$, specifically $L_2$, is employed as an imperceptibility metric. 
    \item \emph{Chamfer Distance}
    \begin{equation}
        D_{CD}(\cX, \cY) \defn \frac{1}{n} \sum_j \min_i \|\vx_i - \vy_j\|_2^2  
    \label{eq:dis-charmfer}
    \end{equation}
    measures the distance between two point sets by averaging the distances of each point to its nearest neighbor from another set. This distance is popularly used in adversarial 3D points~\citep{origin,geoa3,si_adv_pc} but is less effective when a small portion of outlier points exist in the 3D point clouds. 
    \item \emph{Hausdorff Distance}
    \begin{equation}
        D_{HD}(\cX, \cY) \defn \max_j \min_i \|\vx_i - \vy_j\|_2^2
    \label{eq:dis-hausdorff}
    \end{equation}
    is a non-symmetric metric and attaches more importance to the outlier points in $\cY$.
    \item \emph{Consistency of Local Curvature}~\citep{geoa3}
    \begin{align}
        D_{Curv}(\cX,\cY) \defn & \frac{1}{n}\sum_j \|\kappa(\vy_j;\cY) - \kappa(\vx_i;\cX) \|_2^2 \\
        \stt &  \vx_i = \arg \min_i \|\vy_j - \vx_i\|_2,
        \label{eq:dis-curvature}
    \end{align}
    where $\kappa(\vx_i;\cX)$ measures the local geometry of the local $k$ point neighborhoods $\cN_{\vx_i} \subset \cX$ of the point $\vx_i$ and is defined as
    \begin{equation}
        \kappa(\vx_i,\cX) \defn \frac{1}{k} \sum_{\vx_j \in \cN_{\vx_i}} \mid \langle\frac{\vx_j - \vx_i}{\|\vx_j - \vx_i\|_2}, \vn_{\vx_i}\rangle \mid,
        \label{eq:kernel-curv}
    \end{equation}
    where $\langle \cdot,\cdot\rangle$ denotes inner product and $\vn_{\vx_i}$ denotes the unit normal vector to the surface at $\vx_i$.
    \item \emph{KNN Smoothness}~\citep{tsai2020robust}
    \begin{align}
        D_{Smooth}(\cX) &\defn  \frac{1}{n} \sum_i d_i \cdot \mathbbm{1}[d_i > \mu + \gamma\sigma]   \\
        \wrt ~~ d_i &\defn \frac{1}{k}  \left(\sum_{\vx_j \in \cN_{\vx_i}} \|\vx_i - \vx_j\|_2^2 \right),
        \label{eq:knn-smooth}
    \end{align}
    where 
    $\gamma$ is a defined parameter while $\mu$ and $\sigma$ are the mean and standard deviation of the distribution of distances among points, respectively. Unlike other metrics, KNN smoothness is not intended to assess the imperceptibility of modifications, but rather the naturalness of the modified 3D point cloud. This metric encourages every point to be close to its neighbors. 
\end{enumerate}

\subsection{Attacks}

In this part, we discuss related work in generating 3D adversarial point clouds and boundary projection attacks, sharing similar principles with our method in addressing the problem.

\paragraph{Boundary Projection (BP) Attack~\citep{zhang2020walking}}
Let $\vx$ denote an input image of true class $t$ and $\vy$ denote an adversarial example. The Boundary Projection (BP) attack generates adversarial examples by solving the optimization
\begin{align}
    \min_\vy & ~~\|\vx - \vy\|_2^2 \label{eq:bp-dis}\\
    \stt & f(\vy)_t - \max_{k \in [c], k\neq t} f(\vy)_k < 0. \label{eq:bp-mis}
\end{align}

BP attack disentangles (\ref{eq:bp-dis}) and (\ref{eq:bp-mis}) into \textsc{OUT} case and \textsc{IN} case. In \textsc{IN} case, BP aims to find a solution satisfying (\ref{eq:bp-mis}), and searches along the gradient direction,
\begin{equation}
    \vy_{i+1} \defn \vy_i - \alpha \func{n}(\nabla_{\vx} \ell(f(\vy_i), t))),
    \label{eq:bp-gradient}
\end{equation}
where $\ell(f(\vx,t))$ is the negative cross-entropy loss of classifier $f(\cdot)$, $\func{n}(\cdot)\defn \frac{\cdot}{\|\cdot\|_2}$, and $\alpha$ is the step size. At initialization, $\vy_0 \defn \vx$ with fixed step size $\alpha$ to search adversarial examples as soon as possible. Once an adversarial solution is found, BP attack prioritizes reducing distortion with respect to (\ref{eq:bp-dis}). 

Regarding (\ref{eq:bp-dis}), BP attack sets a target distortion $\epsilon = \gamma_i \|\vy_{i} - \vx\|$, where $\gamma_i < 1$ is a parameter that increases linearly with iteration $i$, and then searches in the tangent hyperplane of the level set of the loss at $y_i$, along the direction that is normal to the gradient
\begin{align}
    \vy_{i+1} \defn &(\vy_i - \vv)\sqrt{[ \epsilon^2 - r^2 ]_+}\\
    \vv \defn & \vx + r\func{n}(\nabla_{\vx} \ell(f(\vy_i), t)))\\
    r \defn &\langle \vy_i - \vx, \func{n}(\nabla_{\vx} \ell(f(\vy_i), t))) \rangle, 
\end{align}
where $\vv$ is an auxiliary vector to follow the tangent hyperplane and $[\cdot]_{+}$ takes the positive value.
The above process can be iterative: if the current solution $\vy_i$ crosses the boundary and thus triggers going back to the \textsc{IN} case, BP attack updates with (\ref{eq:bp-gradient}) by setting $\alpha = r + \sqrt{\epsilon^2 - \|\vy_i-\vx\|^2 + r^2}$, which serves as an estimate of the step size required to cross the boundary along the given direction under a linear assumption.

Following the principle of BP~\citep{zhang2020walking}, \ie searching along the gradient of misclassification and minimizing the distortion along its orthogonal directions, we incorporate multiple metrics to assess the imperceptibility of unordered 3D point clouds and introduce Gram-Schmidt (GS) process to orthonormalize the gradient of different optimization terms.
Such a framework allowed exploring various imperceptibility regularizations and identifying a trade-off between efficiency and imperceptibility. 
Our experiments reveal that some metrics conflict with each other. 

\paragraph{Adversarial Attacks on 3D Point Clouds}
Compared to 2D images, a 3D point cloud consists of a set of \emph{unordered} points that represent the surface geometry of an object. Some measurements commonly used in 2D images, such as measuring the magnitude of distortion via $L_2$ norm or evaluating the photo-realistic of adversarial examples via PSNR and SSIM~\citep{wang2004image}, are not faithful when evaluating the imperceptibility of distortion to 3D point clouds.
Several attacks of 3D point clouds are gradient-based methods~\citep{liu2019extending,hamdi2020advpc,geoa3,tsai2020robust,yang2019adversarial,kim2021minimal,shi2022shape}, inspired by the adversarial attacks against image classifiers.
For instance, Su \etal~\citep{liu2019extending} adapted FGSM~\citep{goodfellow2014explaining}, I-FGSM~\citep{kurakin2016physical} and JSMA~\citep{papernot2016limitations} by imposing constraints on the distortion for each point or the entire point cloud and enhanced clipping and gradient projection to preserve the distribution of points on the surface of an object. Building on this, MPG~\citep{yang2019adversarial} introduces a Momentum-Enhanced Point-wise Gradient Method. 
In addition to perturbing point clouds, point addition and subtraction are proposed in the case of unordered point sets. To mislead the model, the attacker adds (removes) a limited number of synthesized points/clusters/objects to (from) a point cloud according to a saliency map~\citep{zheng2019pointcloud,origin,wicker2019robustness} or gradient~\citep{yang2019adversarial}. Within this category, attacks involving the imperceptible insertion or removal of a few points, like ~\citep{zheng2018learning,origin,liu2019extending,wicker2019robustness,liu2020adversarial}, are categorized as distribution attacks. Shape attacks~\citep{liu2020adversarial}, on the other hand, modify multiple points in specific areas of the point cloud. Moreover, adversarial attacks are also explored on mesh representations, for instance, mesh-attack~\citep{mesh-attack} and $\epsilon$-ISO\citep{miao2022isometric}. 

\paragraph{Imperceptible Adversarial Distortions}
We review several recent techniques for generating imperceptible perturbations in 3D point clouds that are resistant to adversarial attacks.
Tsai \etal~\citep{tsai2020robust} proposed an attack based on K-Nearest Neighbor (KNN) loss, while \geoa~\citep{geoa3} introduced consistency of local curvatures as part of geometry loss. Both of them utilize the C\&W~\citep{carlini2017towards} to find adversarial examples, which is expensive. Normal Attack~\citep{tang2022normalattack} incorporates both the gradient and tangent direction at each point as a form of smooth regularization. The normal-tangent attack (NTA)~\citep{tang2022rethinking} employs a directional controlling loss to constrain the distortion along the normal or tangent direction of the gradient. Yeung \etal~\citep{kim2021minimal} consider minimal modification with respect to the $L_0$ norm as the definition of imperceptibility and formalize it into $L_0$-norm optimization problem.
Other approaches focus on manipulating the point cloud representation itself.
Graph Spectral Domain Attack (\gsda)~\citep{gsda} converts the point clouds coordinates into graph spectral domain and perturbs point clouds within that domain. Similarly, \siadv~\citep{si_adv_pc} regards shape-invariant as the definition of imperceptibility and performs a reversible coordinate transformation on the input point cloud to guarantee the preservation of shape. Adversarial attacks, \eg \geoa and \gsda, producing imperceptible adversarial distortions often suffer from low time efficiency. \siadv is an efficient attack addressing imperceptibility. However, it only considers KNN smoothness and often generates outlier points. 
To overcome these limitations, we consider the imperceptibility regularization from \citep{geoa3} and provide insights into the trade-off between misclassification and imperceptibility regularization. 

\section{Method}

\projabbrv\ tackles adversarial optimization efficiently by decomposing the optimization into two phases governed by different objectives, \ie misclassification and imperceptibility. In the \textsc{IN} phase, we aim to find an adversarial point cloud, while in the \textsc{OUT} phase, we aim to optimize the imperceptibility metrics while keeping the current solution adversarial.

Algorithm~\ref{alg:bp-single} outlines our base principle of finding an adversarial example starting from the initial set $\cY_0 = \cX$. Then,  $\cY_i$ is updated iteratively in the direction of the gradient until an adversarial example is found, see \emph{Line 4-5} in Algorithm~\ref{alg:bp-single}. This is phase \textsc{IN}. The loss function is the misclassification loss $\ell(f(\cY_i),t) \defn  f(\cY)_t - \max_{k\in[c], k\neq t} f(\vy)_k $. The normalization function is defined as $\func{n}(\cdot)\defn \frac{\cdot}{\|\cdot\|_2}$, and  $\epsilon$ denotes the step size.
In \emph{line 6-9}, phase \textsc{out} treats the case of $\cY_i$ being adversarial and aims at improving imperceptibility, which can be based on a single metric like $L_p$ norm, while keeping the solution adversarial.
We obtain the normalized direction $\hat{\vd}$ that decreases the imperceptibility regularization, which is then projected onto the tangent hyperplane of the level set of the loss at $\cY_i$, normal to $\hat{\vg}$.

\begin{algorithm}[t]
\caption{\projabbrv\ - Base}
\label{alg:bp-single}
\textbf{Input:} $\cX$: original point cloud to be attacked \\
\textbf{Input:} $t$: true label, $K$: maximum iterations \\
\textbf{Input:} $\epsilon$: step size \\
\textbf{Output:} $\cY^*$
\begin{algorithmic}[1]
        \STATE $\cY^* \gets \cX \cdot \mathbf{0}$, $\cY_0 \gets \cX$
	\WHILE{$i < \IterMax$}	
	\STATE $\hat{\vg} \gets \func{n}(\nabla_{\cX} \ell(f(\cY_i), t)))$ 
        \IF[\COMMENT{$\rhd$ IN phase}]{$f(\cY_i) == t$} \label{alg1:bp-CaseIn} 
            \STATE $\cY_{i+1} \gets \cY_i - \epsilon\hat{\vg}$  
        \ELSE[\COMMENT{$\rhd$ OUT phase}] \label{alg1:bp-CaseOut}
		\STATE $\hat{\vd} \gets \func{n}(\nabla_{\cX} D(\cX,\cY_i))$
		\STATE $\vv \gets \hat{\vd} - \frac{\hat{\vd}\hat{\vg}}{\|\hat{\vg}\|_2^2}\hat{\vg}$ \label{alg1:bp-Z} 
		\STATE $\cY_{i+1} \gets \cY_{i} - \epsilon\vv$ \label{alg:bp-Qout} 
	\ENDIF
        \IF{$D(\cX,\cY_{i+1}) < D(\cX,\cY^*)$}
        \STATE $\cY^* = \cY_{i+1}$
        \ENDIF
	\STATE $i \gets i+1$
	\ENDWHILE
\end{algorithmic}
\end{algorithm}

To account for multiple metrics used to measure imperceptibility, we enhance phase \textsc{out} by optimizing different imperceptibility regularization terms \emph{alternatingly}.
In Algorithm~\ref{alg:bp-alter}, \emph{line 7-9} calculates the normalized direction $\hat{\vd}_i$ for a set $\cD$ of imperceptibility regularizations. In \emph{line 10}, we use the Gram-Schmidt (GS) process to calculate an orthogonal set based on the gradients of misclassification and imperceptibility regularization terms, which allows us to optimize each of them independently. In \emph{line 12-15}, our method searches along the direction that decreases imperceptibility regularization term $D_j$ while saving the best solution with respect to the sum of all imperceptibility regularization terms. This is the core innovation of \our, together with the \textsc{IN-OUT} phase splitting.

\begin{algorithm}[t]
\caption{\projabbrv}
\label{alg:bp-alter}
\textbf{Input:} $\cX$: original point cloud to be attacked \\
\textbf{Input:} $t$: true label, $K$: maximum iterations \\
\textbf{Input:} $\epsilon$: step size\\
\textbf{Input:} $\cD$: a set of imperceptibility regularizations\\
\textbf{Output:} $\cY^*$
\begin{algorithmic}[1]
        \STATE $\cY^* \gets \cX \cdot \mathbf{0}$, $\cY_0 \gets \cX$
	\WHILE{$i < \IterMax$}
	\STATE $\hat{\vg} \gets \func{n}(\nabla_{\cX} \ell(f(\cY_i), t)))$ 
        \IF[\COMMENT{$\rhd$ IN phase}]{$f(\cY_i) == t$}\label{alg2:bp-CaseIn}             
            \STATE $\cY_{i+1} \gets \cY_i - \epsilon \hat{\vg}$ 
        \ELSE[\COMMENT{$\rhd$ OUT phase}] \label{alg2:bp-CaseOut} 
            \FOR{$D_j \in \cD$}
                \STATE $\hat{\vd_j} \gets \func{n}(\nabla_{\cX} D_j(\cX,\cY_i))$
            \ENDFOR
		\STATE $\{\hat{\vv}_1,\cdots,\hat{\vv}_m\} \gets  GS(\{\hat{\vg},\hat{\vd}_1,\cdots,\hat{\vd}_m\})$ \label{alg2:bp-Z}
            \FOR{$\hat{\vv}_j \in \{\hat{\vv}_1,\cdots,\hat{\vv}_m\}$}
                    \STATE $\cY_{i+1} \gets \cY_{i} - \epsilon\hat{\vv}_{j}$ \label{alg2:bp-Qout}
                \IF{$\sum_j D_{j}(\cX,\cY_{i+1})<\sum_j D_{j}(\cX,\cY^*)$}
                \STATE $\cY^* = \cY_{i+1}$
                \ENDIF
            \ENDFOR
		 
	\ENDIF
	\STATE $i \gets i+1$
	\ENDWHILE
\end{algorithmic}
\end{algorithm}
\section{Experiments}

In this section, we first present our experimental setting, and then carry out an ablation study on coordinate transformation, imperceptibility regularization, and step size. More ablation studies on step size and the number of iterations, as well as the discussion of convergence are in the supplementary materials. 
We assess the performance of our method relative to \gsda and \siadv as the current state-of-the-art techniques regarding imperceptibility in this setting. Additionally, we compare with \geoa which incorporates the fusion of various imperceptibility metrics as a form of regularization.
Last but not least, we evaluate our method against three common defense techniques and test our attack in a black-box setting.

\subsection{Experiments Setting}

\paragraph{Dataset} The ModelNet40~\citep{wu20153d} dataset contains 12,311 CAD models from the 40 most common object categories in the world. There are 9,843 training objects and 2,468 testing objects. Following the previous setting~\citep{origin,pointnet}, we uniformly sample 1,024 points from the surface of each object and re-scale them into a unit ball. For attacks, we randomly select 100 test examples from each of the 10 largest classes, \ie airplane, bed, bookshelf, bottle, chair, monitor, sofa, table, toilet, and vase, as original point clouds to generate adversarial point clouds. Given our focus on untargeted attacks, we conduct experiments with 1,000 point clouds.

\paragraph{Networks} We take the pre-trained model of PointNet~\citep{pointnet}\footnote{\url{https://github.com/shikiw/SI-Adv}} trained with several data augmentations such as random point-dropping and rotation. 
{We also use the pre-trained model of DGCNN~\citep{dgcnn}\footnote{\url{https://github.com/WoodwindHu/GSDA}} trained by Hu \etal. To verify our approach works on various architectures, we provide a comparison on Point-transformer\footnote{\url{https://github.com/lulutang0608/Point-BERT}} as well.}

\paragraph{Attacks} Following the original setting of \geoa and \gsda, we use Adam optimizer~\citep{papernot2016limitations} with a fixed learning schedule of $500$ iterations. The learning rate and momentum are set as $0.01$ and $0.9$, respectively. For the weights of geometry-aware regularization of \geoa, we let $\lambda_1=0.1, \lambda_2=1.0, \lambda_3=1.0$. The penalty parameter is initialized as $\beta=2,500$ and automatically adjusted by conducting $10$ runs of binary search. We set $k=16$ to define local point neighborhoods in \geoa. In \gsda, we let $k=10$ for building the KNN graph, and let the penalty parameter $\beta=10$ at the beginning and adjust it after $10$ runs of binary search. The weights of Chamfer distance loss and Hausdorff distance loss in the regularization term are set to $5.0$ and $0.5$, respectively. 
We use the white-box version of \siadv with the step size $0.007$ and maximum iterations $100$. Adversarial examples are constrained by the $L_\infty$ norm ball with $0.16$ radius. For our method, we set maximum iterations as $100$.  

\paragraph{Evaluation Metrics} To quantitatively compare adversarial results across different methods, we use attack the success rate \asr and the imperceptibility metrics outlined in Section~\ref{sec:imper}. We use $k=16$ to define local point neighborhoods for $D_{Curv}$ and $D_{Smooth}$ and $\gamma = 1.05$ for $D_{Smooth}$. As evaluation metrics, we denote $L_2$ norm as \lt, Chamfer distance as \cd, Hausdorff distance as \hd, consistency of local curvature as \curv, and KNN smoothness as \smooth. We report time in seconds, noted T, of each attack measured on a TITAN V 250W+Intel(R) Xeon(R) CPU E5-2630 v4 \_ 2.20GHz.

To compare the different attacks fairly, we follow the evaluation protocol~\citep{zhang2020smooth} for operating characteristics: for $D \in [0,D_{max}]$,
\begin{equation}
    \func{P} \defn \frac{1}{N_{suc}}|\{\cY \in X_{suc} : D(\cY,\cX) < D\}|,
    \label{eq:curve-eval}
\end{equation}
where $N_{suc}$ is the total number of the subset of adversarial point cloud $X_{suc}$ that succeeded to deceive the network. This function varies from $\func{P}(0)=0$ to $\func{P}(D_{max}) = P_{suc}$.


\begin{table}[hptb]

\caption{\projabbrv success rate \asr (\%), average values of \lt ($1e-1$), \cd ($1e-4$), \hd ($1e-2$), \curv ($1e-2$), \smooth ($1e-3$) and time (s) of our attack constrained by $D_{L_2}$ with different coordinate transformation against PointNet. 
}

\centering
\footnotesize
\setlength{\tabcolsep}{8pt}

\begin{tabular}{p{1.5cm} c | ccccc | c}\toprule
\Th{method}& \asr$\uparrow$& \lt$\downarrow$&\cd$\downarrow$&\hd$\downarrow$&\curv$\downarrow$&\smooth$\downarrow$&Time$\downarrow$\\\midrule
$T_{ori}$ & \tb{100} & \tb{0.71} & \tb{0.61} & \tb{1.41} & \tb{0.12} & 1.48 & 1.56\\
$T_{rsi}$ & \tb{100} & 2.03 & 1.40 & 3.17 & 0.14 & \tb{1.25} & 2.21\\
$T_{gft}$ & \tb{100}  & 0.49 & 0.47  & 1.05 & 0.12 & 1.55 & 10.28 \\
\bottomrule
\end{tabular}

\label{tab:abl-coor-trans}
\end{table}

\subsection{Ablation}
Before studying the impact of various imperceptibility factors, we start off by an ablation analysis on coordinate transformation and imperceptibility regularization. Specifically, coordinate transformation are central to the manner in which \siadv and \gsda incorporate their imperceptibility constraints.
Further ablation studies then explore the impact of various parameters of our method, including coordinate transformation, imperceptibility regularization terms, step size, and the maximum number of iterations. Due to page limits, the experimental results on step size, and maximum number of iterations are in the supplement.

\paragraph{Influence of Coordinate Transformation}
To improve imperceptibility, \siadv and \gsda transform  coordinates. To evaluate how much these coordinate transformations contribute to the final results, we carry out the attack with fixed step size $0.06$ for $100$ iterations with $L_2$ norm constraint and:
\begin{itemize}[itemsep=1pt,topsep=1pt,parsep=1pt]
    \item[$T_{ori}$:] without any coordinate transformation;
    \item[$T_{rsi}$:] with the reversible shape-invariant coordinate transformation of \siadv;
    \item[$T_{gft}$:] with the Graph Fourier Transform (GFT) of \gsda before and after the optimization process.
\end{itemize}

Table~\ref{tab:abl-coor-trans} shows $T_{rsi}$ and $T_{gft}$ performing well only in the \smooth metric. In a nutshell, coordinate transformations are of little value for \projabbrv  and thus not considered in the sequel.


\begin{table}[hptb]

\caption{\projabbrv success rate \asr (\%), average values of \lt ($1e-1$), \cd ($1e-4$), \hd ($1e-2$), \curv ($1e-2$), \smooth ($1e-3$) and time (s) of our attack constrained by different imperceptibility regularization with $T_{ori}$ and $T_{rsi}$ coordinate transformations against PointNet. 
}

\centering
\footnotesize
\setlength{\tabcolsep}{8pt}

\begin{tabular}{p{1.5cm} c | ccccc | c}\toprule
\Th{method}& \asr$\uparrow$& \lt$\downarrow$&\cd$\downarrow$&\hd$\downarrow$&\curv$\downarrow$&\smooth$\downarrow$&Time$\downarrow$\\\midrule
$D_{L_2}$ & \tb{100} & \tb{0.71} & 0.61 & 1.41 & \tb{0.12} & 1.48 & 1.56\\
$D_{CD}$ & \tb{100} & 5.07 & \tb{0.37} & 0.93 & 0.16 & 1.56 & 1.84\\
$D_{HD}$ & \tb{100} & 5.76 & 1.58 & \tb{0.26} & 0.35 & 1.67 & 1.75\\
$D_{Curv}$ & \tb{100} & 7.26 & 3.47 & 5.07 & 0.20 & \tb{1.20} & 2.02\\
\bottomrule
\end{tabular}

\label{tab:abl-single-sloss}
\end{table}

\paragraph{Influence of Different Imperceptibility Regularizations} 
Initially, we examine performance by separately optimizing imperceptibility regularization terms, such as $D_{L_2}$, $D_{CD}$, $D_{HD}$, or $D_{Curv}$, in Algorithm \ref{alg:bp-single} with a fixed step size of $0.06$.
As shown in Table~\ref{tab:abl-single-sloss}, we find that using the imperceptibility regularization induces optimality in the corresponding metric, except for $D_{Curv}$ that gives the optimal results for \smooth. 
Figure~\ref{fig:abl-single-sloss} shows operating characteristics for Table~\ref{tab:abl-single-sloss}. \our with $D_{L_2}$ generates adversarial examples with extremely small \lt , with a few exceptions that have large \lt (around $2$). 
Also, examples generated with $D_{L_2}$ often have similar \cd as if using $D_{CD}$ instead.
However, a few cases with a large value increase the average value in Table~\ref{tab:abl-single-sloss}. With $D_{HD}$ we obtain the worst results on \curv and vice versa. The correlation between these two imperceptibility regularization terms is relatively small, and thus it is difficult to optimize both of them together.

\begin{figure}
\centering

\setlength{\tabcolsep}{0.75pt}
\begin{tabular}{c c}
\extfig{l2}{
\begin{tikzpicture}
\begin{axis}[
	height=3cm,
	width=6cm,
        yticklabel style={font=\tiny},
	xlabel={\lt},
	ylabel={\asr},
        legend pos=outer north east]
    \addplot[blue] table{figures/eval/single_reg/l2/BP-CD_5.txt};\leg{$D_{CD}$}
    \addplot[black] table{figures/eval/single_reg/l2/BP-HD_5.txt};\leg{$D_{HD}$}
    \addplot[red] table{figures/eval/single_reg/l2/BP-L2_5.txt};\leg{$D_{L_2}$}
    \addplot[green] table{figures/eval/single_reg/l2/BP-CURV_5.txt};\leg{$D_{Curv}$}
    \legend{};
\end{axis}
\end{tikzpicture}
}
& 
\extfig{CD}{
\begin{tikzpicture}
\begin{axis}[
	height=3cm,
	width=6cm,
	xlabel={\cd},
	ylabel={\asr},
        yticklabel style={font=\tiny},
        legend pos=outer north east]
    \addplot[blue] table{figures/eval/single_reg/cd/BP-CD_5.txt};\leg{$D_{CD}$}
    \addplot[black] table{figures/eval/single_reg/cd/BP-HD_5.txt};\leg{$D_{HD}$}
    \addplot[red] table{figures/eval/single_reg/cd/BP-L2_5.txt};\leg{$D_{L_2}$}
    \addplot[green] table{figures/eval/single_reg/cd/BP-CURV_5.txt};\leg{$D_{Curv}$}
    \legend{};
\end{axis}
\end{tikzpicture}
}
\\
\extfig{HD}{
\begin{tikzpicture}
\begin{axis}[
	height=3cm,
	width=6cm,
        yticklabel style={font=\tiny},
        scaled x ticks = base 10:2,
	xlabel={\hd},
	ylabel={\asr},
        legend pos=outer north east]
    \addplot[blue] table{figures/eval/single_reg/hd/BP-CD_5.txt};\leg{$D_{CD}$}
    \addplot[black] table{figures/eval/single_reg/hd/BP-HD_5.txt};\leg{$D_{HD}$}
    \addplot[red] table{figures/eval/single_reg/hd/BP-L2_5.txt};\leg{$D_{L_2}$}
    \addplot[green] table{figures/eval/single_reg/hd/BP-CURV_5.txt};\leg{$D_{Curv}$}
    \legend{};
\end{axis}
\end{tikzpicture}
}
& 
\extfig{Curv}{
\begin{tikzpicture}
\begin{axis}[
	height=3cm,
	width=6cm,
        yticklabel style={font=\tiny},
	xlabel={\curv},
	ylabel={\asr},
        legend pos=outer north east]
    \addplot[blue] table{figures/eval/single_reg/curv/BP-CD_5.txt};\leg{$D_{CD}$}
    \addplot[black] table{figures/eval/single_reg/curv/BP-HD_5.txt};\leg{$D_{HD}$}
    \addplot[red] table{figures/eval/single_reg/curv/BP-L2_5.txt};\leg{$D_{L_2}$}
    \addplot[green] table{figures/eval/single_reg/curv/BP-CURV_5.txt};\leg{$D_{Curv}$}
    \legend{};
\end{axis}
\end{tikzpicture}
}
\\
\mc{2}{
\extfig{smooth}{
\begin{tikzpicture}
\begin{axis}[
	height=3cm,
	width=6cm,
	xlabel={\smooth},
	ylabel={\asr},
        yticklabel style={font=\tiny},
        legend style={font=\tiny},
        legend pos=outer north east]
    \addplot[blue] table{figures/eval/single_reg/smooth/BP-CD_5.txt};\leg{$D_{CD}$}
    \addplot[black] table{figures/eval/single_reg/smooth/BP-HD_5.txt};\leg{$D_{HD}$}
    \addplot[red] table{figures/eval/single_reg/smooth/BP-L2_5.txt};\leg{$D_{L_2}$}
    \addplot[green] table{figures/eval/single_reg/smooth/BP-CURV_5.txt};\leg{$D_{Curv}$}
\end{axis}
\end{tikzpicture}
}}
\\
    \end{tabular}
    \caption{Operating charateristics on PointNet for our attack constrained by different imperceptibility regularization.
    }
    \label{fig:abl-single-sloss}
\end{figure}

In Table~\ref{tab:abl-combine-sloss}, we investigate the combination of different imperceptibility regularization terms with adaptive step size, following Algorithm \ref{alg:bp-alter}.  We observe that combining different imperceptibility regularization terms does not provide a dominant solution on all the metrics but provides a trade-off solution for the imperceptibility regularization we optimize for. 
We identify the significance of $D_{L_2}$ in enhancing all imperceptibility metrics for adversarial distortions. Combining $D_{HD}$ and $D_{Curv}$ in \our yields the poorest results in \lt, \cd, and \curv, aligning with the observation in Figure~\ref{fig:abl-single-sloss}.

\begin{table}[hptb]

\caption{\projabbrv success rate \asr (\%), average values of \lt ($1e-1$), \cd ($1e-4$), \hd ($1e-2$), \curv ($1e-2$), \smooth ($1e-3$) and time (s) of our attack constrained by different imperceptibility regularization without coordinate transformation against PointNet. 
}

\centering
\footnotesize
\setlength{\tabcolsep}{7pt}

\begin{tabular}{l c | ccccc | c}\toprule
\Th{method}& \asr$\uparrow$& \lt$\downarrow$&\cd$\downarrow$&\hd$\downarrow$&\curv$\downarrow$&\smooth$\downarrow$&Time$\downarrow$\\\midrule
\footnotesize
$D_{L_2+HD}$ & \tb{100} &\tb{ 2.23} & 1.15 & \tb{0.52} & 0.16 & 1.54 & 1.92 \\
\footnotesize
$D_{L_2+Curv}$ & \tb{100} & 2.40 & 1.31 & 2.11 & \tb{0.09} & \tb{1.33} & 2.37 \\
\footnotesize
$D_{HD+Curv}$ & \tb{100} & 5.83 & 2.22 & 0.99 & 0.17 & 1.45 & 2.58 \\
\footnotesize
$D_{L_2+HD+Curv}$ & \tb{100} & 2.74 & 1.33 & 0.78 & 0.11 & 1.47 & 2.75 \\
\footnotesize
$D_{all}$  & \tb{100} &  3.37 &  \tb{0.88} & 0.71 &  0.11 & 1.50 & 2.46\\
\bottomrule
\end{tabular}

\label{tab:abl-combine-sloss}
\end{table}

\paragraph{Step Size and Iterations}
In the supplementary material, we report on the efficiency and sensitivity of our attack with respect to step size, testing six different fixed step sizes ($\epsilon \in [0.001,0.01,0.02,0.03,0.04,0.05,0.06,0.1]$), as well as adaptive step size $\epsilon_{i+1} \defn \epsilon (1-\gamma)$. In general, our approach turns out not to be sensitive to step size. A fixed step size works better with a single imperceptibility regularization term while adaptive step size performs better with multiple terms.
We also conduct experiments with varying maximum iterations ($K \in [20, 40, \cdots, 200]$). The findings indicate a correlation between convergence and the imperceptibility regularization term. Details are available in the supplement.

\begin{table}[hptb]
\caption{Success rate \asr (\%), average values of \lt ($1e-1$), \cd ($1e-4$), \hd ($1e-2$), \curv ($1e-2$), \smooth ($1e-3$) and time (s) of our attack compared with baseline attacks against three networks.}
\centering
\footnotesize
\setlength{\tabcolsep}{7.5pt}

\begin{tabular}{p{1.5cm} c | ccccc | c}\toprule
\Th{method} & \asr $\uparrow$ & \lt $\downarrow$ & \cd $\downarrow$ & \hd $\downarrow$ & \curv $\downarrow$ & \smooth $\downarrow$ &Time $\downarrow$ \\\midrule
\mc{8}{PointNet}\\ \midrule
\geoa & 97 &         44.86 &         4.81 &         0.73 &         0.36 &         1.73 &   125.48 \\ 
\siadv & \bf{100} &         6.51 &         3.08 &         4.25 &         0.27 &         \bf{1.20} & {0.08}\\ 
\gsda & 97 &         54.47 &         4.70 &         2.65 &         0.51 &         1.54 & 138.73 \\ 
\rowcolor{LightCyan}
\our  & \tb{100} &         \tb{3.37} &        \tb{0.88} &         \bf{0.71} &         \tb{0.11} &         1.50 &  2.46 \\\midrule
\mc{8}{DGCNN}\\ \midrule
\geoa & 100 &         195.45 &         6.95 &         0.46 &         {0.23} &         2.04 &  307.05\\
\siadv & 100 &         31.14 &         9.17 &         2.95 &         1.02 &         \tb{1.72} & {0.49}\\
\gsda & 100 &         145.36 &         6.06 &         0.54 &         0.27 &         1.94 &  329.83\\
\rowcolor{LightCyan}
\our  & 100 &         \tb{23.65} &         \tb{4.48} &         \tb{0.45} &         0.53 &         1.80 &  6.77\\\midrule
\mc{8}{{Point-Transformer}}\\ \midrule
\geoa & 61.4 &         47.50 &         5.81 &         0.43 &         0.69 &         1.91 &         267.79 \\
\siadv &  \tb{100.0} & 12.13 & 6.26 & 3.61 & 0.65 & \tb{1.47}&  0.30\\
\gsda & 61.8 &         57.03 &         5.83 &         2.13 &         0.84 &         1.76 &         291.32 \\
\rowcolor{LightCyan}
\our  & \tb{100.0} & \tb{6.86} & \tb{1.78} & \tb{0.37} & \tb{0.24} & 1.64 &  3.61\\\bottomrule
\end{tabular}

\label{tab:baseline}
\end{table}

\begin{figure*}[ptb]
\newcommand{\sizeS}{.21}  
\centering
\setlength{\tabcolsep}{0pt}
\!\!\!\!\!\!\begin{tabular}{@{}ccccc@{}} 
\textbf{Clean} &\geoa &\gsda &\siadv & \projabbrv\\
\figt[\sizeS]{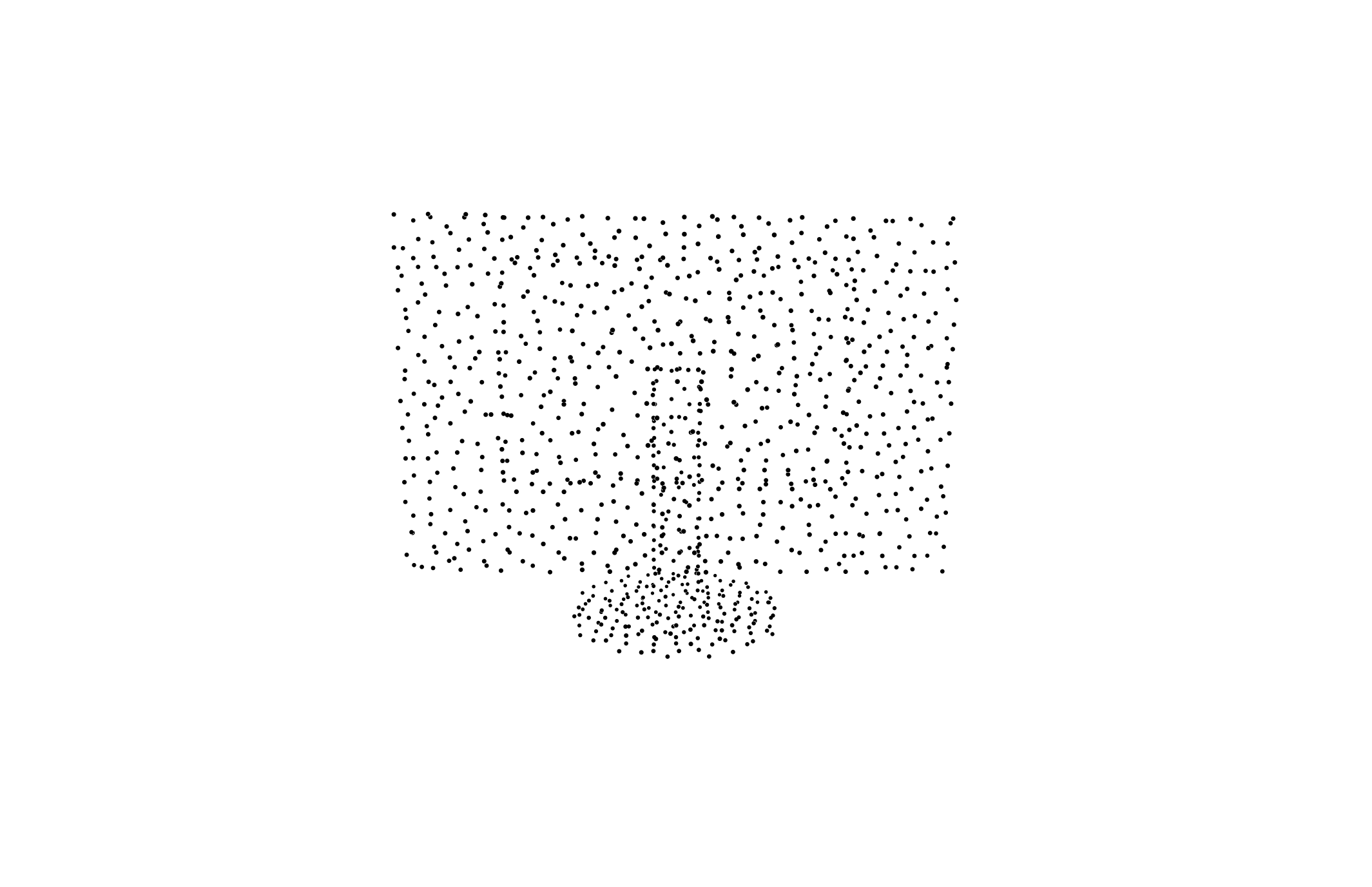} &
\figt[\sizeS]{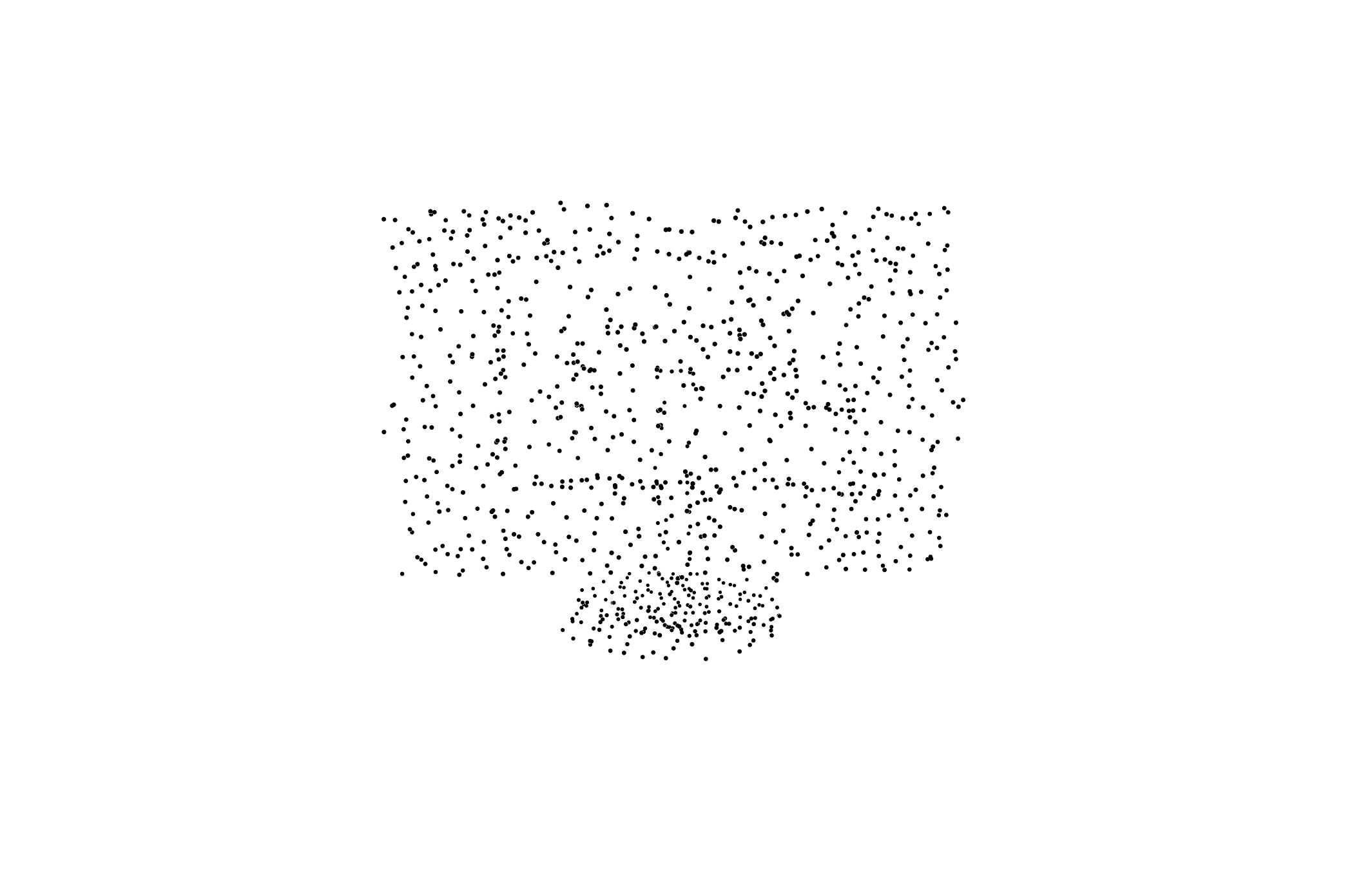} &
\figt[\sizeS]{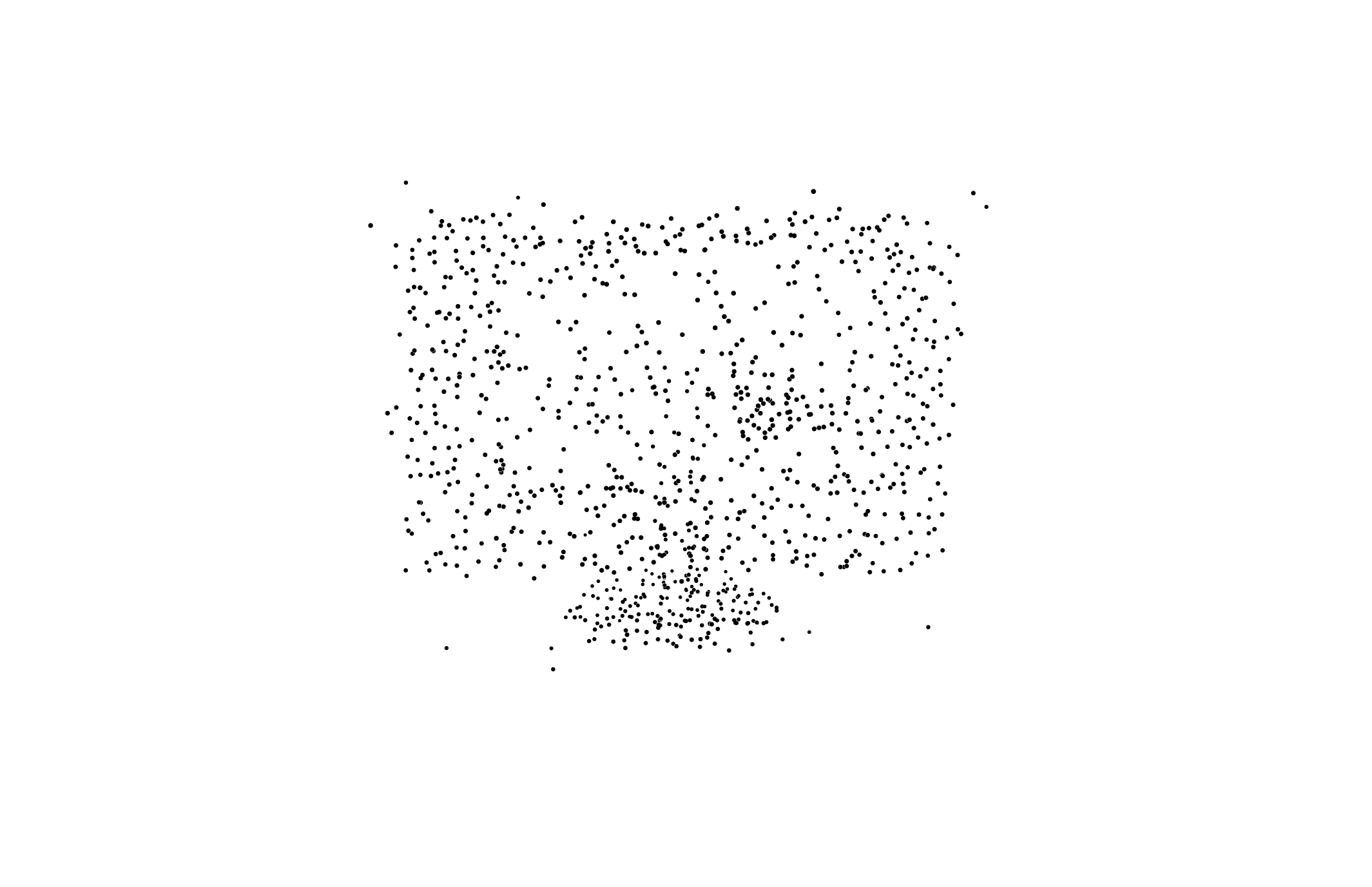} &
\figt[\sizeS]{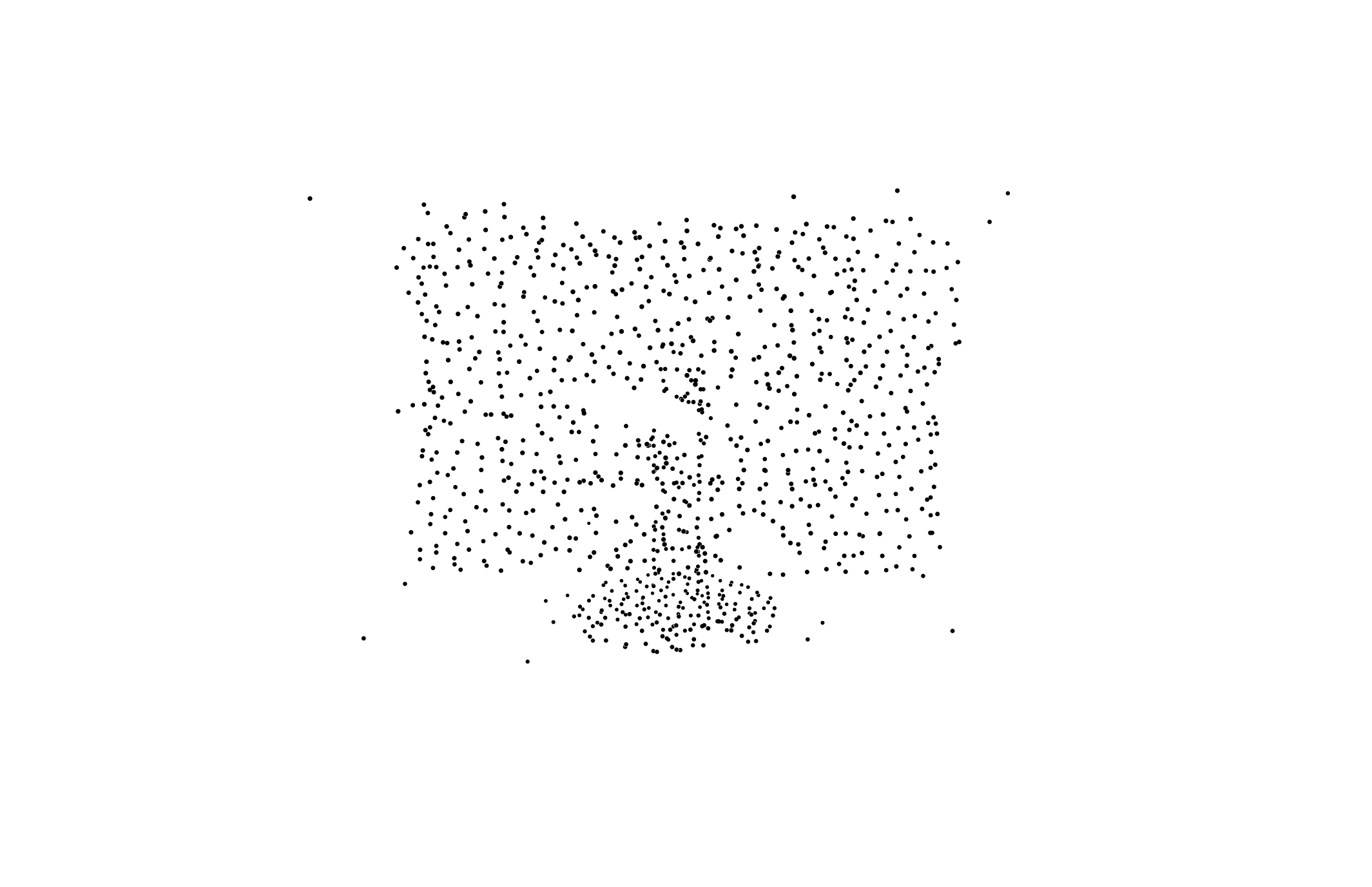} &
\figt[\sizeS]{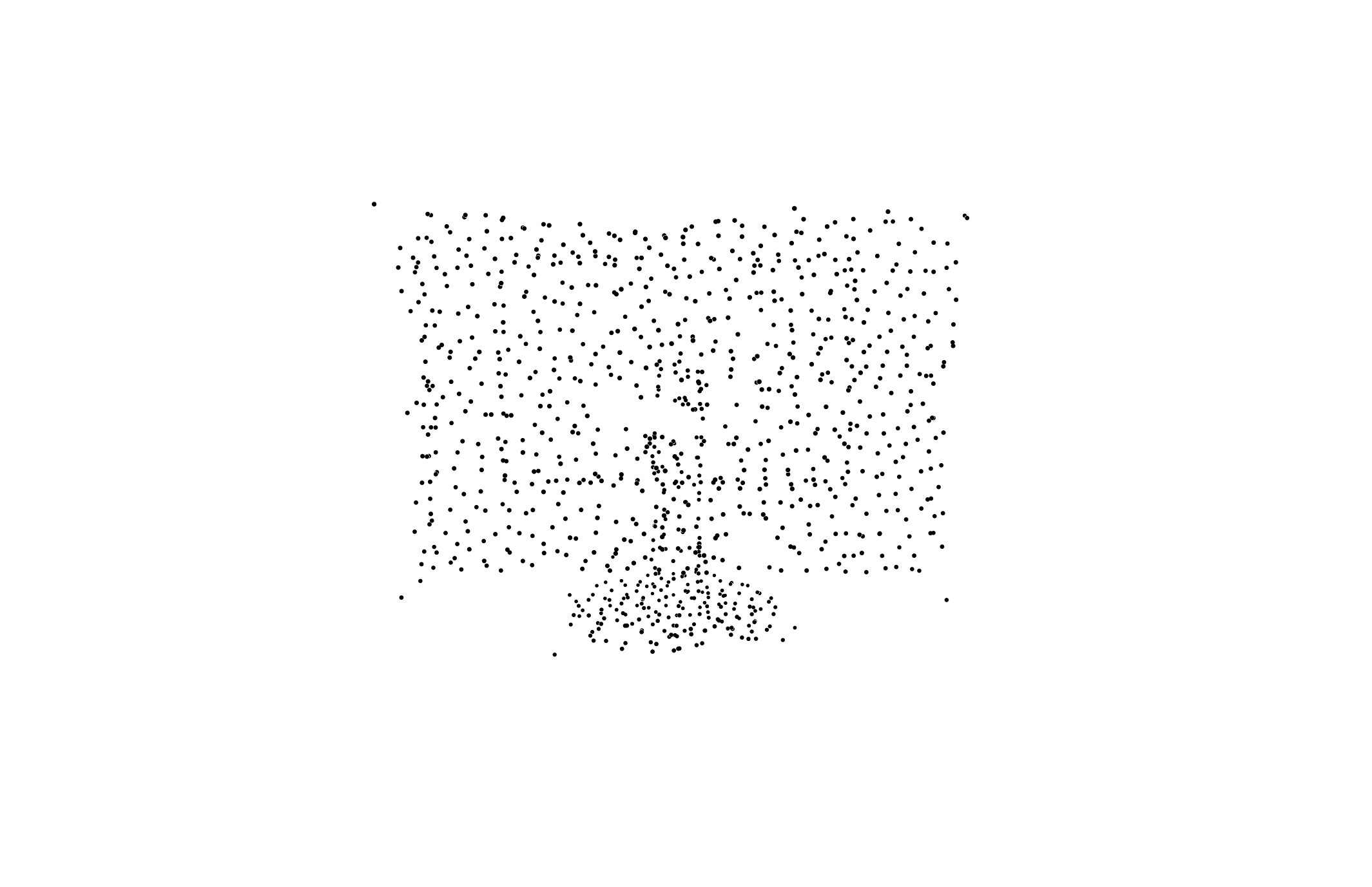} \\[-5mm] 
(a) \blue{\cmark monitor} &\red{\xmark flower-pot} &\red{\xmark mantel} &\red{\xmark mantel} &\red{\xmark mantel} \\



\figt[\sizeS]{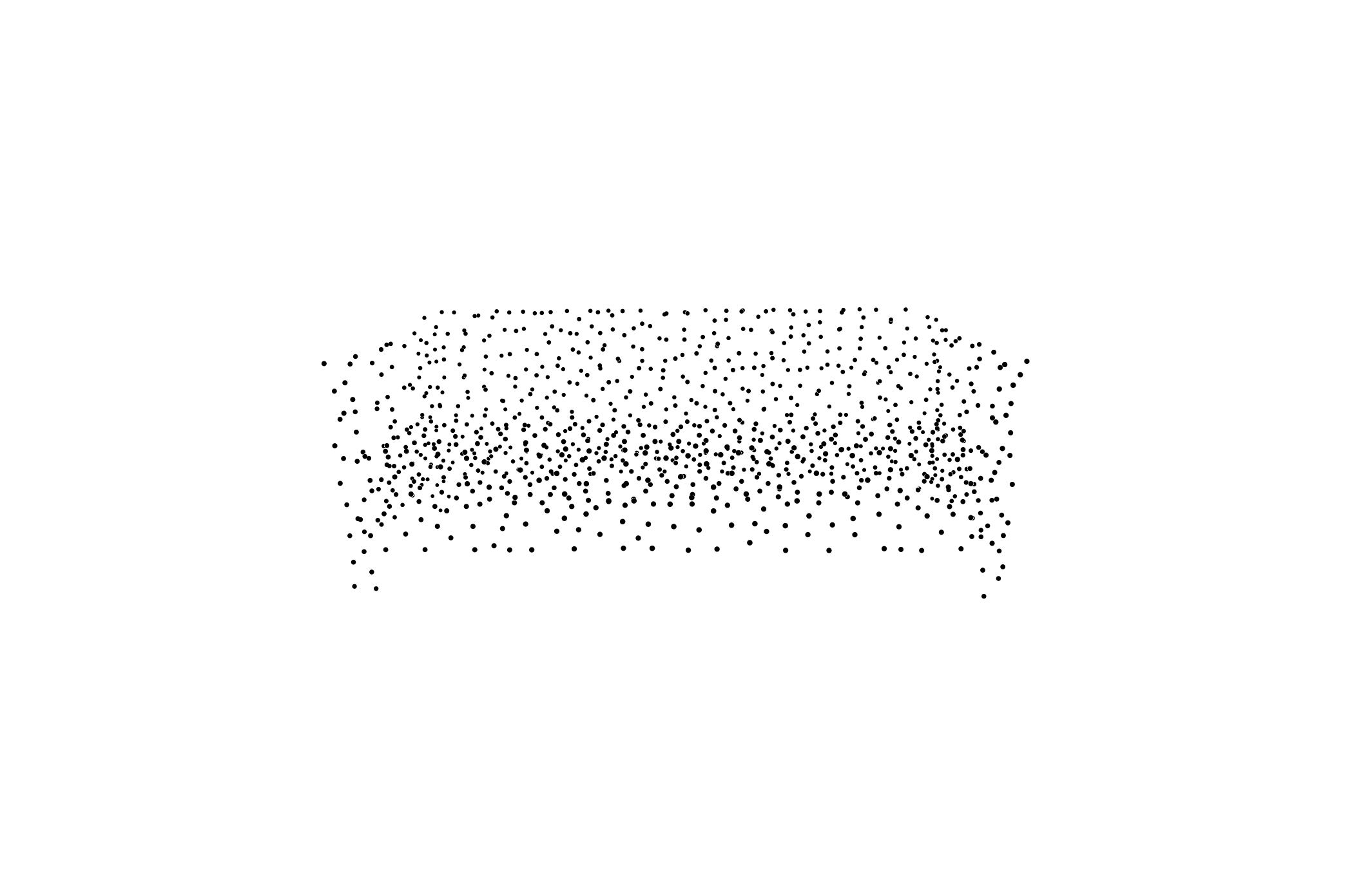} &
\figt[\sizeS]{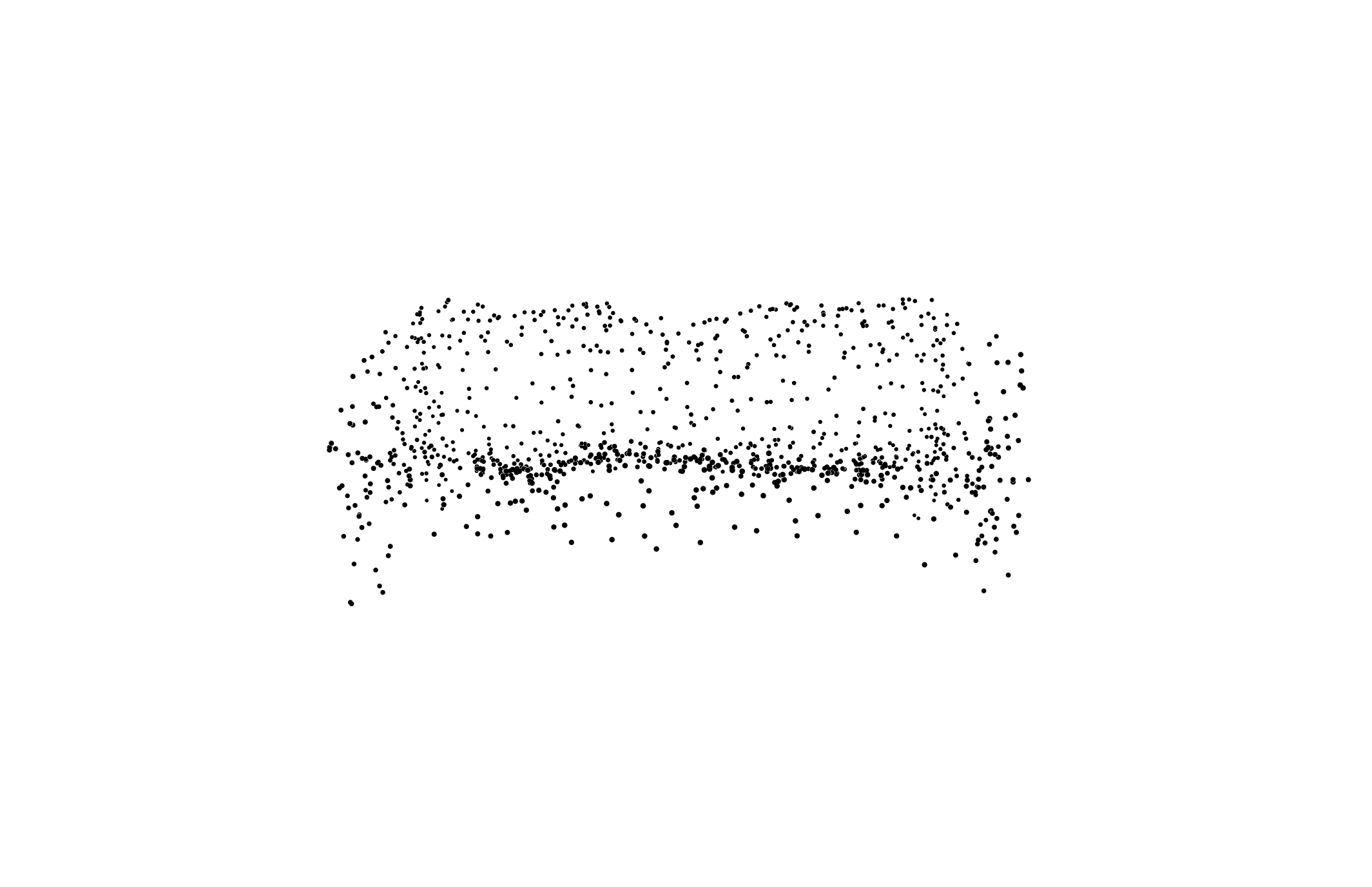} &
\figt[\sizeS]{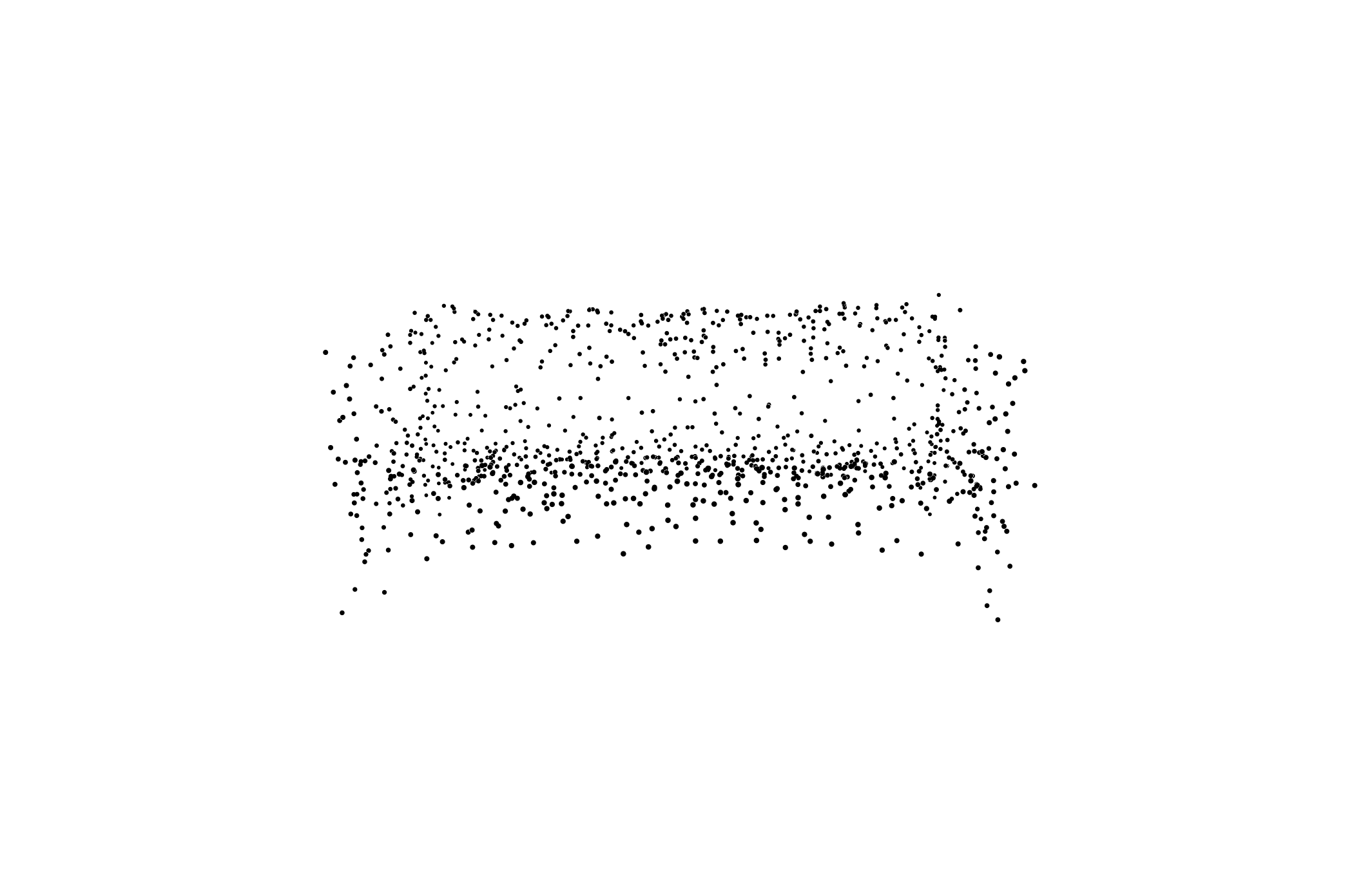} &
\figt[\sizeS]{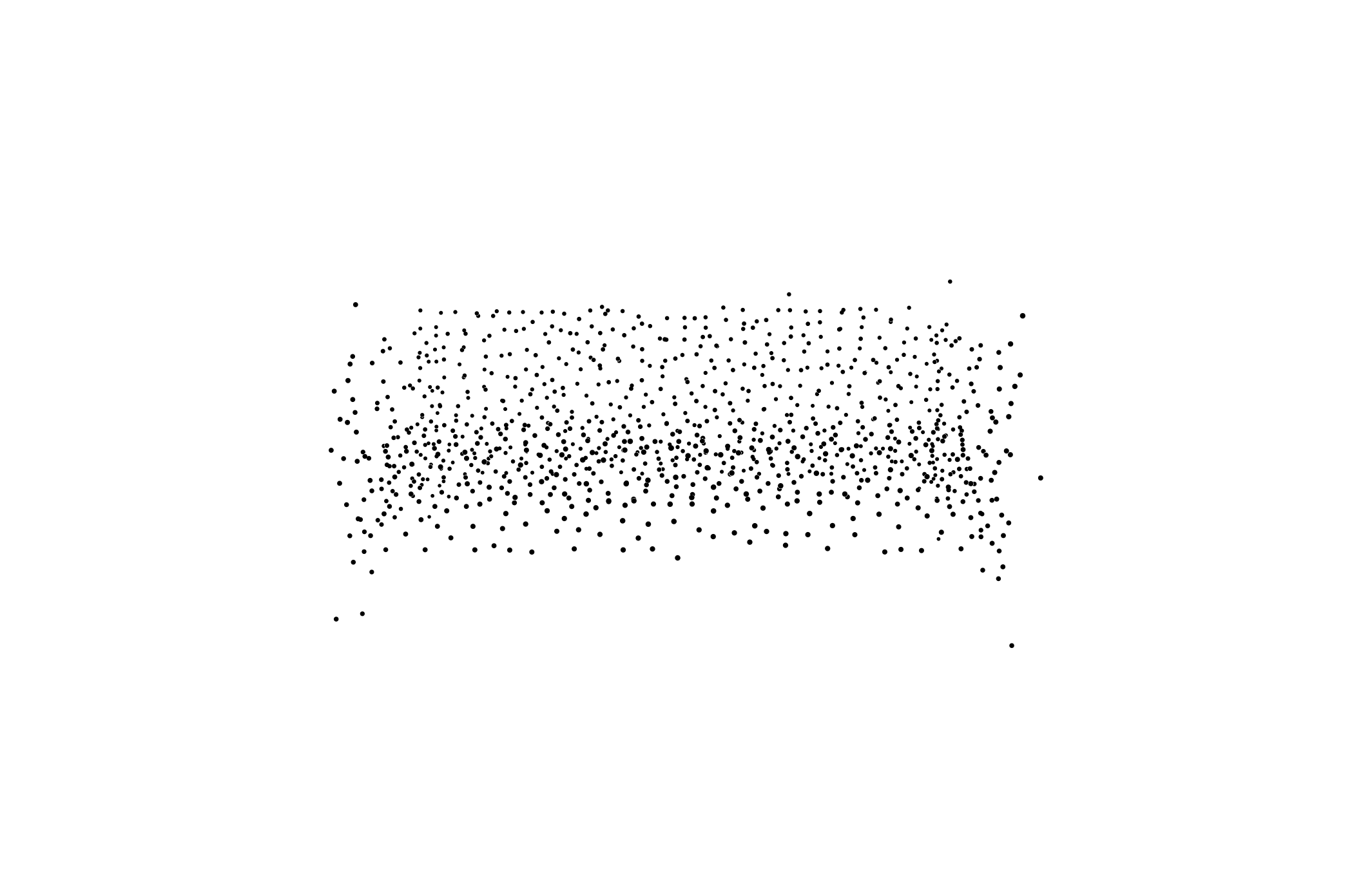} &
\figt[\sizeS]{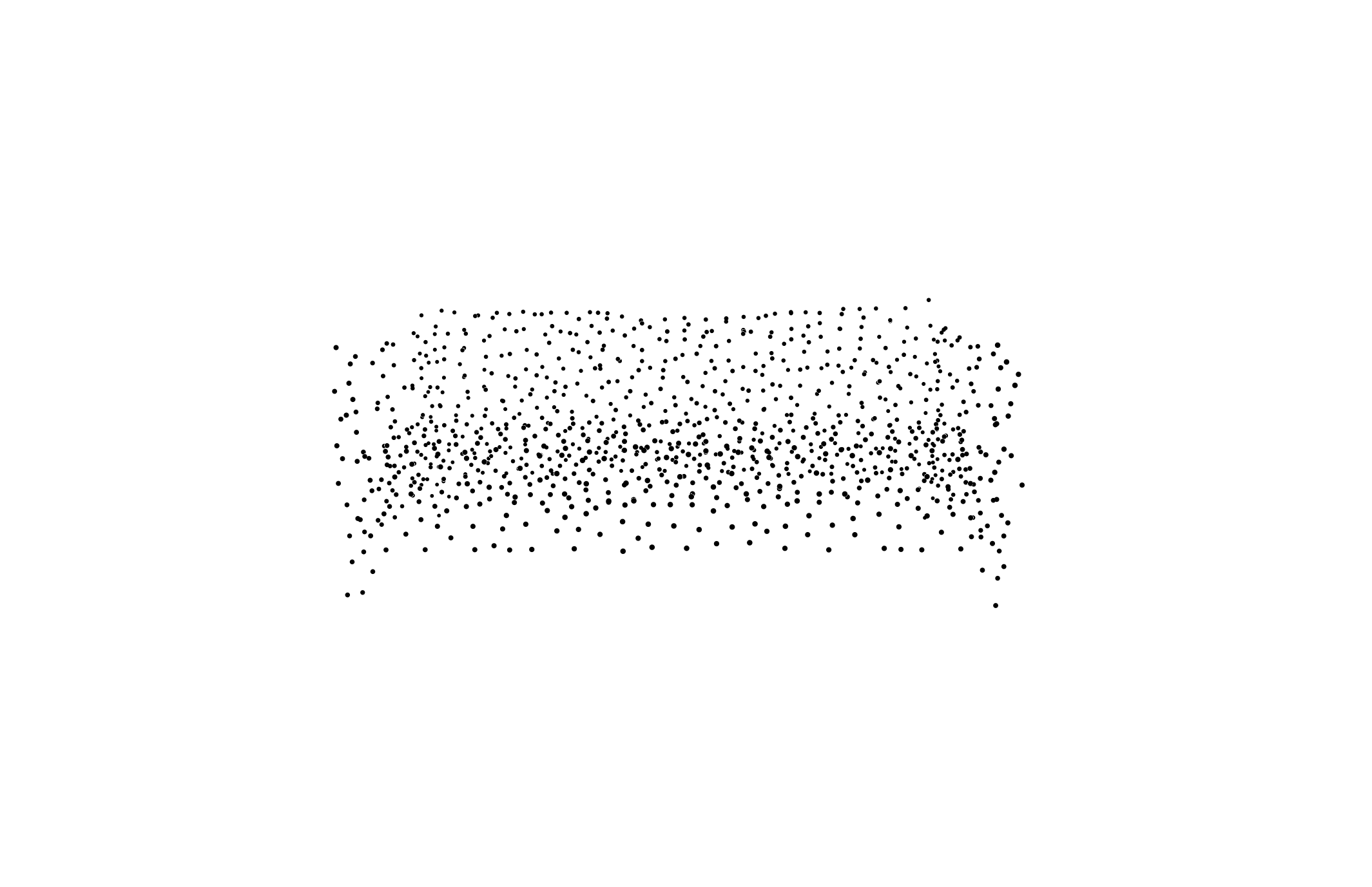} \\[-5mm] 
(b) \blue{\cmark sofa} &\red{\xmark flower-pot} &\red{\xmark bed} &\red{\xmark bed} &\red{\xmark bed} \\


\end{tabular}
\caption{Visualization of adversarial distortions produced by baseline methods and \our.}
\label{fig:vis-3d-pointnet-baseline}
\end{figure*}

\subsection{Comparison}

We now compare our method directly to the mentioned state-of-the-art adversarial attacks. According to Table~\ref{tab:baseline}, our method outperforms the baseline methods, where \our refers to the version with all the imperceptibility regularizations incorporated.  
In PointNet and Point-Transformer, our method performs best on all the metrics except \smooth. 
In DGCNN, we outperform almost all combinations of baseline method and metric.
\geoa and \gsda use a line search to balance misclassification and imperceptibility in adversarial optimization. However, they struggle to find an optimal parameter within reasonable computation budget for Point-Transformer (500 iterations for \gsda while 100 iterations for \siadv and \our). 

\begin{table}[hptb]

\caption{Success rate \asr (\%), average values of $L_2$ norm ($1e-1$), Chamfer distance ($1e-4$), Hausdorff distance ($1e-2$), consistency of  local curvature ($1e-2$), KNN smoothness ($1e-3$) and time (s) of our attack and the baseline attacks against PointNet with defense.}

\centering
\footnotesize
\setlength{\tabcolsep}{7.8pt}

\begin{tabular}{p{1.5cm}c | ccccc | c}\toprule
\Th{method}& \asr$\uparrow$& \lt$\downarrow$&\cd$\downarrow$&\hd$\downarrow$&\curv$\downarrow$&\smooth$\downarrow$&Time$\downarrow$\\\midrule
\mc{8}{PointNet with \srs}\\\midrule
\siadv & 77.3 & 8.39 & 4.36 & 3.87 & 0.36 & \tb{1.40} &{8.78} \\
\rowcolor{LightCyan}
\our & \tb{82.0} & \tb{4.68} & \tb{1.65} & \tb{0.84} & \tb{0.18} & 1.55 & 156.71\\\midrule
\mc{8}{PointNet with \sor}\\\midrule
\siadv & \tb{100} & 17.06 & 7.08 & 3.71 & 0.65 & 1.88 & 0.19 \\
\rowcolor{LightCyan}
\our & \tb{100} & \tb{9.63} & \tb{1.80} & \tb{0.39} & \tb{0.22} & \tb{1.76} & 3.22 \\\midrule
\mc{8}{PointNet with  DUP-Net}\\\midrule
\siadv & \tb{90.8} & 20.85 & 8.85 & 3.88 & 0.78 & 1.96 & 14.37 \\
\rowcolor{LightCyan}
\our & 76.6 & \tb{12.9} & \tb{2.78} & \tb{0.58} & \tb{0.29} & \tb{1.79} & 98.77 \\
\bottomrule
\end{tabular}

\label{tab:defense}
\end{table}

\paragraph{Visualization}
We visualize the adversarial point clouds generated by different attacks in Figure~\ref{fig:vis-3d-pointnet-baseline}. The adversarial distortions of \geoa and \gsda are perceivable due to unbalanced point distributions. \siadv seems closer to our attacks, but tends to introduce more outlier points.

\begin{figure}[htb]
\centering
\footnotesize
\setlength{\tabcolsep}{0pt}
\begin{tabular}{c c}
\extfig{HD}{
\begin{tikzpicture}
\begin{axis}[
	height=3cm,
	width=5.8cm,
	xlabel={\hd},
	ylabel={\asr},
        yticklabel style={font=\tiny},
        legend pos=outer north east]
    \addplot[black] table{figures/eval/defense/hd/bp3_con_dupnet.txt};\leg{our dupnet}
    \addplot[blue] table{figures/eval/defense/hd/si_adv_dupnet.txt};\leg{si dupnet}
    \addplot +[mark=none, red, dashed] coordinates {(0.05, -0.01) (0.05, 0.81)};
    \legend{};
\end{axis}
\end{tikzpicture}
}
& 
\extfig{Curv}{
\begin{tikzpicture}
\begin{axis}[
	height=3cm,
	width=5.8cm,
	xlabel={\curv},
	ylabel={\asr},
        yticklabel style={font=\tiny},
        legend style={font=\tiny},
        legend pos=outer north east]
    \addplot[black] table{figures/eval/defense/curve/bp3_con_dupnet.txt};\leg{\our}
    \addplot[blue] table{figures/eval/defense/curve/si_adv_dupnet.txt};\leg{\siadv}
    \addplot +[mark=none, red, dashed] coordinates {(0.011, -0.01) (0.011, 0.81)};
\end{axis}
\end{tikzpicture}
}
\\
    \end{tabular}
    \caption{Operating characteristics of \asr \vs \hd and \curv on PointNet with  DUP-Net \wrt Table~\ref{tab:defense}.}
    \label{fig:defense-dupnet}
\end{figure}

\paragraph{Defenses}
We now study attack success rates under several state-of-the-art defense methods: Statistical Outlier Removal (SOR)~\citep{rusu2008towards} with $k=2, \alpha=1.1$; Simple Random Sampling (SRS)~\citep{yang2019adversarial} with drop number $500$; Denoiser and UPsampler Network (DUP-Net) ~\citep{zhou2019dup} with $k=2, \alpha=1.1$, number of points as $1024$ and up-sampling rate as $4$.
The results are shown in Table~\ref{tab:defense}, where we only compare with \siadv as that is known to outperform    \geoa and \gsda in defense \citep{si_adv_pc,gsda}. \our clearly generates less perceptible adversarial distortions. 
{To attack against randomness (SRS), we enhance the attacks by forwarding $100$ times and averaging the gradients, which is commonly performed on attacks on images~\citep{athalye2018obfuscated}. We see our method outperforming \siadv on SRS and SOR. With a limited distortion budget, \eg left side of the red line in Figure~\ref{fig:defense-dupnet}, our method achieves a higher success rate. Noteworthy, our method fails on hard examples, which require large distortions that are  not alligned with the imperceptibility constraints.}

\begin{table}[hptb]

\caption{Success rate \asr (\%), average values of \lt ($1e-1$), \cd ($1e-4$), \hd ($1e-2$), \curv ($1e-2$), \smooth ($1e-3$) and time (s) of our attack compared with baseline attacks under black-box setting against PACov with DGCNN as a surrogate model.}

\centering
\footnotesize
\setlength{\tabcolsep}{7.8pt}

\begin{tabular}{p{1.5cm}c | ccccc | c}\toprule
\Th{method}& \asr$\uparrow$& \lt$\downarrow$&\cd$\downarrow$&\hd$\downarrow$&\curv$\downarrow$&\smooth$\downarrow$&Time$\downarrow$\\\midrule
\siadv & 100.0 & 37.30 & 7.63 & 8.76 & 0.46 & 1.45 & 96.34 \\
\simba & 100.0 & 39.26 & 11.11 & 10.40 & 0.60 & 1.64 & 100.07 \\
\simbapp & 100.0 & 40.89 & 12.40 & 21.86 & 0.57 & 1.68 & {87.11} \\
\rowcolor{LightCyan}
\our & 100.0 & \tb{35.17} & \tb{7.35} & \tb{8.56} & \tb{0.45} & \tb{1.44} & 108.77 \\
\bottomrule
\end{tabular}

\label{tab:black}
\end{table}

{\paragraph{Black-Box Attacks} Our framework aims to efficiently solve adversarial optimization with various imperceptibility regularizations. 
The method we compare with always uses the white-box scenario and only \siadv has a black-box version.
Like \siadv, we adapt our method to the black-box scenario and compare it with \siadv and two famous query-based black-box attack algorithms, \ie, \simba~\citep{guo2019simple} and \simbapp~\citep{yang2020learning} in Table~\ref{tab:black}, where we perform better. In our experiments, we choose step size $0.32$ because according to \siadv, it is the best step size for \siadv. Our method outperforms \siadv in its best parameter. }

\section{Conclusion}

This paper has presented \projabbrv, a method substantially improving the efficiency and imperceptibility of attacks on 3D point cloud classification tasks. It disentangles  misclassification and imperceptibility, and supports generating adversarial point clouds across a range of imperceptibility metrics. 

The empirical results show that \our clearly outperforms the state-of-the-art baseline methods that aim for imperceptibility. Interestingly, \our with the $D_{L_2}$ metric often is efficient in improving across all the imperceptibility metrics. Further, our experiments on the defense model implicitely show that adversarial examples with small magnitudes are easy to defend against by using randomization. 

~

\noindent \textbf{Acknowledgements}~This work has received financial support from CAS Project for Young Scientists in Basic Research (YSBR-040) and ISCAS New Cultivation Project (ISCAS-PYFX-202201). This work also has received financial support from VolkswagenStiftung as part of Grant AZ 98514 -- \href{https://explainable-intelligent.systems}{EIS} and by DFG under grant No.~389792660 as part of TRR~248 -- \href{https://perspicuous-computing.science}{CPEC}. \protect\includegraphics[height=8pt]{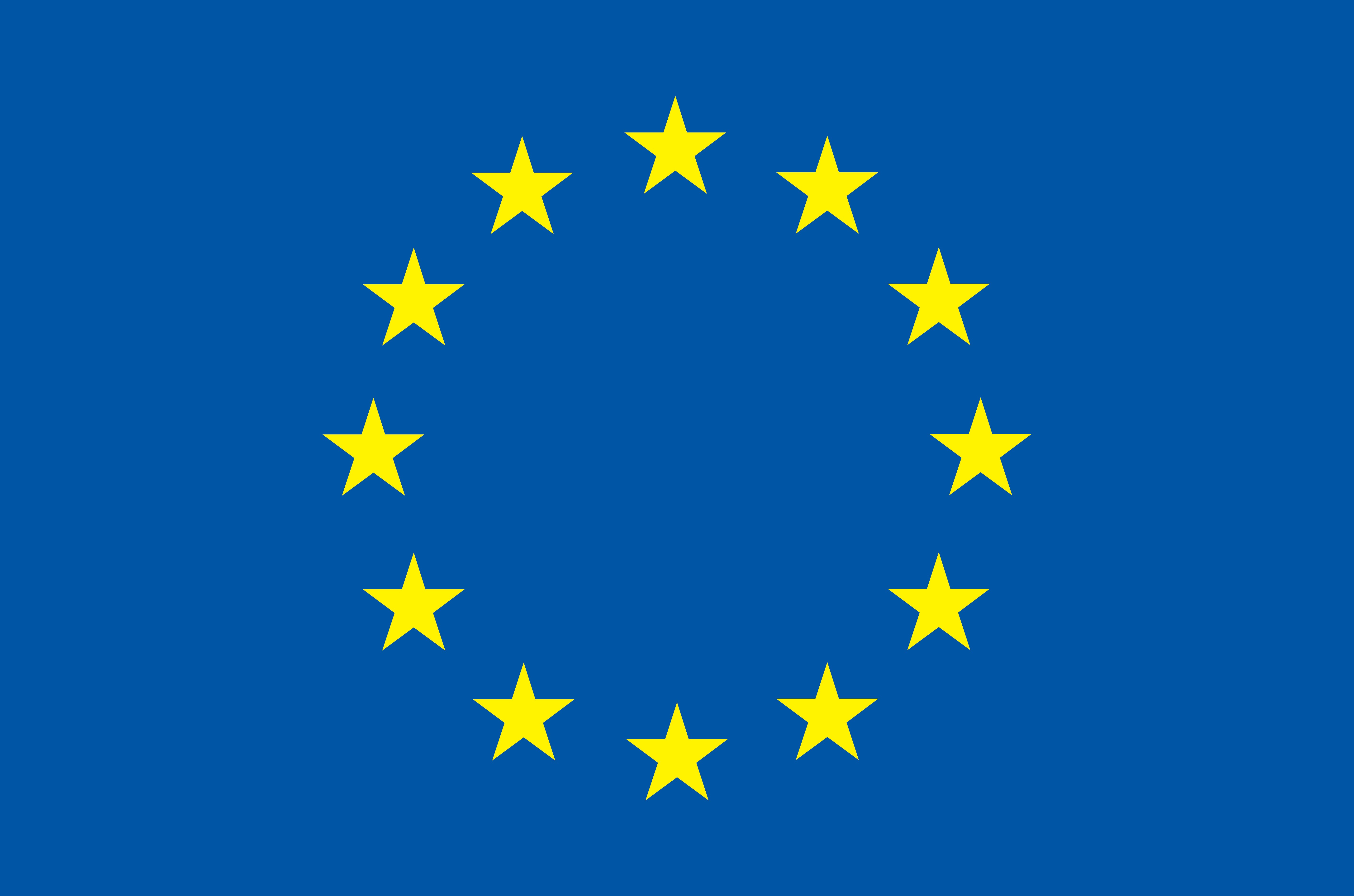} This work is part of the European Union’s Horizon 2020 research and innovation programme under the Marie Sk\l{}odowska-Curie grant no.\@ 101008233.

%
%
%
\bibliographystyle{splncs04}
\bibliography{egbib}
\clearpage





\appendix

\renewcommand{\theequation}{A\arabic{equation}}
\renewcommand{\thetable}{A\arabic{table}}
\renewcommand{\thefigure}{A\arabic{figure}}


\section*{Supplementary material}
We provide results of ablation on step size and iterations. Besides, we provide more experimental results, \eg experiments on the extra networks, providing more operating characteristics, discussion of convergence, and visualization. In the end, we provide more details on our black-box attack.

\begin{table}[hptb]
\caption{\our success probability \asr (\%), average values of \lt ($1e-1$), \cd ($1e-4$), \hd ($1e-2$), \curv ($1e-2$), \smooth ($1e-3$) and time (s) of our attack constrained by different imperceptibility regularization without coordinate transformation against PointNet. Step size: $\epsilon_{i+1} \defn \epsilon_i (1 - \gamma)$; Maximum iteration: $100$.}
\centering
\footnotesize
\setlength{\tabcolsep}{2pt}
\begin{tabular}{p{1.5cm} c|ccccc|c}\toprule
\Th{method}& \asr$\uparrow$& \lt$\downarrow$&\cd$\downarrow$&\hd$\downarrow$&\curv$\downarrow$&\smooth$\downarrow$&T$\downarrow$\\\midrule
$D_{L_2}$ & \tb{100} & \tb{2.00} & 1.14 & 1.51 & \tb{0.13} & 1.42 & 1.73 \\
$D_{CD}$ & \tb{100} & 6.50 & 2.93 & 4.54 & 0.16 & \tb{1.17} & 2.05 \\
$D_{HD}$ & \tb{100} & 5.67 & 2.21 & \tb{0.82} & 0.27 & 1.47 & 1.92 \\
$D_{Curv}$ & \tb{100} & 6.51 & 2.93 & 4.54 & 0.16 & \tb{1.17} & 2.00 \\
\bottomrule
\end{tabular}

\label{tab:abl-stepsize-sloss-supp}
\end{table}

\begin{figure}[htb]
\centering
\setlength{\tabcolsep}{2pt}
\begin{tabular}{c c}
\extfig{l2}{
\begin{tikzpicture}
\begin{axis}[
	height=3cm,
	width=4.2cm,
        yticklabel style={font=\tiny},
        legend style={font=\tiny},
        scaled x ticks = base 10:2,
	xlabel={$\epsilon$},
	ylabel={\lt},
        legend pos=outer north east]
    \addplot[blue] table{figures/eval/f1/l2/bp_cd.txt};\leg{$D_{CD}$}
    \addplot[black] table{figures/eval/f1/l2/bp_hd.txt};\leg{$D_{HD}$}
    \addplot[red] table{figures/eval/f1/l2/bp_l2.txt};\leg{$D_{L_2}$}
    \addplot[green] table{figures/eval/f1/l2/bp_curv.txt};\leg{$D_{Curv}$}
    \addplot[orange]table{figures/eval/combine_reg_fix/l2.txt};\leg{$D_{All}$}
    \legend{};
\end{axis}
\end{tikzpicture}
}
& 
\extfig{CD}{
\begin{tikzpicture}
\begin{axis}[
	height=3cm,
	width=4.2cm,
        yticklabel style={font=\tiny},
        legend style={font=\tiny},
        scaled x ticks = base 10:2,
	xlabel={$\epsilon$},
	ylabel={\cd},
        legend pos=outer north east]
    \addplot[blue] table{figures/eval/f1/cd/bp_cd.txt};\leg{$D_{CD}$}
    \addplot[black] table{figures/eval/f1/cd/bp_cd.txt};\leg{$D_{HD}$}
    \addplot[red] table{figures/eval/f1/cd/bp_l2.txt};\leg{$D_{L_2}$}
    \addplot[green] table{figures/eval/f1/cd/bp_curv.txt};\leg{$D_{Curv}$}
    \addplot[orange]table{figures/eval/combine_reg_fix/cd.txt};\leg{$D_{All}$}
    \legend{};
\end{axis}
\end{tikzpicture}
}
\\
\extfig{HD}{
\begin{tikzpicture}
\begin{axis}[
	height=3cm,
	width=4.2cm,
        yticklabel style={font=\tiny},
        legend style={font=\tiny},
        scaled x ticks = base 10:2,
	xlabel={$\epsilon$},
	ylabel={\hd},
        legend pos=outer north east]
    \addplot[blue] table{figures/eval/f1/hd/bp_cd.txt};\leg{$D_{CD}$}
    \addplot[black] table{figures/eval/f1/hd/bp_hd.txt};\leg{$D_{HD}$}
    \addplot[red] table{figures/eval/f1/hd/bp_l2.txt};\leg{$D_{L_2}$}
    \addplot[green] table{figures/eval/f1/hd/bp_curv.txt};\leg{$D_{Curv}$}
    \addplot[orange]table{figures/eval/combine_reg_fix/hd.txt};\leg{$D_{All}$}
    \legend{};
\end{axis}
\end{tikzpicture}
}
&
\extfig{Curv}{
\begin{tikzpicture}
\begin{axis}[
	height=3cm,
	width=4.2cm,
        yticklabel style={font=\tiny},
        legend style={font=\tiny},
        scaled x ticks = base 10:2,
	xlabel={$\epsilon$},
	ylabel={\curv},
        legend pos=outer north east]
    \addplot[blue] table{figures/eval/f1/curv/bp_cd.txt};\leg{$D_{CD}$}
    \addplot[black] table{figures/eval/f1/curv/bp_hd.txt};\leg{$D_{HD}$}
    \addplot[red] table{figures/eval/f1/curv/bp_l2.txt};\leg{$D_{L_2}$}
    \addplot[green] table{figures/eval/f1/curv/bp_curv.txt};\leg{$D_{Curv}$}
    \addplot[orange]table{figures/eval/combine_reg_fix/curv.txt};\leg{$D_{All}$}
    \legend{};
\end{axis}
\end{tikzpicture}
}
\\
\mc{2}{
\extfig{smooth}{
\begin{tikzpicture}
\begin{axis}[
	height=3cm,
	width=4.2cm,
        yticklabel style={font=\tiny},
        legend style={font=\tiny},
        scaled x ticks = base 10:2,
	xlabel={$\epsilon$},
	ylabel={\smooth},
        legend pos=outer north east]
    \addplot[blue] table{figures/eval/f1/smooth/bp_cd.txt};\leg{$D_{CD}$}
    \addplot[black] table{figures/eval/f1/smooth/bp_hd.txt};\leg{$D_{HD}$}
    \addplot[red] table{figures/eval/f1/smooth/bp_l2.txt};\leg{$D_{L_2}$}
    \addplot[green] table{figures/eval/f1/smooth/bp_curv.txt};\leg{$D_{Curv}$}
    \addplot[orange]table{figures/eval/combine_reg_fix/smooth.txt};\leg{$D_{All}$}
\end{axis}
\end{tikzpicture}
}}
\\
    \end{tabular}
    \caption{Various indicators under different step sizes with imperceptibility regularization of \our. Maximum iteration: $100$; Step size: $\epsilon \in [0.001, 0.01, 0.02, \cdots, 0.06, 0.1]$.}
    \label{fig:abl-step-size}
\end{figure}

\section{Step size}

For step size, we have two schemes: fixed step size and adaptive step size, \ie $\epsilon_{i+1} \defn \epsilon_i (1 - \gamma)$. 

We first carry out experiments on fixed step size. We test the step size $\epsilon \in [0.001, 0.01, 0.02, \cdots, 0.06, 0.1]$ since the \siadv chooses the step size around this range. According to Figure~\ref{fig:abl-step-size}, we find that our method is not very sensitive to step size. It is also hard to choose the best step size. For instance, with $D_{HD}$, small step size, \eg $0.001$ gives better \curv and \smooth but worse \lt and \hd.

We test adaptive step size with single imperceptibility regularization as Table~\ref{tab:abl-stepsize-sloss-supp}.
Compared with fixed step $0.06$ in {Table 2}, adaptive step size help to optimize $D_{Curv}$ but it decreases the performance for other imperceptibility metrics. However, with several imperceptibility regularization terms, adaptive step size performs better than fixed step size. Thus in the main paper, we show the fixed step size with single imperceptibility regularization while the adaptive step size with combined imperceptibility regularization.

\begin{figure}
\centering
\begin{tabular}{c c}
\extfig{iter}{
\begin{tikzpicture}
\begin{axis}[
	height=3cm,
	width=4.2cm,
        yticklabel style={font=\tiny},
        legend style={font=\tiny},
	xlabel={iteration},
	ylabel={\lt},
        legend pos=outer north east]
    \addplot[blue]table{figures/eval/combine_iteration/l2.txt};
\end{axis}
\end{tikzpicture}} &
\extfig{iter}{
\begin{tikzpicture}
\begin{axis}[
	height=3cm,
	width=4.2cm,
        yticklabel style={font=\tiny},
        legend style={font=\tiny},
	xlabel={iteration},
	ylabel={\cd},
        legend pos=outer north east]
    \addplot[black]table{figures/eval/combine_iteration/cd.txt};
\end{axis}
\end{tikzpicture}}\\
\extfig{iter}{
\begin{tikzpicture}
\begin{axis}[
	height=3cm,
	width=4.2cm,
        yticklabel style={font=\tiny},
        legend style={font=\tiny},
	xlabel={iteration},
	ylabel={\hd},
        legend pos=outer north east]
    \addplot[green]table{figures/eval/combine_iteration/hd.txt};
\end{axis}
\end{tikzpicture}}&
\extfig{iter}{
\begin{tikzpicture}
\begin{axis}[
	height=3cm,
	width=4.2cm,
        yticklabel style={font=\tiny},
        legend style={font=\tiny},
	xlabel={iteration},
	ylabel={\curv},
        legend pos=outer north east]
    \addplot[red]table{figures/eval/combine_iteration/curv.txt};
\end{axis}
\end{tikzpicture}}\\
\mc{2}{
\extfig{iter}{
\begin{tikzpicture}
\begin{axis}[
	height=3cm,
	width=4.2cm,
        yticklabel style={font=\tiny},
        legend style={font=\tiny},
	xlabel={iteration},
	ylabel={\smooth},
        legend pos=outer north east]
    \addplot[orange]table{figures/eval/combine_iteration/smooth.txt};
\end{axis}
\end{tikzpicture}}}

\end{tabular}
\caption{Various indicators under different iterations with all imperceptibility regularization and adaptive step size.}
\label{fig:iteration}
\end{figure}

\section{Iterations}

We test the influence on the maximum iteration on \our with all imperceptibility regularization and adaptive step size. As Figure~\ref{fig:iteration}, more iteration helps in optimizing \lt, \cd, \hd, and \curv but makes \smooth worse. Plus, our method converges within the $100$ iteration. Thus in the main paper, we always choose $100$ as the maximum iteration.

\section{More results}

\paragraph{More network} Since most baseline methods verify their performance on PointNet++, we also carry out the experiments on the pre-trained model of PointNet++\footnote{\url{https://github.com/Gorilla-Lab-SCUT/GeoA3}} with single scale grouping. We find that with the given parameters, none of the attacks achieve $100 \%$ \asr. A large distortion is needed to deceive the PointNet++ while our attack considers large distortions against our imperceptibility constraints. Our method reaches the same \asr with lower distortion.

\begin{table}[hptb]
\caption{Success probability \asr (\%), average values of \lt ($1e-1$), \cd ($1e-4$), \hd ($1e-2$), \curv ($1e-2$), \smooth ($1e-3$) and time (s) of our attack compared with baseline attacks against PointNet++ SSG.}
\centering
\footnotesize
\setlength{\tabcolsep}{2pt}
\begin{tabular}{p{1.5cm} c|ccccc|c}\toprule
\Th{method} & \asr $\uparrow$ & \lt $\downarrow$ & \cd $\downarrow$ & \hd $\downarrow$ & \curv $\downarrow$ & \smooth $\downarrow$ &T $\downarrow$ \\\midrule
\geoa & \tb{99} &         76.73 &         5.67 &         0.41 &         0.28 &         1.94 &  198.91\\
\siadv & 96 &         13.23 &         6.32 &         2.36 &         0.60 &         \tb{1.61} &  \tb{0.19}\\
\gsda & 98 &         72.77 &         4.91 &         0.62 &         0.33 &         1.86 & 239.40\\
\rowcolor{LightCyan}
\our  & 96 &         \tb{7.64} &         \tb{2.04} &         \tb{0.19} &         \tb{0.20} &         1.73 &  4.74\\
\bottomrule
\end{tabular}

\label{tab:baseline-pointnet++}
\end{table}

\begin{figure}
\centering
\begin{tabular}{cc}

\extfig{l2}{
\begin{tikzpicture}
\begin{axis}[
	height=3cm,
	width=4.2cm,
        yticklabel style={font=\tiny},
	xlabel={\lt},
	ylabel={\asr},
        legend pos=outer north east]

    \addplot[blue] table{figures/eval/baseline/PointNet2SSG/l2/geoa3.txt};\leg{\geoa}
    \addplot[black] table{figures/eval/baseline/PointNet2SSG/l2/gsda.txt};\leg{\gsda}
    \addplot[green] table{figures/eval/baseline/PointNet2SSG/l2/si_adv.txt};\leg{\siadv}
    \addplot[red] table{figures/eval/baseline/PointNet2SSG/l2/ours.txt};\leg{\our}

    \legend{};
\end{axis}
\end{tikzpicture}
}
& 
\extfig{CD}{
\begin{tikzpicture}
\begin{axis}[
	height=3cm,
	width=4.2cm,
        yticklabel style={font=\tiny},
        restrict x to domain=0:0.01,
	xlabel={\cd},
	ylabel={\asr},
        legend pos=outer north east]

    \addplot[blue] table{figures/eval/baseline/PointNet2SSG/cd/geoa3.txt};\leg{\geoa}
    \addplot[black] table{figures/eval/baseline/PointNet2SSG/cd/gsda.txt};\leg{\gsda}
    \addplot[green] table{figures/eval/baseline/PointNet2SSG/cd/si_adv.txt};\leg{\siadv}
    \addplot[red] table{figures/eval/baseline/PointNet2SSG/cd/ours.txt};\leg{\our}

    \legend{};
\end{axis}
\end{tikzpicture}
}
\\
\extfig{HD}{
\begin{tikzpicture}
\begin{axis}[
	height=3cm,
	width=4.2cm,
        yticklabel style={font=\tiny},
        restrict x to domain=0:0.1,
        scaled x ticks = base 10:2,
	xlabel={\hd},
	ylabel={\asr},
        legend pos=outer north east]

    \addplot[blue] table{figures/eval/baseline/PointNet2SSG/hd/geoa3.txt};\leg{\geoa}
    \addplot[black] table{figures/eval/baseline/PointNet2SSG/hd/gsda.txt};\leg{\gsda}
    \addplot[green] table{figures/eval/baseline/PointNet2SSG/hd/si_adv.txt};\leg{\siadv}
    \addplot[red] table{figures/eval/baseline/PointNet2SSG/hd/ours.txt};\leg{\our}

    \legend{};
\end{axis}
\end{tikzpicture}
}
& 
\extfig{Curv}{
\begin{tikzpicture}
\begin{axis}[
	height=3cm,
	width=4.2cm,
        yticklabel style={font=\tiny},
        restrict x to domain=0:0.01,
        scaled x ticks = base 10:3,
	xlabel={\curv},
	ylabel={\asr},
        legend pos=outer north east]

    \addplot[blue] table{figures/eval/baseline/PointNet2SSG/curv/geoa3.txt};\leg{\geoa}
    \addplot[black] table{figures/eval/baseline/PointNet2SSG/curv/gsda.txt};\leg{\gsda}
    \addplot[green] table{figures/eval/baseline/PointNet2SSG/curv/si_adv.txt};\leg{\siadv}
    \addplot[red] table{figures/eval/baseline/PointNet2SSG/curv/ours.txt};\leg{\our}

    \legend{};
\end{axis}
\end{tikzpicture}
}
\\
\mc{2}{
\extfig{smooth}{
\begin{tikzpicture}
\begin{axis}[
	height=3cm,
	width=4.2cm,
        yticklabel style={font=\tiny},
        legend style={font=\tiny},
	xlabel={\smooth},
	ylabel={\asr},
        legend pos=outer north east]

    \addplot[blue] table{figures/eval/baseline/PointNet2SSG/smooth/geoa3.txt};\leg{\geoa}
    \addplot[black] table{figures/eval/baseline/PointNet2SSG/smooth/gsda.txt};\leg{\gsda}
    \addplot[green] table{figures/eval/baseline/PointNet2SSG/smooth/si_adv.txt};\leg{\siadv}
    \addplot[red] table{figures/eval/baseline/PointNet2SSG/smooth/ours.txt};\leg{\our}

\end{axis}
\end{tikzpicture}
}}
\\
    \end{tabular}
    \caption{Operating characteristics on PointNet++ for our attack and baseline attacks, \wrt Table~\ref{tab:baseline-pointnet++}. 
    }
    \label{fig:baseline-comparison}
\end{figure}

Figure~\ref{fig:baseline-comparison} shows the operating characteristics on PointNet++ \wrt Table~\ref{tab:baseline-pointnet++}. We clearly see that the red line (\our) is on the left side of all other curves except on \smooth. It indicates that with a given distortion budget, \our is better than the others.

\begin{figure}
\centering
\begin{tabular}{cc}
\extfig{l2}{
\begin{tikzpicture}
\begin{axis}[
	height=3cm,
	width=4.2cm,
        yticklabel style={font=\tiny},
        legend style={font=\tiny},
	xlabel={\lt},
	ylabel={\asr},
        legend pos=outer north east]
    \addplot[blue] table{figures/eval/combine_reg/l2/bp3_hd_curv_exp.txt};\leg{$D_{HD+Curv}$}
    \addplot[black] table{figures/eval/combine_reg/l2/bp3_l2_hd_exp.txt};\leg{$D_{L_2+HD}$}
    \addplot[red] table{figures/eval/combine_reg/l2/bp3_l2_curv_exp.txt};\leg{$D_{L_2+Curv}$}
    \addplot[green] table{figures/eval/combine_reg/l2/bp3_l2_hd_curv_exp.txt};\leg{$D_{L_2+HD+Curv}$}
    \addplot[orange] table{figures/eval/baseline/PointNet/l2/ours.txt};\leg{$D_{All}$}
    \legend{};
\end{axis}
\end{tikzpicture}
}
& 
\extfig{CD}{
\begin{tikzpicture}
\begin{axis}[
	height=3cm,
	width=4.2cm,
        yticklabel style={font=\tiny},
        legend style={font=\tiny},
	xlabel={\cd},
	ylabel={\asr},
        legend pos=outer north east]
    \addplot[blue] table{figures/eval/combine_reg/cd/bp3_hd_curv_exp.txt};\leg{$D_{HD+Curv}$}
    \addplot[black] table{figures/eval/combine_reg/cd/bp3_l2_hd_exp.txt};\leg{$D_{L_2+HD}$}
    \addplot[red] table{figures/eval/combine_reg/cd/bp3_l2_curv_exp.txt};\leg{$D_{L_2+Curv}$}
    \addplot[green] table{figures/eval/combine_reg/cd/bp3_l2_hd_curv_exp.txt};\leg{$D_{L_2+HD+Curv}$}
    \addplot[orange] table{figures/eval/baseline/PointNet/cd/ours.txt};\leg{$D_{All}$}
    \legend{};
\end{axis}
\end{tikzpicture}
}
\\
\extfig{HD}{
\begin{tikzpicture}
\begin{axis}[
	height=3cm,
	width=4.2cm,
        yticklabel style={font=\tiny},
        legend style={font=\tiny},
        scaled x ticks = base 10:2,
	xlabel={\hd},
	ylabel={\asr},
        legend pos=outer north east]
    \addplot[blue] table{figures/eval/combine_reg/hd/bp3_hd_curv_exp.txt};\leg{$D_{HD+Curv}$}
    \addplot[black] table{figures/eval/combine_reg/hd/bp3_l2_hd_exp.txt};\leg{$D_{L_2+HD}$}
    \addplot[red] table{figures/eval/combine_reg/hd/bp3_l2_curv_exp.txt};\leg{$D_{L_2+Curv}$}
    \addplot[green] table{figures/eval/combine_reg/hd/bp3_l2_hd_curv_exp.txt};\leg{$D_{L_2+HD+Curv}$}
    \addplot[orange] table{figures/eval/baseline/PointNet/hd/ours.txt};\leg{$D_{All}$}
    \legend{};
\end{axis}
\end{tikzpicture}
}
& 
\extfig{Curv}{
\begin{tikzpicture}
\begin{axis}[
	height=3cm,
	width=4.2cm,
        yticklabel style={font=\tiny},
        legend style={font=\tiny},
	xlabel={\curv},
	ylabel={\asr},
        legend pos=outer north east]
    \addplot[blue] table{figures/eval/combine_reg/curve/bp3_hd_curv_exp.txt};\leg{$D_{HD+Curv}$}
    \addplot[black] table{figures/eval/combine_reg/curve/bp3_l2_hd_exp.txt};\leg{$D_{L_2+HD}$}
    \addplot[red] table{figures/eval/combine_reg/curve/bp3_l2_curv_exp.txt};\leg{$D_{L_2+Curv}$}
    \addplot[green] table{figures/eval/combine_reg/curve/bp3_l2_hd_curv_exp.txt};\leg{$D_{L_2+HD+Curv}$}
    \addplot[orange] table{figures/eval/baseline/PointNet/curv/ours.txt};\leg{$D_{All}$}
    \legend{};
\end{axis}
\end{tikzpicture}
}
\\
\mc{2}{
\extfig{smooth}{
\begin{tikzpicture}
\begin{axis}[
	height=3cm,
	width=4.2cm,
        yticklabel style={font=\tiny},
        legend style={font=\tiny},
	xlabel={\smooth},
	ylabel={\asr},
        legend style={font=\tiny},
        legend pos=outer north east]
    \addplot[blue] table{figures/eval/combine_reg/smooth/bp3_hd_curv_exp.txt};\leg{$D_{HD+Curv}$}
    \addplot[black] table{figures/eval/combine_reg/smooth/bp3_l2_hd_exp.txt};\leg{$D_{L_2+HD}$}
    \addplot[red] table{figures/eval/combine_reg/smooth/bp3_l2_curv_exp.txt};\leg{$D_{L_2+Curv}$}
    \addplot[green] table{figures/eval/combine_reg/smooth/bp3_l2_hd_curv_exp.txt};\leg{$D_{L_2+HD+Curv}$}
    \addplot[orange] table{figures/eval/baseline/PointNet/smooth/ours.txt};\leg{$D_{All}$}
\end{axis}
\end{tikzpicture}
}}
\\
    \end{tabular}
    \caption{Operating characteristics on PointNet for \our with different imperceptibility regularization, \wrt Table 3.}
    \label{fig:abl-combine-sloss-supp}
\end{figure}

\begin{figure}
\centering
\begin{tabular}{cc}

\extfig{l2}{
\begin{tikzpicture}
\begin{axis}[
	height=3cm,
	width=4.2cm,
        yticklabel style={font=\tiny},
        legend style={font=\tiny},
        restrict x to domain=0:10,
	xlabel={\lt},
	ylabel={\asr},
        legend pos=outer north east]
    \addplot[blue] table{figures/eval/baseline/PointNet/l2/geoa3.txt};\leg{\geoa}
    \addplot[black] table{figures/eval/baseline/PointNet/l2/gsda.txt};\leg{\gsda}
    \addplot[green] table{figures/eval/baseline/PointNet/l2/si_adv.txt};\leg{\siadv}
    \addplot[red] table{figures/eval/baseline/PointNet/l2/ours.txt};\leg{\our}

    \legend{};
\end{axis}
\end{tikzpicture}
}
& 
\extfig{CD}{
\begin{tikzpicture}
\begin{axis}[
	height=3cm,
	width=4.2cm,
        yticklabel style={font=\tiny},
        legend style={font=\tiny},
        restrict x to domain=0:0.01,
	xlabel={\cd},
	ylabel={\asr},
        legend pos=outer north east]
    \addplot[blue] table{figures/eval/baseline/PointNet/cd/geoa3.txt};\leg{\geoa}
    \addplot[black] table{figures/eval/baseline/PointNet/cd/gsda.txt};\leg{\gsda}
    \addplot[green] table{figures/eval/baseline/PointNet/cd/si_adv.txt};\leg{\siadv}
    \addplot[red] table{figures/eval/baseline/PointNet/cd/ours.txt};\leg{\our}
\legend{};

\end{axis}
\end{tikzpicture}
}
\\
\extfig{HD}{
\begin{tikzpicture}
\begin{axis}[
	height=3cm,
	width=4.2cm,
        yticklabel style={font=\tiny},
        legend style={font=\tiny},
        restrict x to domain=0:0.1,
        scaled x ticks = base 10:2,
	xlabel={\hd},
	ylabel={\asr},
        legend pos=outer north east]
    \addplot[blue] table{figures/eval/baseline/PointNet/hd/geoa3.txt};\leg{\geoa}
    \addplot[black] table{figures/eval/baseline/PointNet/hd/gsda.txt};\leg{\gsda}
    \addplot[green] table{figures/eval/baseline/PointNet/hd/si_adv.txt};\leg{\siadv}
    \addplot[red] table{figures/eval/baseline/PointNet/hd/ours.txt};\leg{\our}

    \legend{};
\end{axis}
\end{tikzpicture}
}
& 
\extfig{Curv}{
\begin{tikzpicture}
\begin{axis}[
	height=3cm,
	width=4.2cm,
        yticklabel style={font=\tiny},
        legend style={font=\tiny},
        restrict x to domain=0:0.01,
        scaled x ticks = base 10:3,
	xlabel={\curv},
	ylabel={\asr},
        legend pos=outer north east]
    \addplot[blue] table{figures/eval/baseline/PointNet/curv/geoa3.txt};\leg{\geoa}
    \addplot[black] table{figures/eval/baseline/PointNet/curv/gsda.txt};\leg{\gsda}
    \addplot[green] table{figures/eval/baseline/PointNet/curv/si_adv.txt};\leg{\siadv}
    \addplot[red] table{figures/eval/baseline/PointNet/curv/ours.txt};\leg{\our}

    \legend{};
\end{axis}
\end{tikzpicture}
}
\\
\mc{2}{
\extfig{smooth}{
\begin{tikzpicture}
\begin{axis}[
	height=3cm,
	width=4.2cm,
        yticklabel style={font=\tiny},
        legend style={font=\tiny},
	xlabel={\smooth},
	ylabel={\asr},
        legend pos=outer north east]
    \addplot[blue] table{figures/eval/baseline/PointNet/smooth/geoa3.txt};\leg{\geoa}
    \addplot[black] table{figures/eval/baseline/PointNet/smooth/gsda.txt};\leg{\gsda}
    \addplot[green] table{figures/eval/baseline/PointNet/smooth/si_adv.txt};\leg{\siadv}
    \addplot[red] table{figures/eval/baseline/PointNet/smooth/ours.txt};\leg{\our}

\end{axis}
\end{tikzpicture}
}}
\\
    \end{tabular}
    \caption{Operating characteristics on PointNet for our attack and baseline attacks, \wrt {Table 4}.}
    \label{fig:baseline-comparison-supp-pointnet}
\end{figure}

\begin{figure}
\centering
\begin{tabular}{cc}

\extfig{l2}{
\begin{tikzpicture}
\begin{axis}[
	height=3cm,
	width=4.2cm,
        yticklabel style={font=\tiny},
        legend style={font=\tiny},
        restrict x to domain=0:10,
	xlabel={\lt},
	ylabel={\asr},
        legend pos=outer north east]

    \addplot[blue] table{figures/eval/baseline/DGCNN/l2/geoa3.txt};\leg{\geoa}
    \addplot[black] table{figures/eval/baseline/DGCNN/l2/gsda.txt};\leg{\gsda}
    \addplot[green] table{figures/eval/baseline/DGCNN/l2/si_adv.txt};\leg{\siadv}
    \addplot[red] table{figures/eval/baseline/DGCNN/l2/ours.txt};\leg{\our}
    \legend{};
\end{axis}
\end{tikzpicture}
}
& 
\extfig{CD}{
\begin{tikzpicture}
\begin{axis}[
	height=3cm,
	width=4.2cm,
        yticklabel style={font=\tiny},
        legend style={font=\tiny},
        restrict x to domain=0:0.01,
	xlabel={\cd},
	ylabel={\asr},
        legend pos=outer north east]

    \addplot[blue] table{figures/eval/baseline/DGCNN/cd/geoa3.txt};\leg{\geoa}
    \addplot[black] table{figures/eval/baseline/DGCNN/cd/gsda.txt};\leg{\gsda}
    \addplot[green] table{figures/eval/baseline/DGCNN/cd/si_adv.txt};\leg{\siadv}
    \addplot[red] table{figures/eval/baseline/DGCNN/cd/ours.txt};\leg{\our}
    \legend{};
\end{axis}
\end{tikzpicture}
}
\\
\extfig{HD}{
\begin{tikzpicture}
\begin{axis}[
	height=3cm,
	width=4.2cm,
        yticklabel style={font=\tiny},
        legend style={font=\tiny},
        restrict x to domain=0:0.1,
        scaled x ticks = base 10:2,
	xlabel={\hd},
	ylabel={\asr},
        legend pos=outer north east]

    \addplot[blue] table{figures/eval/baseline/DGCNN/hd/geoa3.txt};\leg{\geoa}
    \addplot[black] table{figures/eval/baseline/DGCNN/hd/gsda.txt};\leg{\gsda}
    \addplot[green] table{figures/eval/baseline/DGCNN/hd/si_adv.txt};\leg{\siadv}
    \addplot[red] table{figures/eval/baseline/DGCNN/hd/ours.txt};\leg{\our}
    \legend{};
\end{axis}
\end{tikzpicture}
}
& 
\extfig{Curv}{
\begin{tikzpicture}
\begin{axis}[
	height=3cm,
	width=4.2cm,
        yticklabel style={font=\tiny},
        legend style={font=\tiny},
        restrict x to domain=0:0.01,
        scaled x ticks = base 10:3,
	xlabel={\curv},
	ylabel={\asr},
        legend pos=outer north east]

    \addplot[blue] table{figures/eval/baseline/DGCNN/curv/geoa3.txt};\leg{\geoa}
    \addplot[black] table{figures/eval/baseline/DGCNN/curv/gsda.txt};\leg{\gsda}
    \addplot[green] table{figures/eval/baseline/DGCNN/curv/si_adv.txt};\leg{\siadv}
    \addplot[red] table{figures/eval/baseline/DGCNN/curv/ours.txt};\leg{\our}
    \legend{};
\end{axis}
\end{tikzpicture}
}
\\
\mc{2}{ 
\extfig{smooth}{
\begin{tikzpicture}
\begin{axis}[
	height=3cm,
	width=4.2cm,
        yticklabel style={font=\tiny},
        legend style={font=\tiny},
	xlabel={\smooth},
	ylabel={\asr},
        legend pos=outer north east]

    \addplot[blue] table{figures/eval/baseline/DGCNN/smooth/geoa3.txt};\leg{\geoa}
    \addplot[black] table{figures/eval/baseline/DGCNN/smooth/gsda.txt};\leg{\gsda}
    \addplot[green] table{figures/eval/baseline/DGCNN/smooth/si_adv.txt};\leg{\siadv}
    \addplot[red] table{figures/eval/baseline/DGCNN/smooth/ours.txt};\leg{\our}
\end{axis}
\end{tikzpicture}
}}
\\
    \end{tabular}
    \caption{Operating characteristics on DGCNN for our attack and baseline attacks, \wrt {Table 4}.}
    \label{fig:baseline-comparison-supp-dgcnn}
\end{figure}

\begin{figure}
\centering
\begin{tabular}{cc}

\extfig{l2t}{
\begin{tikzpicture}
\begin{axis}[
	height=3cm,
	width=4.2cm,
        yticklabel style={font=\tiny},
        legend style={font=\tiny},
        restrict x to domain=0:10,
	xlabel={\lt},
	ylabel={\asr},
        legend pos=outer north east]

    \addplot[blue] table{figures/eval/baseline/PTrans/l2/geoa3.txt};\leg{\geoa}
    \addplot[black] table{figures/eval/baseline/PTrans/l2/gsda.txt};\leg{\gsda}
    \addplot[green] table{figures/eval/baseline/PTrans/l2/si_adv.txt};\leg{\siadv}
    \addplot[red] table{figures/eval/baseline/PTrans/l2/ours.txt};\leg{\our}
    \legend{};
\end{axis}
\end{tikzpicture}
}
& 
\extfig{CD}{
\begin{tikzpicture}
\begin{axis}[
	height=3cm,
	width=4.2cm,
        yticklabel style={font=\tiny},
        legend style={font=\tiny},
        restrict x to domain=0:0.01,
	xlabel={\cd},
	ylabel={\asr},
        legend pos=outer north east]

    \addplot[blue] table{figures/eval/baseline/PTrans/cd/geoa3.txt};\leg{\geoa}
    \addplot[black] table{figures/eval/baseline/PTrans/cd/gsda.txt};\leg{\gsda}
    \addplot[green] table{figures/eval/baseline/PTrans/cd/si_adv.txt};\leg{\siadv}
    \addplot[red] table{figures/eval/baseline/PTrans/cd/ours.txt};\leg{\our}
    \legend{};
\end{axis}
\end{tikzpicture}
}
\\
\extfig{HDt}{
\begin{tikzpicture}
\begin{axis}[
	height=3cm,
	width=4.2cm,
        yticklabel style={font=\tiny},
        legend style={font=\tiny},
        restrict x to domain=0:0.1,
        scaled x ticks = base 10:2,
	xlabel={\hd},
	ylabel={\asr},
        legend pos=outer north east]

    \addplot[blue] table{figures/eval/baseline/PTrans/hd/geoa3.txt};\leg{\geoa}
    \addplot[black] table{figures/eval/baseline/PTrans/hd/gsda.txt};\leg{\gsda}
    \addplot[green] table{figures/eval/baseline/PTrans/hd/si_adv.txt};\leg{\siadv}
    \addplot[red] table{figures/eval/baseline/PTrans/hd/ours.txt};\leg{\our}
    \legend{};
\end{axis}
\end{tikzpicture}
}
& 
\extfig{Curvt}{
\begin{tikzpicture}
\begin{axis}[
	height=3cm,
	width=4.2cm,
        yticklabel style={font=\tiny},
        legend style={font=\tiny},
        restrict x to domain=0:0.01,
        scaled x ticks = base 10:3,
	xlabel={\curv},
	ylabel={\asr},
        legend pos=outer north east]

    \addplot[blue] table{figures/eval/baseline/PTrans/curv/geoa3.txt};\leg{\geoa}
    \addplot[black] table{figures/eval/baseline/PTrans/curv/gsda.txt};\leg{\gsda}
    \addplot[green] table{figures/eval/baseline/PTrans/curv/si_adv.txt};\leg{\siadv}
    \addplot[red] table{figures/eval/baseline/PTrans/curv/ours.txt};\leg{\our}
    \legend{};
\end{axis}
\end{tikzpicture}
}
\\
\mc{2}{
\extfig{smootht}{
\begin{tikzpicture}
\begin{axis}[
	height=3cm,
	width=4.2cm,
        yticklabel style={font=\tiny},
        legend style={font=\tiny},
	xlabel={\smooth},
	ylabel={\asr},
        legend pos=outer north east]

    \addplot[blue] table{figures/eval/baseline/PTrans/smooth/geoa3.txt};\leg{\geoa}
    \addplot[black] table{figures/eval/baseline/PTrans/smooth/gsda.txt};\leg{\gsda}
    \addplot[green] table{figures/eval/baseline/PTrans/smooth/si_adv.txt};\leg{\siadv}
    \addplot[red] table{figures/eval/baseline/PTrans/smooth/ours.txt};\leg{\our}
\end{axis}
\end{tikzpicture}
}}
\\
    \end{tabular}
    \caption{Operating characteristics on Point-Transformer for our attack and baseline attacks, \wrt Table 4.}
    \label{fig:baseline-comparison-supp-transformer}
\end{figure}

\begin{figure}
\centering
\begin{tabular}{cc}
\extfig{l2}{
\begin{tikzpicture}
\begin{axis}[
	height=3cm,
	width=4.2cm,
        yticklabel style={font=\tiny},
        legend style={font=\tiny},
	xlabel={\lt},
	ylabel={\asr},
        legend pos=outer north east]
    \addplot[red] table{figures/eval/defense/l2/bp3_con_sor.txt};\leg{our sor}
    \addplot[blue] table{figures/eval/defense/l2/bp3_con_srs.txt};\leg{our srs}
    \addplot[black] table{figures/eval/defense/l2/bp3_con_dupnet.txt};\leg{our dupnet}

    \addplot[red, dashed] table{figures/eval/defense/l2/si_adv_sor.txt};\leg{si sor}
    \addplot[blue, dashed] table{figures/eval/defense/l2/si_adv_srs.txt};\leg{si srs}
    \addplot[black, dashed] table{figures/eval/defense/l2/si_adv_dupnet.txt};\leg{si dupnet}

    \legend{};
\end{axis}
\end{tikzpicture}
}
& 
\extfig{CD}{
\begin{tikzpicture}
\begin{axis}[
	height=3cm,
	width=4.2cm,
        yticklabel style={font=\tiny},
        legend style={font=\tiny},
	xlabel={\cd},
	ylabel={\asr},
        legend pos=outer north east]
    \addplot[red] table{figures/eval/defense/cd/bp3_con_sor.txt};\leg{our srs}
    \addplot[blue] table{figures/eval/defense/cd/bp3_con_srs.txt};\leg{our srs}
    \addplot[black] table{figures/eval/defense/cd/bp3_con_dupnet.txt};\leg{our dupnet}

    \addplot[red, dashed] table{figures/eval/defense/cd/si_adv_sor.txt};\leg{si srs}
    \addplot[blue, dashed] table{figures/eval/defense/cd/si_adv_srs.txt};\leg{si srs}
    \addplot[black, dashed] table{figures/eval/defense/cd/si_adv_dupnet.txt};\leg{si dupnet}
    \legend{};
\end{axis}
\end{tikzpicture}
}
\\
\extfig{HD}{
\begin{tikzpicture}
\begin{axis}[
	height=3cm,
	width=4.2cm,
        yticklabel style={font=\tiny},
        legend style={font=\tiny},
	xlabel={\hd},
	ylabel={\asr},
        legend pos=outer north east]
    \addplot[red] table{figures/eval/defense/hd/bp3_con_sor.txt};\leg{our srs}
    \addplot[blue] table{figures/eval/defense/hd/bp3_con_srs.txt};\leg{our srs}
    \addplot[black] table{figures/eval/defense/hd/bp3_con_dupnet.txt};\leg{our dupnet}

    \addplot[red, dashed] table{figures/eval/defense/hd/si_adv_sor.txt};\leg{si srs}
    \addplot[blue, dashed] table{figures/eval/defense/hd/si_adv_srs.txt};\leg{si srs}
    \addplot[black, dashed] table{figures/eval/defense/hd/si_adv_dupnet.txt};\leg{si dupnet}
    \legend{};
\end{axis}
\end{tikzpicture}
}
& 
\extfig{Curv}{
\begin{tikzpicture}
\begin{axis}[
	height=3cm,
	width=4.2cm,
        yticklabel style={font=\tiny},
        legend style={font=\tiny},
	xlabel={\curv},
	ylabel={\asr},
        legend style={font=\tiny},
        legend pos=outer north east]
    \addplot[red] table{figures/eval/defense/curve/bp3_con_sor.txt};\leg{our srs}
    \addplot[blue] table{figures/eval/defense/curve/bp3_con_srs.txt};\leg{our srs}
    \addplot[black] table{figures/eval/defense/curve/bp3_con_dupnet.txt};\leg{our dupnet}

    \addplot[red, dashed] table{figures/eval/defense/curve/si_adv_sor.txt};\leg{si srs}
    \addplot[blue, dashed] table{figures/eval/defense/curve/si_adv_srs.txt};\leg{si srs}
    \addplot[black, dashed] table{figures/eval/defense/curve/si_adv_dupnet.txt};\leg{si dupnet}
    \legend{};
\end{axis}
\end{tikzpicture}
}
\\
\mc{2}{
\extfig{smooth}{
\begin{tikzpicture}
\begin{axis}[
	height=3cm,
	width=4.2cm,
        yticklabel style={font=\tiny},
        legend style={font=\tiny},
	xlabel={\smooth},
	ylabel={\asr},
        legend pos=outer north east]
    \addplot[red] table{figures/eval/defense/smooth/bp3_con_sor.txt};\leg{SOR}
    \addplot[blue] table{figures/eval/defense/smooth/bp3_con_srs.txt};\leg{SRS}
    \addplot[black] table{figures/eval/defense/smooth/bp3_con_dupnet.txt};\leg{DupNet}

    \addplot[red, dashed] table{figures/eval/defense/smooth/si_adv_sor.txt};
    \addplot[blue, dashed] table{figures/eval/defense/smooth/si_adv_srs.txt};
    \addplot[black, dashed] table{figures/eval/defense/smooth/si_adv_dupnet.txt};
\end{axis}
\end{tikzpicture}
}}
\\
    \end{tabular}
    \caption{Operating characteristics on defense, \wrt Table 5. Solid line: \our; Dashed line: \siadv; Step size: $0.06$; Maximum iteration: $100$.}
    \label{fig:abl-defense-supp}
\end{figure}

\paragraph{More operating characteristics} Operating characteristics provide more information than the average value. Thus, here we provide all corresponding operating characteristics.
Figure~\ref{fig:abl-combine-sloss-supp} operates characteristics for {Table 3}. We see it more intuitively that with $D_{L_2 + Curv}$, we get the worst result on \hd, and with $D_{HD +Curv}$, we get the worst results on \lt and \cd. In \curv, all the combinations have similar results, with $D_{L_2 +HD}$ and $D_{HD + Curv}$ getting worse average number due to a few extremely bad examples. It implies the direction to reduce each imperceptibility regularization is not correlated with each other. Using the combination of all imperceptibility regularizations for attack is a trade-off solution according to the performance of each imperceptibility metric.

Here, we show the corresponding figures of PointNet, DGCNN, and Point-Transformer in Figure~\ref{fig:baseline-comparison-supp-pointnet},  Figure~\ref{fig:baseline-comparison-supp-dgcnn}, and Figure~\ref{fig:baseline-comparison-supp-transformer}. 
Compared to Figure~\ref{fig:baseline-comparison} and Figure~\ref{fig:baseline-comparison-supp-dgcnn}, \siadv works better on PointNet, especially on \cd and \curv. \our always outperforms on \lt and \cd. \siadv always performs best on \smooth but worst on \hd.

We also provide our attack and the baseline
attacks against PointNet on $1000$ samples from ModelNet40, Set
the step size to $0.0075$, forwarding $100$ times, and averaging the
gradients, set max step to $100$ for SRS in {Table 5}.  According to Figure~\ref{fig:abl-defense-supp}, we clearly observe that with a reasonable given distortion budget, our attack outperforms \siadv. DupNet is the best defense model among the three defenses.

\begin{figure}[htp]
\centering
\setlength{\tabcolsep}{2pt}
\begin{tabular}{cc}
\extfig{grad_cd}{
\begin{tikzpicture}
\begin{axis}[
	height=3.5cm,
	width=4.2cm,
	xlabel={Iterations},
	ylabel={$cos(\hat{g},D_{CD})$},
        legend pos=outer north east]
    \addplot[blue] table{figures/eval/cos_data/ours/grad_cd_cos.txt};\leg{\our}
    \addplot[black] table{figures/eval/cos_data/siadv/grad_cd_cos.txt};\leg{\our}
    \legend{};
\end{axis}
\end{tikzpicture}
}
&
\extfig{grad_cd}{
\begin{tikzpicture}
\begin{axis}[
	height=3.5cm,
	width=4.2cm,
	xlabel={Iterations},
	ylabel={$cos(\hat{g},D_{Curv})$},
        legend pos=outer north east]
    \addplot[blue] table{figures/eval/cos_data/ours/grad_curv_cos.txt};\leg{\our}
    \addplot[black] table{figures/eval/cos_data/siadv/grad_curv_cos.txt};\leg{\our}
    \legend{};
\end{axis}
\end{tikzpicture}
}
\\
\extfig{grad_hd}{
\begin{tikzpicture}
\begin{axis}[
	height=3.5cm,
	width=4.2cm,
	xlabel={Iterations},
	ylabel={$cos(\hat{g},D_{HD})$},
        legend pos=outer north east]
    \addplot[blue] table{figures/eval/cos_data/ours/grad_hd_cos.txt};\leg{\our}
    \addplot[black] table{figures/eval/cos_data/siadv/grad_hd_cos.txt};\leg{\our}
    \legend{};
\end{axis}
\end{tikzpicture}
}
&
\extfig{grad_cd}{
\begin{tikzpicture}
\begin{axis}[
	height=3.5cm,
	width=4.2cm,
	xlabel={Iterations},
	ylabel={$cos(\hat{g},D_{L2})$},
        legend pos=outer north east]
    \addplot[blue] table{figures/eval/cos_data/ours/grad_l2_cos.txt};\leg{\our}
    \addplot[black] table{figures/eval/cos_data/siadv/grad_l2_cos.txt};\leg{\our}
    \legend{};
\end{axis}
\end{tikzpicture}
}
\\
\extfig{grad_cd}{
\begin{tikzpicture}
\begin{axis}[
	height=3.5cm,
	width=4.2cm,
	xlabel={Iterations},
	ylabel={$cos(D_{CD},D_{Curv})$},
        legend pos=outer north east]
    \addplot[blue] table{figures/eval/cos_data/ours/cd_curv_cos.txt};\leg{\our}
    \addplot[black] table{figures/eval/cos_data/siadv/cd_curv_cos.txt};\leg{\our}
    \legend{};
\end{axis}
\end{tikzpicture}
}
&
\extfig{grad_cd}{
\begin{tikzpicture}
\begin{axis}[
	height=3.5cm,
	width=4.2cm,
	xlabel={Iterations},
	ylabel={$cos(D_{CD},D_{HD})$},
        legend pos=outer north east]
    \addplot[blue] table{figures/eval/cos_data/ours/hd_cd_cos.txt};\leg{\our}
    \addplot[black] table{figures/eval/cos_data/siadv/hd_cd_cos.txt};\leg{\our}
    \legend{};
\end{axis}
\end{tikzpicture}
}
\\
\extfig{grad_cd}{
\begin{tikzpicture}
\begin{axis}[
	height=3.5cm,
	width=4.2cm,
	xlabel={Iterations},
	ylabel={$cos(D_{Curv},D_{HD})$},
        legend pos=outer north east]
    \addplot[blue] table{figures/eval/cos_data/ours/hd_curv_cos.txt};\leg{\our}
    \addplot[black] table{figures/eval/cos_data/siadv/hd_curv_cos.txt};\leg{\our}
    \legend{};
\end{axis}
\end{tikzpicture}
}
&
\extfig{grad_cd}{
\begin{tikzpicture}
\begin{axis}[
	height=3.5cm,
	width=4.2cm,
	xlabel={Iterations},
	ylabel={$cos(D_{Curv},D_{L2})$},
        legend pos=outer north east]
    \addplot[blue] table{figures/eval/cos_data/ours/l2_curv_cos.txt};\leg{\our}
    \addplot[black] table{figures/eval/cos_data/siadv/l2_curv_cos.txt};\leg{\our}
    \legend{};
\end{axis}
\end{tikzpicture}
}
\\
\extfig{grad_cd}{
\begin{tikzpicture}
\begin{axis}[
	height=3.5cm,
	width=4.2cm,
        xlabel={Iterations},
	ylabel={$cos(D_{CD},D_{L2})$},
        legend pos=outer north east]
    \addplot[blue] table{figures/eval/cos_data/ours/l2_cd_cos.txt};\leg{\our}
    \addplot[black] table{figures/eval/cos_data/siadv/l2_cd_cos.txt};\leg{\our}
    \legend{};
\end{axis}
\end{tikzpicture}
}
&
\extfig{grad_cd}{
\begin{tikzpicture}
\begin{axis}[
	height=3.5cm,
	width=4.2cm,
	xlabel={Iterations},
	ylabel={$cos(D_{HD},D_{L2})$},
        legend pos=outer north east]
    \addplot[blue] table{figures/eval/cos_data/ours/l2_hd_cos.txt};\leg{\our}
    \addplot[black] table{figures/eval/cos_data/siadv/l2_hd_cos.txt};\leg{\our}
    \legend{};
\end{axis}
\end{tikzpicture}
}
\\
\end{tabular}
\caption{Cosine distance among gradients of misclassification and imperceptibility regularizations when iteration increasing. Test in $1000$ images on PointNet. Blue lines: \our; Black lines: \siadv.}
\label{fig:cos_dis}
\end{figure}
\begin{figure}
\centering
\begin{tabular}{cc}
\extfig{l2}{
\begin{tikzpicture}
\begin{axis}[
	height=3cm,
	width=4.2cm,
        yticklabel style={font=\tiny},
        legend style={font=\tiny},
	xlabel={\lt},
	ylabel={\asr},
        legend pos=outer north east]
    \addplot[red] table{figures/eval/black_box/l2/ours.txt};\leg{\our}
    \addplot[blue] table{figures/eval/black_box/l2/simba.txt};\leg{\simba}
    \addplot[black] table{figures/eval/black_box/l2/simbapp.txt};\leg{\simbapp}
    \addplot[green] table{figures/eval/black_box/l2/si-adv.txt};\leg{\siadv}
    \legend{};
\end{axis}
\end{tikzpicture}
}
& 
\extfig{CD}{
\begin{tikzpicture}
\begin{axis}[
	height=3cm,
	width=4.2cm,
        yticklabel style={font=\tiny},
        legend style={font=\tiny},
	xlabel={\cd},
	ylabel={\asr},
        legend pos=outer north east]
    \addplot[red] table{figures/eval/black_box/cd/ours.txt};\leg{\our}
    \addplot[blue] table{figures/eval/black_box/cd/simba.txt};\leg{\simba}
    \addplot[black] table{figures/eval/black_box/cd/simbapp.txt};\leg{\simbapp}
    \addplot[green] table{figures/eval/black_box/cd/si-adv.txt};\leg{\siadv}
    \legend{};
\end{axis}
\end{tikzpicture}
}
\\
\extfig{HD}{
\begin{tikzpicture}
\begin{axis}[
	height=3cm,
	width=4.2cm,
        yticklabel style={font=\tiny},
        legend style={font=\tiny},
        scaled x ticks = base 10:1,
	xlabel={\hd},
	ylabel={\asr},
        legend pos=outer north east]
    \addplot[red] table{figures/eval/black_box/hd/ours.txt};\leg{\our}
    \addplot[blue] table{figures/eval/black_box/hd/simba.txt};\leg{\simba}
    \addplot[black] table{figures/eval/black_box/hd/simbapp.txt};\leg{\simbapp}
    \addplot[green] table{figures/eval/black_box/hd/si-adv.txt};\leg{\siadv}
    \legend{};
\end{axis}
\end{tikzpicture}
}
& 
\extfig{Curv}{
\begin{tikzpicture}
\begin{axis}[
	height=3cm,
	width=4.2cm,
        yticklabel style={font=\tiny},
        legend style={font=\tiny},
	xlabel={\curv},
	ylabel={\asr},
        legend style={font=\tiny},
        legend pos=outer north east]
    \addplot[red] table{figures/eval/black_box/curv/ours.txt};\leg{\our}
    \addplot[blue] table{figures/eval/black_box/curv/simba.txt};\leg{\simba}
    \addplot[black] table{figures/eval/black_box/curv/simbapp.txt};\leg{\simbapp}
    \addplot[green] table{figures/eval/black_box/curv/si-adv.txt};\leg{\siadv}
    \legend{};
\end{axis}
\end{tikzpicture}
}
\\
\mc{2}{
\extfig{smooth}{
\begin{tikzpicture}
\begin{axis}[
	height=3cm,
	width=4.2cm,
        yticklabel style={font=\tiny},
        legend style={font=\tiny},
	xlabel={\smooth},
	ylabel={\asr},
        legend pos=outer north east]
    \addplot[red] table{figures/eval/black_box/smooth/ours.txt};\leg{\our}
    \addplot[blue] table{figures/eval/black_box/smooth/simba.txt};\leg{\simba}
    \addplot[black] table{figures/eval/black_box/smooth/simbapp.txt};\leg{\simbapp}
    \addplot[green] table{figures/eval/black_box/smooth/si-adv.txt};\leg{\siadv}
\end{axis}
\end{tikzpicture}
}}
\\
    \end{tabular}
    \caption{Operating characteristics on defense, \wrt Table 6.}
    \label{fig:abl-black-supp}
\end{figure}

\paragraph{Discussion of convergence}
To demonstrate that our attack optimizes each imperceptibility regularization and converges, we plot the cosine distance among the gradient of classification and the descending direction of each regularization. As Figure~\ref{fig:cos_dis}, the cosine distance roughly converges to a stable value in the end. Checking the curves for \siadv, we observe that the cosine distance between the gradient of classification and the descending direction of each regularization converges quickly, \ie the first few iterations. Except for the cosine distance between the gradient and the descending direction of \lt is one, the rest cosine distance values are all negative. It means for each iteration of \siadv, when moving along the direction of the gradient, the $D_{L_2}$ decreases while the rest increases. However, for our attack, we optimize the misclassification and imperceptibility regularization with GS process, so we manage to force the descending direction between \cd and \curv, \cd and \hd, \cd and \lt to zeros (orthogonal). Our attack also leads to the cosine distance between the gradient and the rest imperceptibility regularization being negative, which implies our attack is searching on the boundary of the classifier.

\begin{algorithm}[bt]
\caption{\projabbrv in Black-box setting}
\label{alg:bp-black}
\textbf{Input:} $\cX$: original point cloud to be attacked \\
\textbf{Input:} $t$: true label; $\epsilon_1$ : step size for stage 1; $\epsilon_2$ : step size for stage 2 \\
\textbf{Input:} $\cD$ : a set of imperceptibility regularization\\
\textbf{Input:} $f_s$: surrogate model; $f_b$: target model\\
\textbf{Input:} $\cD$: a set of imperceptibility regularization\\
\textbf{Output:} $\cY^*$
\begin{algorithmic}[1]
\STATE $\cY^* \gets \cX \cdot \mathbf{0}$, $\cY' \gets T_{rsi}(\cX)$
\STATE $\vg \gets \nabla_\cY' {\ell(f_s(T'_{rsi}(\cY')), t)}$
\STATE $\hat{\vg} \gets \func{n}(\vg)$
\STATE $\cS \gets SM(\vg)$ 
\STATE $\cS \gets  Sort(\cS)$
\WHILE{$f_b(T_{rsi}(\cY')) == t$ and $\cS \neq \varnothing$}
    \STATE $\mathbf{s} \gets \cS(0)$
    \STATE $\cS \gets \cS/\mathbf{s}$
    \STATE $\theta \gets \arctan(g_{i2}/g_{i1})$
    \STATE $\mathbf{s} \gets \mathbf{s} \cdot (\cos \theta, \sin \theta, 0)$
    \FOR{$\eta \in \{-\epsilon_1, \epsilon_1\}$}
        \STATE $\cY' \gets \cY' - \eta \vq$
        \IF{$f_b(T'_{rsi}(\cY')) \neq t$}
            \FOR{$D_j \in \cD$}
                \STATE $\hat{\vd_j} \gets \func{n}(\nabla_{\cY'} D_j(\cX,T'_{rsi}(\cY')))$
            \ENDFOR
	\STATE $\{\hat{\vv}_1,\cdots,\hat{\vv}_m\} \gets  GS(\{\hat{\vg},\hat{\vd}_1,\cdots,\hat{\vd}_m\})$ 
            \FOR{$D_j \in \cD$}
            \STATE $\cY' \gets \cY' - \epsilon_2\hat{\vv}_{j}$ 
                \IF{$\sum_j D_{j}(\cX,T'_{rsi}(\cY'))<\sum_j D_{j}(\cX,\cY^*)$}
                \STATE $\cY^* =T'_{rsi}(\cY')$
                \ENDIF
            \ENDFOR
        \ENDIF
    \ENDFOR
\ENDWHILE
    
\end{algorithmic}
\label{alg:black-box}
\end{algorithm}

\paragraph{More visualization}

\begin{figure*}[hptb]
\newcommand{\sizeS}{.20}
\centering
\setlength{\tabcolsep}{0pt}
\begin{tabular}{ccccc}
\textbf{Clean} &\geoa &\gsda &\siadv & \projabbrv\\
\figt[\sizeS]{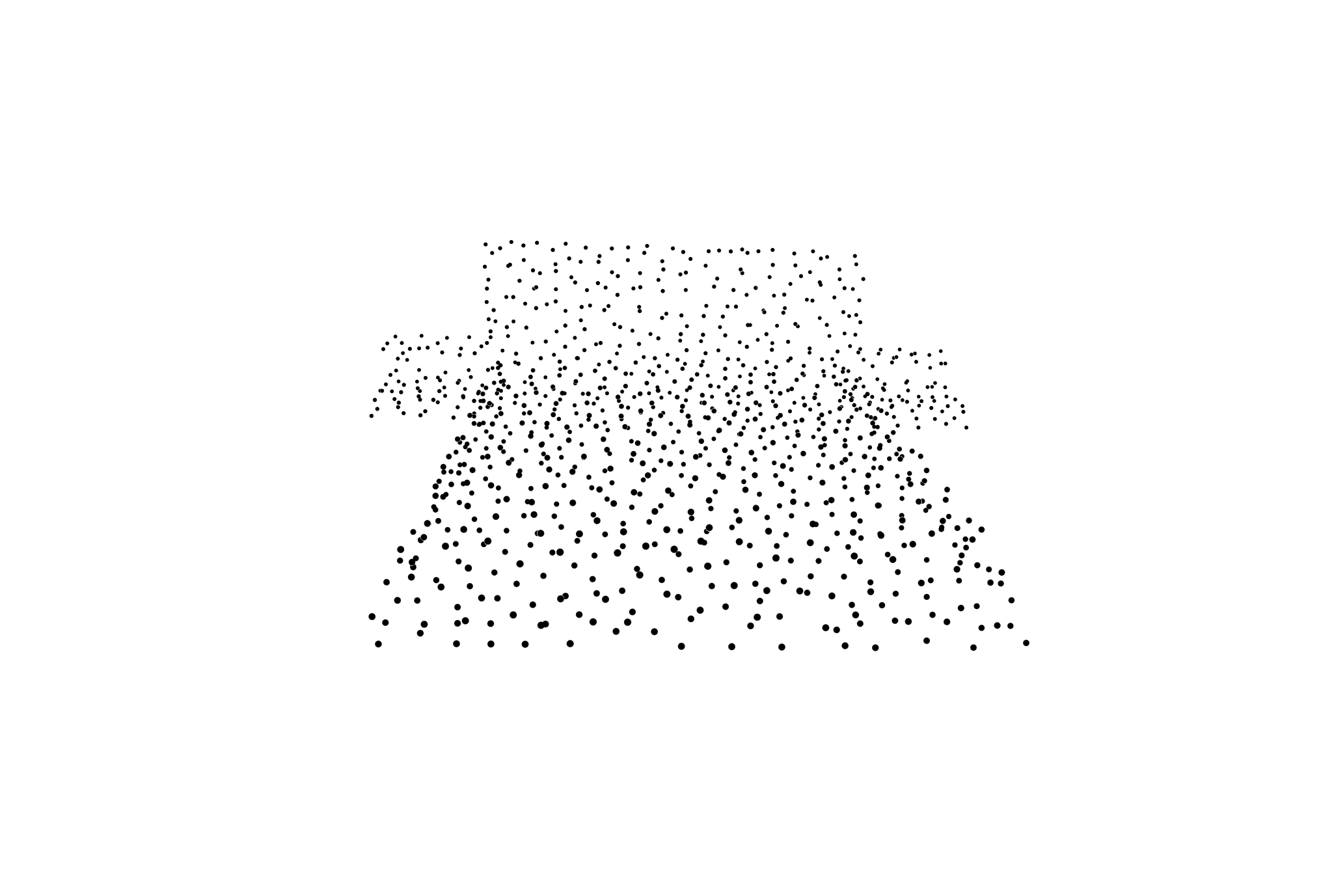} &
\figt[\sizeS]{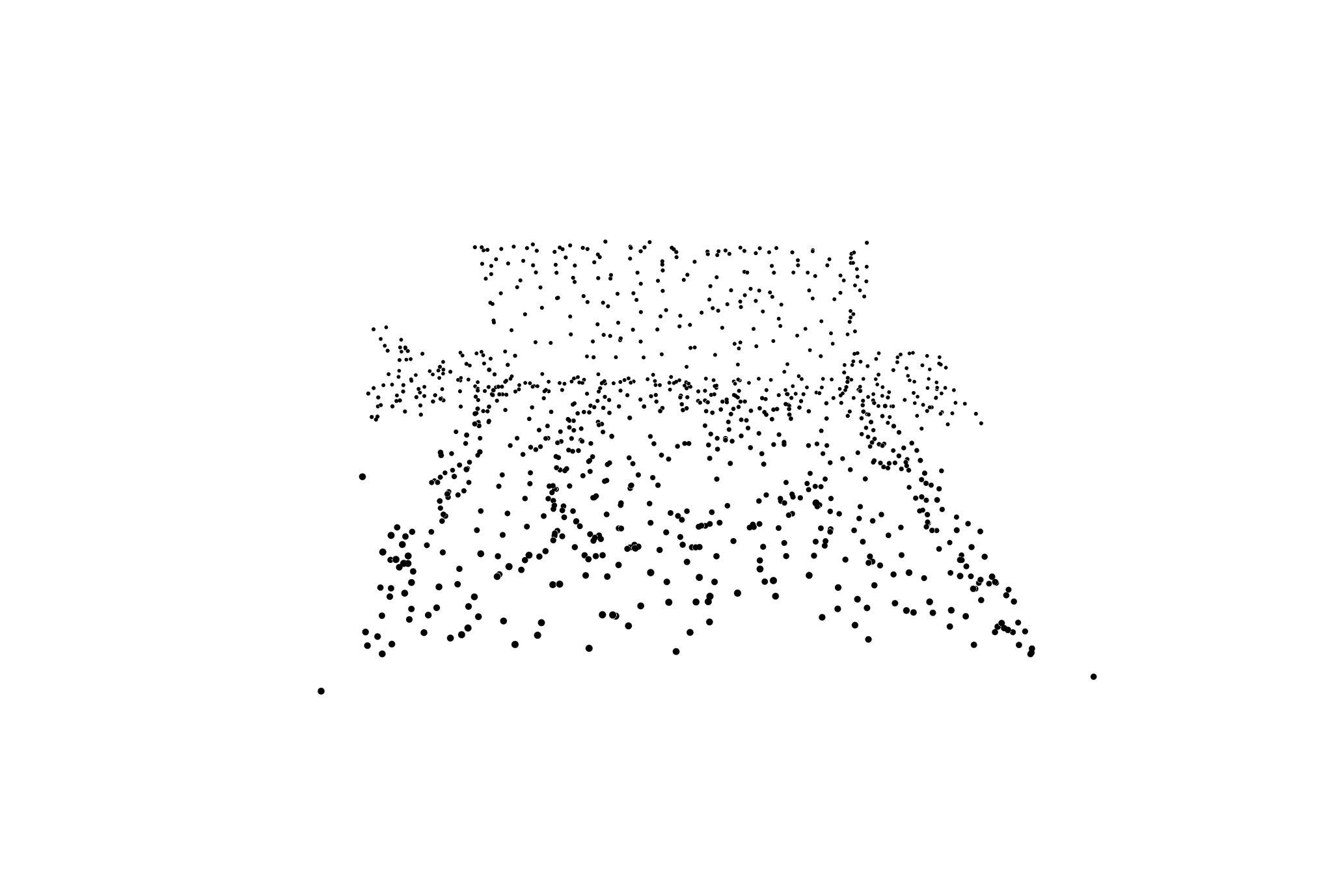} &
\figt[\sizeS]{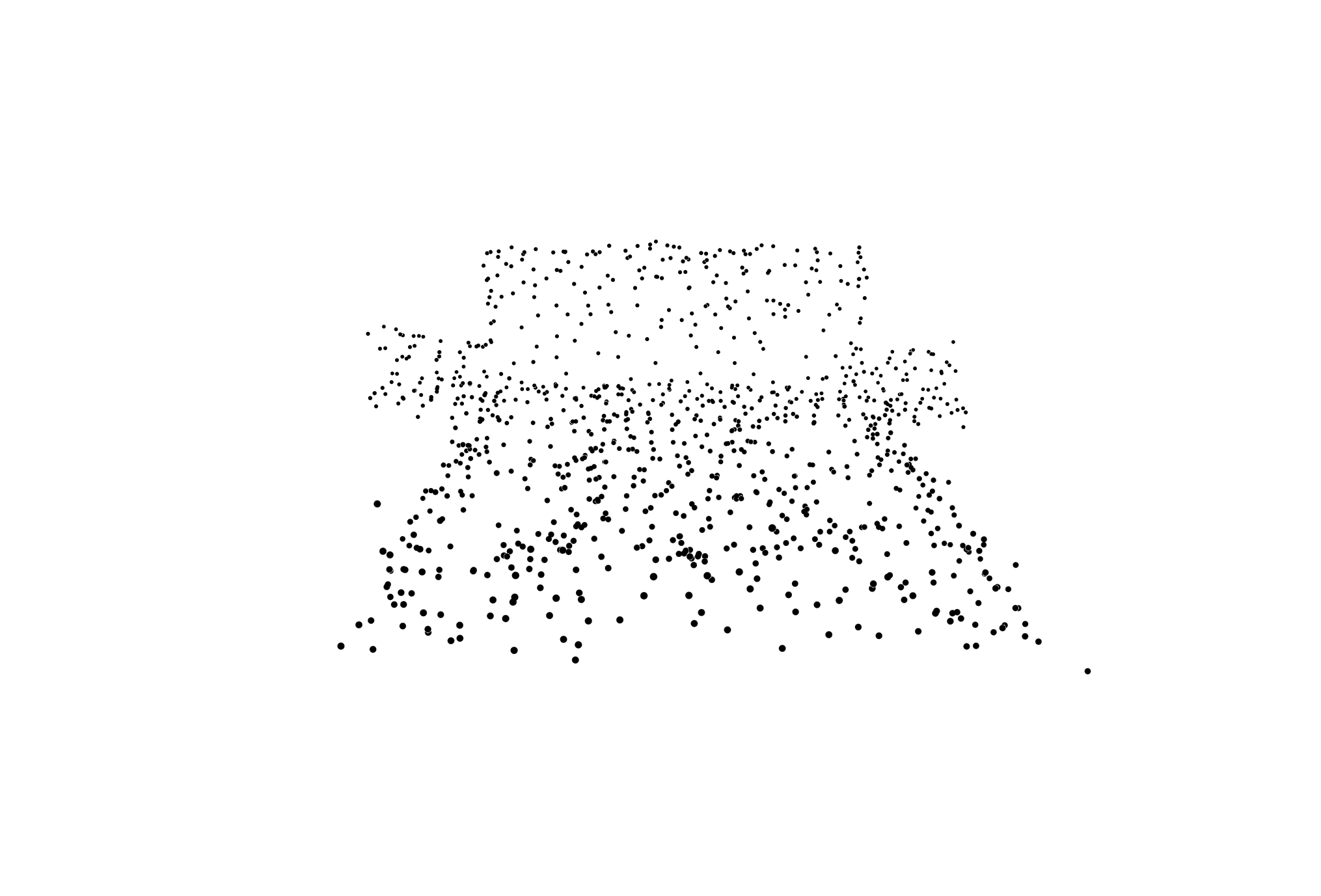} &
\figt[\sizeS]{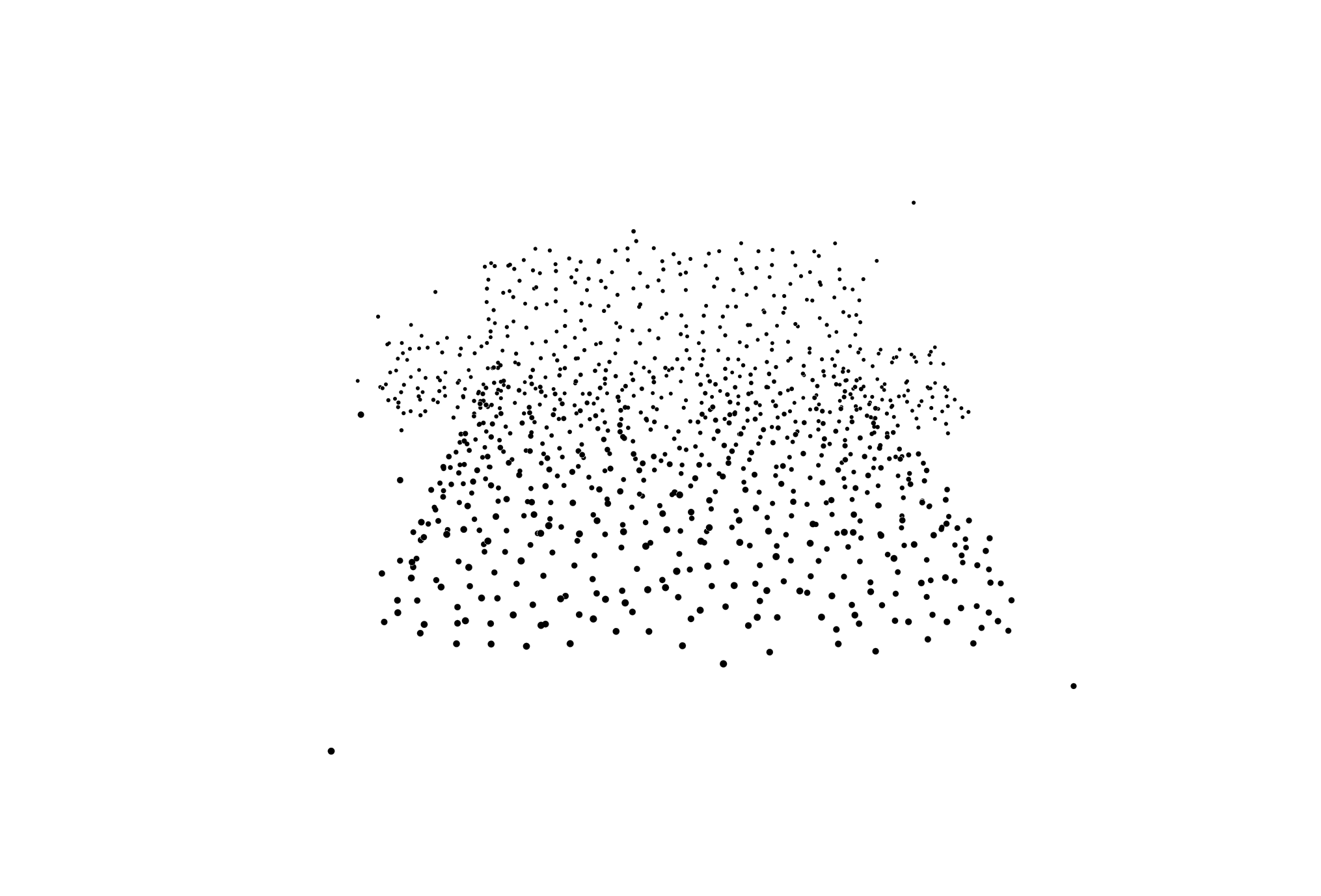} &
\figt[\sizeS]{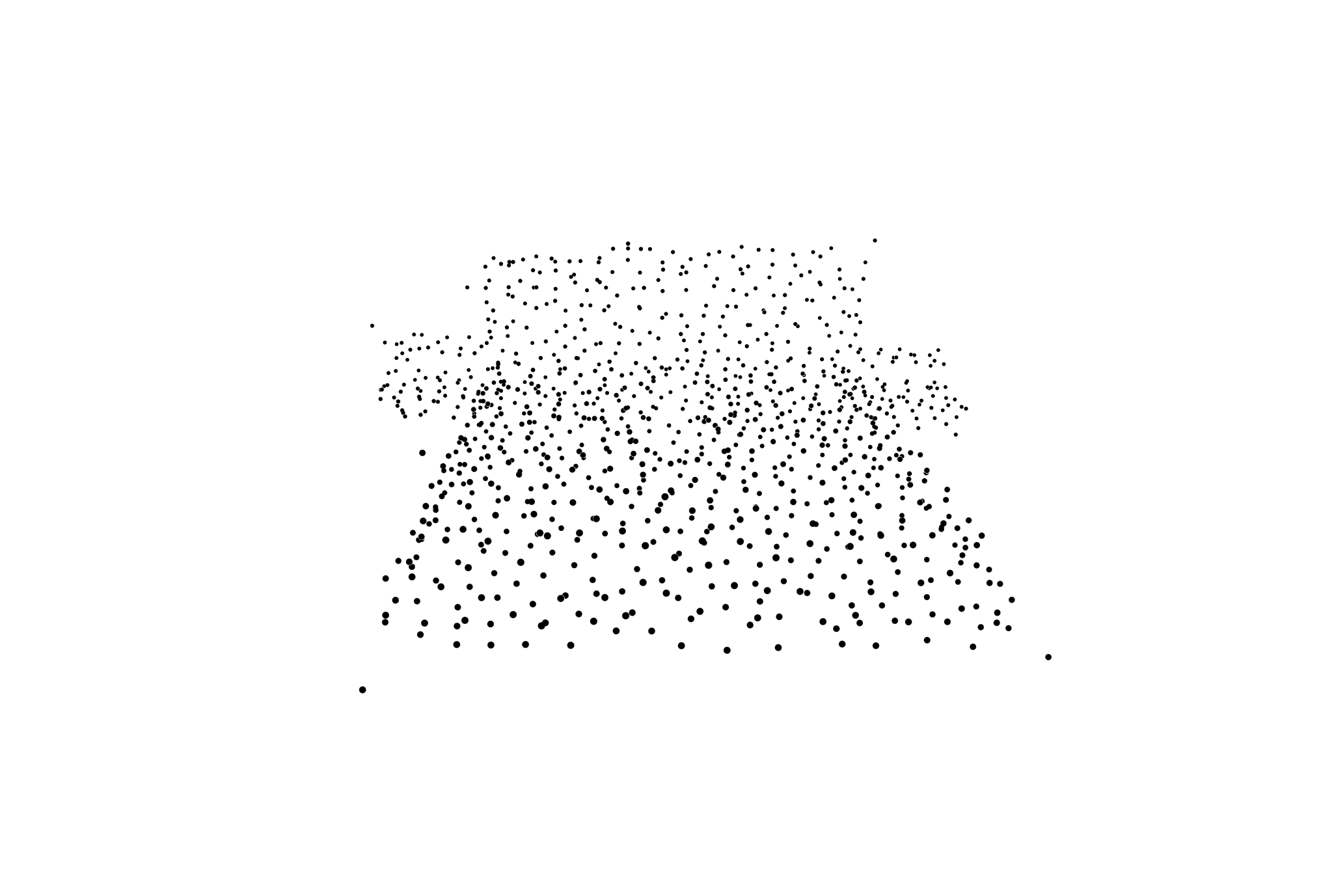} \\
(a) \blue{\cmark bed} &\red{\xmark bathtub} &\red{\xmark bathtub} &\red{\xmark bathtub} &\red{\xmark bathtub} \\

\figt[\sizeS]{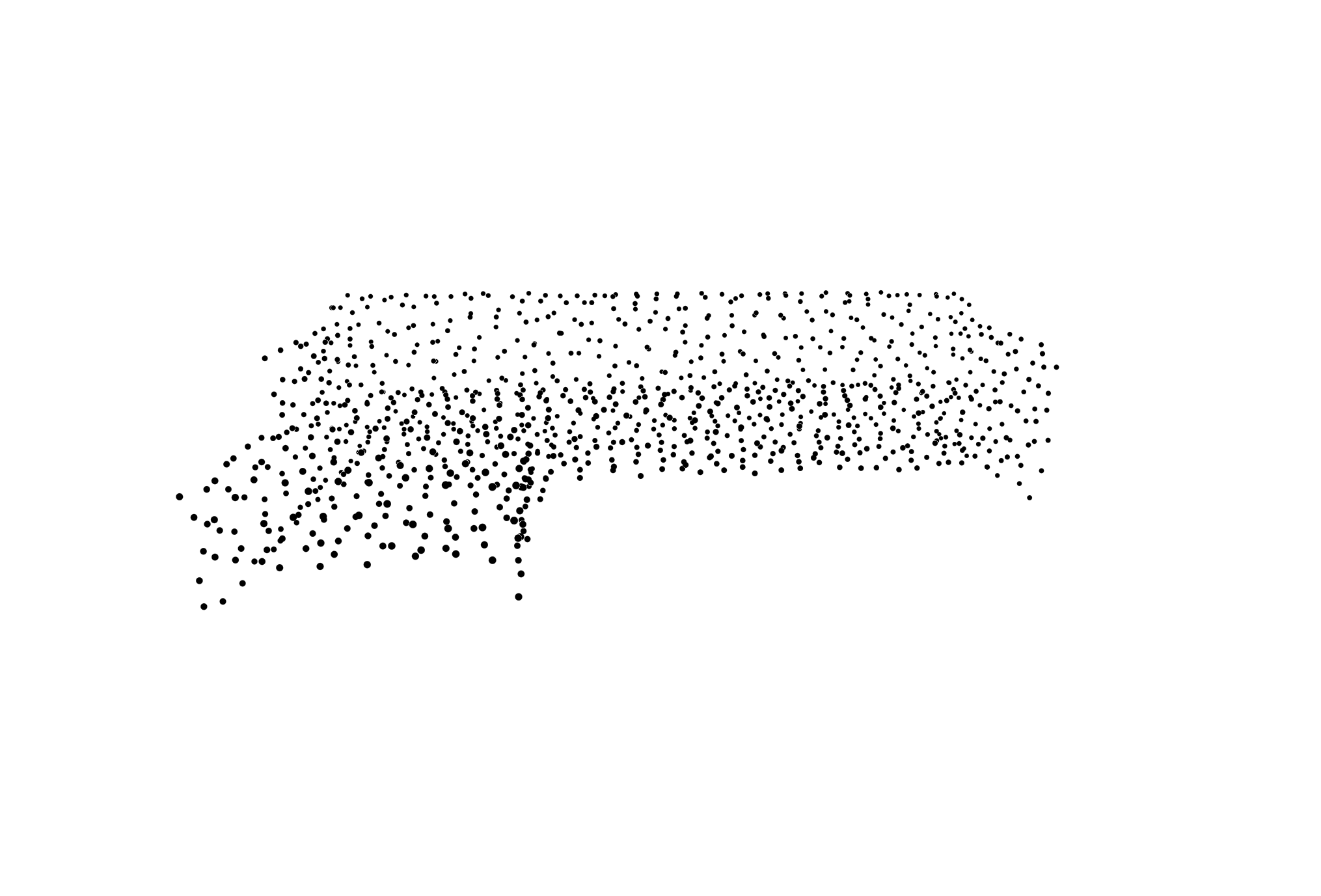} &
\figt[\sizeS]{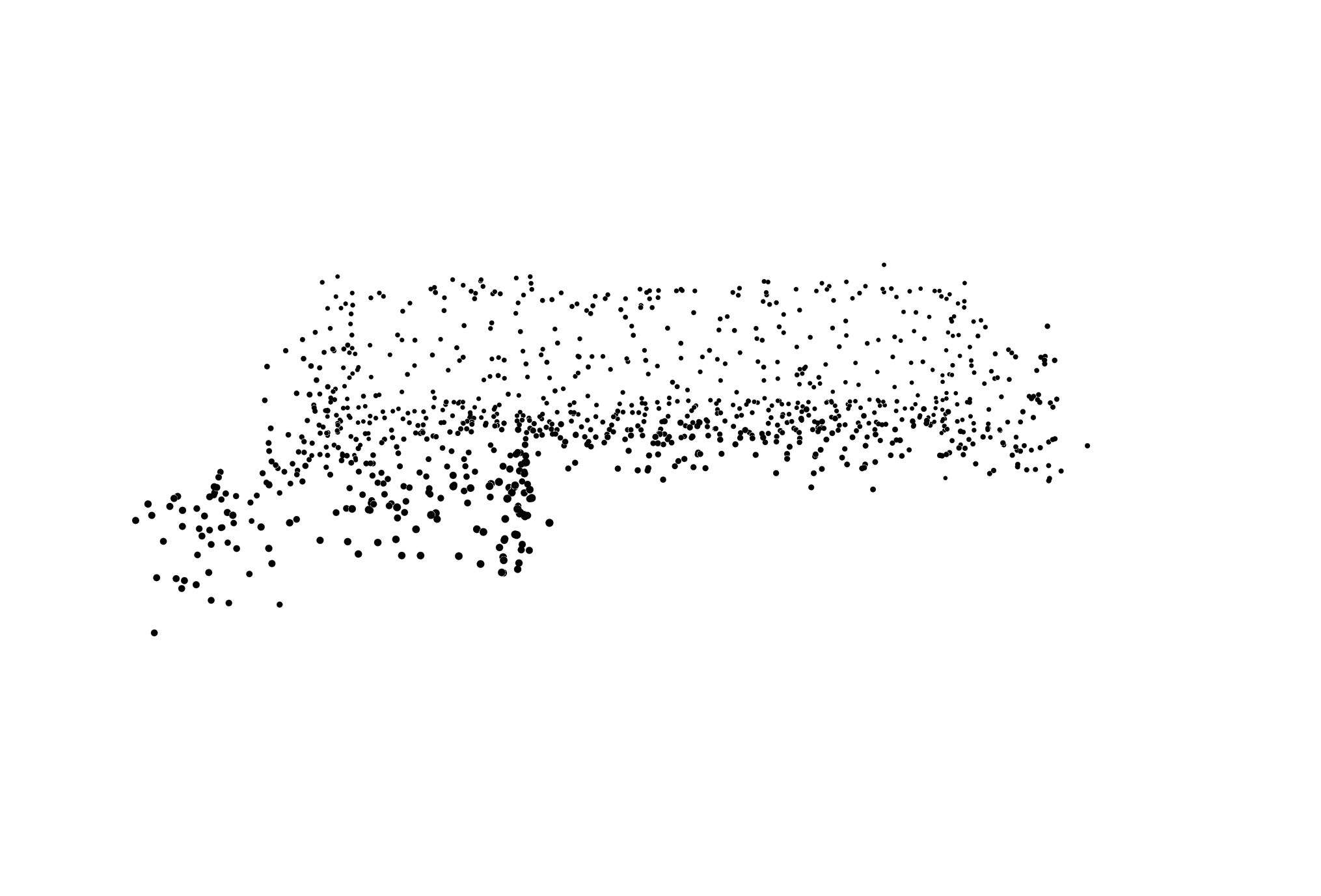} &
\figt[\sizeS]{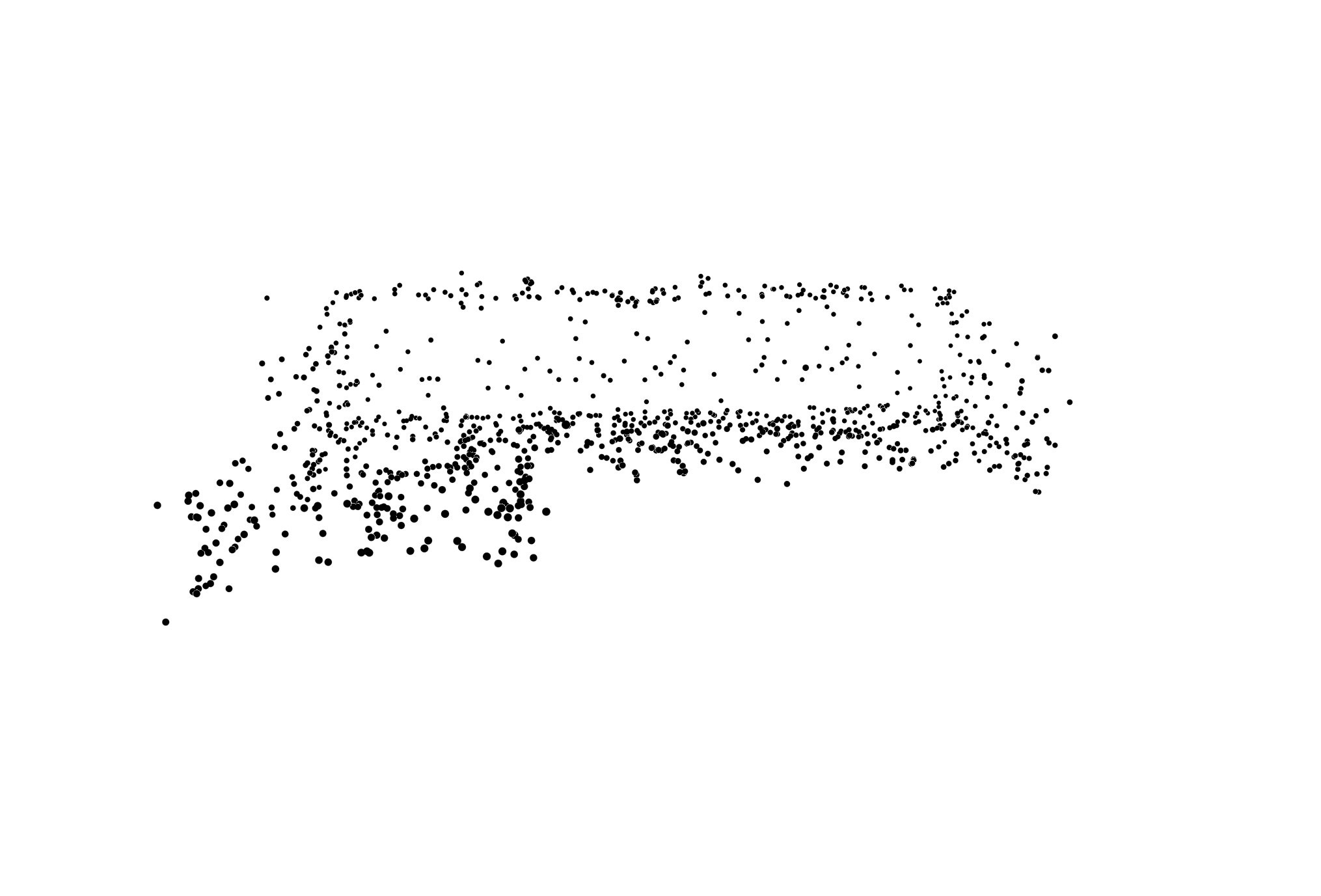} &
\figt[\sizeS]{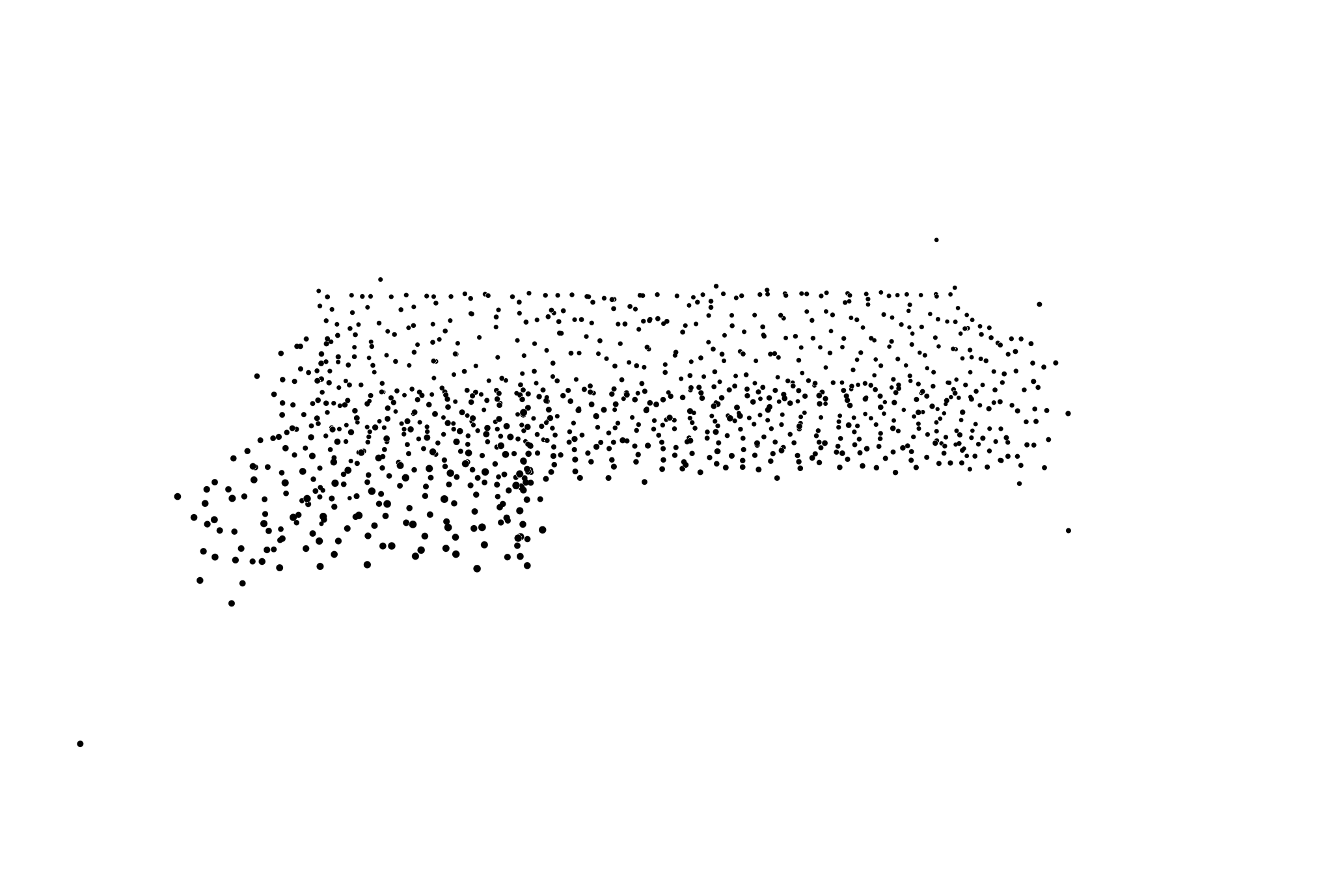} &
\figt[\sizeS]{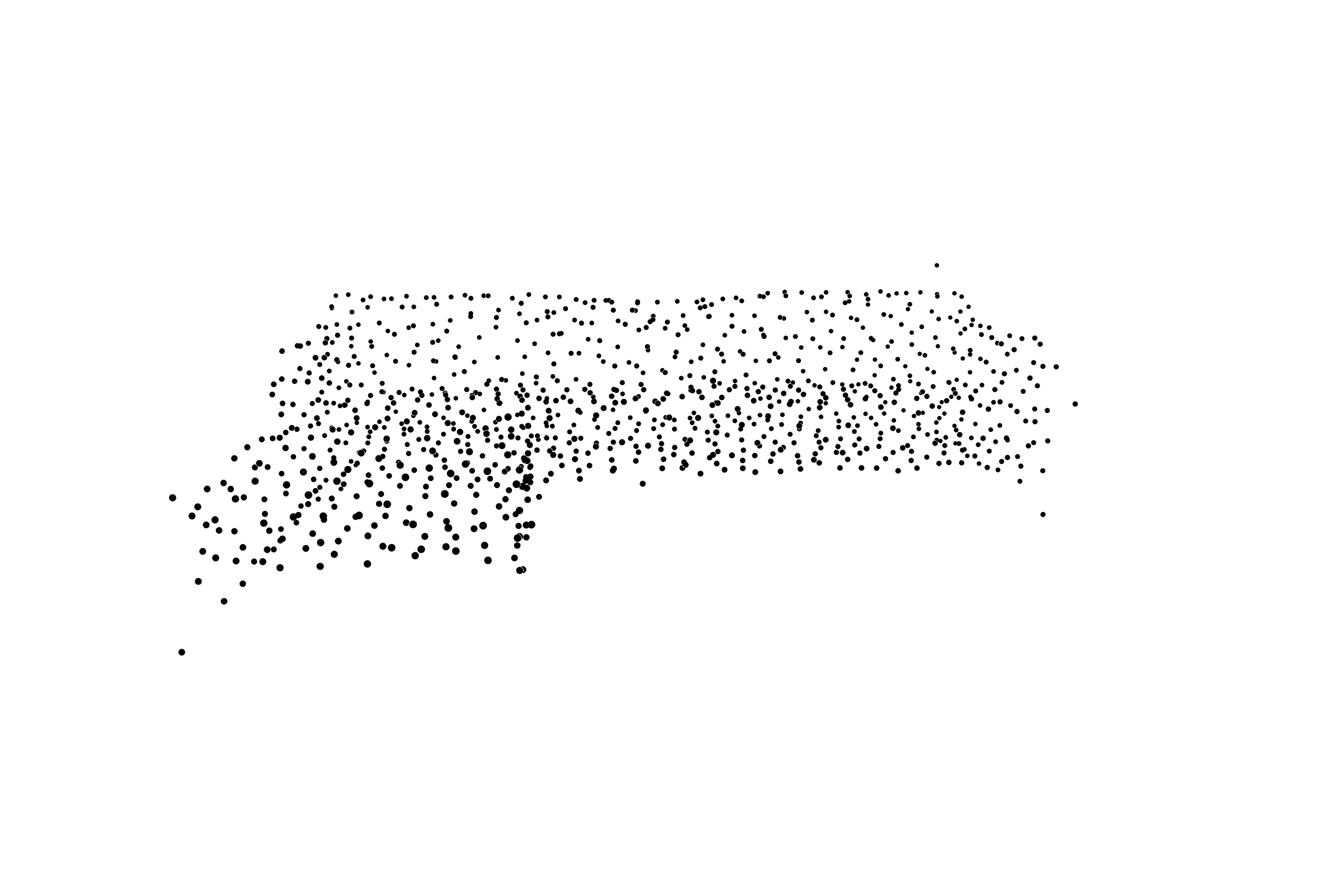} \\
(b) \blue{\cmark sofa} &\red{\xmark bed} &\red{\xmark bathtub} &\red{\xmark bed} &\red{\xmark bed} \\

\figt[\sizeS]{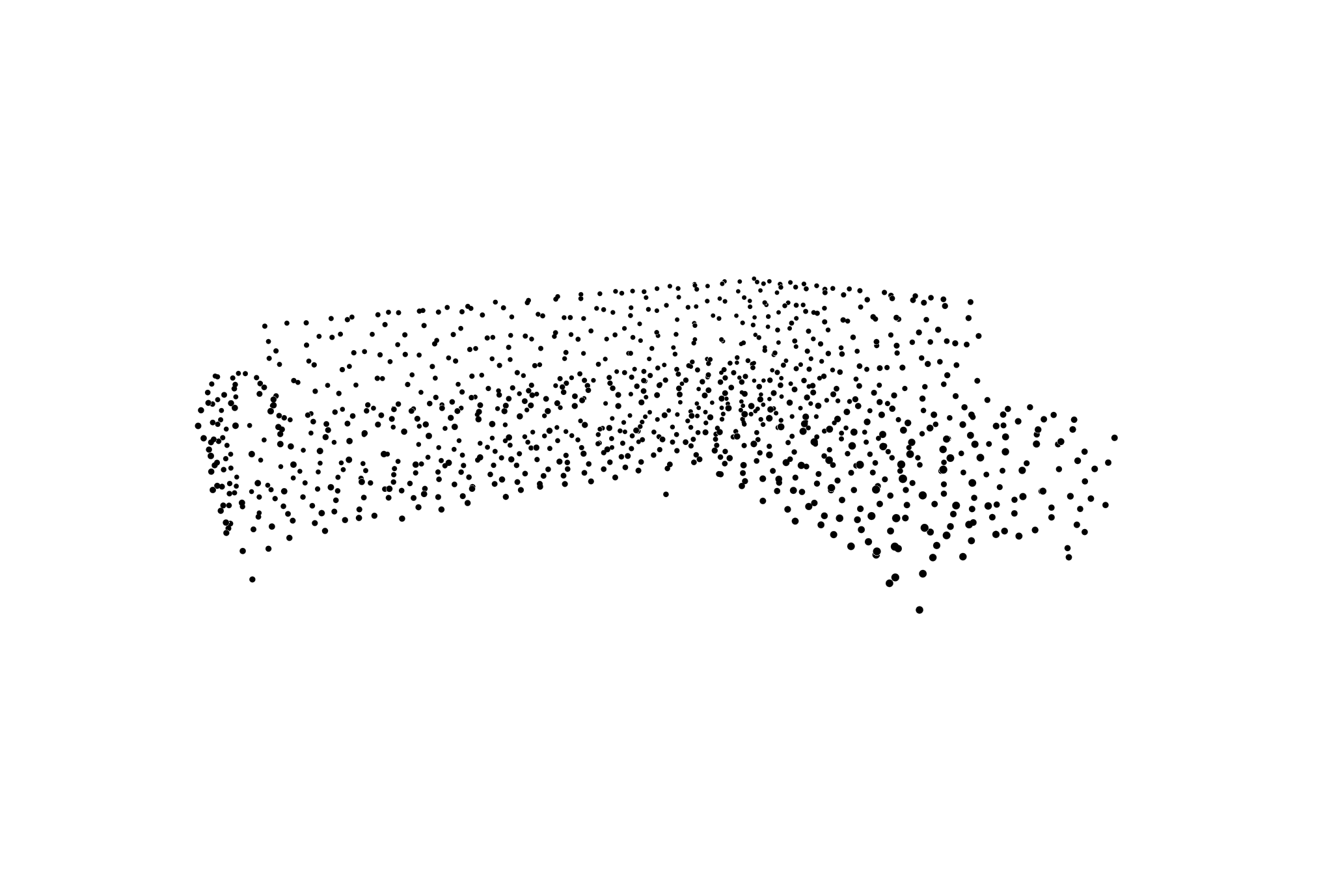} &
\figt[\sizeS]{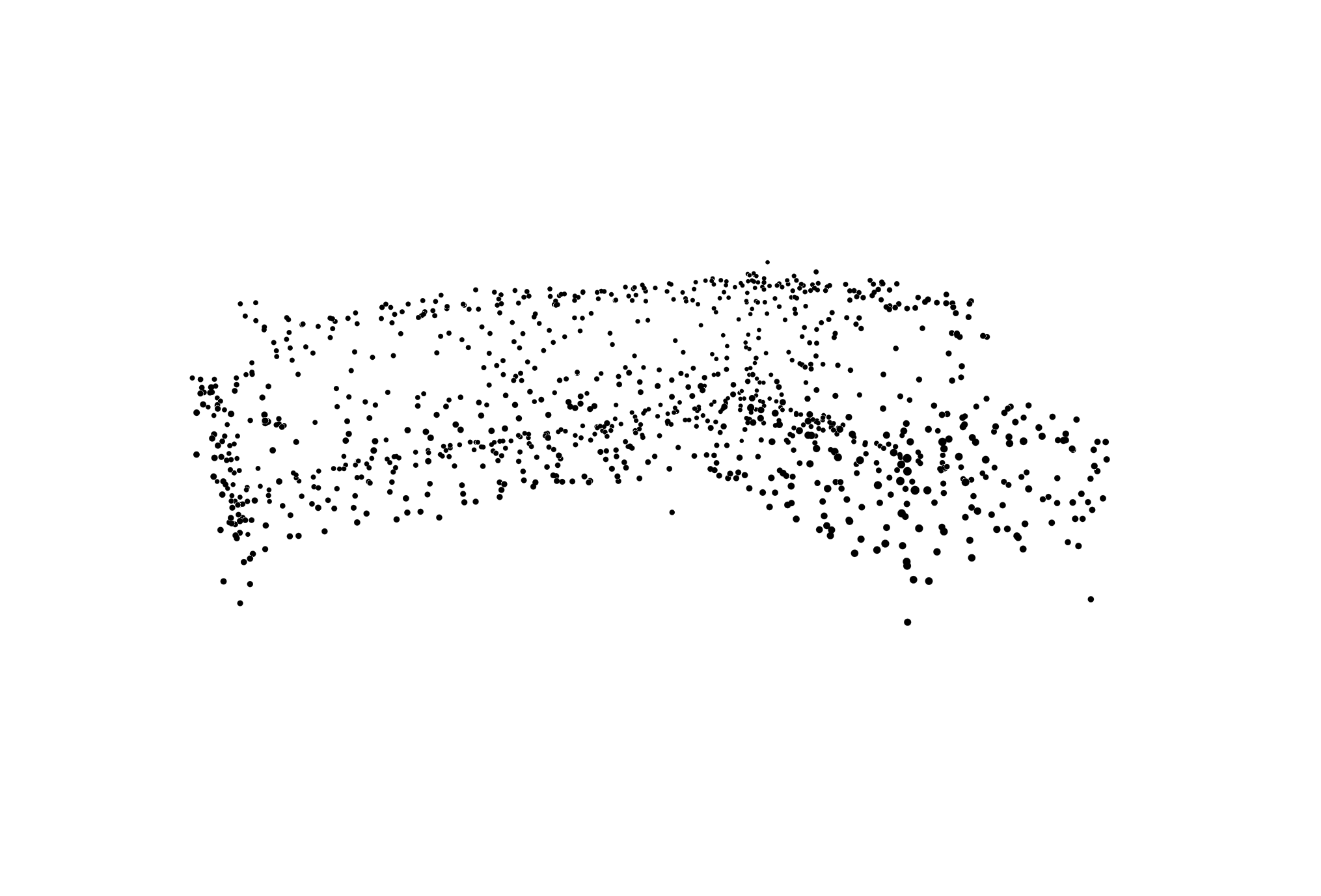} &
\figt[\sizeS]{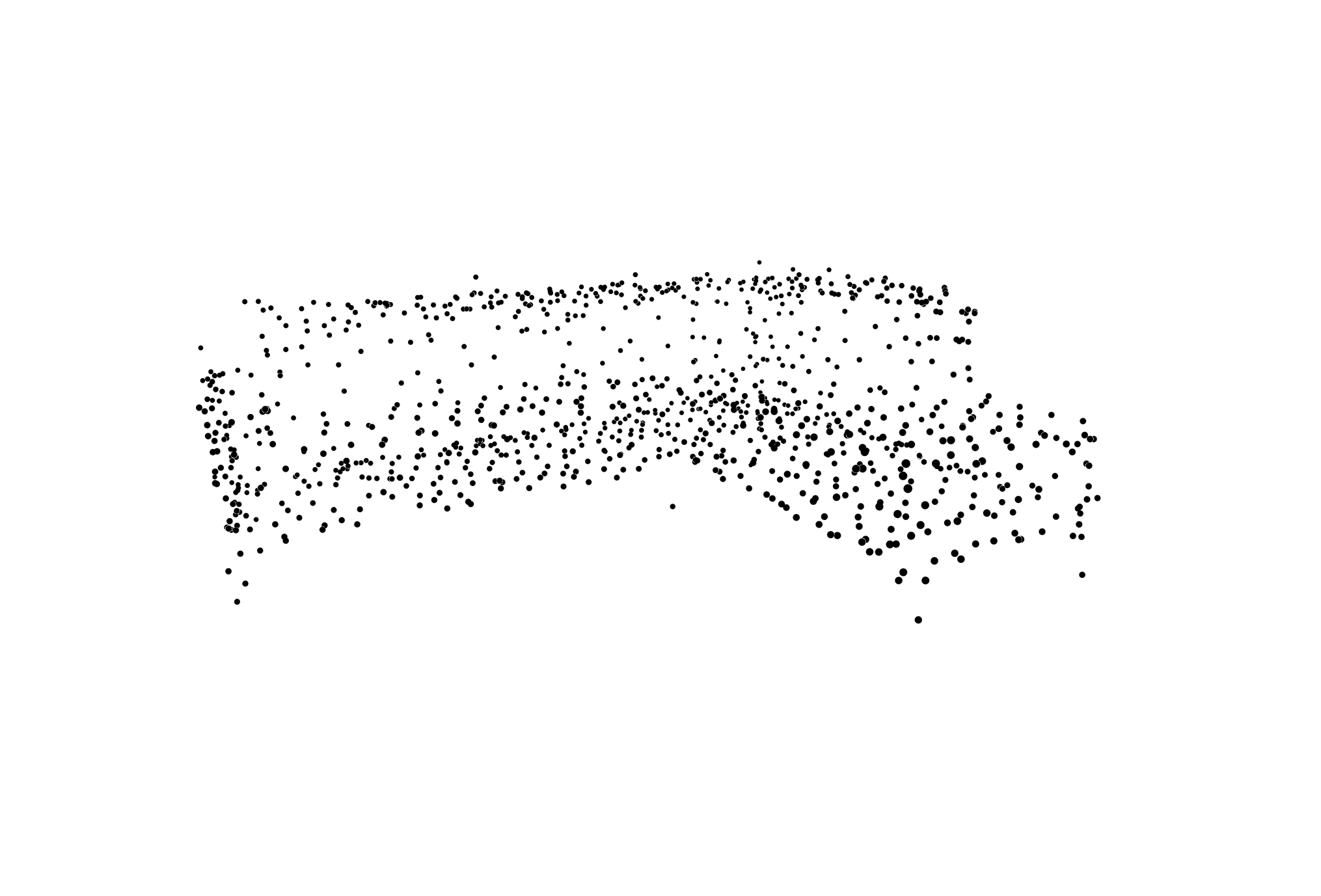} &
\figt[\sizeS]{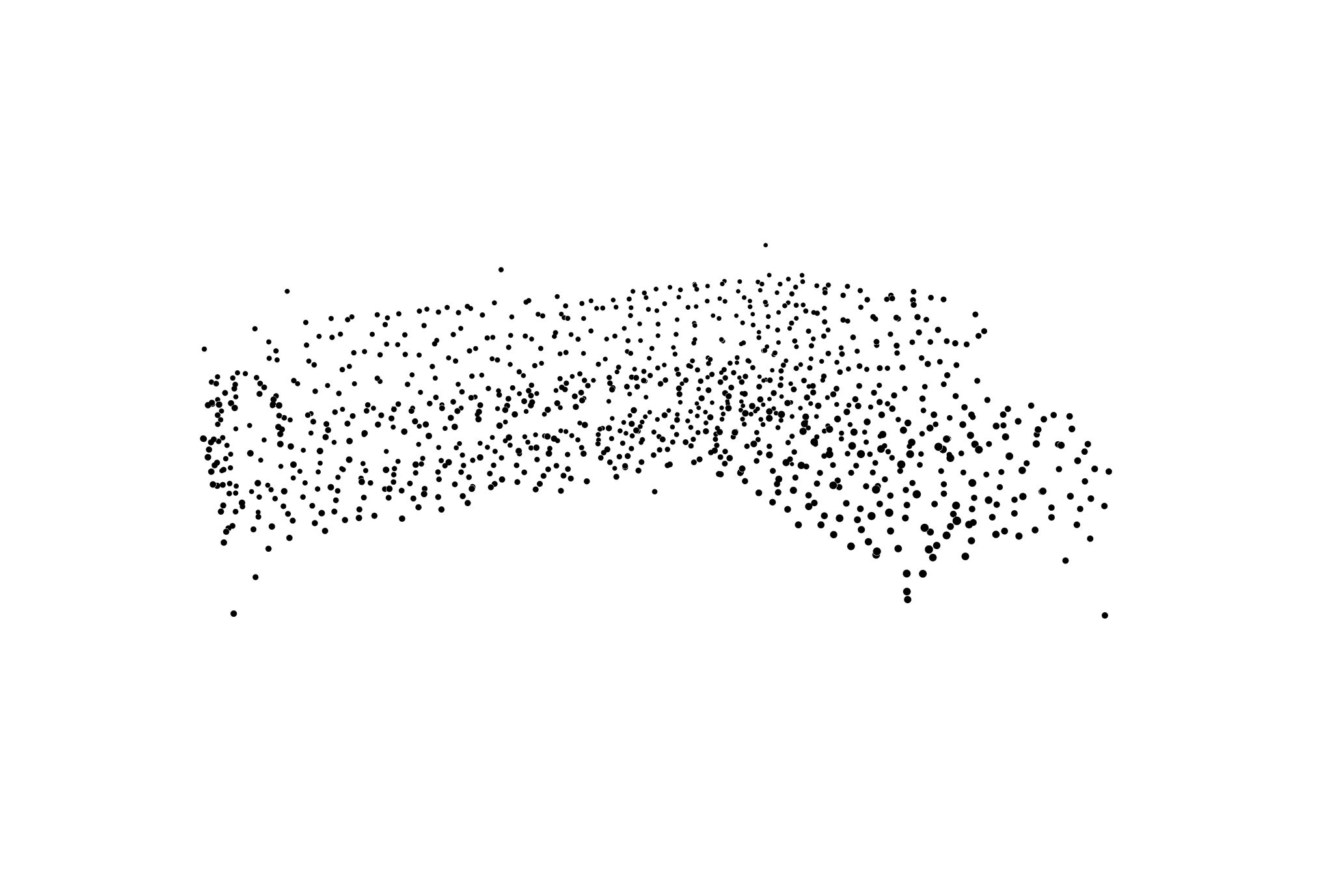} &
\figt[\sizeS]{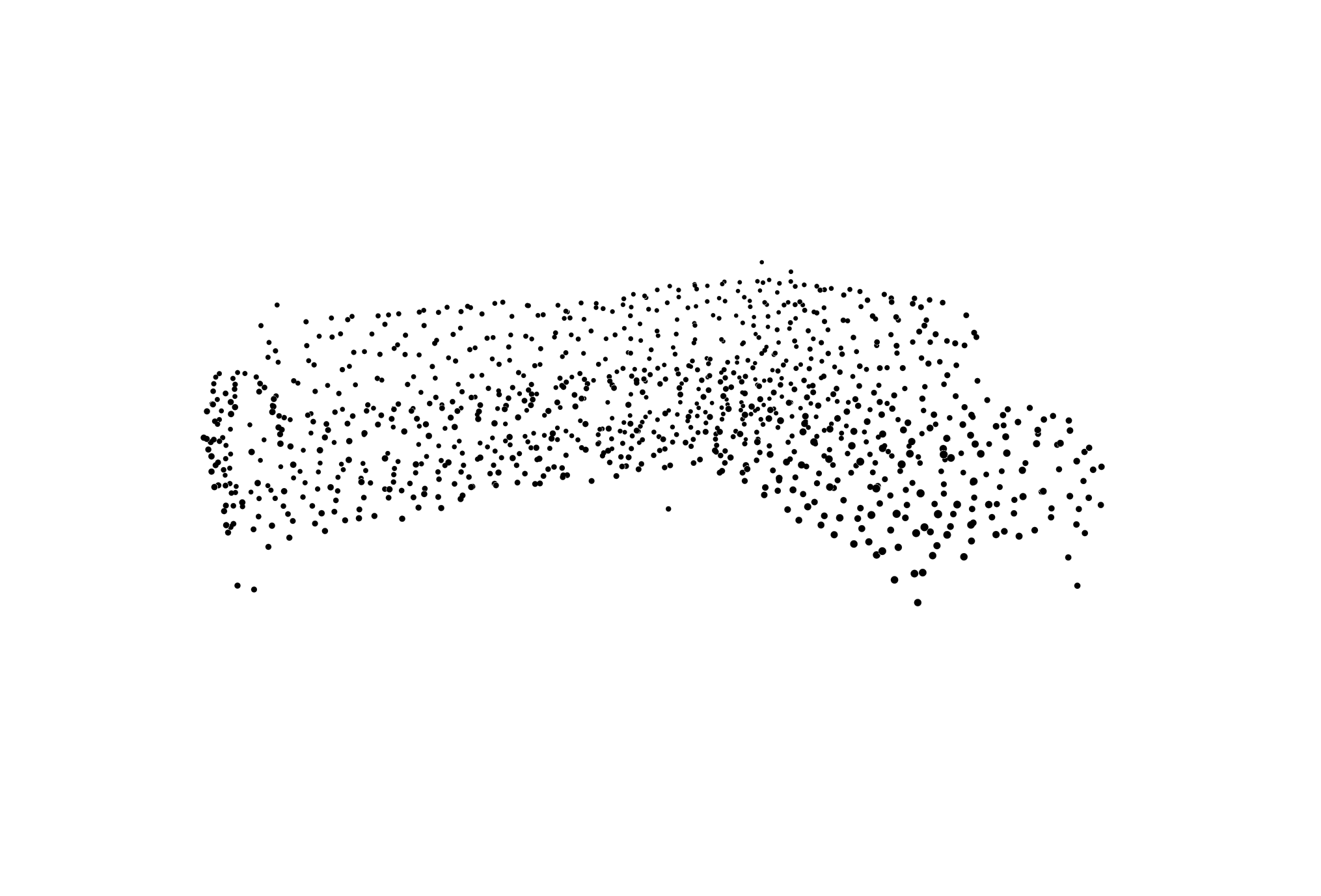} \\
(c) \blue{\cmark sofa} &\red{\xmark desk} &\red{\xmark desk} &\red{\xmark desk} &\red{\xmark desk} \\

\figt[\sizeS]{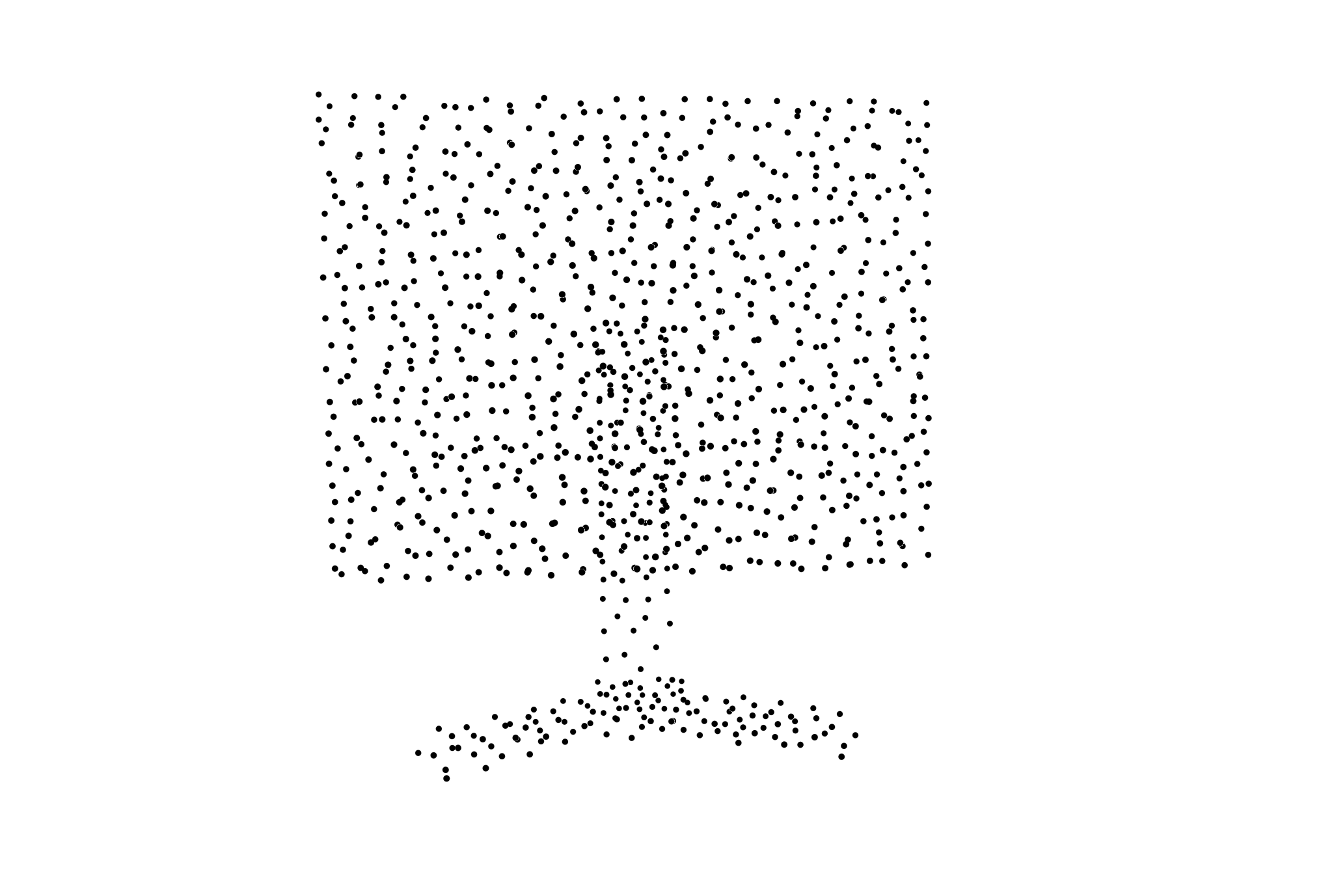} &
\figt[\sizeS]{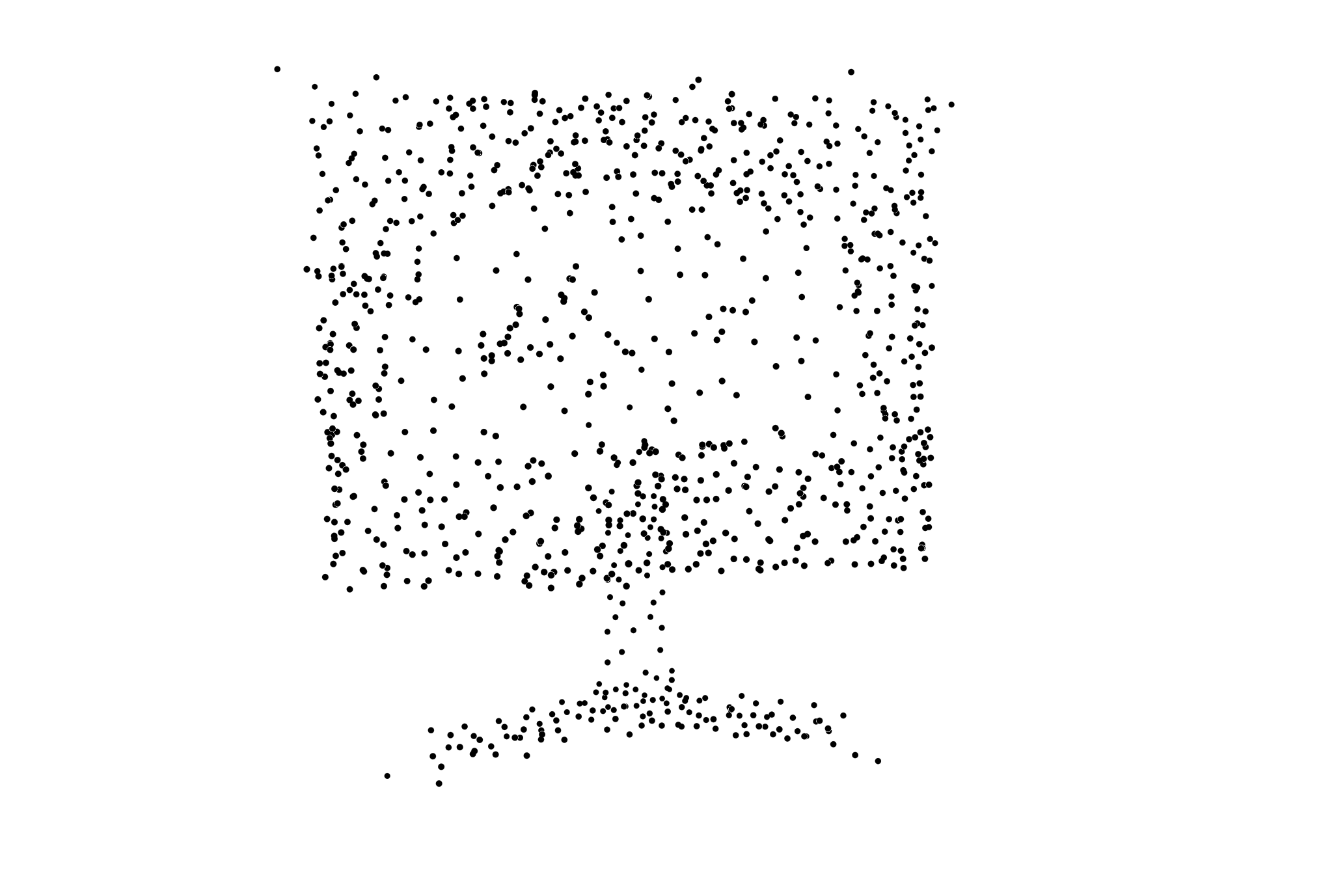} &
\figt[\sizeS]{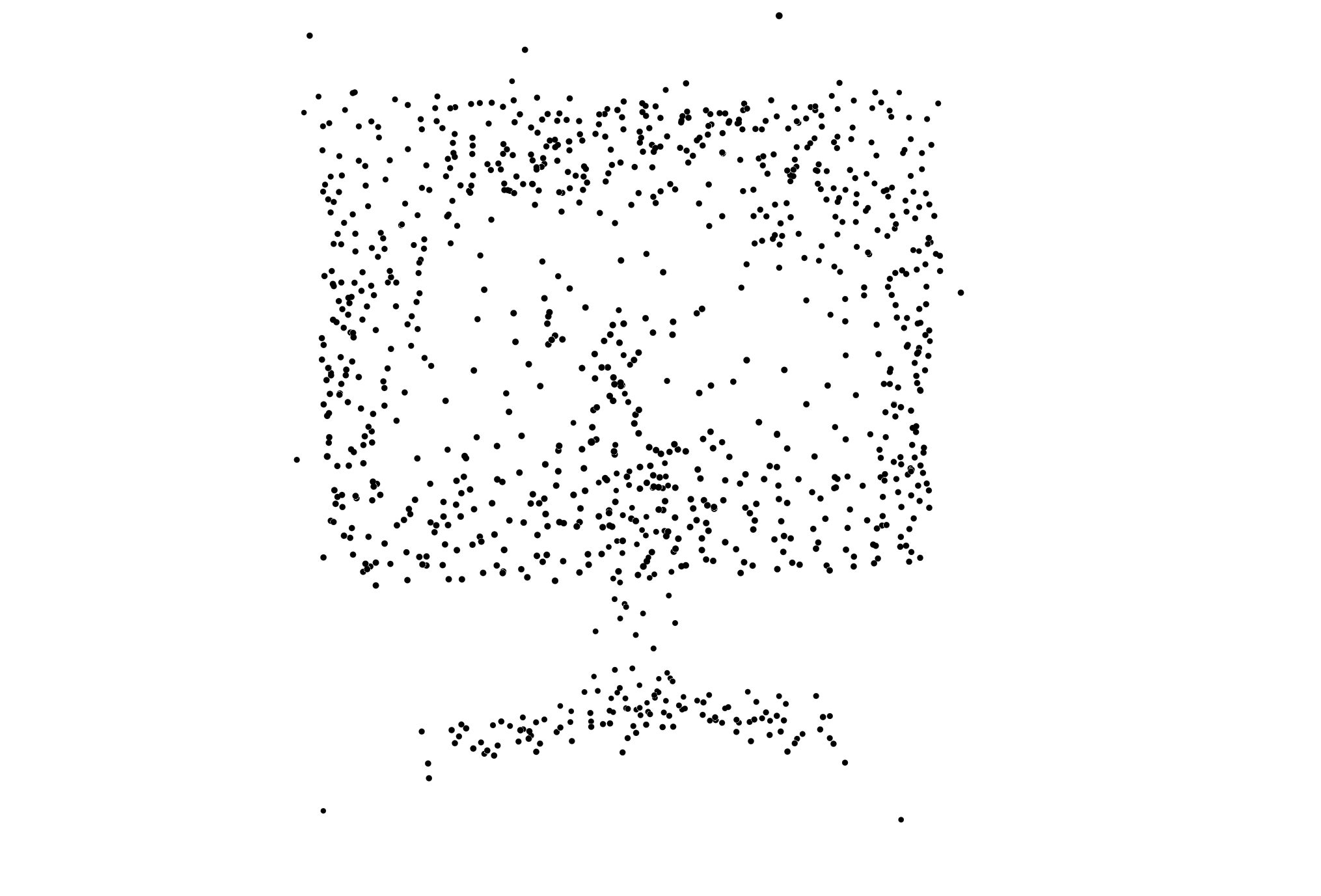} &
\figt[\sizeS]{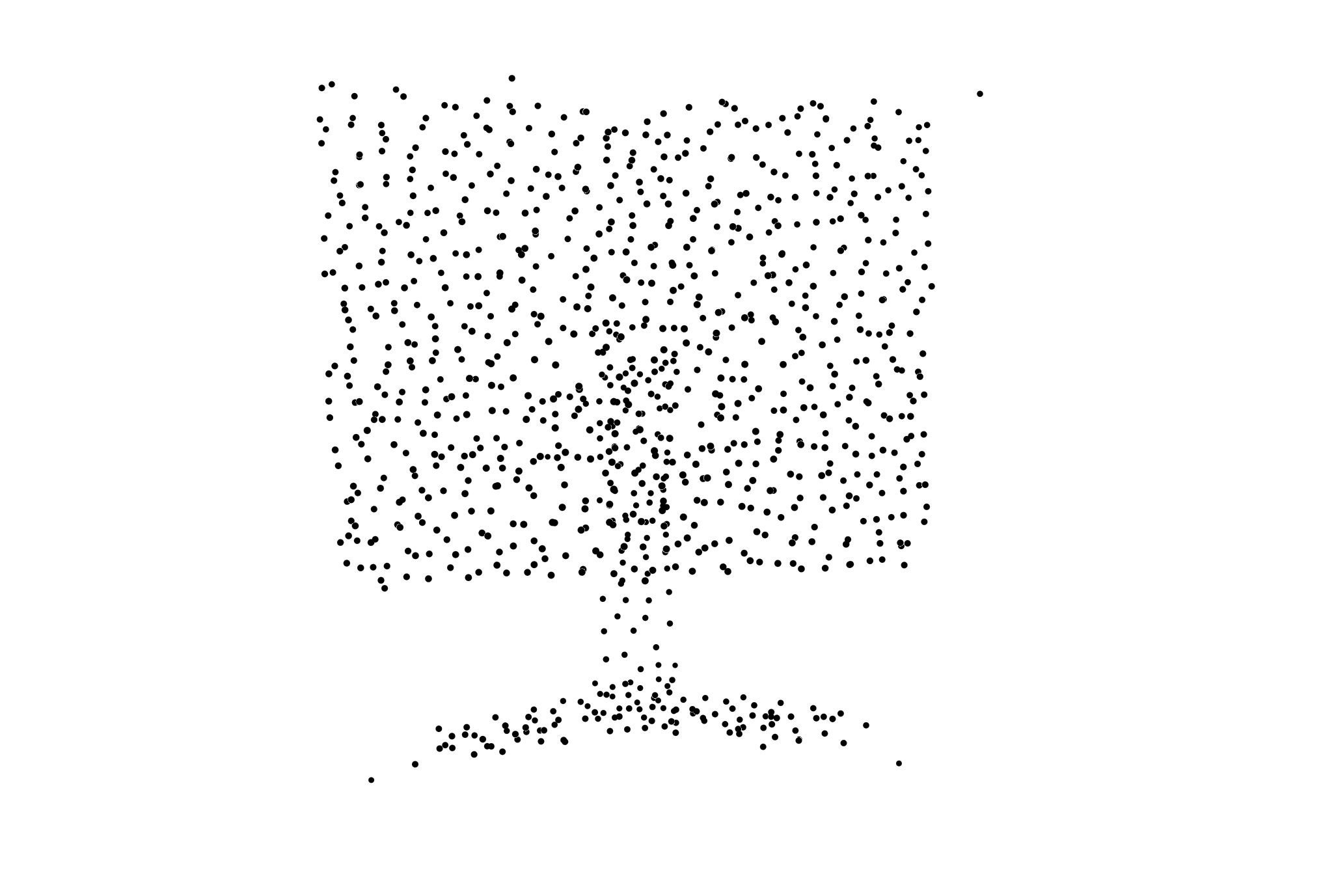} &
\figt[\sizeS]{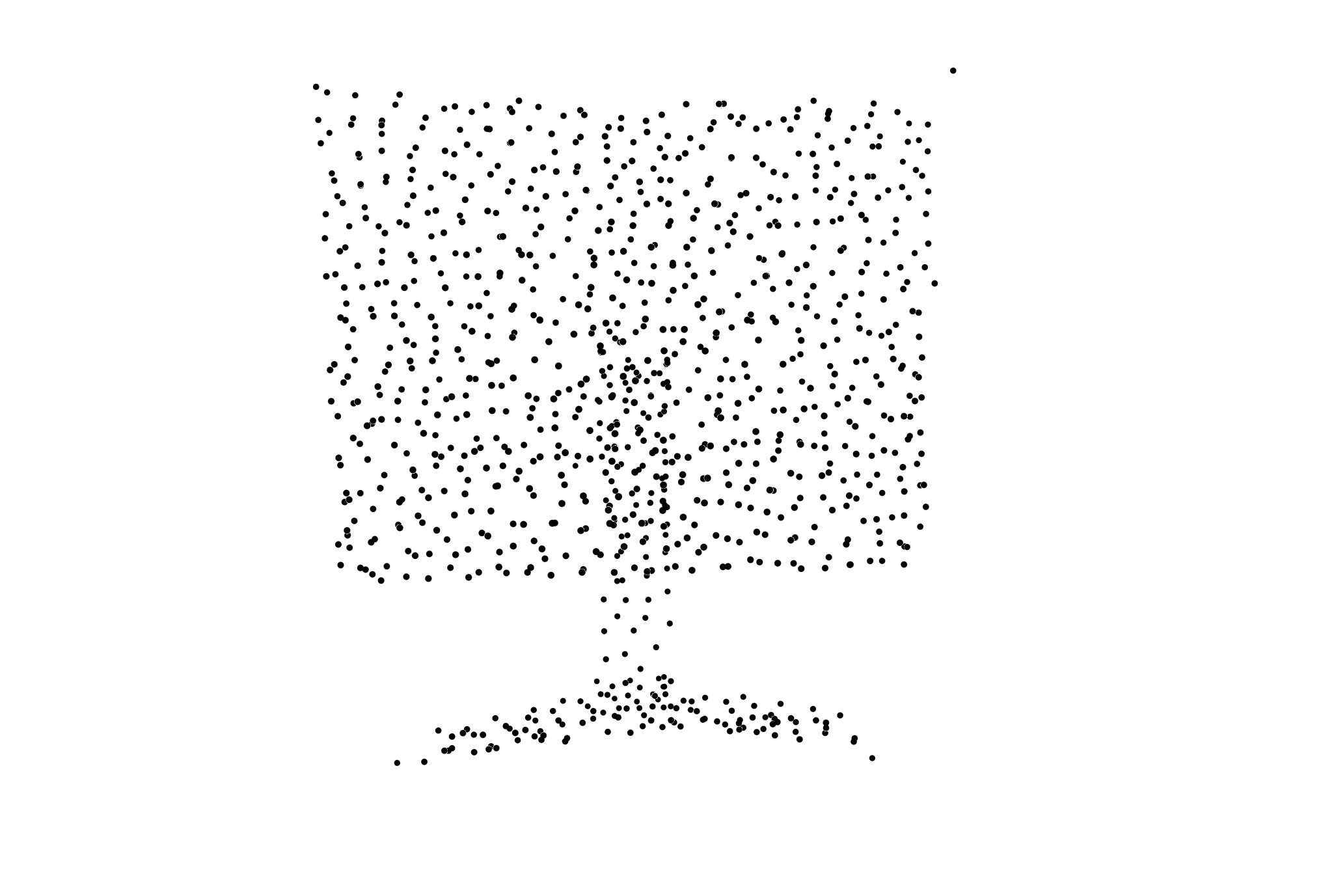} \\
(d) \blue{\cmark monitor} &\red{\xmark mantel} &\red{\xmark mantel} &\red{\xmark mantel} &\red{\xmark mantel} \\

\figt[\sizeS]{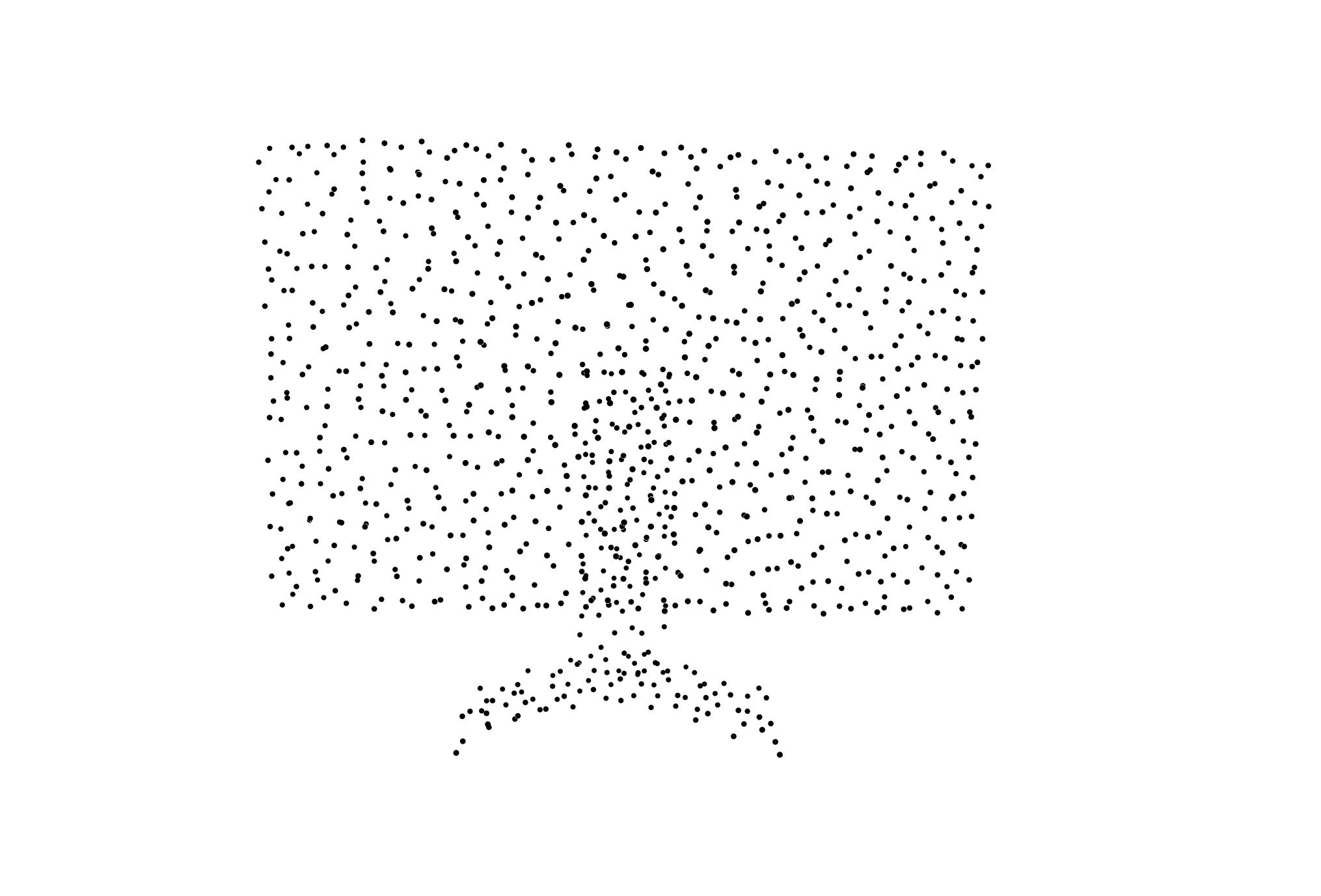} &
\figt[\sizeS]{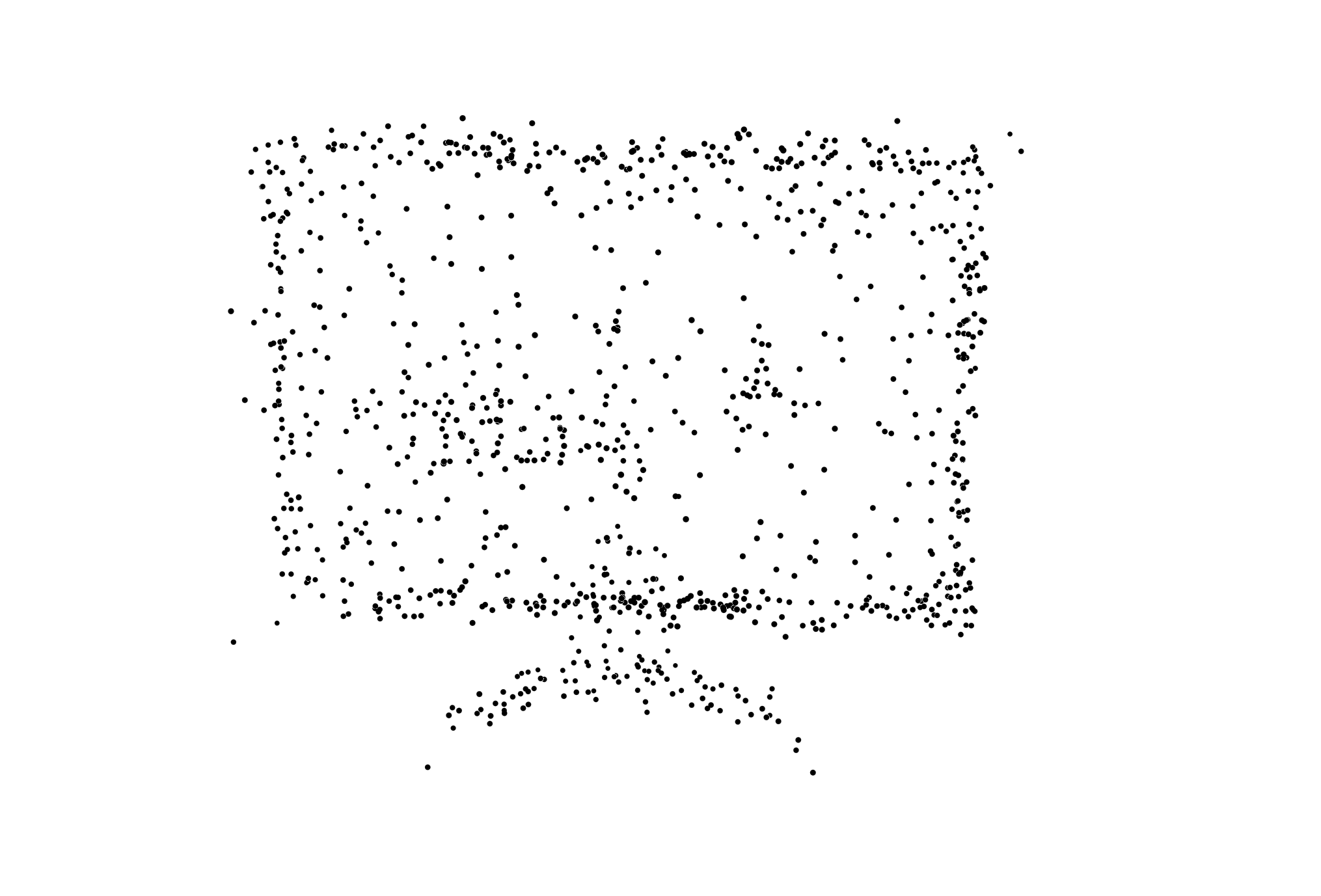} &
\figt[\sizeS]{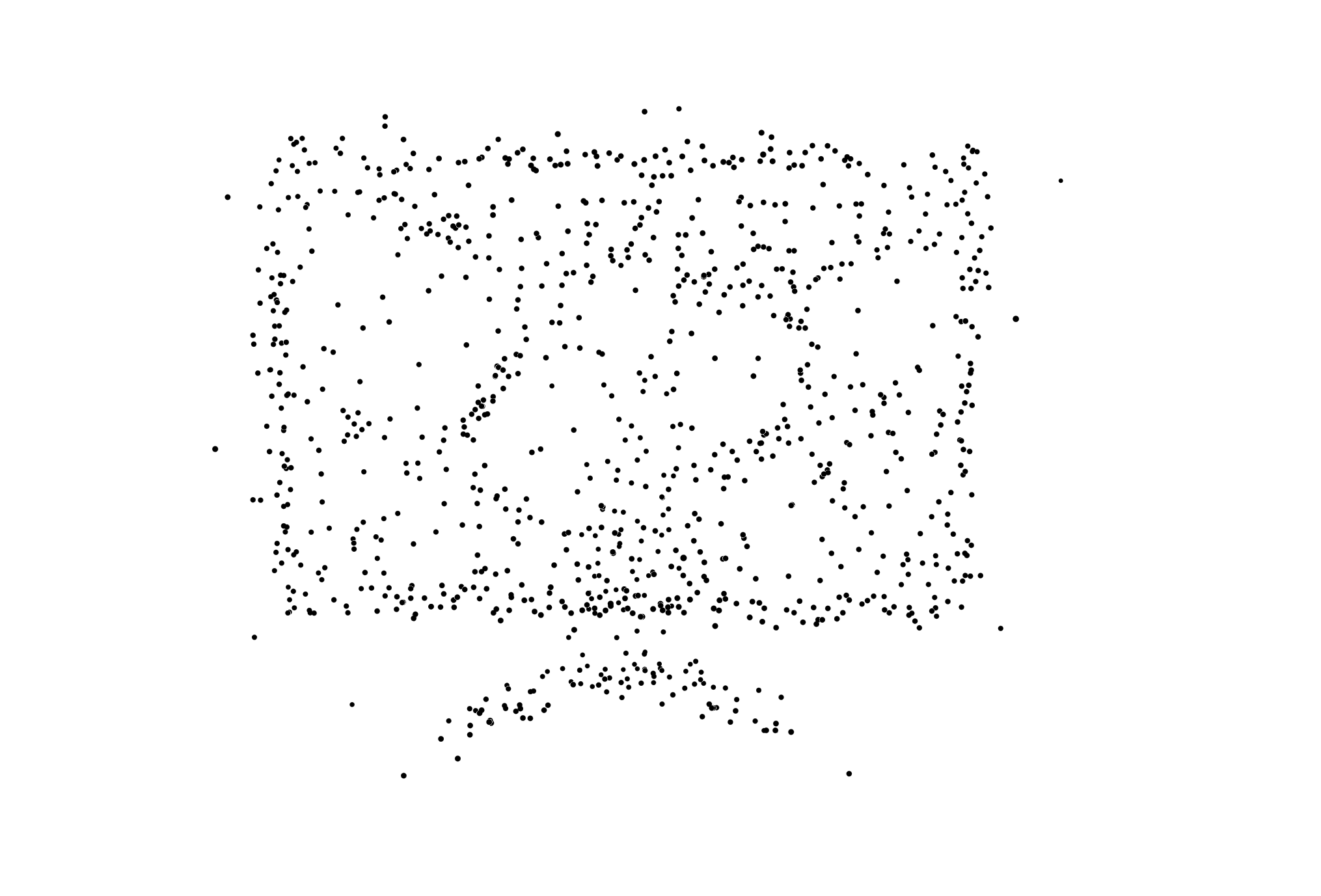} &
\figt[\sizeS]{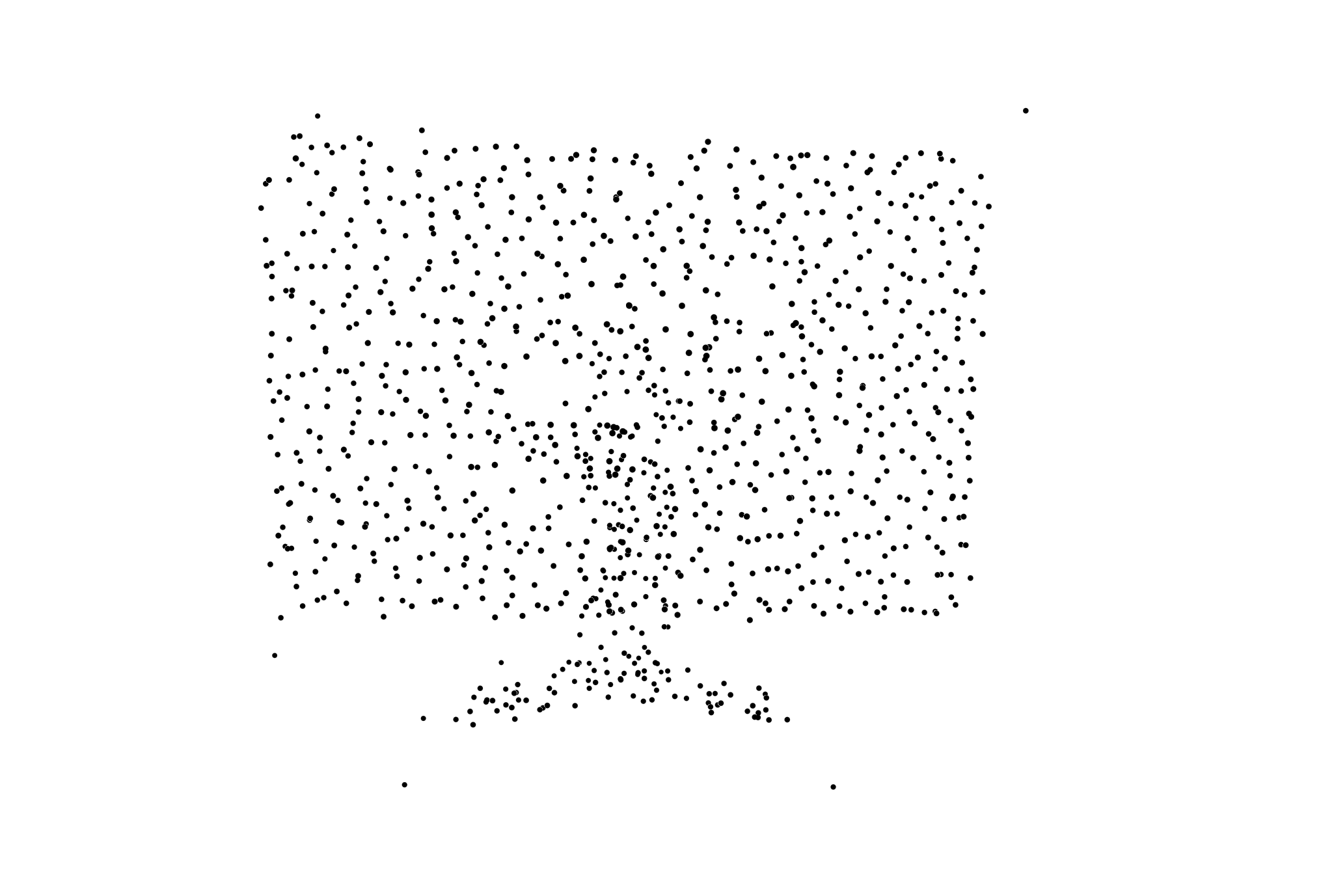} &
\figt[\sizeS]{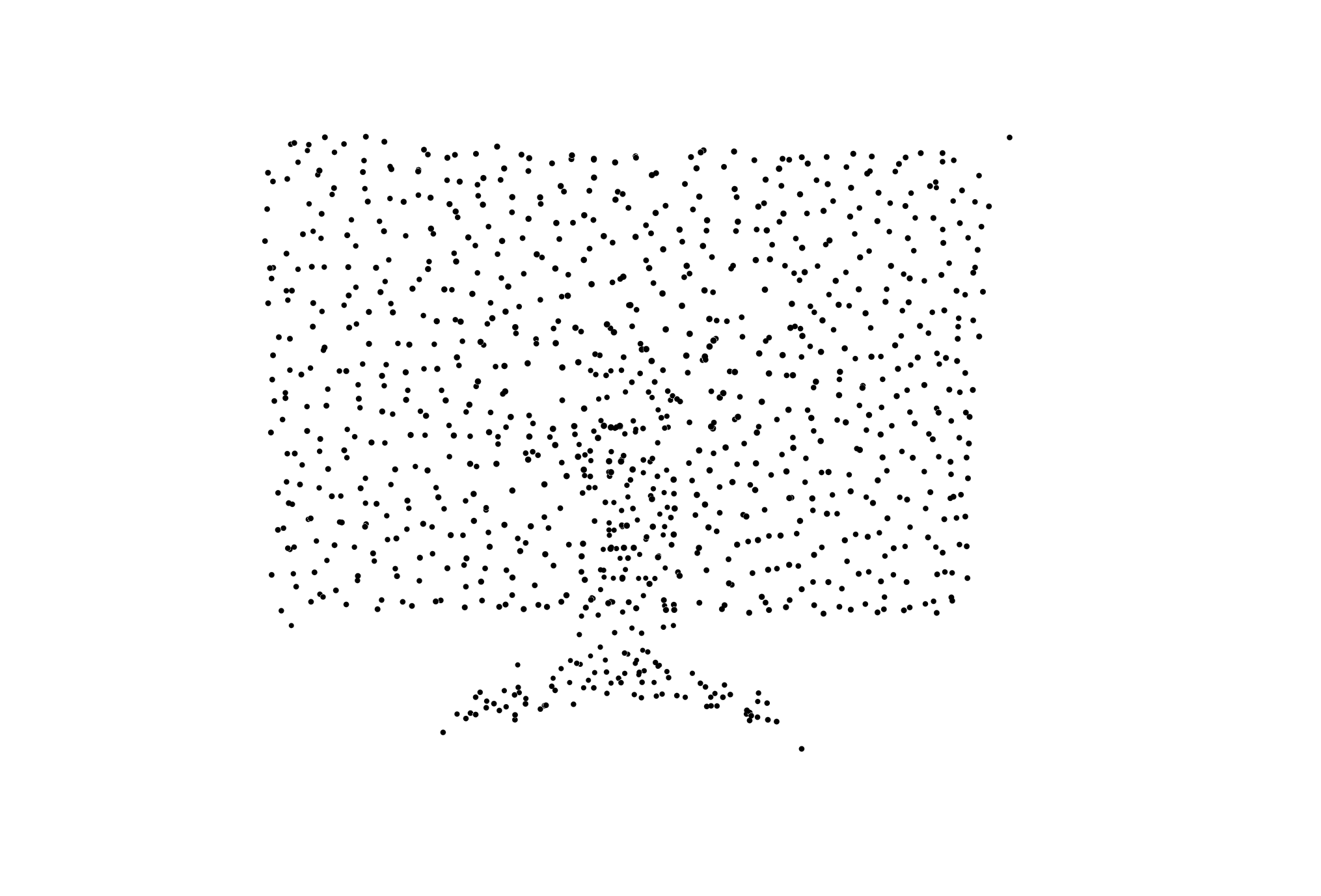} \\
(e) \blue{\cmark monitor} &\red{\xmark tv-stand} &\red{\xmark desk} &\red{\xmark mantel} &\red{\xmark mantel} \\

\end{tabular}
\caption{Visualization for adversarial point clouds produced by baseline methods on PointNet.}
\label{fig:vis-3d-pointnet-supp}
\end{figure*}

\begin{figure*}[hptb]
\newcommand{\sizeS}{.20}
\centering
\setlength{\tabcolsep}{0pt}
\begin{tabular}{ccccc}
\textbf{Clean} &\geoa &\gsda &\siadv & \projabbrv\\
\figt[\sizeS]{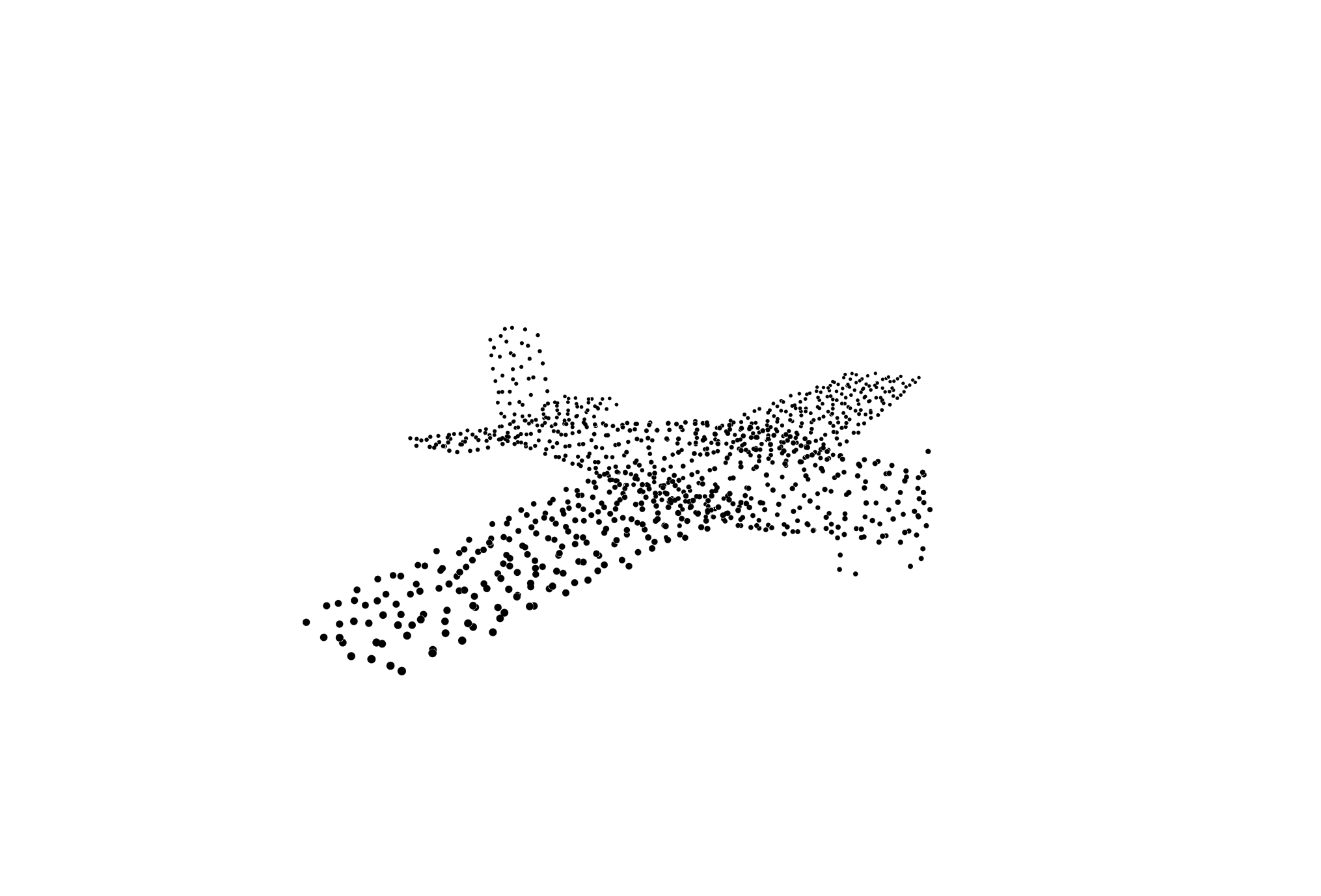} &
\figt[\sizeS]{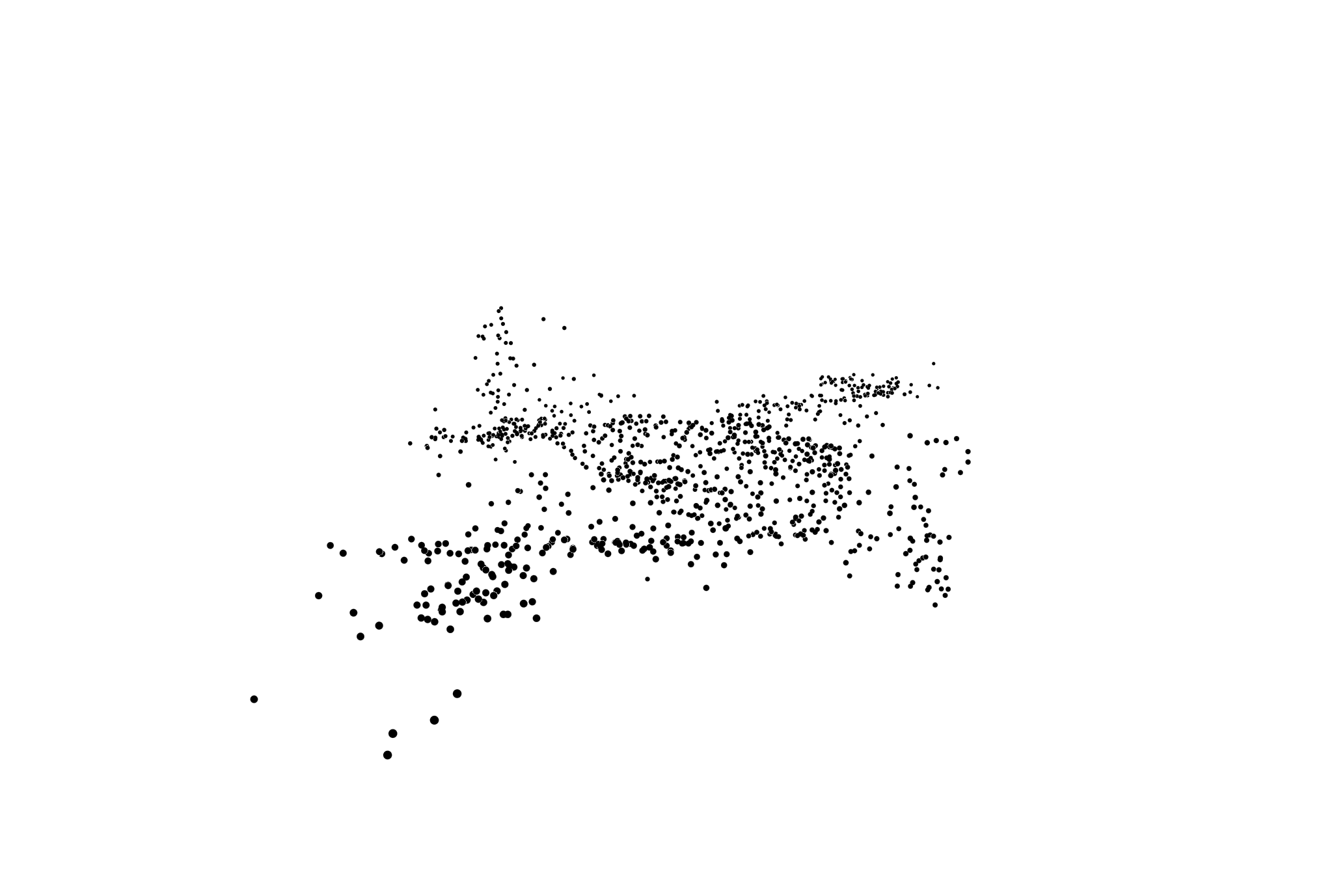} &
\figt[\sizeS]{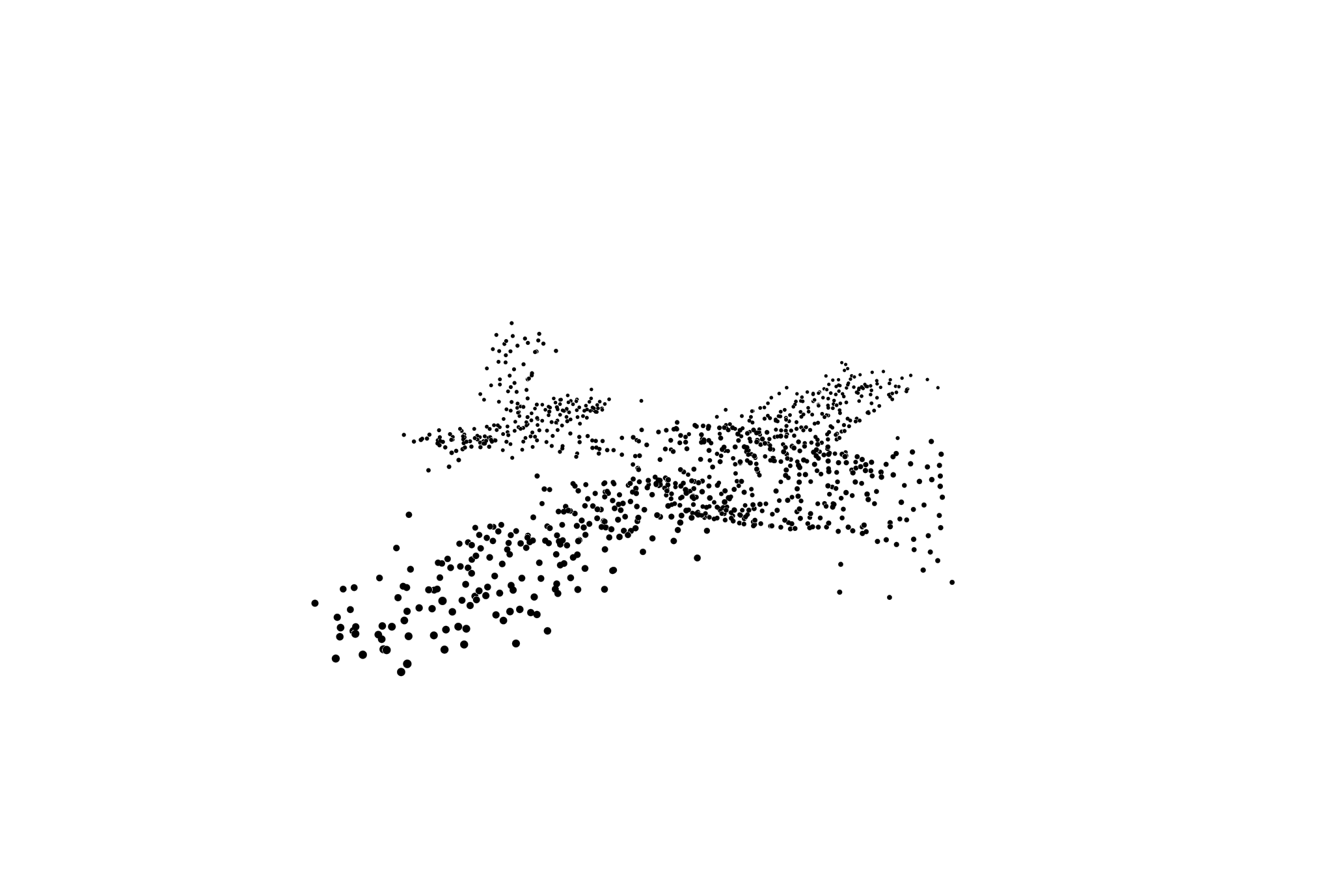} &
\figt[\sizeS]{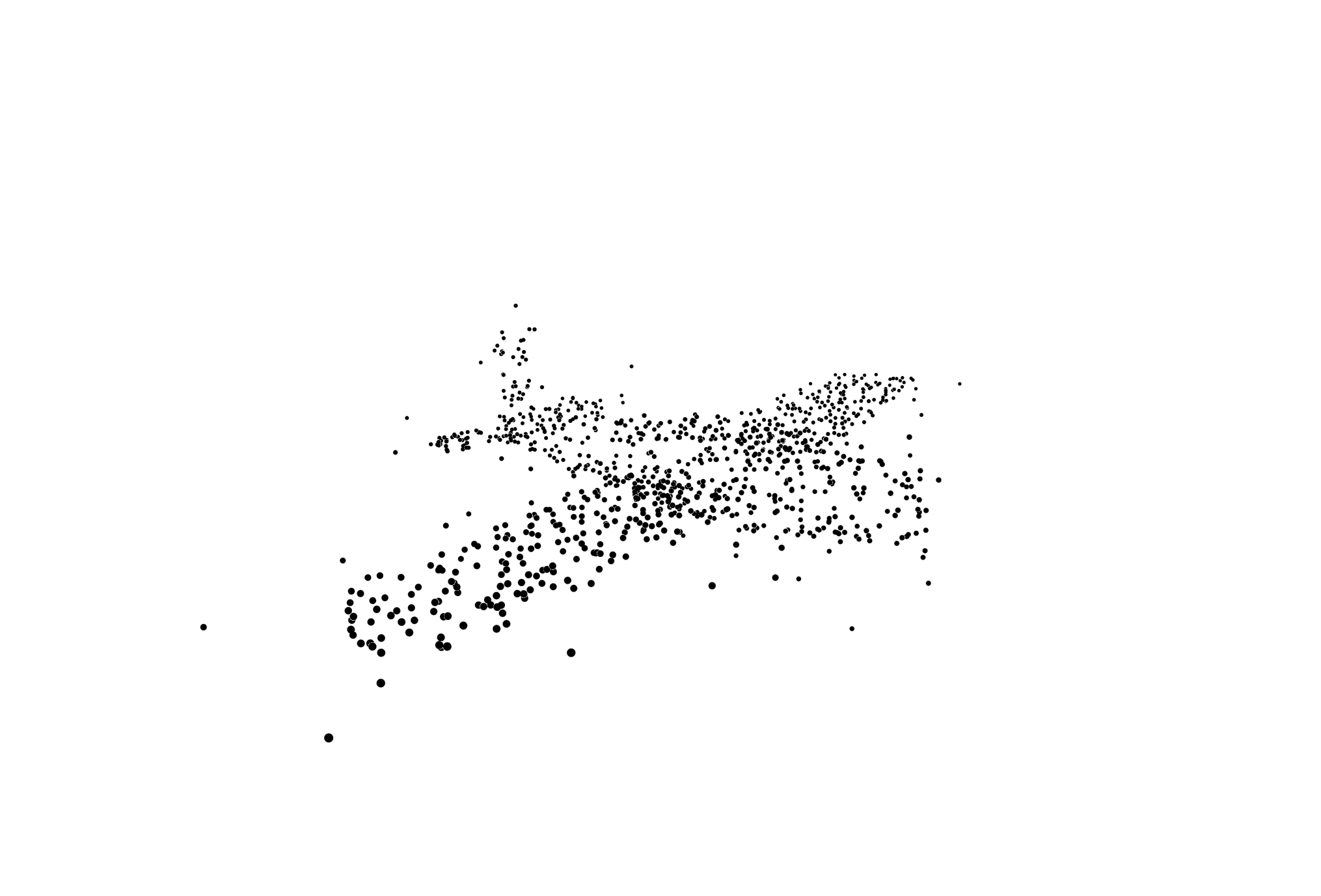} &
\figt[\sizeS]{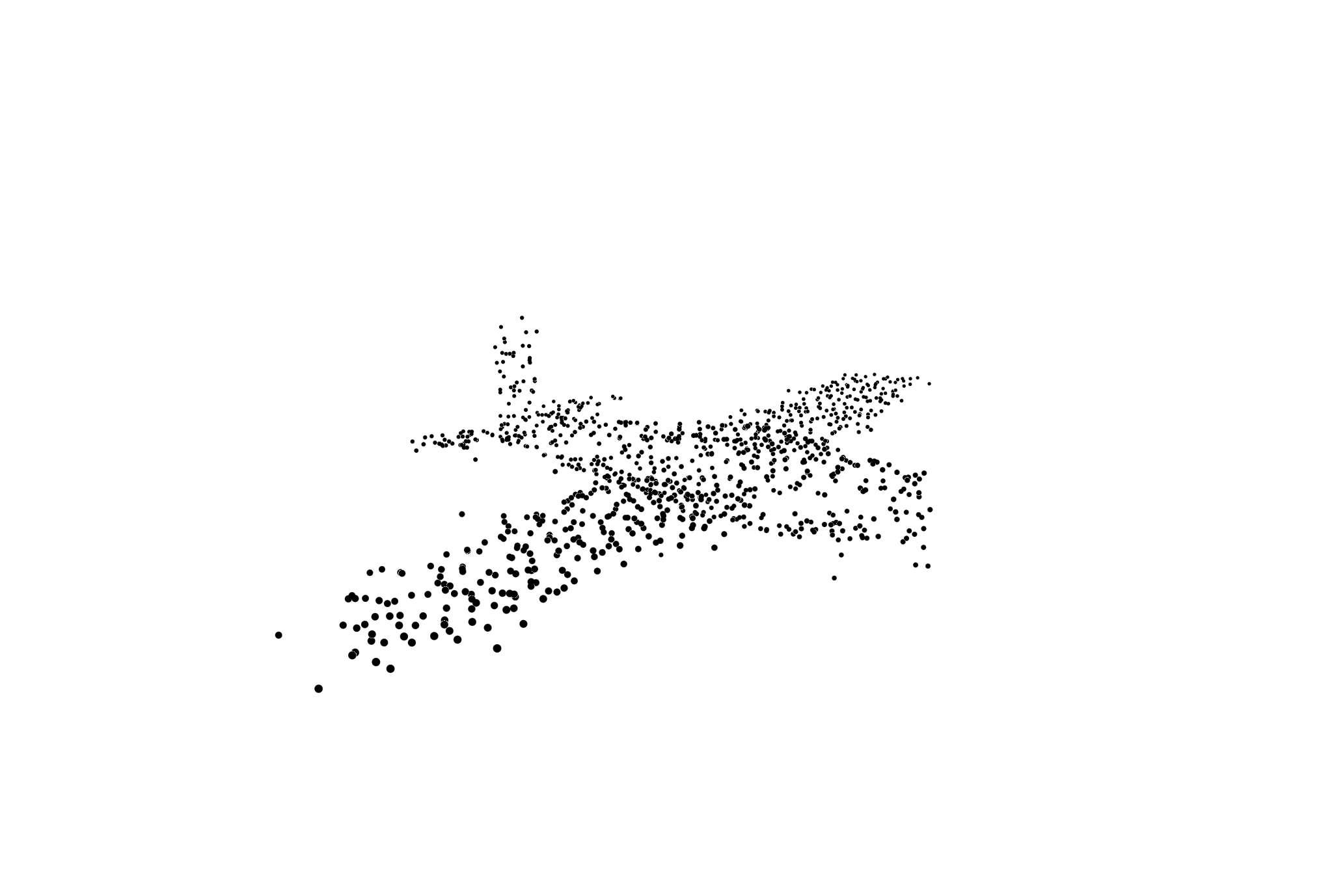} \\
(a) \blue{\cmark airplane} &\red{\xmark bed} &\red{\xmark bed} &\red{\xmark bed} &\red{\xmark bed} \\

\figt[\sizeS]{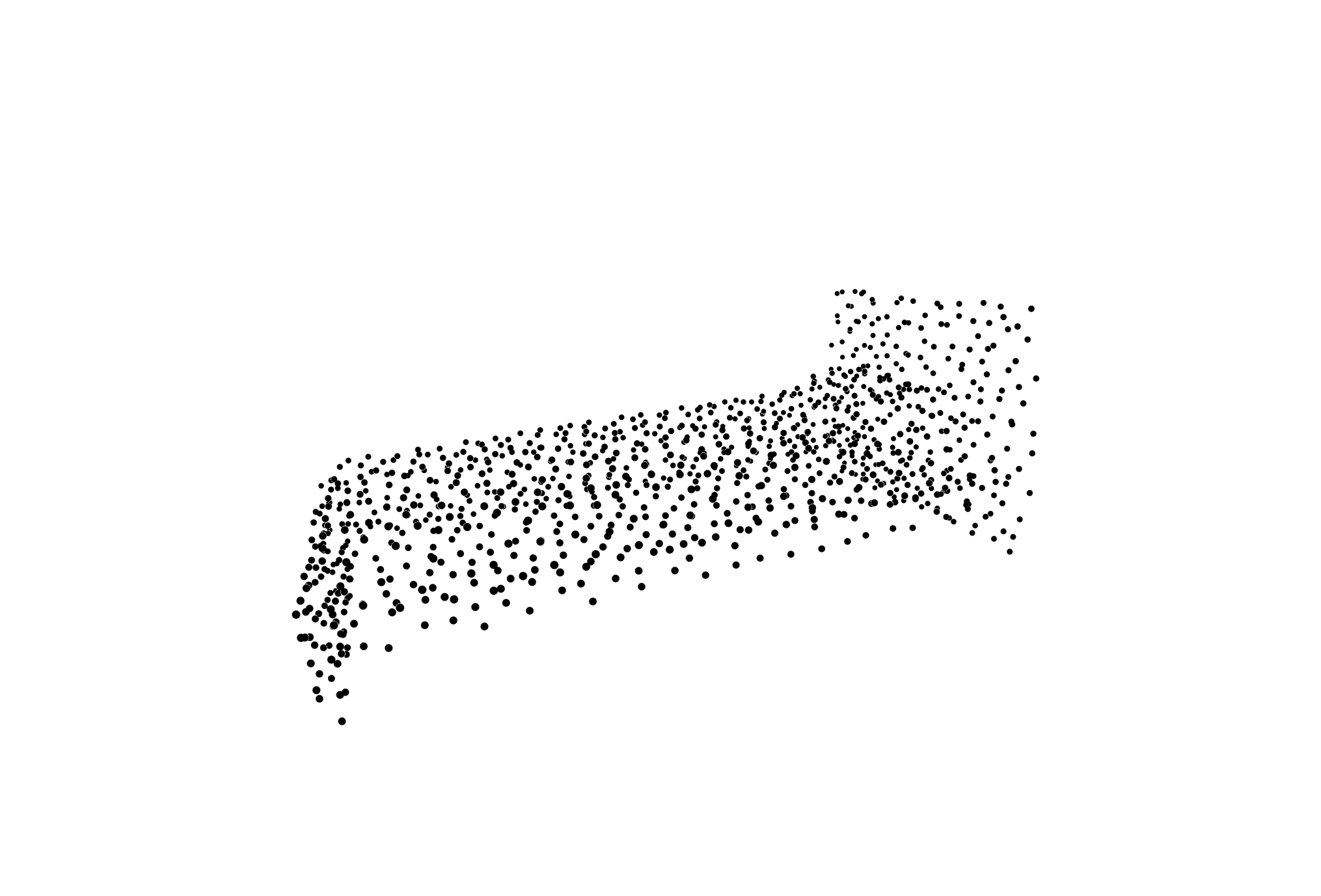} &
\figt[\sizeS]{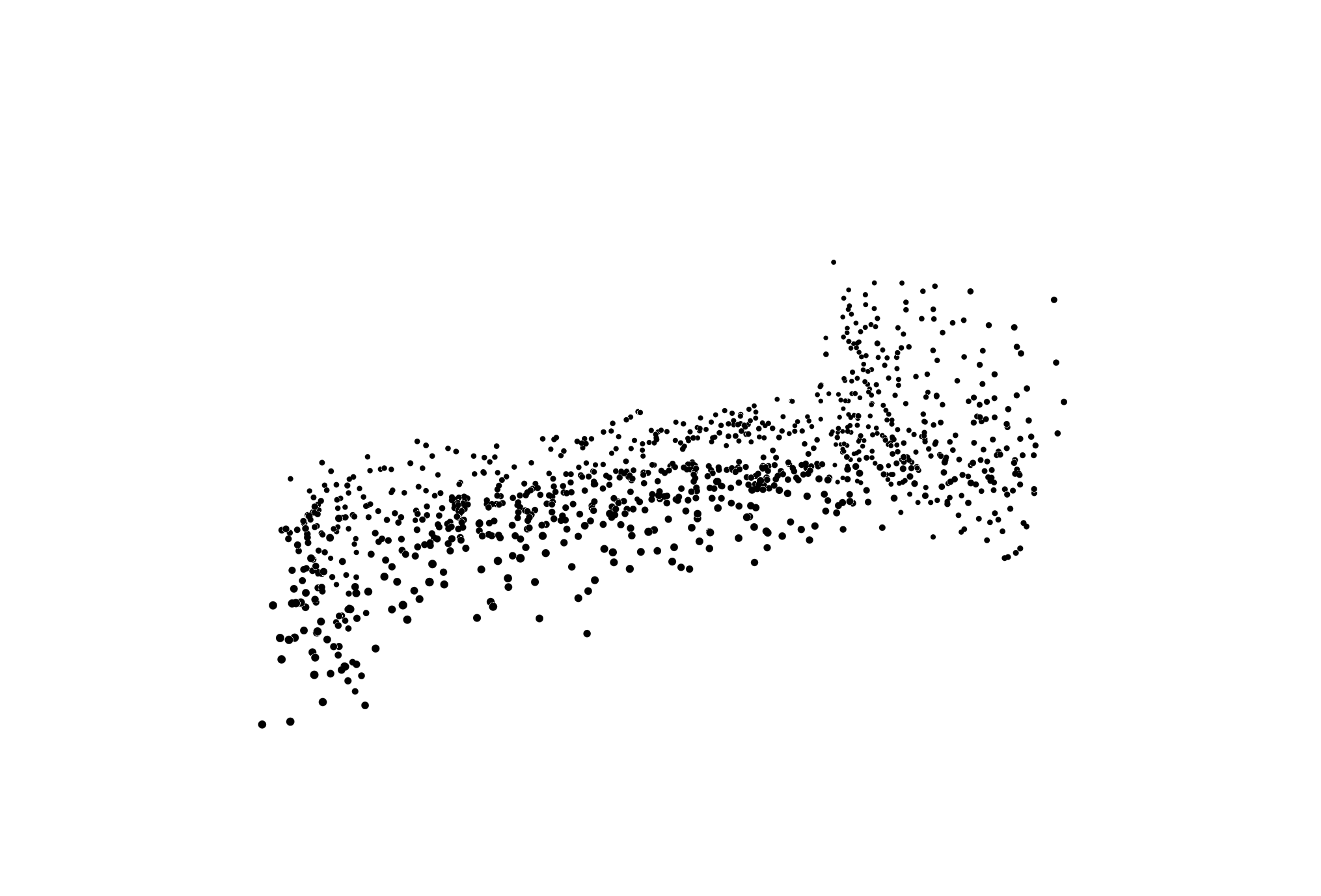} &
\figt[\sizeS]{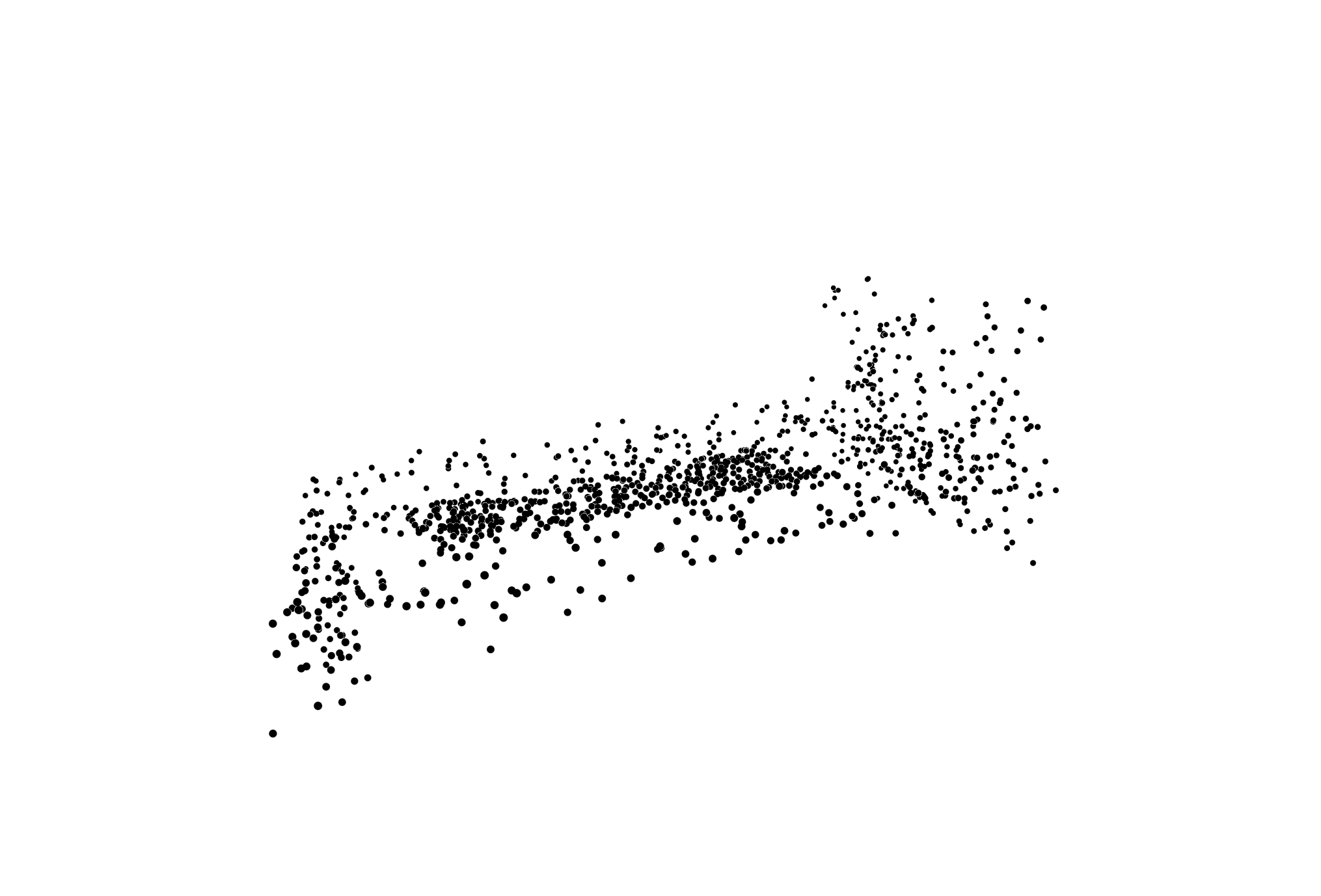} &
\figt[\sizeS]{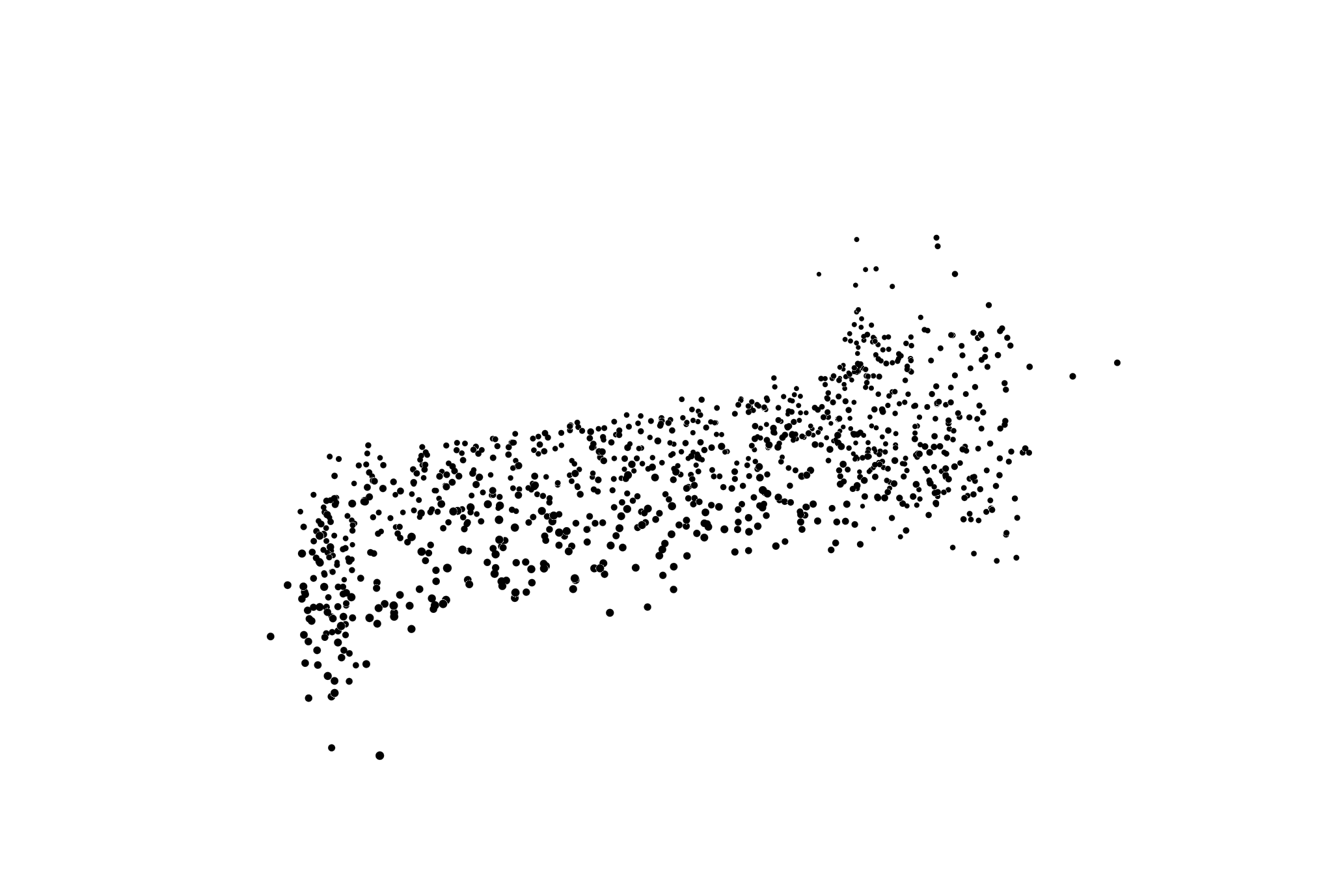} &
\figt[\sizeS]{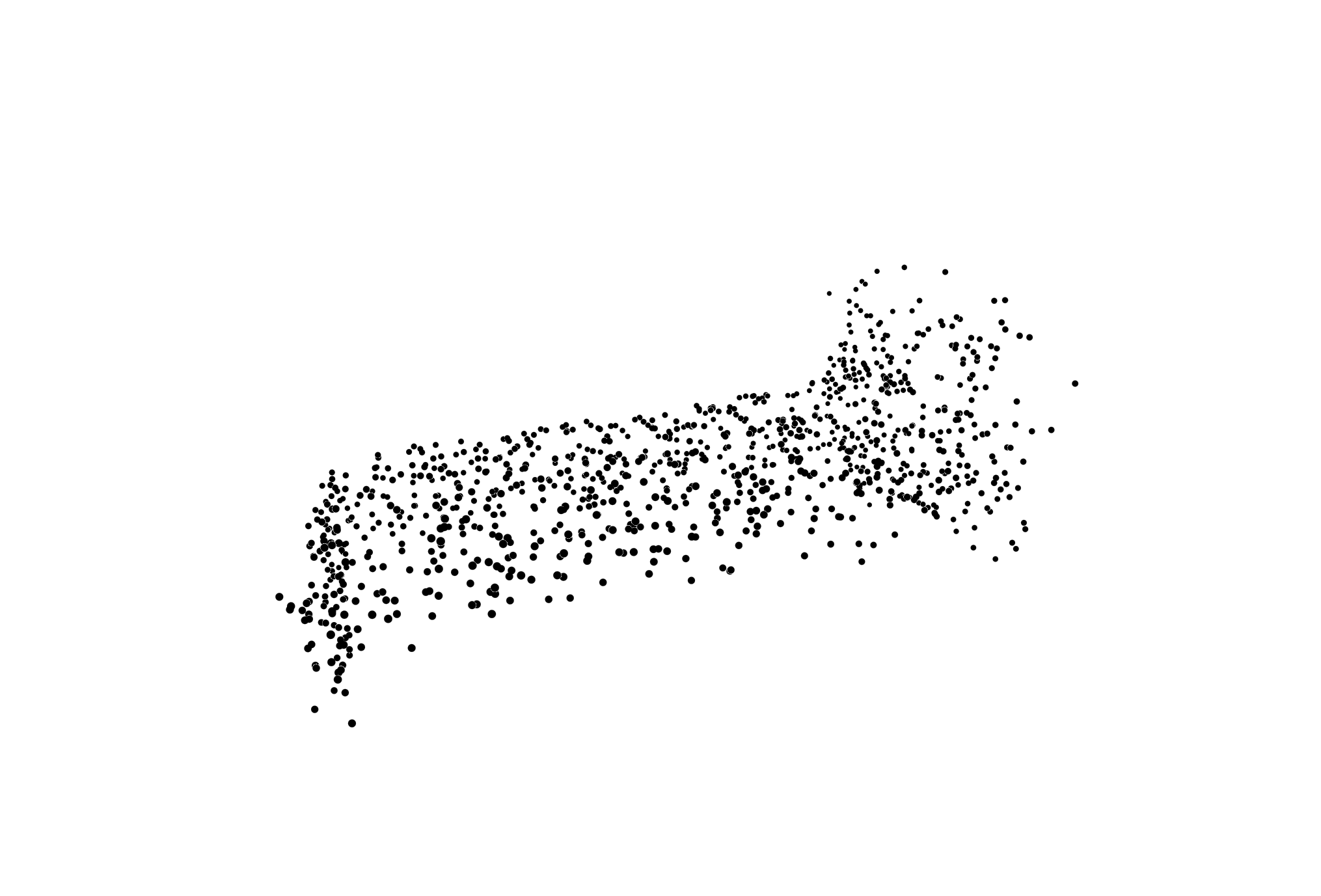} \\
(b) \blue{\cmark bed} &\red{\xmark desk} &\red{\xmark desk} &\red{\xmark desk} &\red{\xmark desk} \\

\figt[\sizeS]{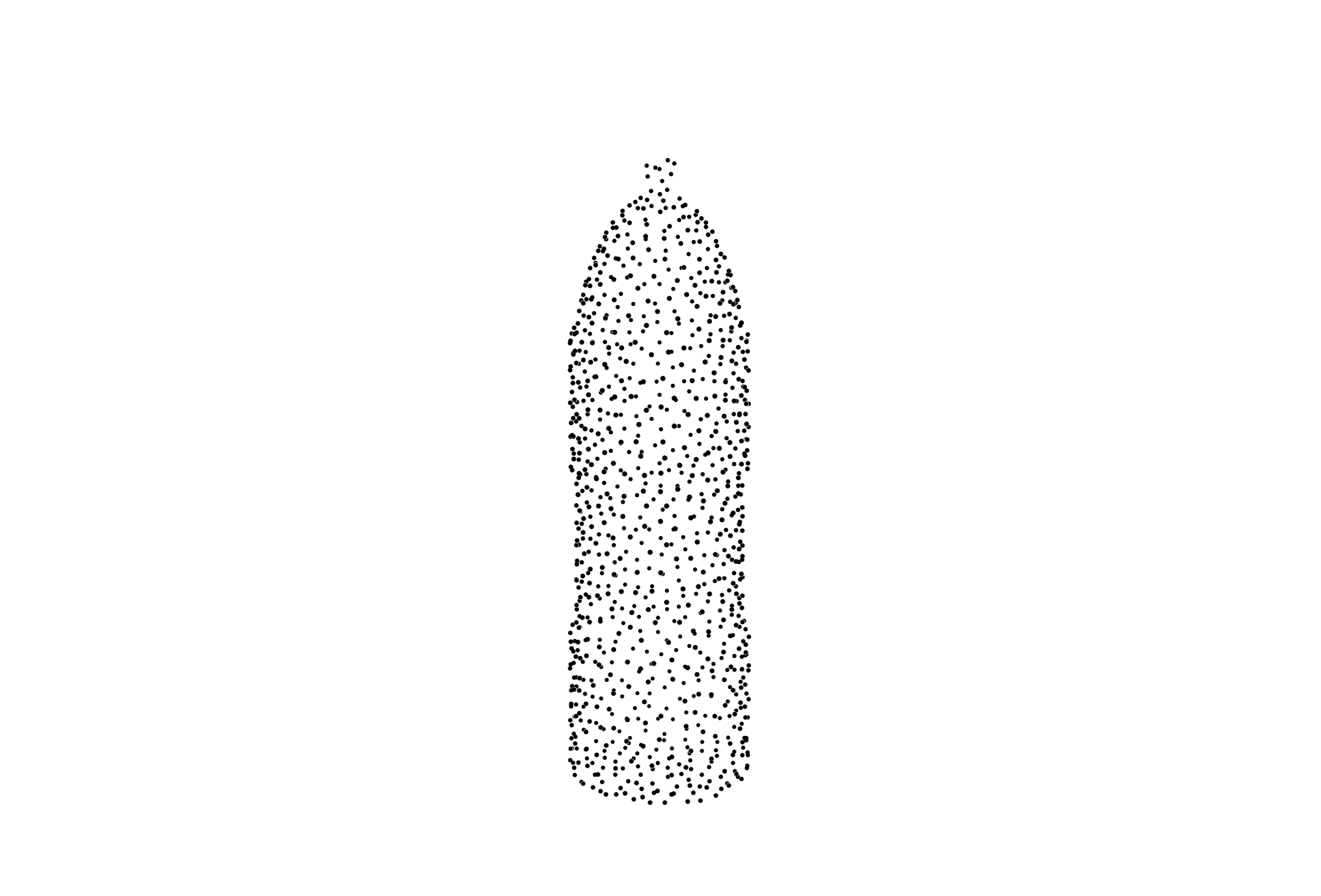} &
\figt[\sizeS]{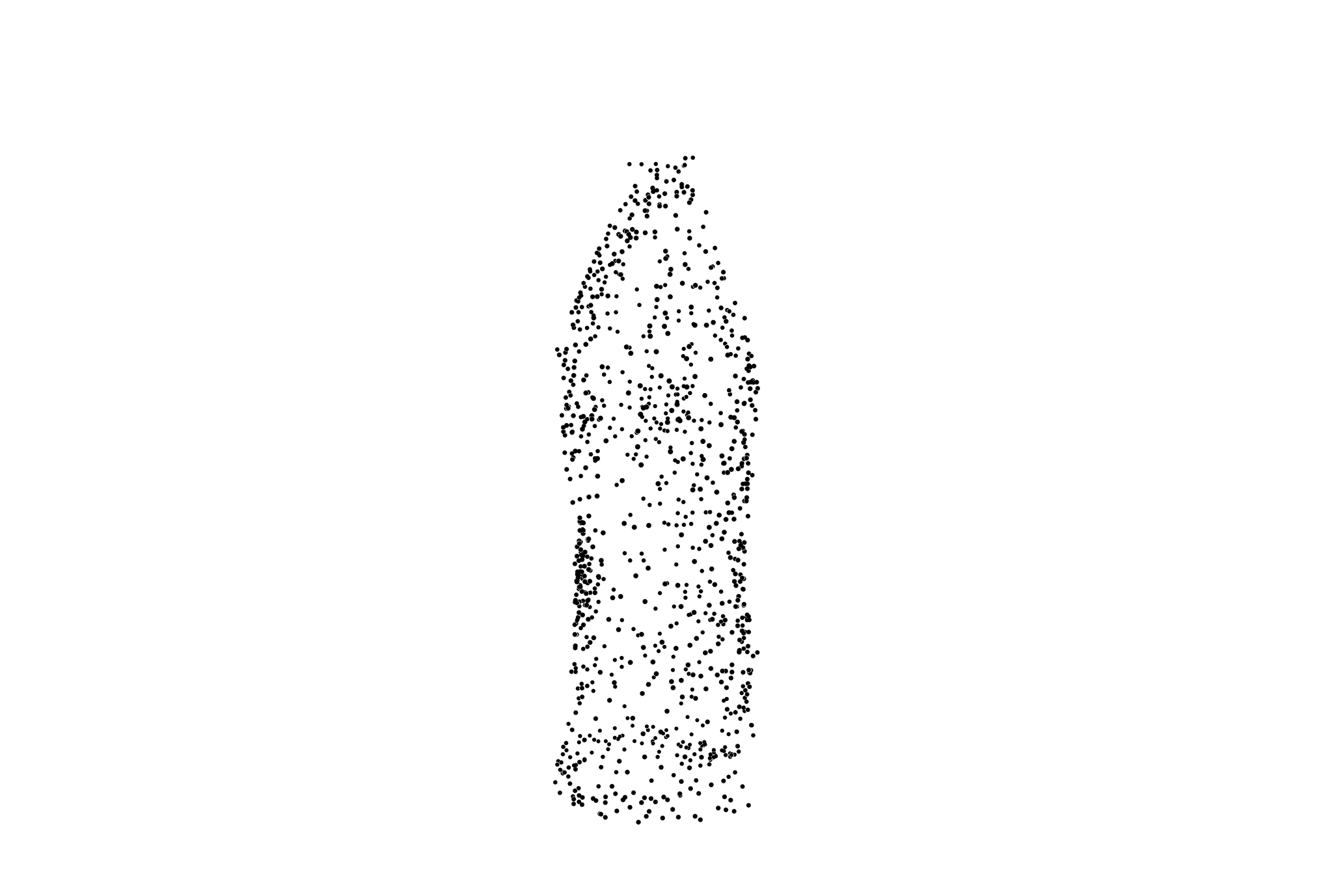} &
\figt[\sizeS]{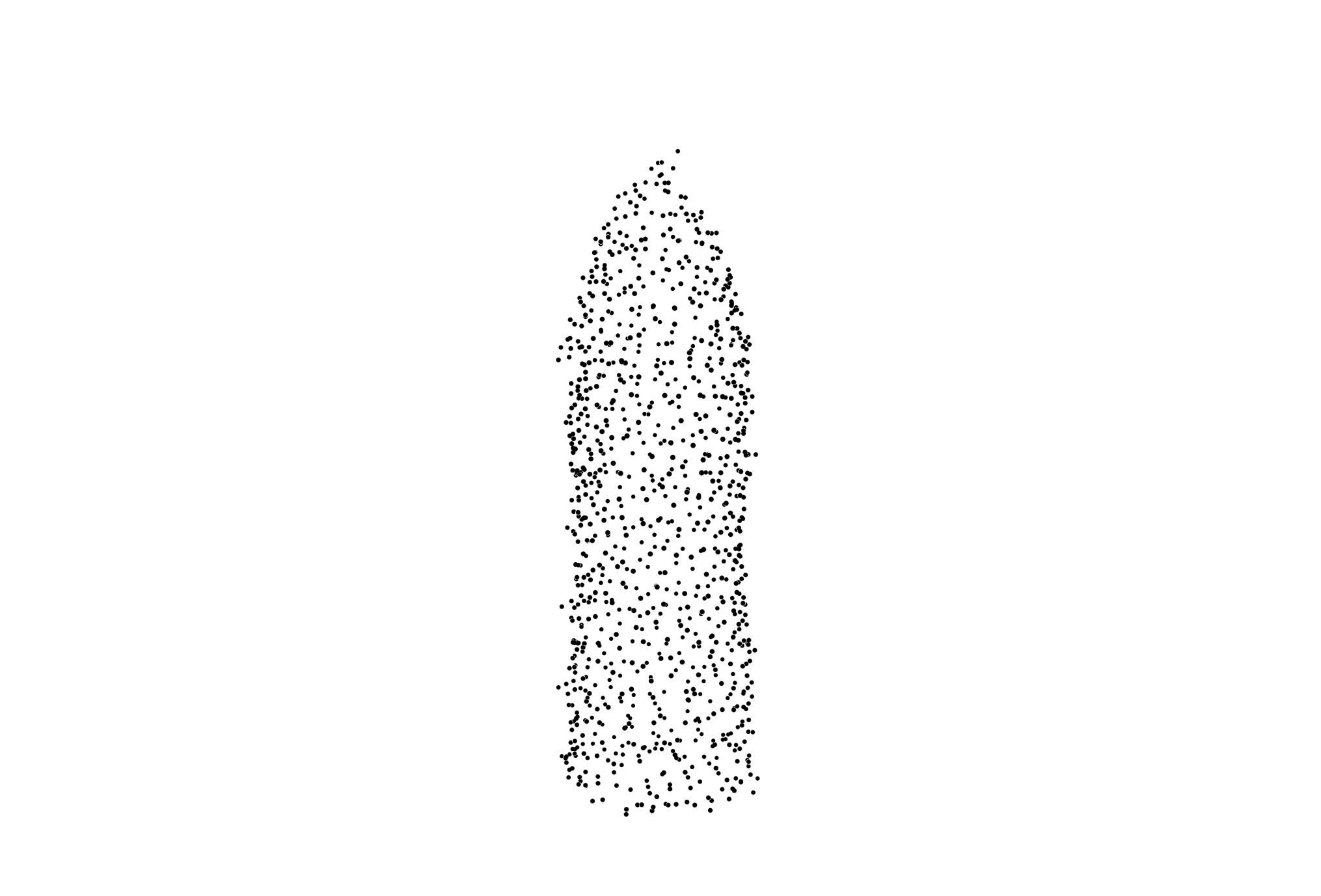} &
\figt[\sizeS]{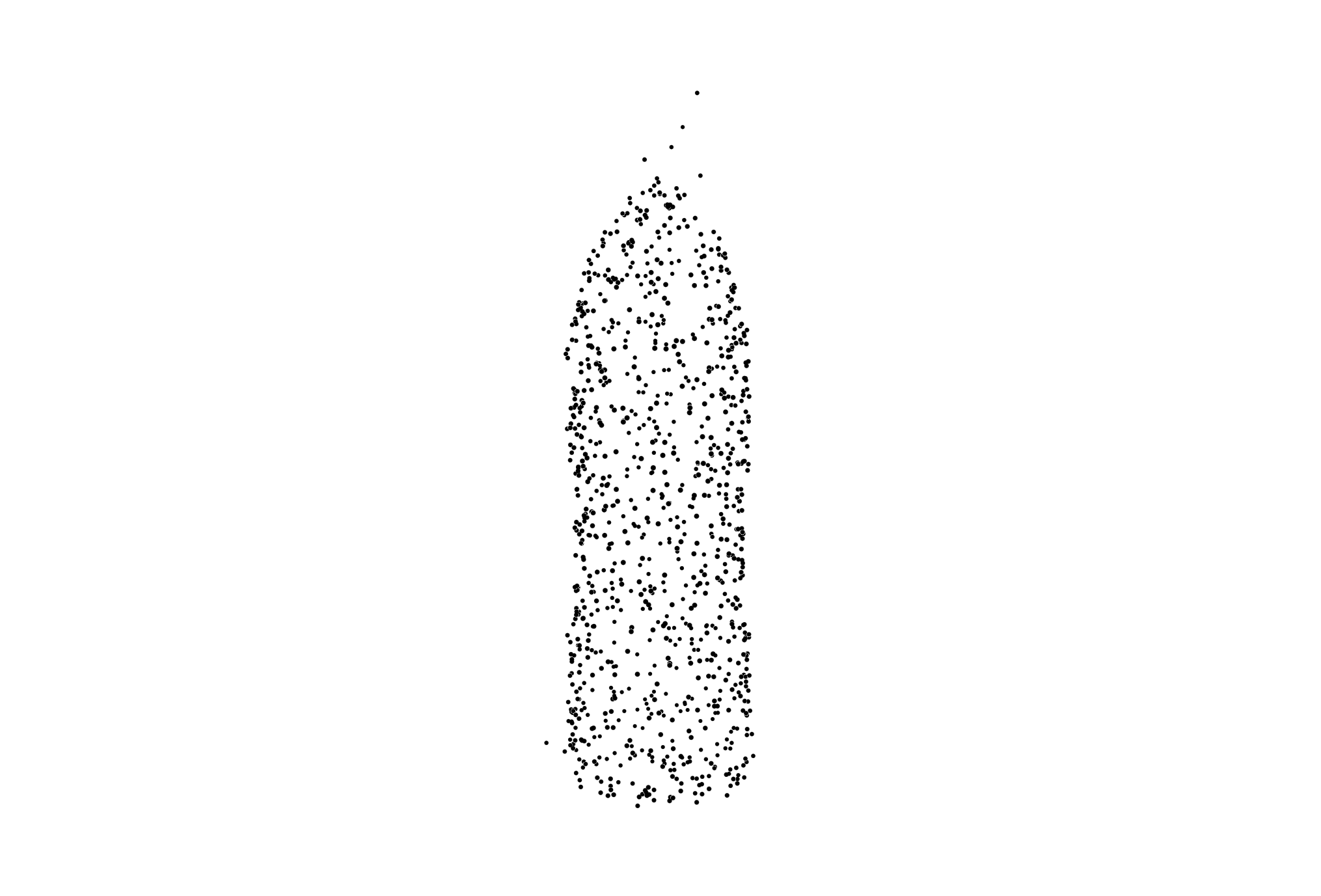} &
\figt[\sizeS]{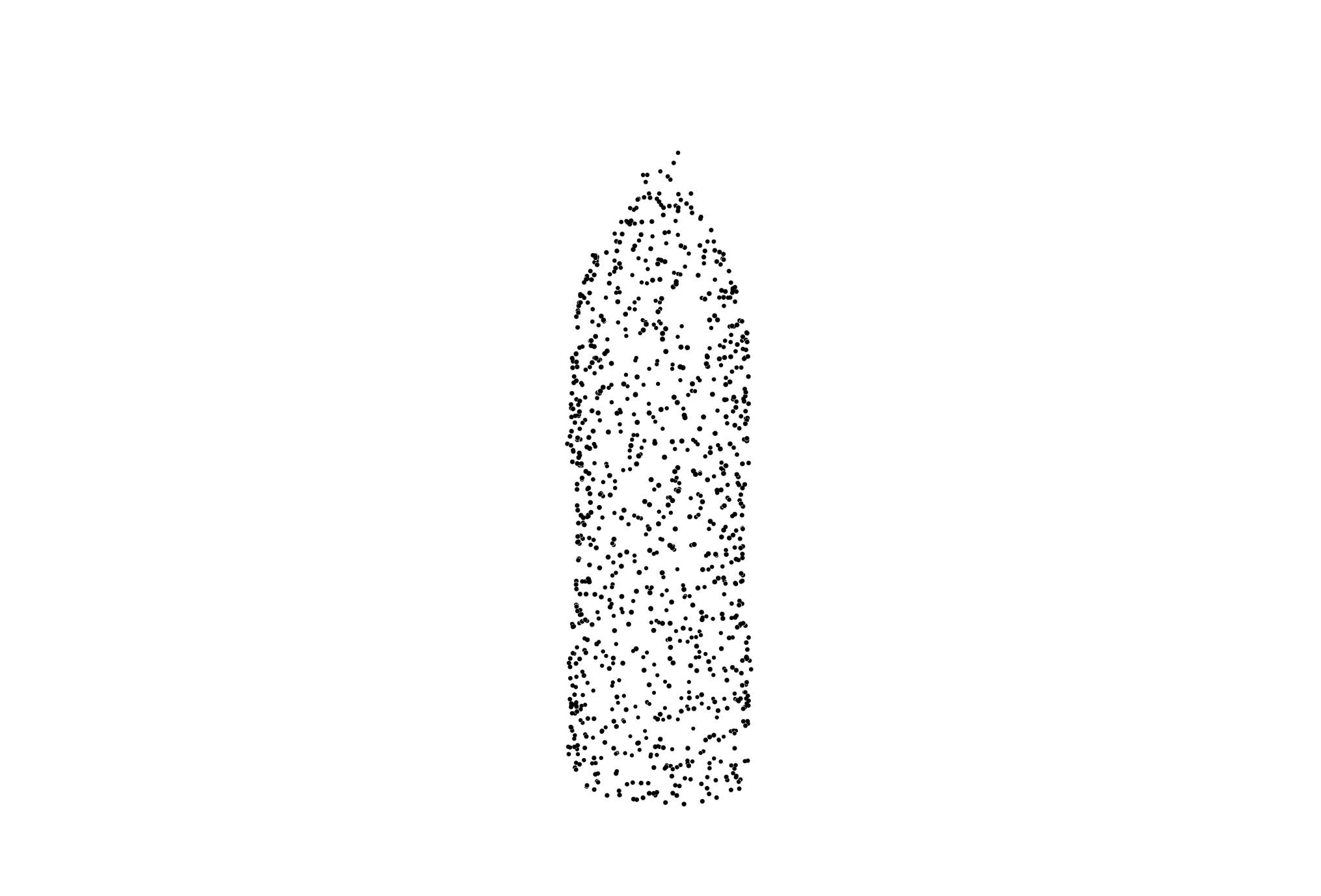} \\
(c) \blue{\cmark bottle} &\red{\xmark flower pot} &\red{\xmark flower pot} &\red{\xmark cup} &\red{\xmark flower pot} \\

\figt[\sizeS]{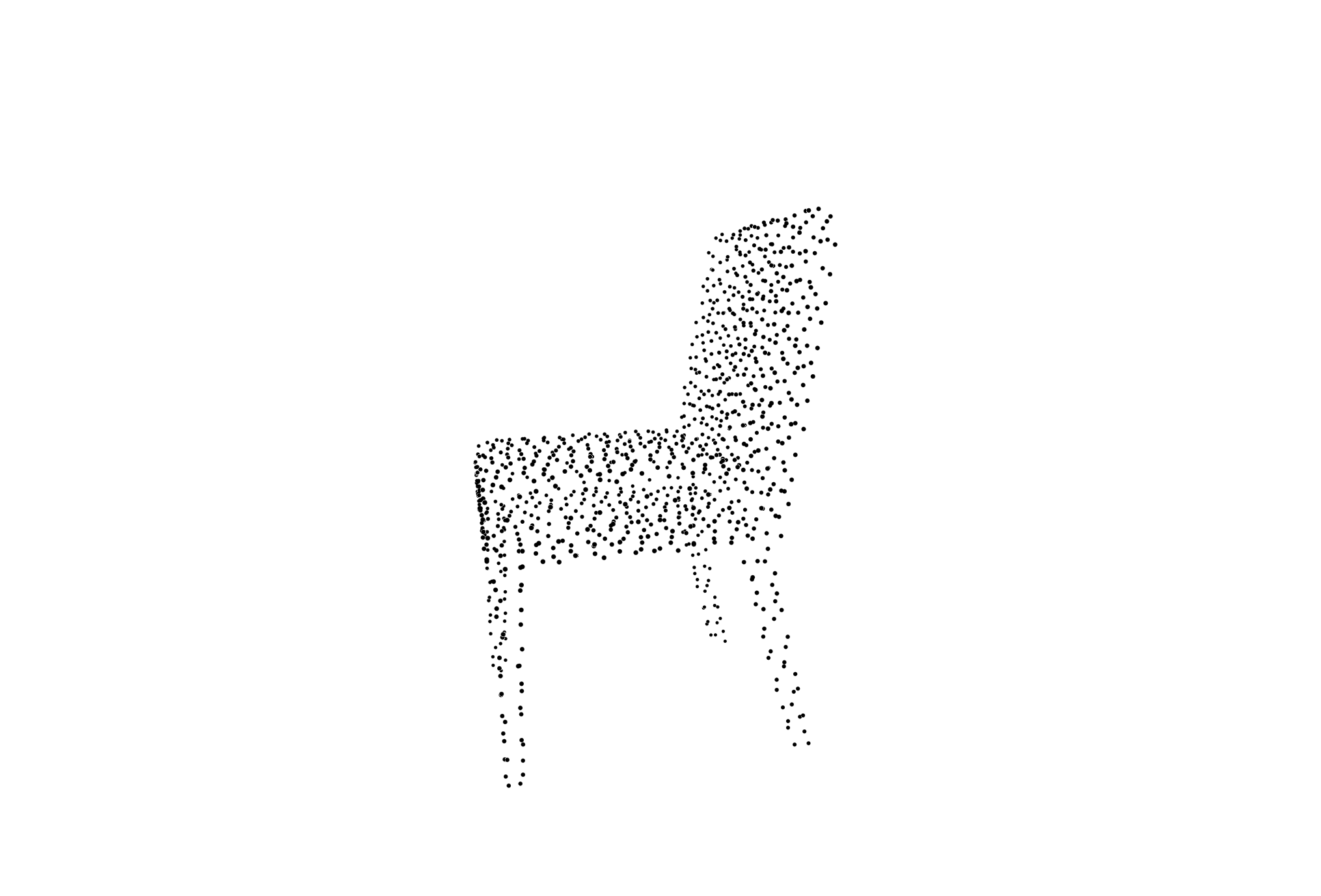} &
\figt[\sizeS]{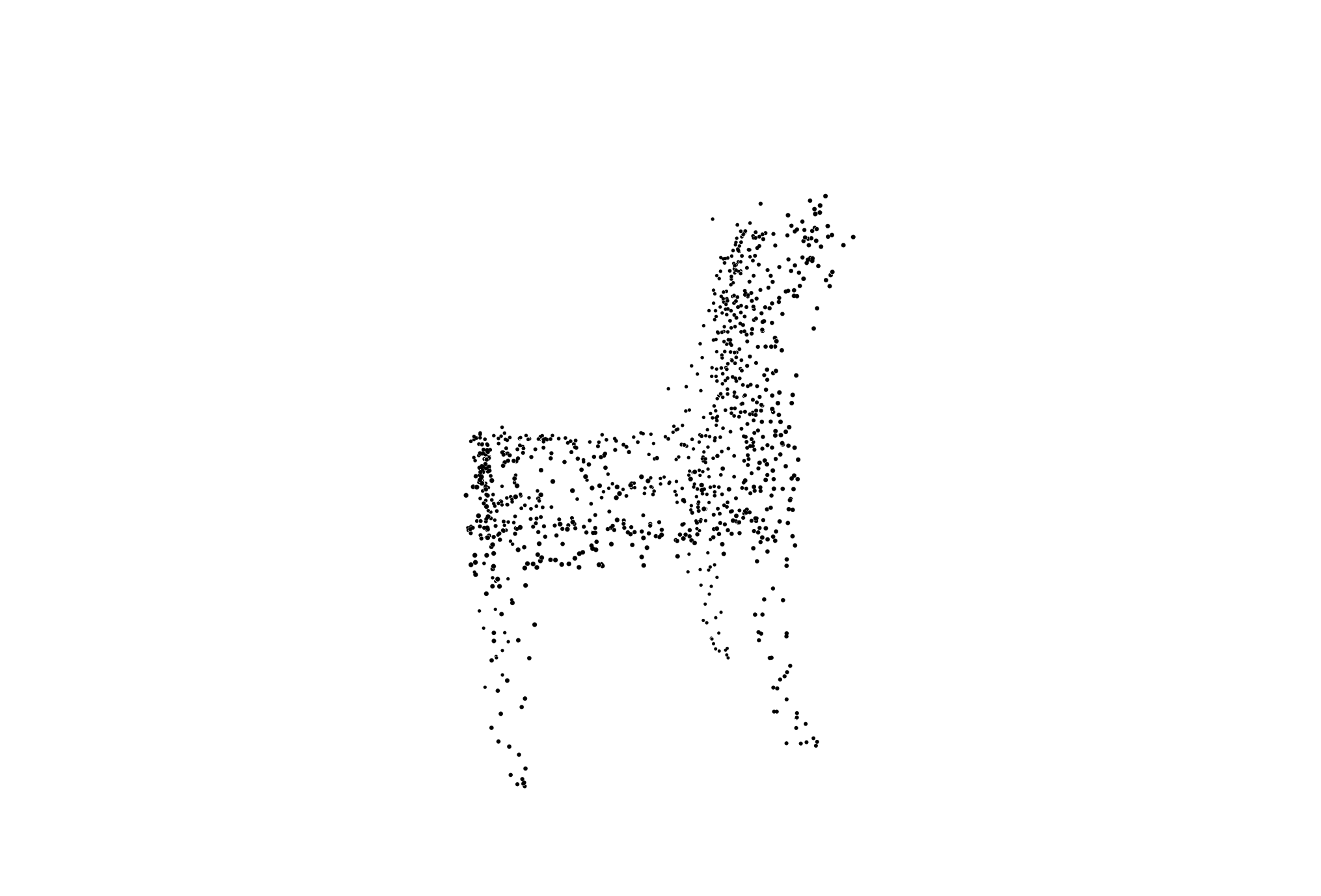} &
\figt[\sizeS]{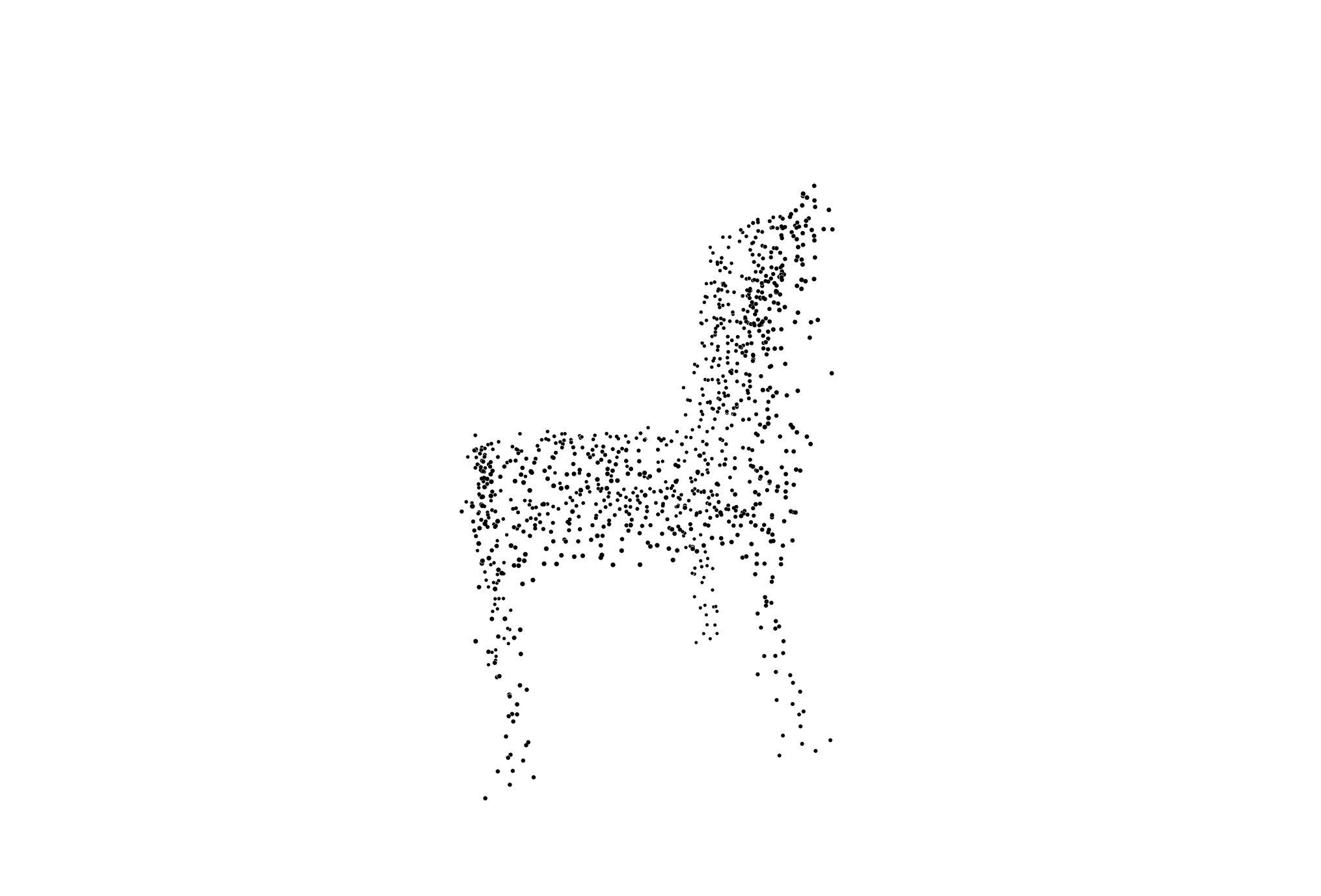} &
\figt[\sizeS]{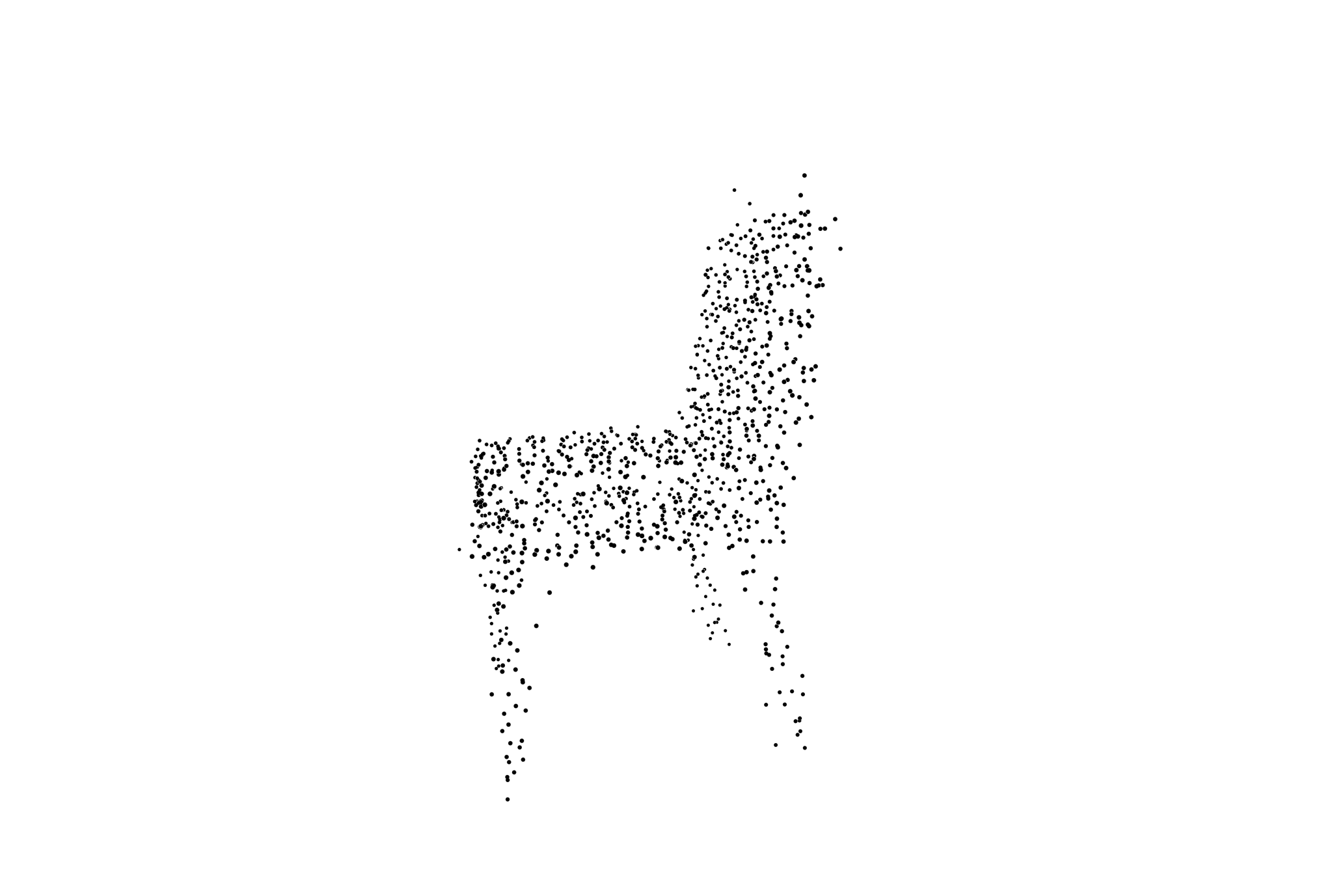} &
\figt[\sizeS]{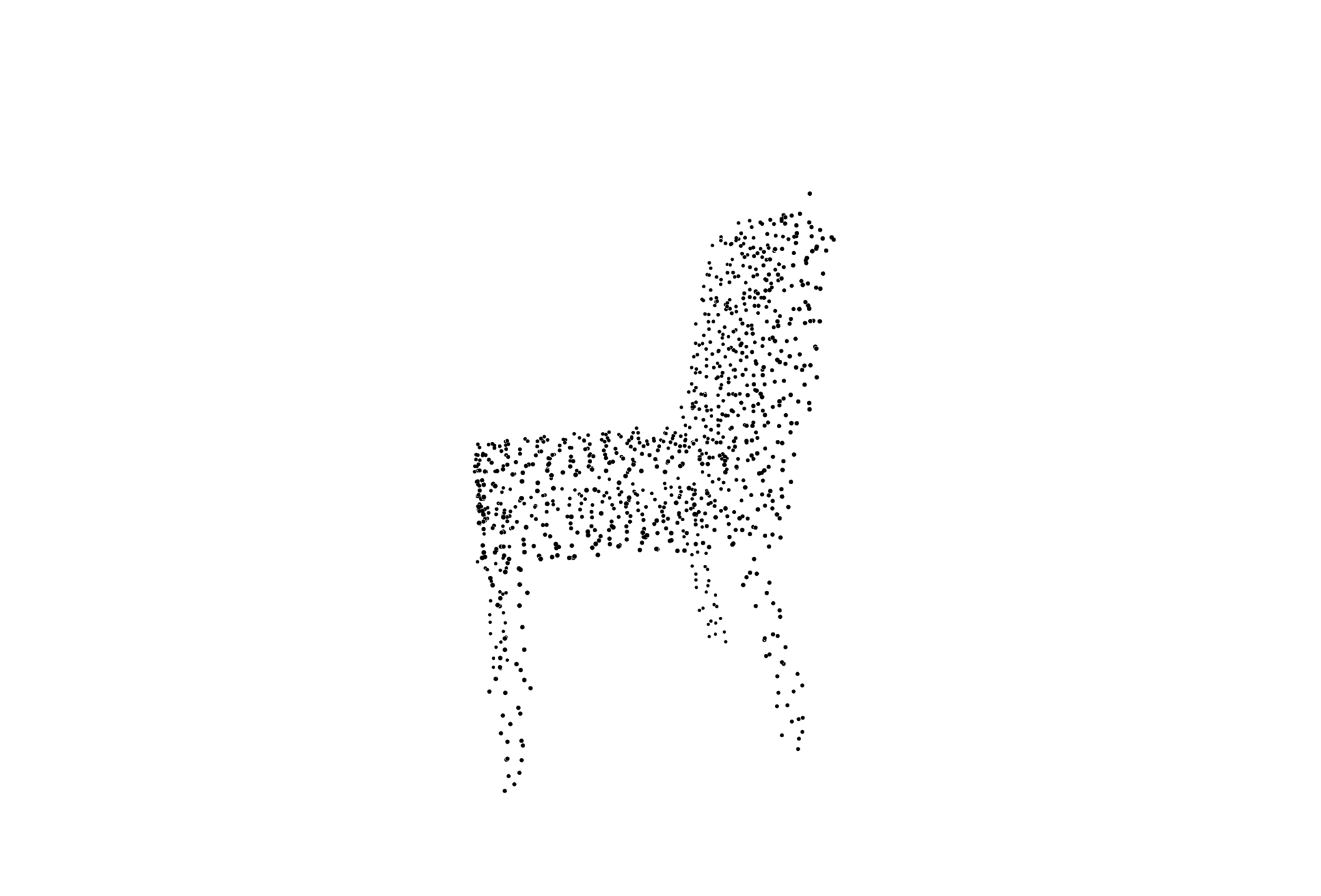} \\
(d) \blue{\cmark chair} &\red{\xmark bed} &\red{\xmark bowl} &\red{\xmark bed} &\red{\xmark bed} \\

\figt[\sizeS]{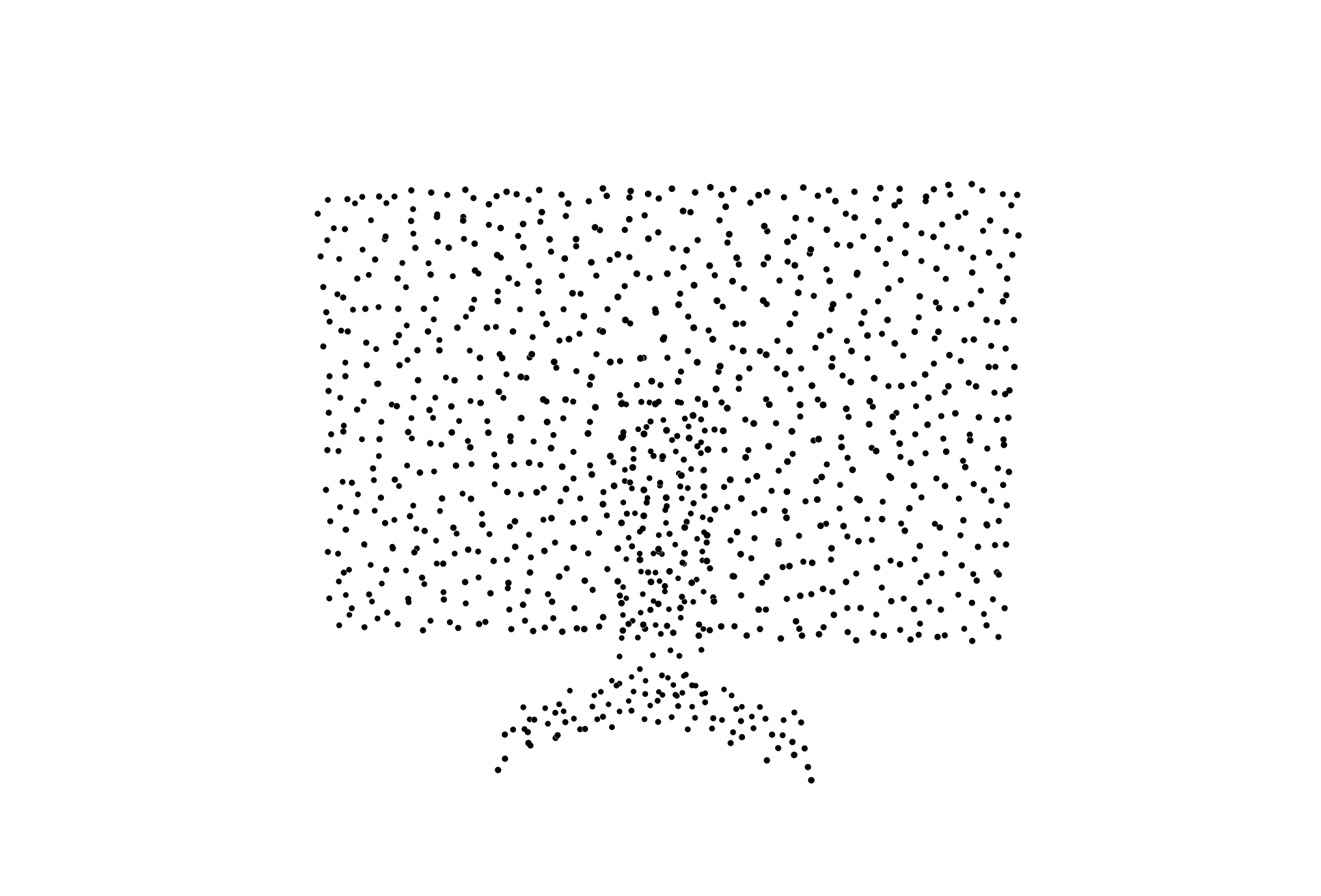} &
\figt[\sizeS]{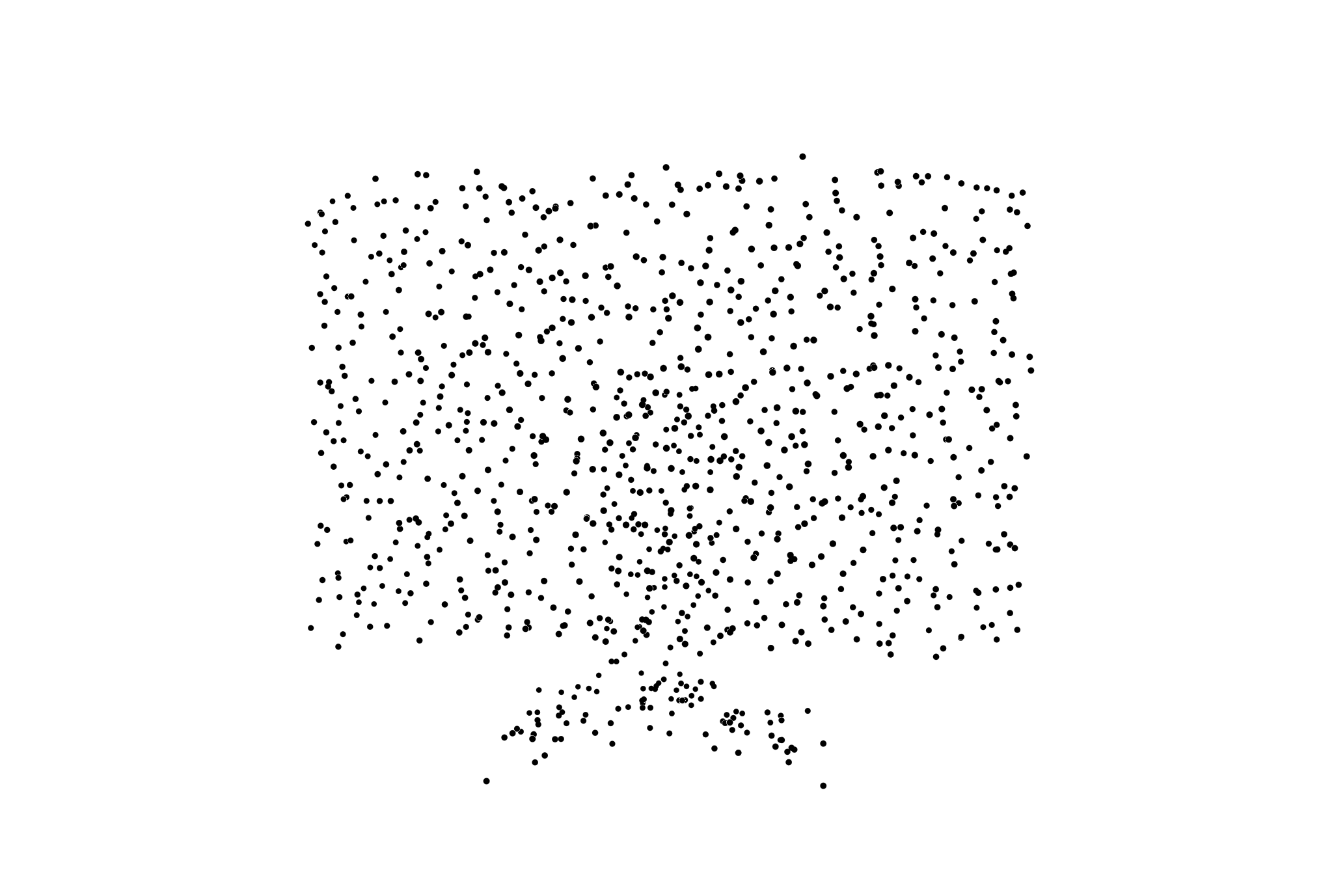} &
\figt[\sizeS]{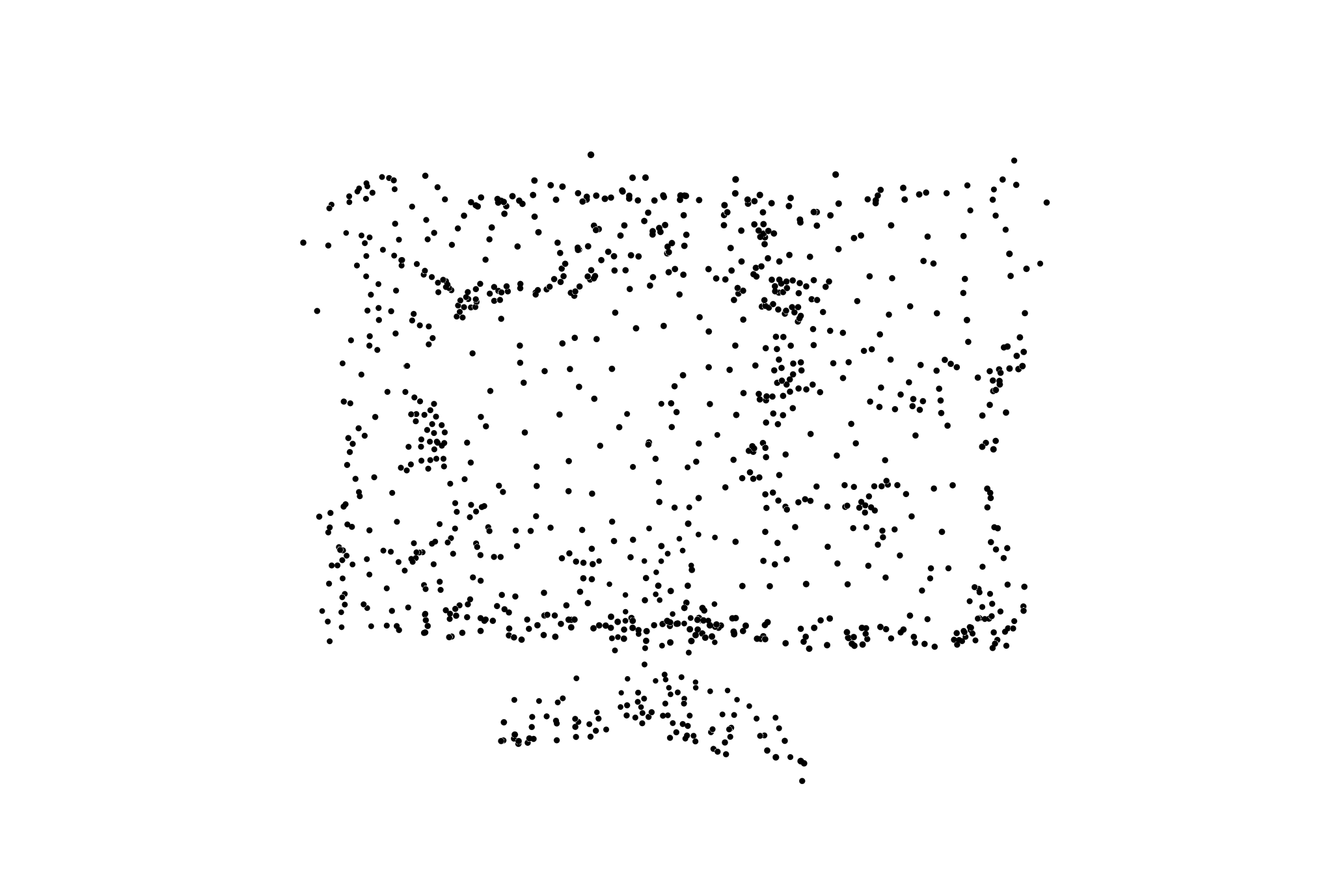} &
\figt[\sizeS]{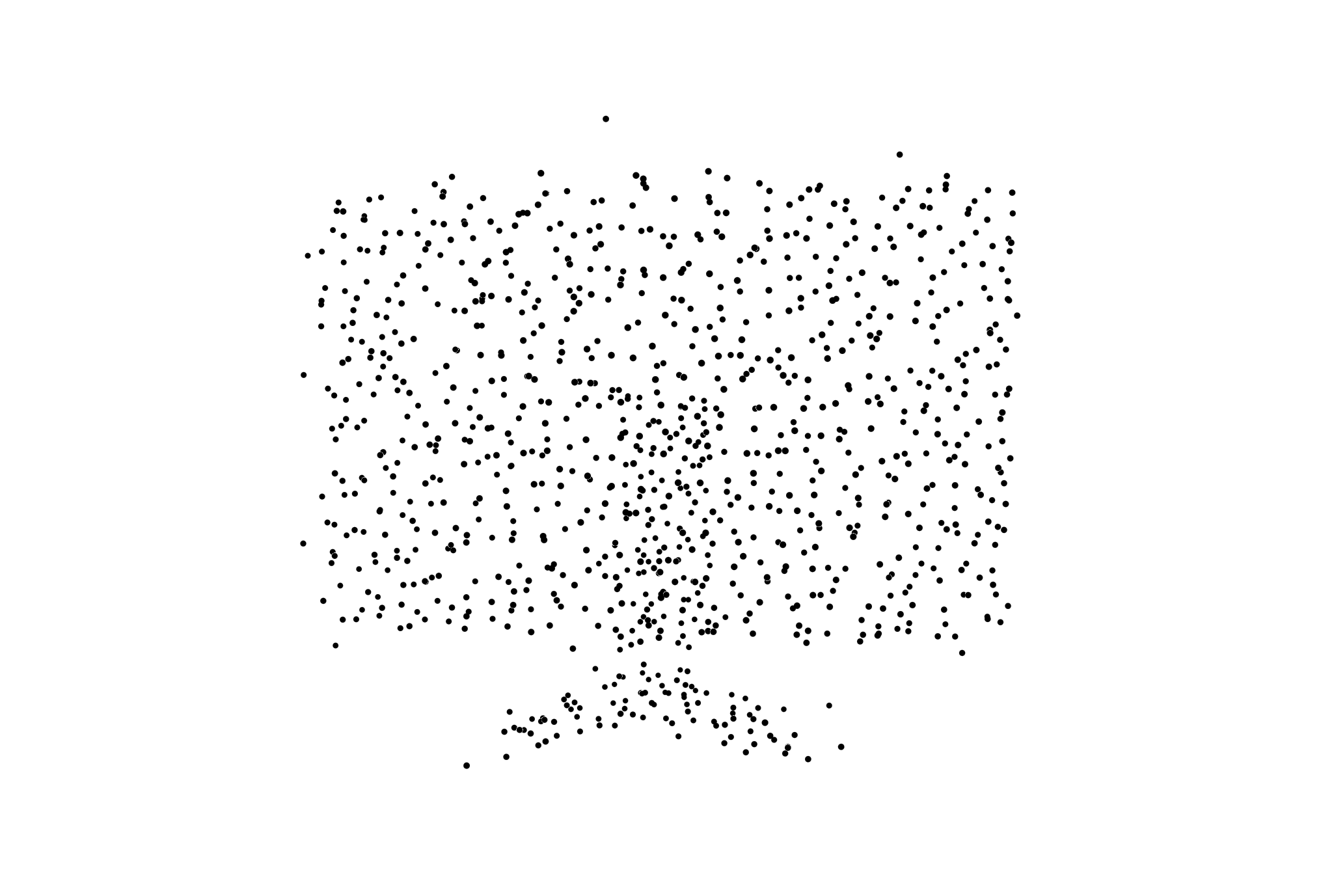} &
\figt[\sizeS]{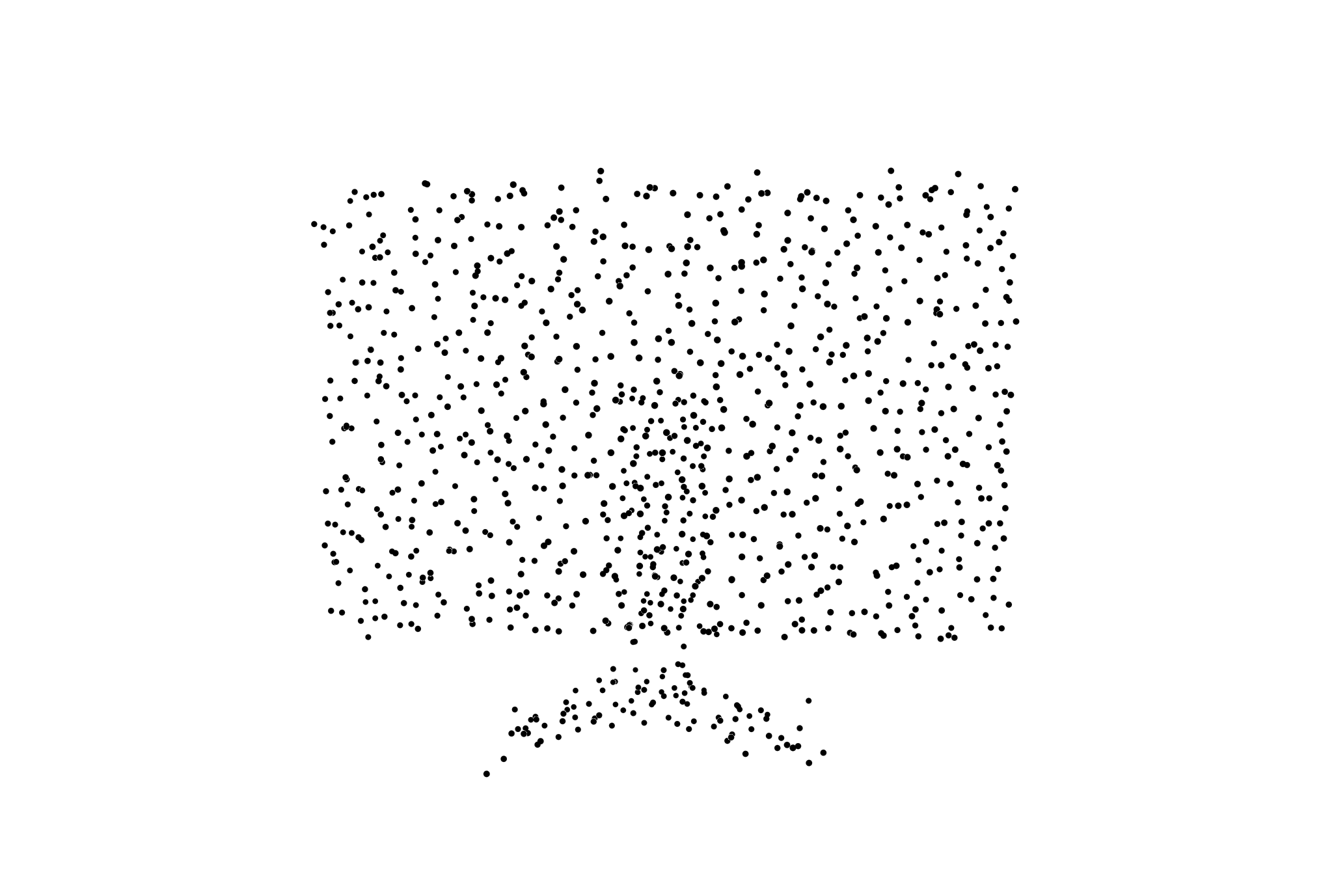} \\
(e) \blue{\cmark monitor} &\red{\xmark table} &\red{\xmark bed} &\red{\xmark bed} &\red{\xmark bottle} \\

\end{tabular}
\caption{Visualization for adversarial point clouds produced by baseline methods on PointNet++.}
\label{fig:vis-3d-pointnet2-supp}
\end{figure*}

\begin{figure*}[hptb]
\newcommand{\sizeS}{.20}
\centering
\setlength{\tabcolsep}{0pt}
\begin{tabular}{ccccc}
\textbf{Clean} &\geoa &\gsda &\siadv & \projabbrv\\
\figt[\sizeS]{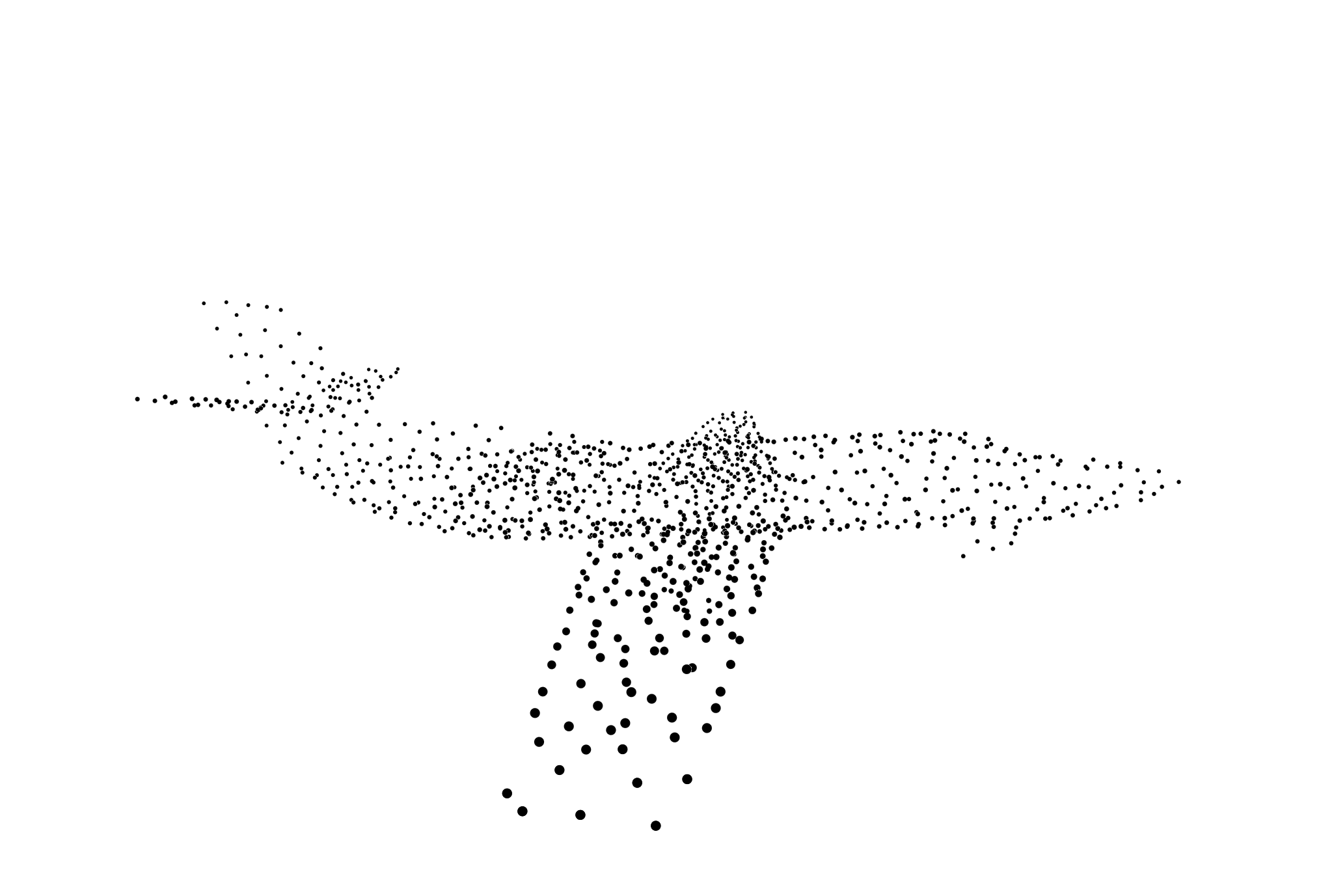} &
\figt[\sizeS]{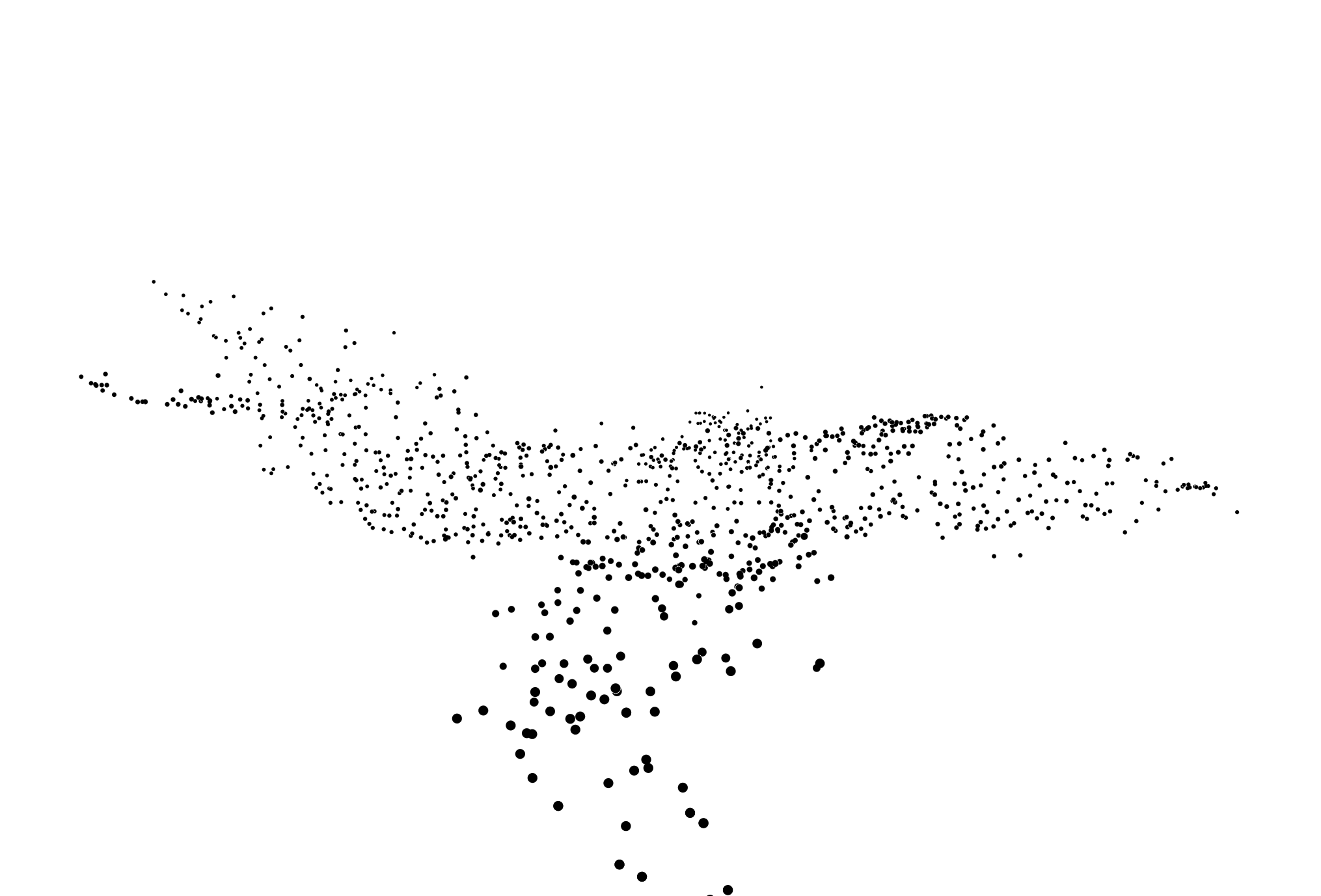} &
\figt[\sizeS]{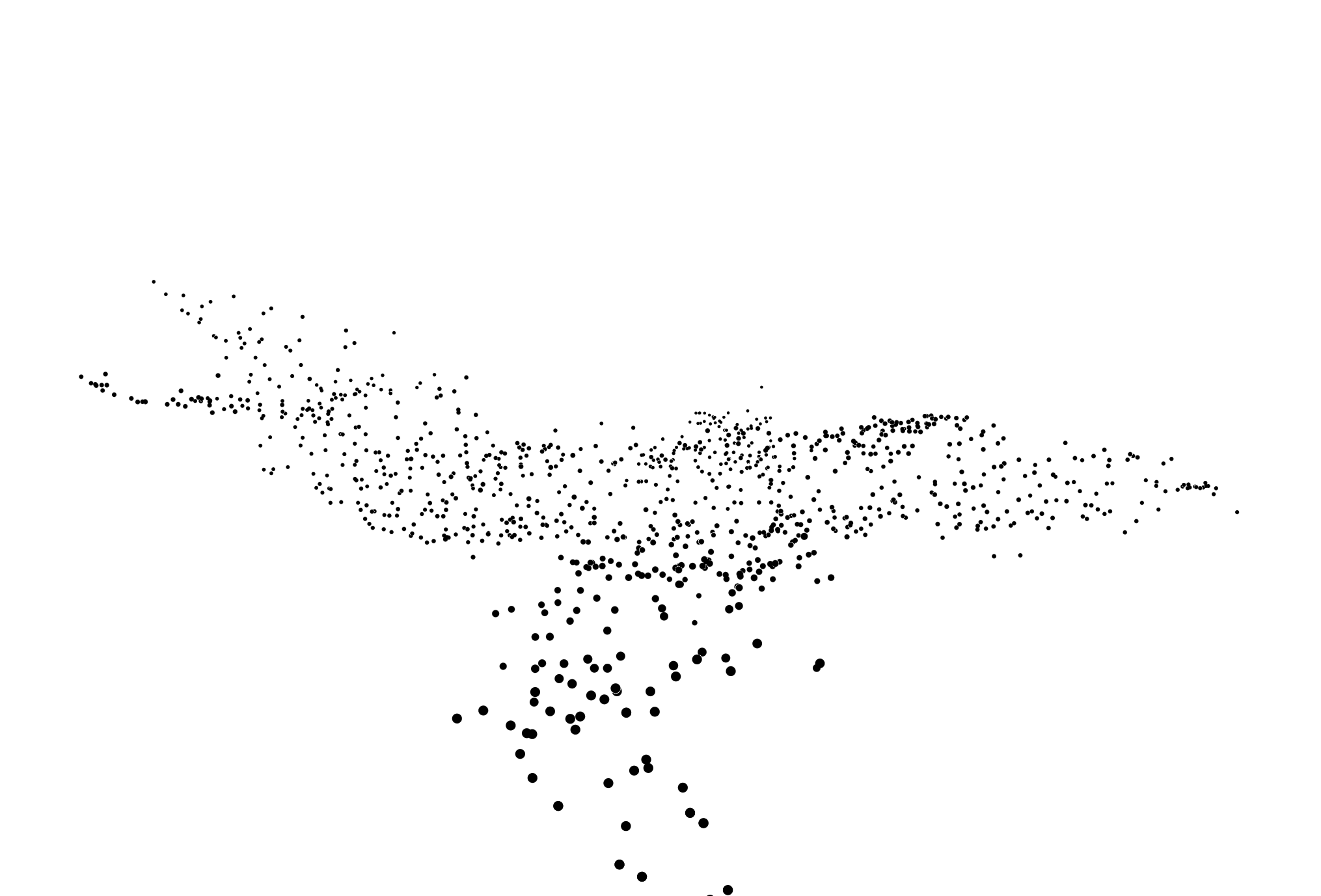} &
\figt[\sizeS]{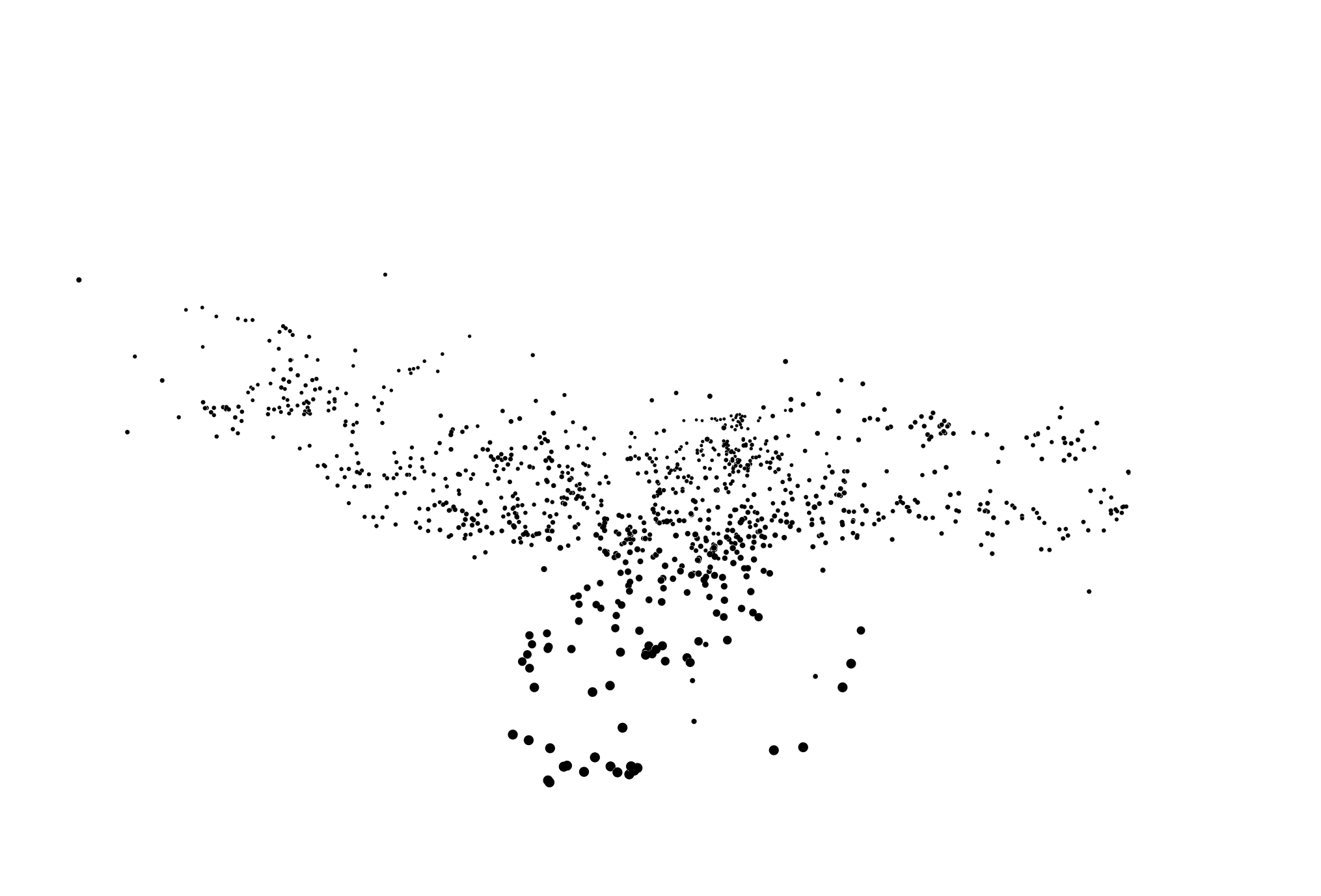} &
\figt[\sizeS]{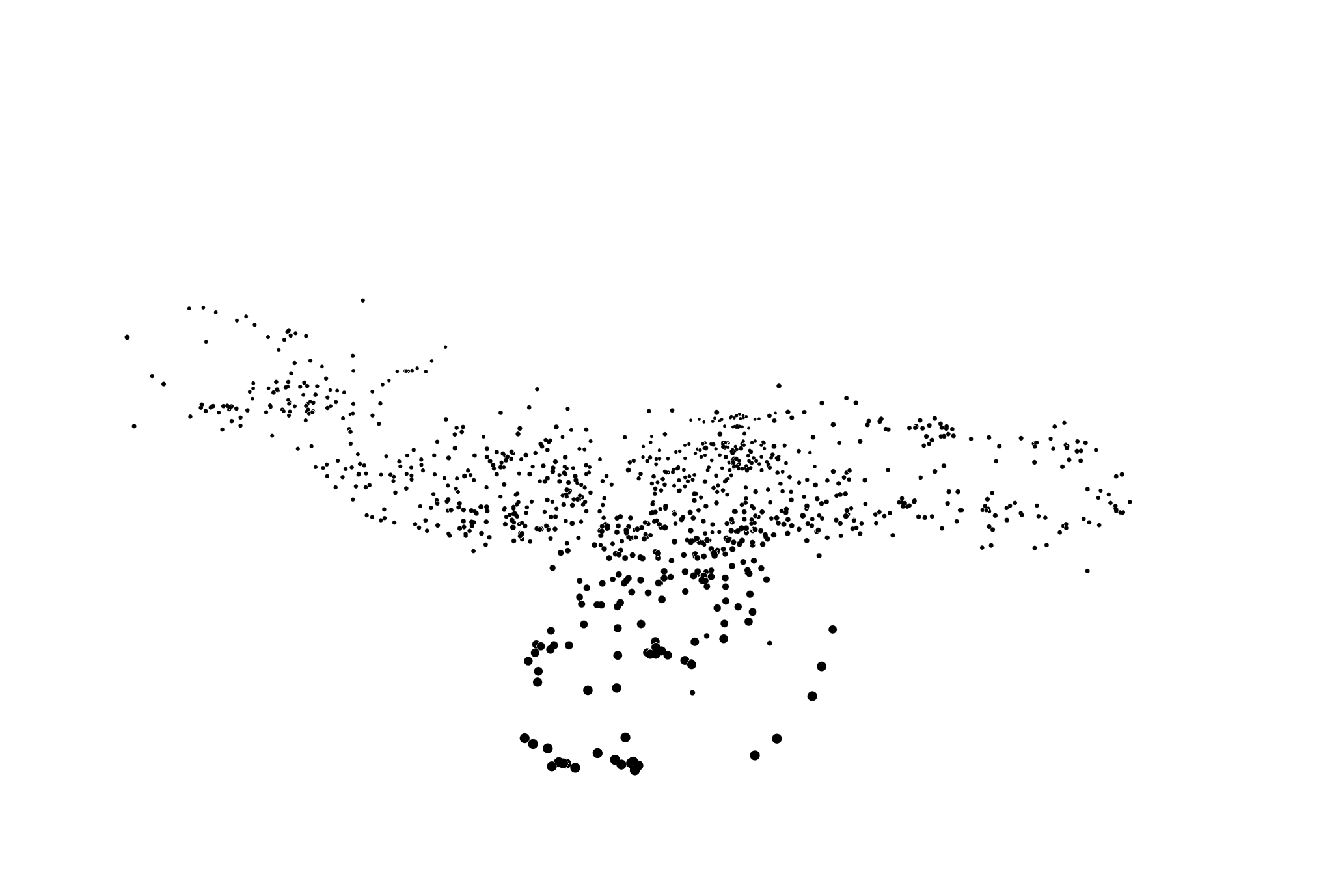} \\
(a) \blue{\cmark airplane} &\red{\xmark bench} &\red{\xmark bench} &\red{\xmark stairs} &\red{\xmark stairs} \\

\figt[\sizeS]{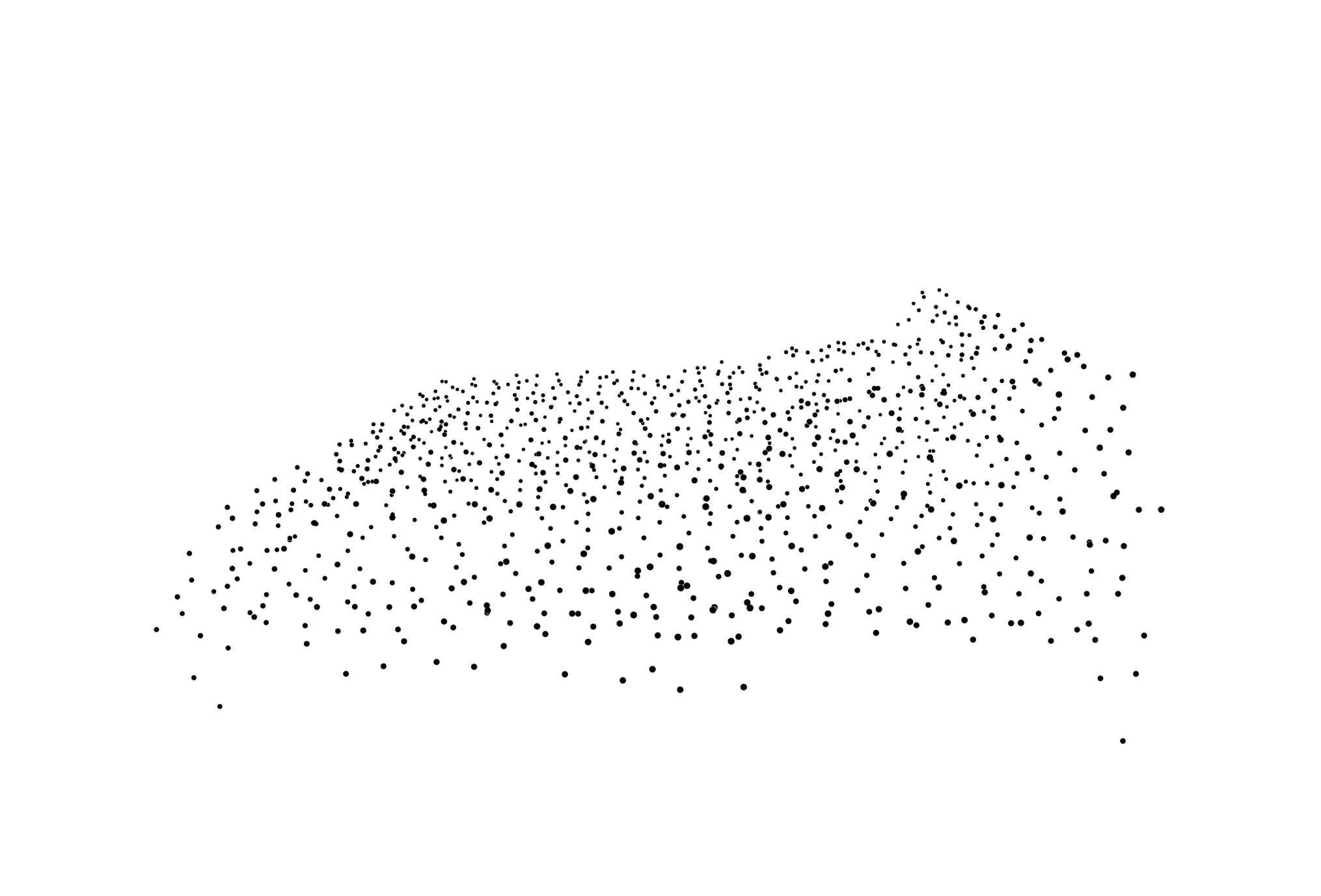} &
\figt[\sizeS]{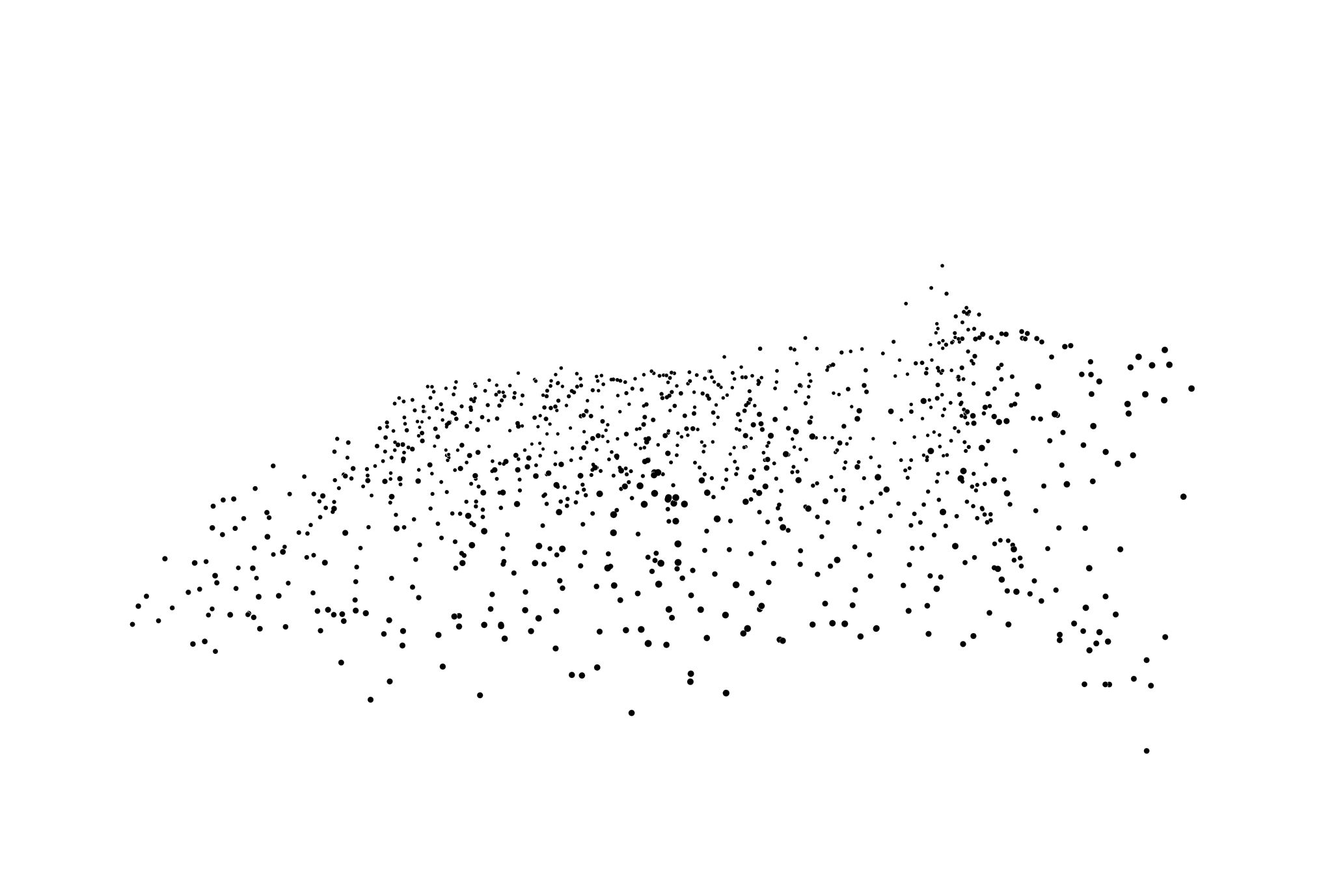} &
\figt[\sizeS]{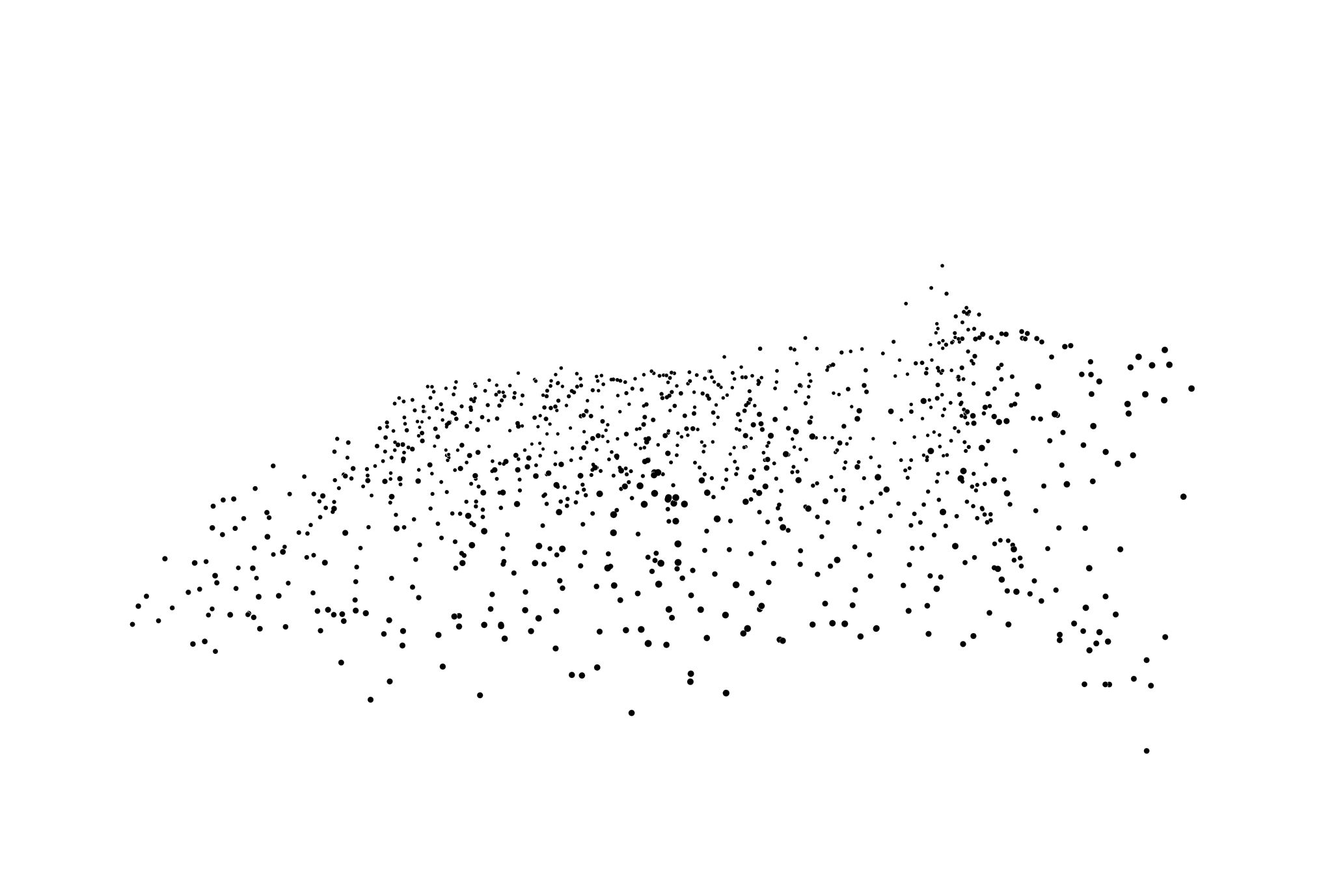} &
\figt[\sizeS]{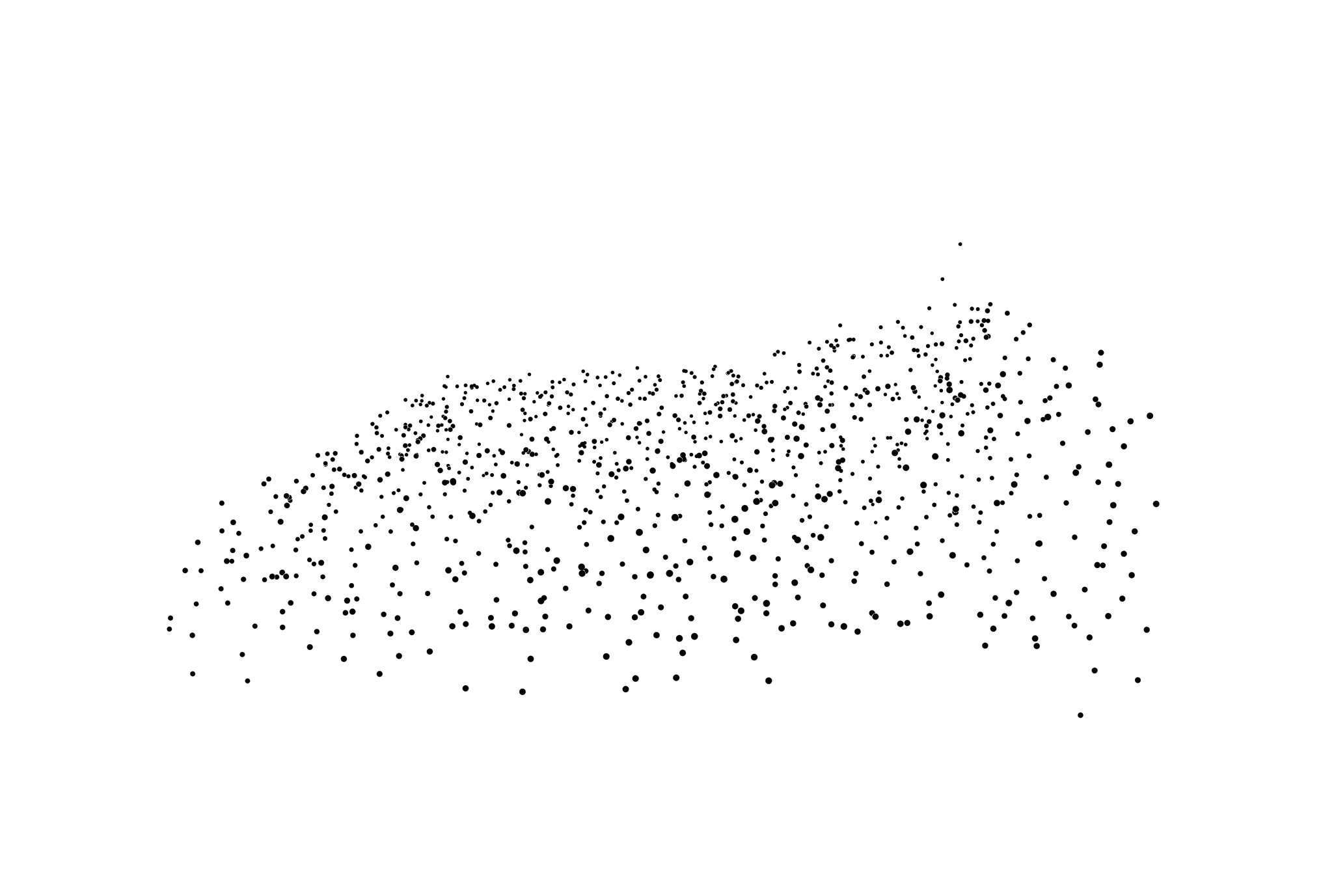} &
\figt[\sizeS]{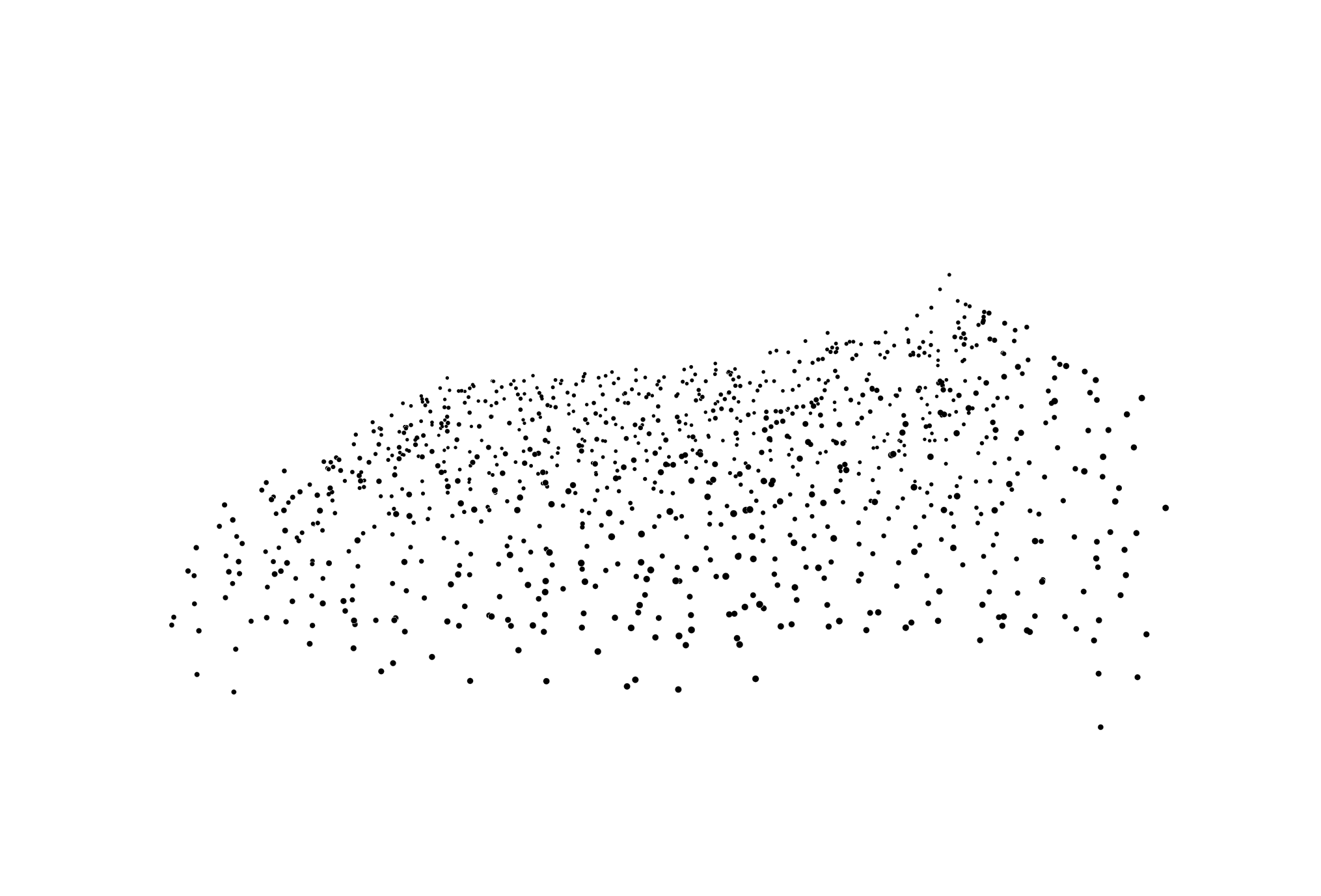} \\
(b) \blue{\cmark bed} &\red{\xmark table} &\red{\xmark table} &\red{\xmark piano} &\red{\xmark piano} \\

\figt[\sizeS]{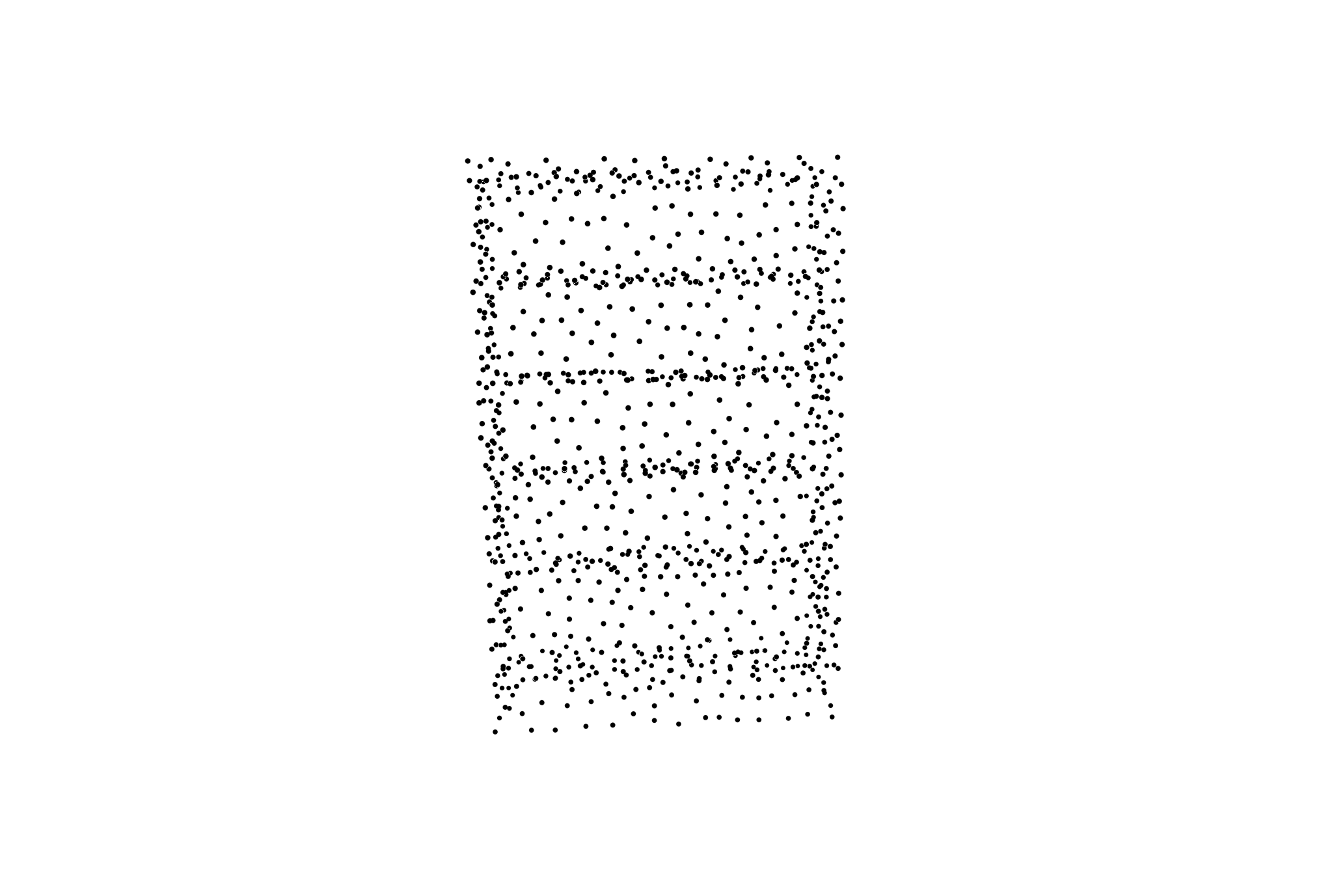} &
\figt[\sizeS]{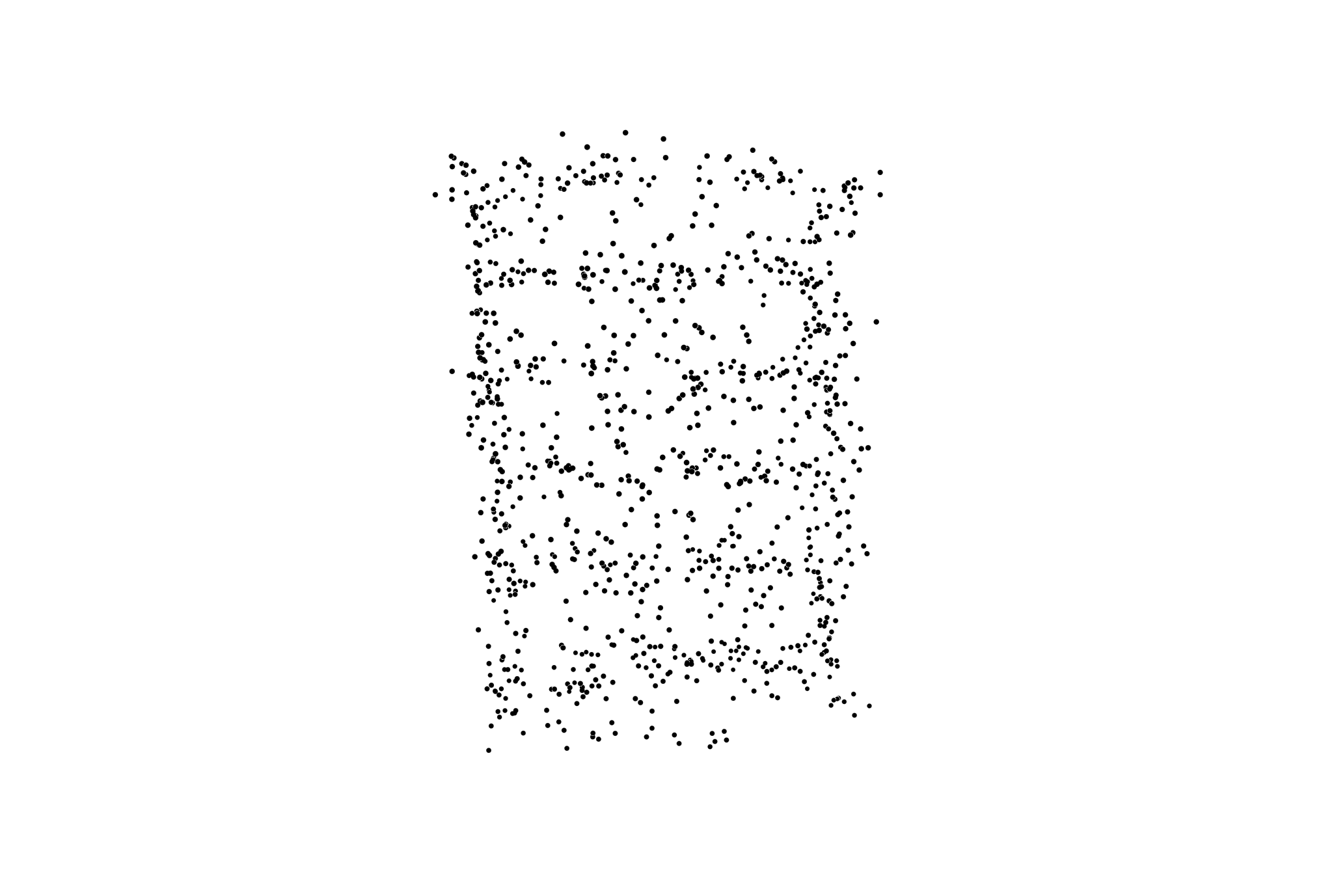} &
\figt[\sizeS]{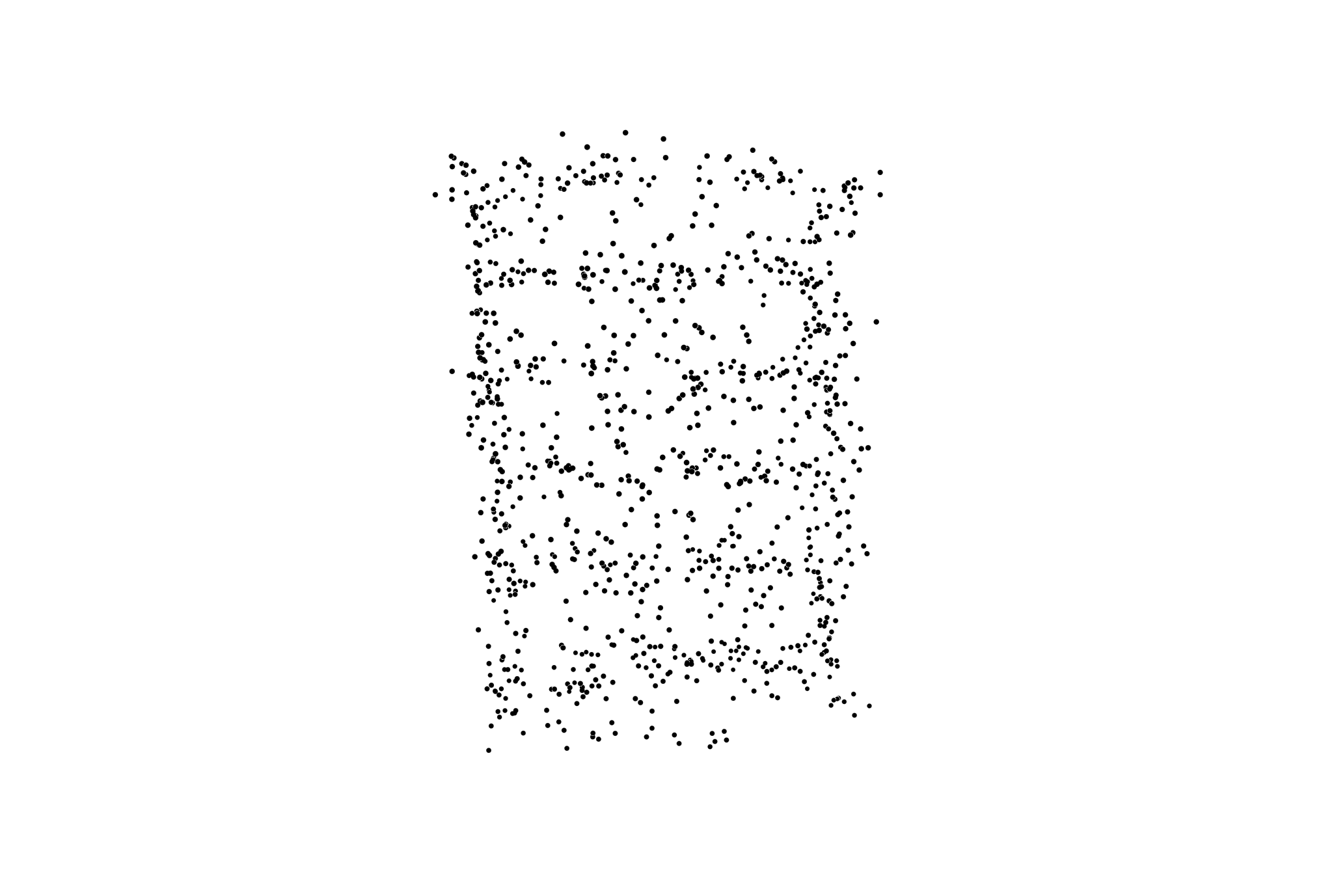} &
\figt[\sizeS]{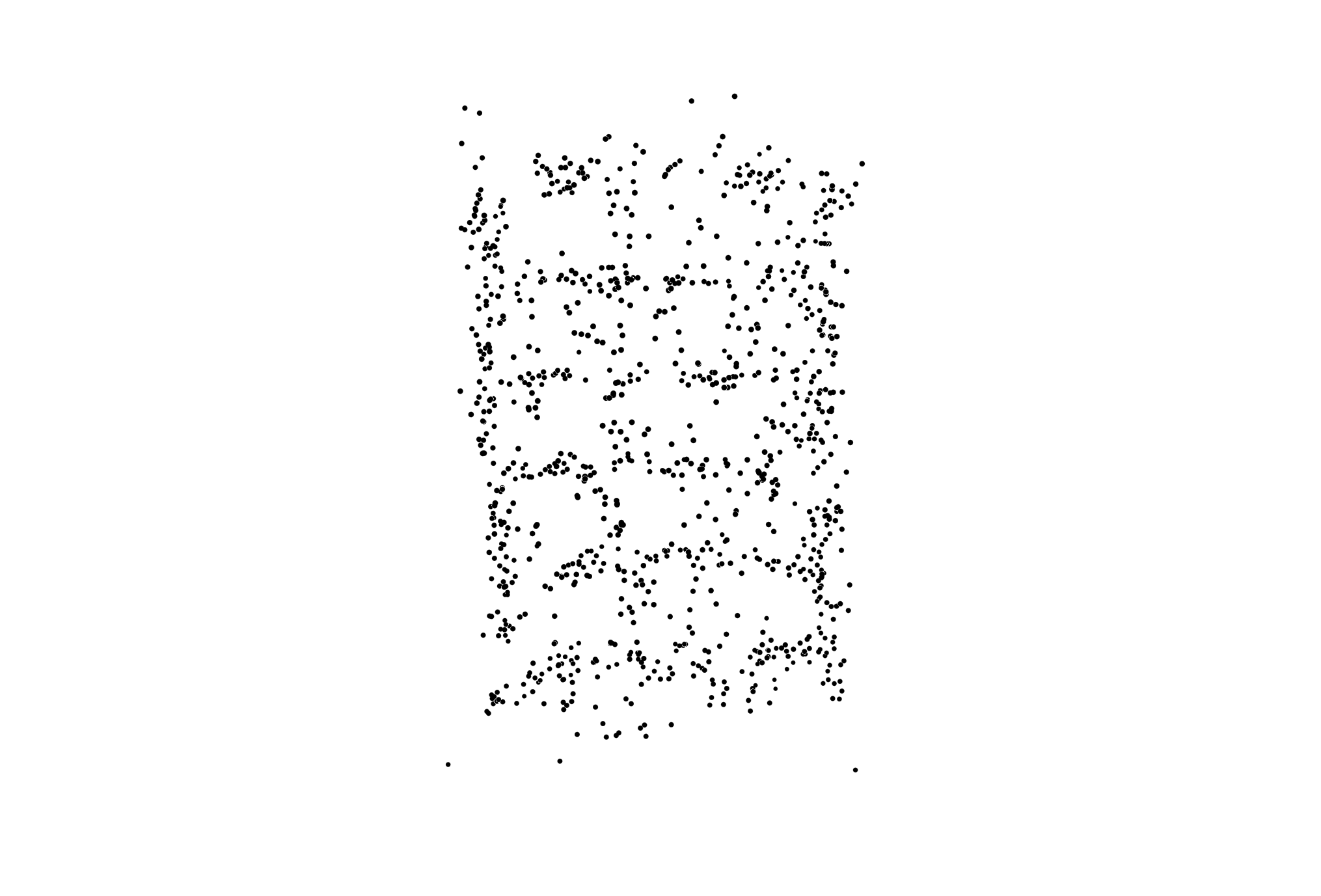} &
\figt[\sizeS]{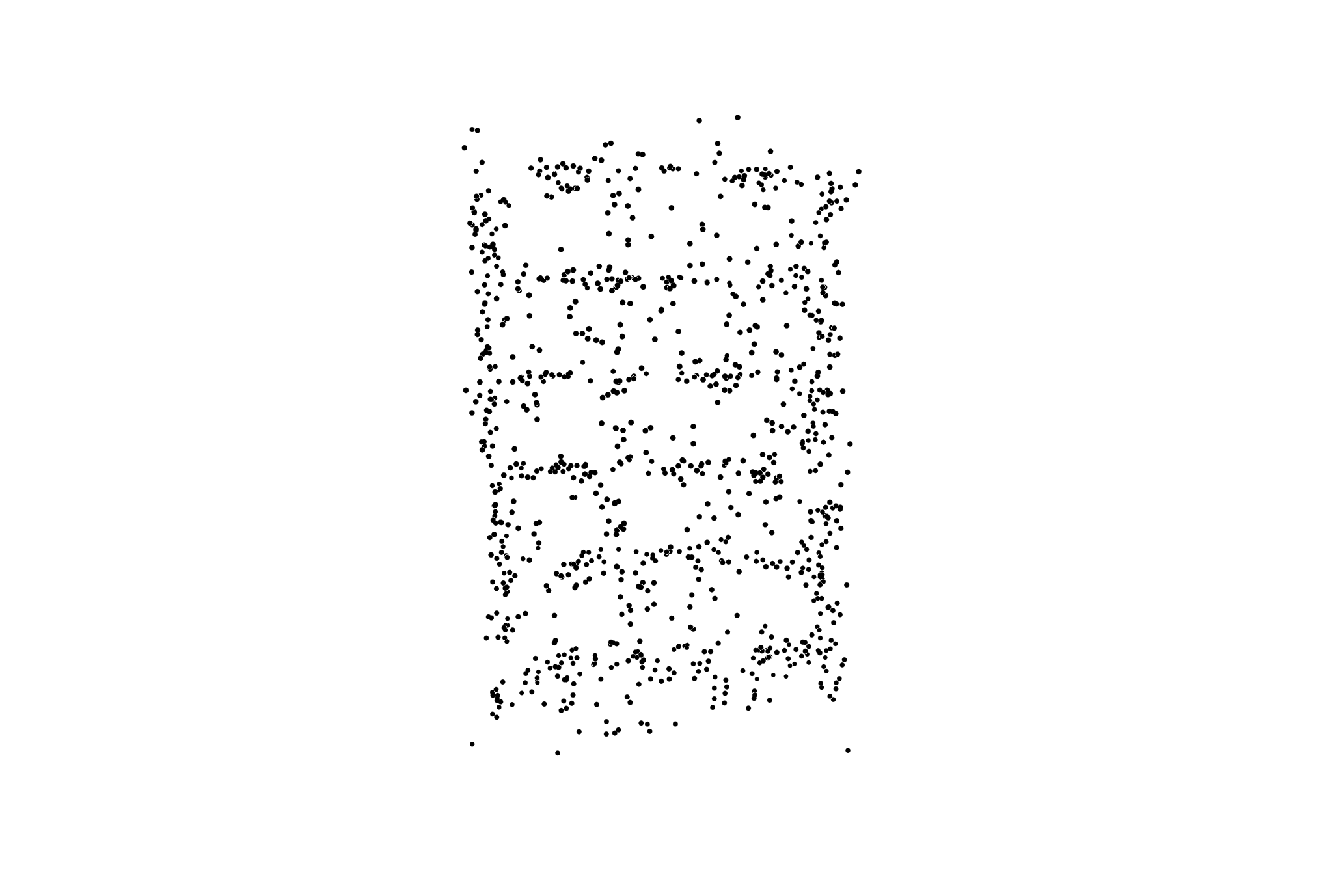} \\
(c) \blue{\cmark bookshelf} &\red{\xmark chair} &\red{\xmark chair} &\red{\xmark chair} &\red{\xmark chair} \\

\figt[\sizeS]{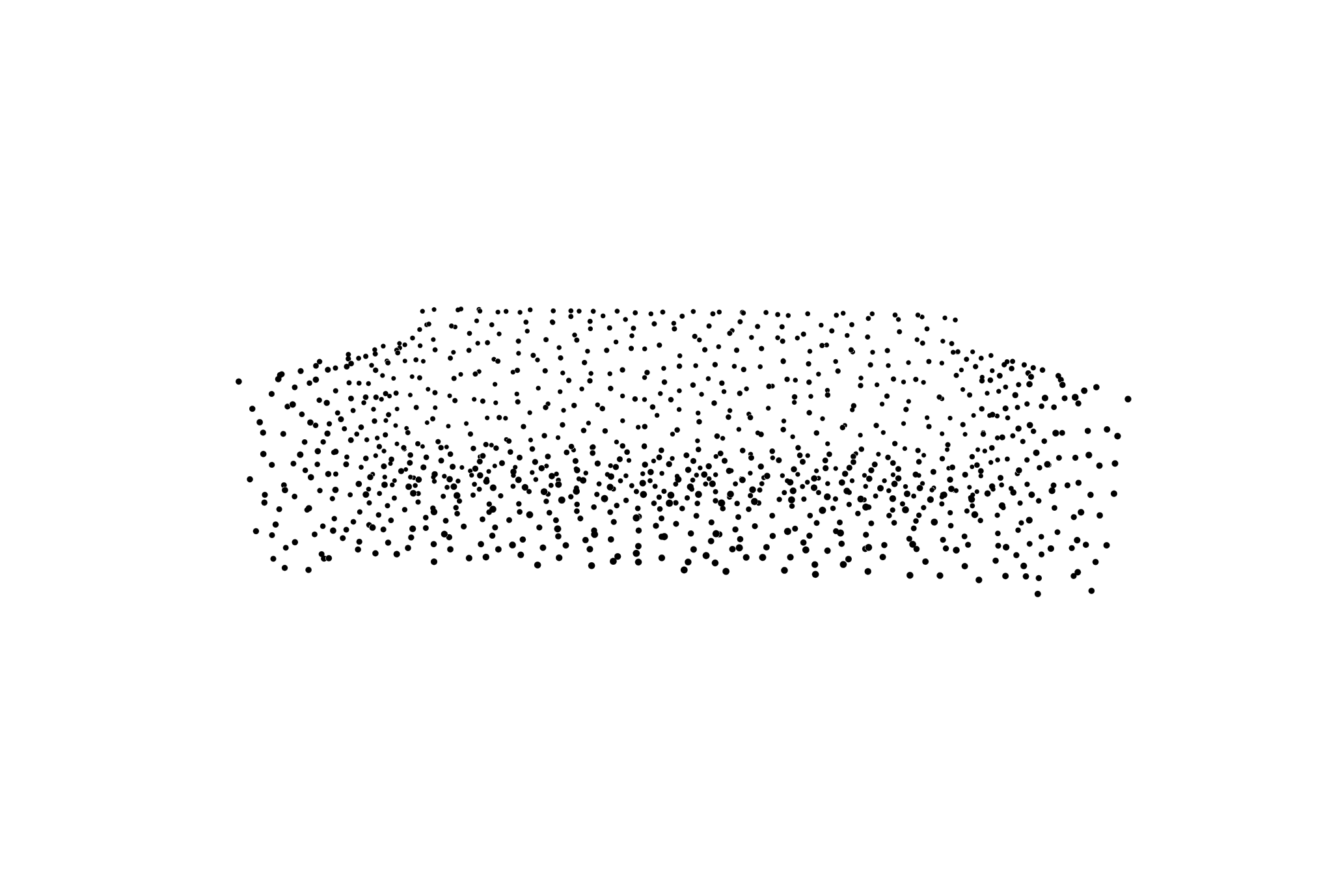} &
\figt[\sizeS]{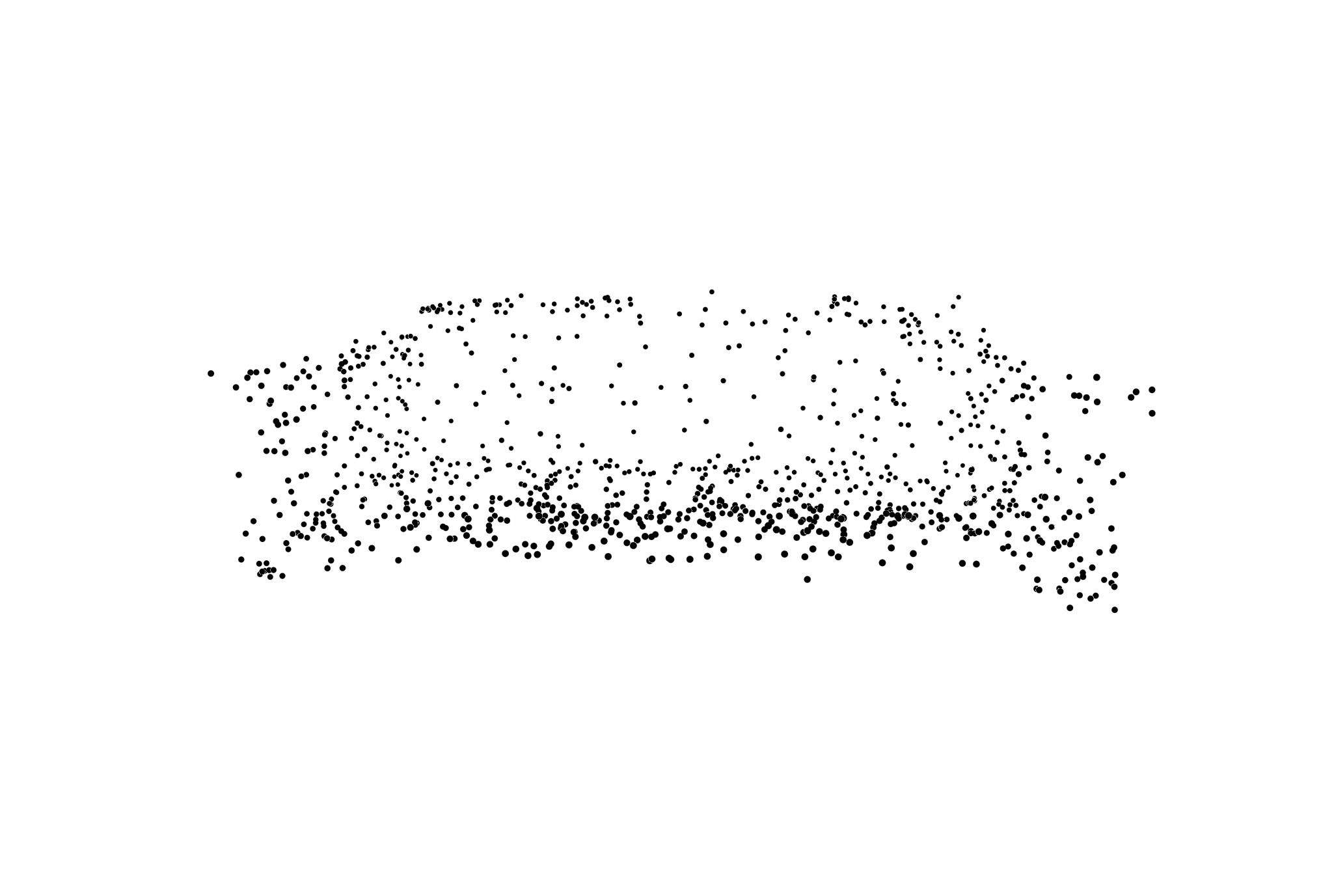} &
\figt[\sizeS]{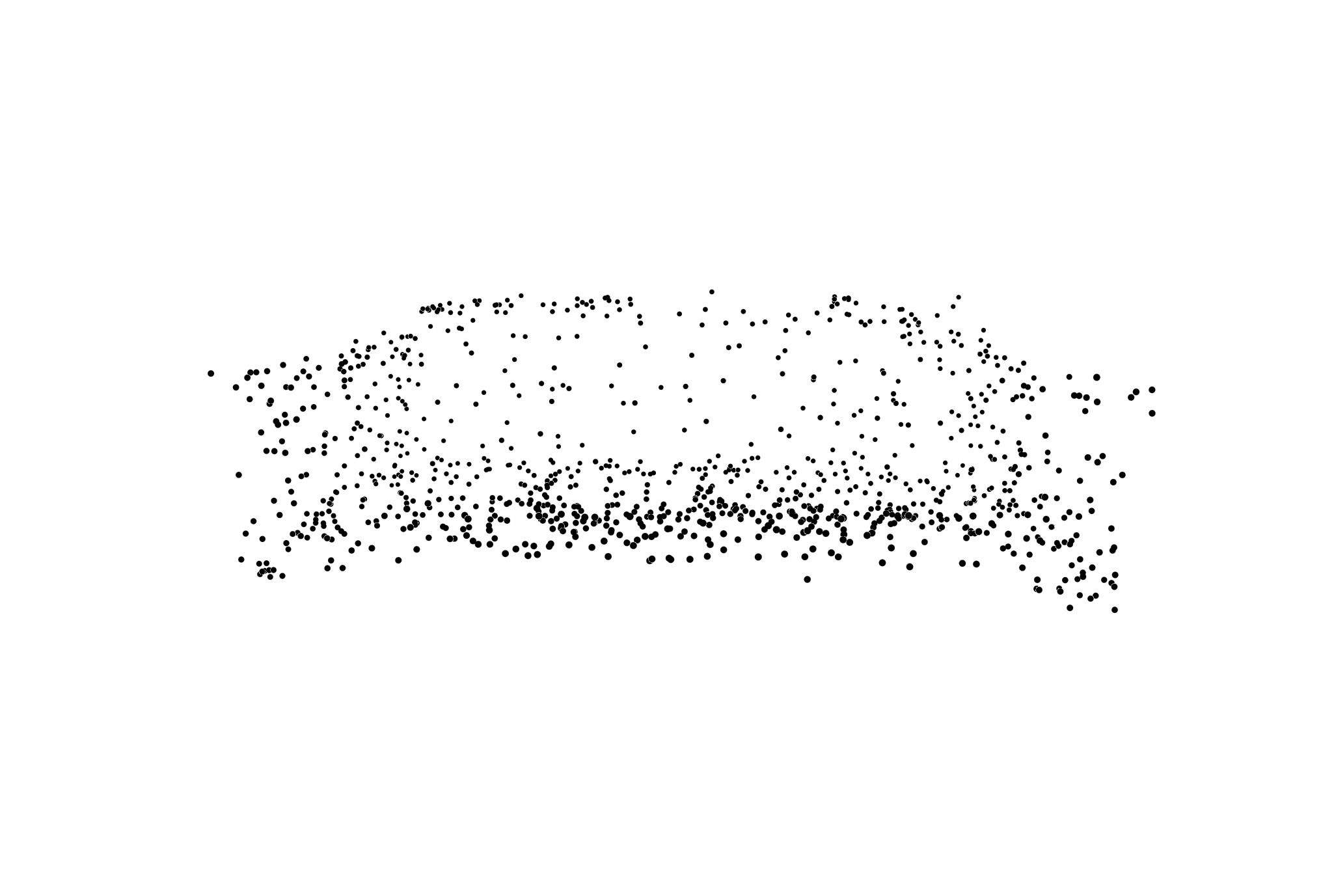} &
\figt[\sizeS]{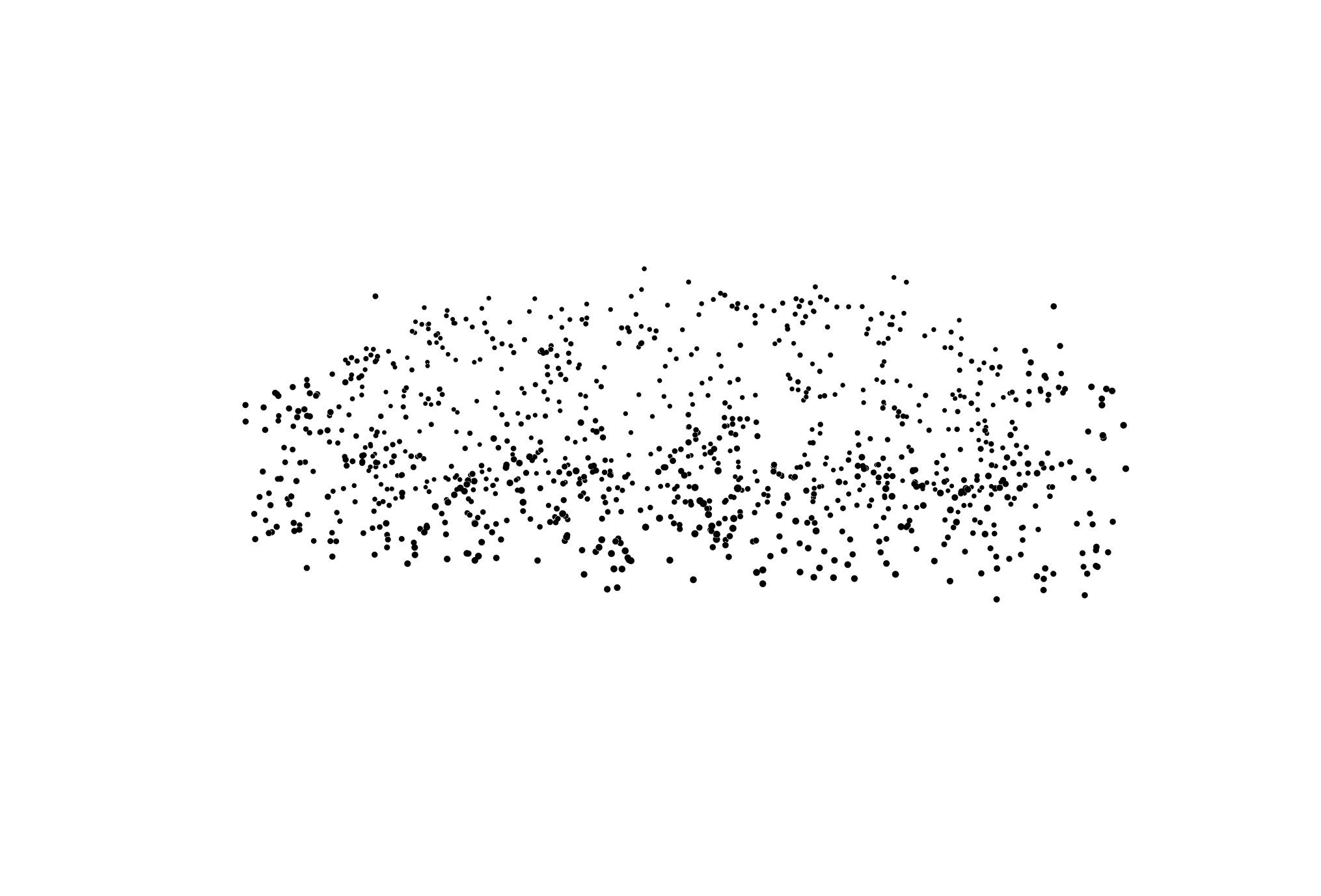} &
\figt[\sizeS]{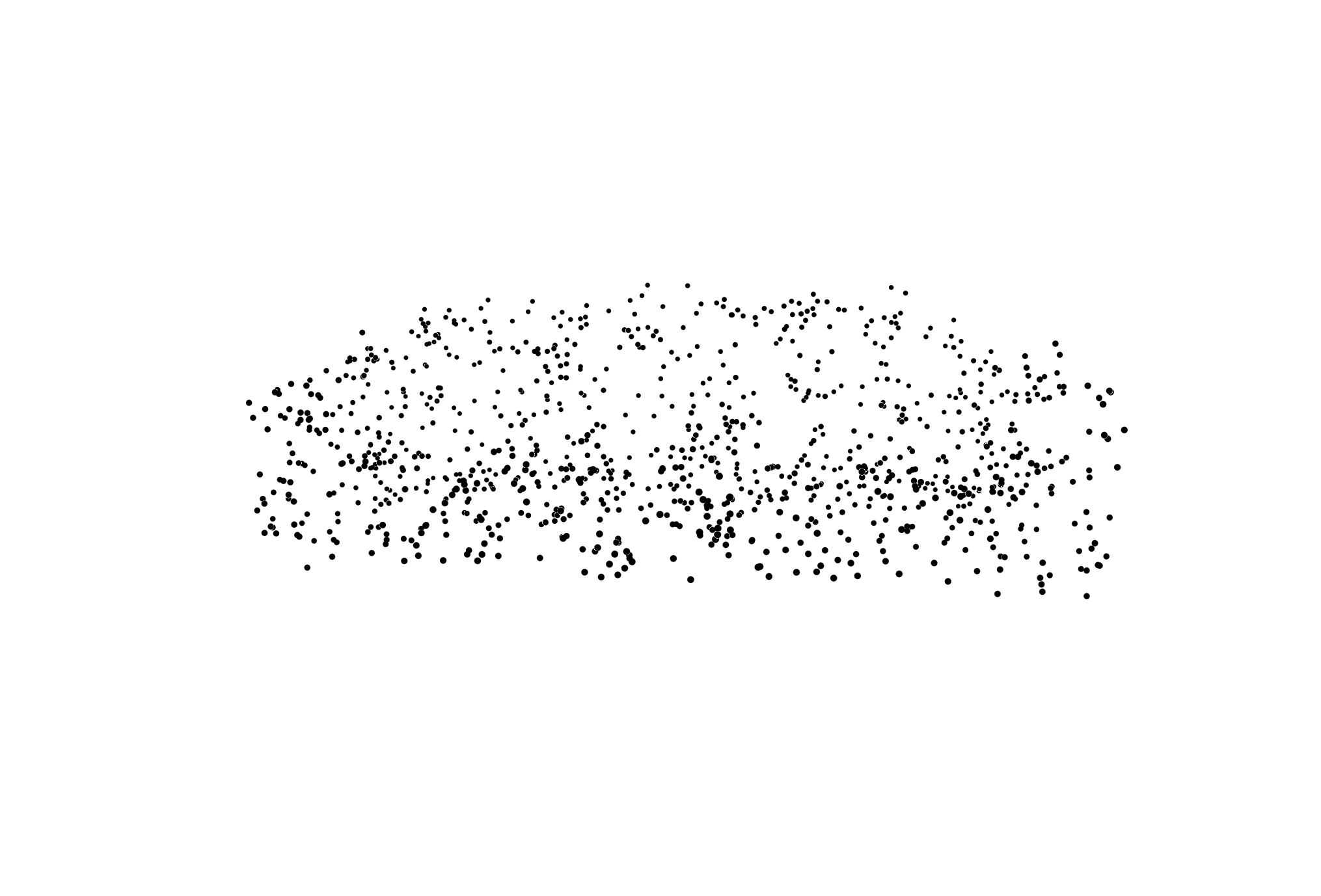} \\
(d) \blue{\cmark sofa} &\red{\xmark bed} &\red{\xmark bed} &\red{\xmark plant} &\red{\xmark plant} \\

\figt[\sizeS]{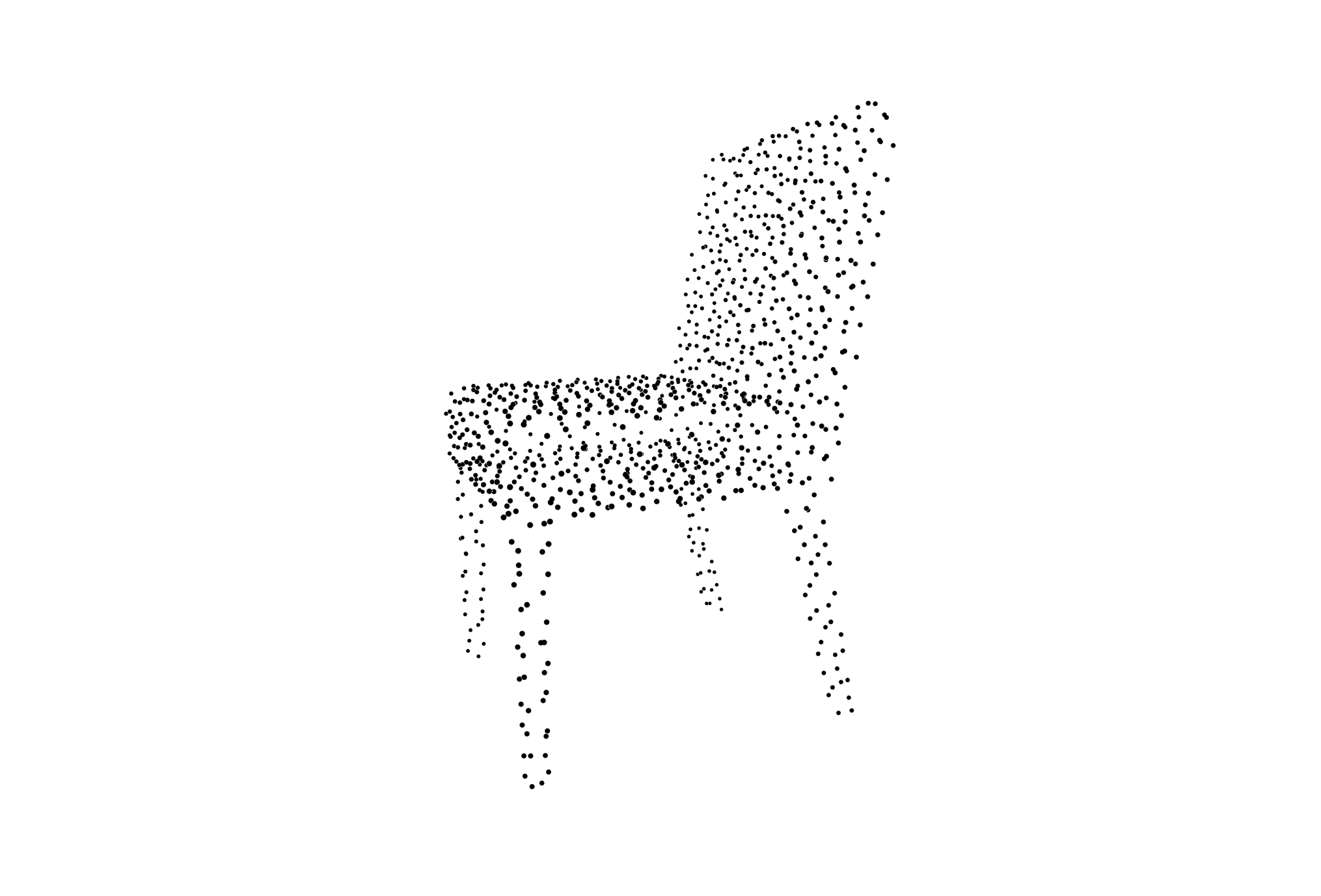} &
\figt[\sizeS]{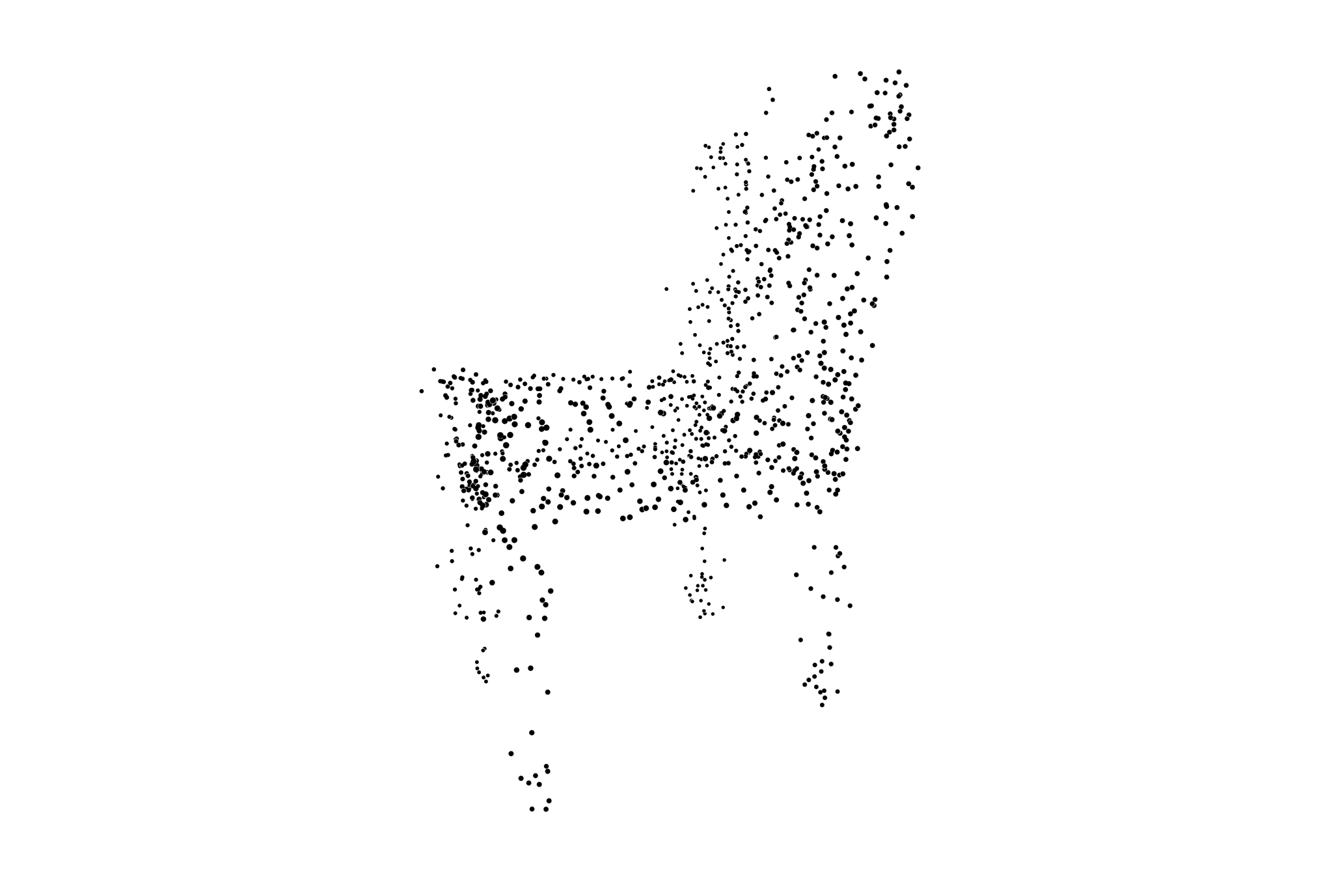} &
\figt[\sizeS]{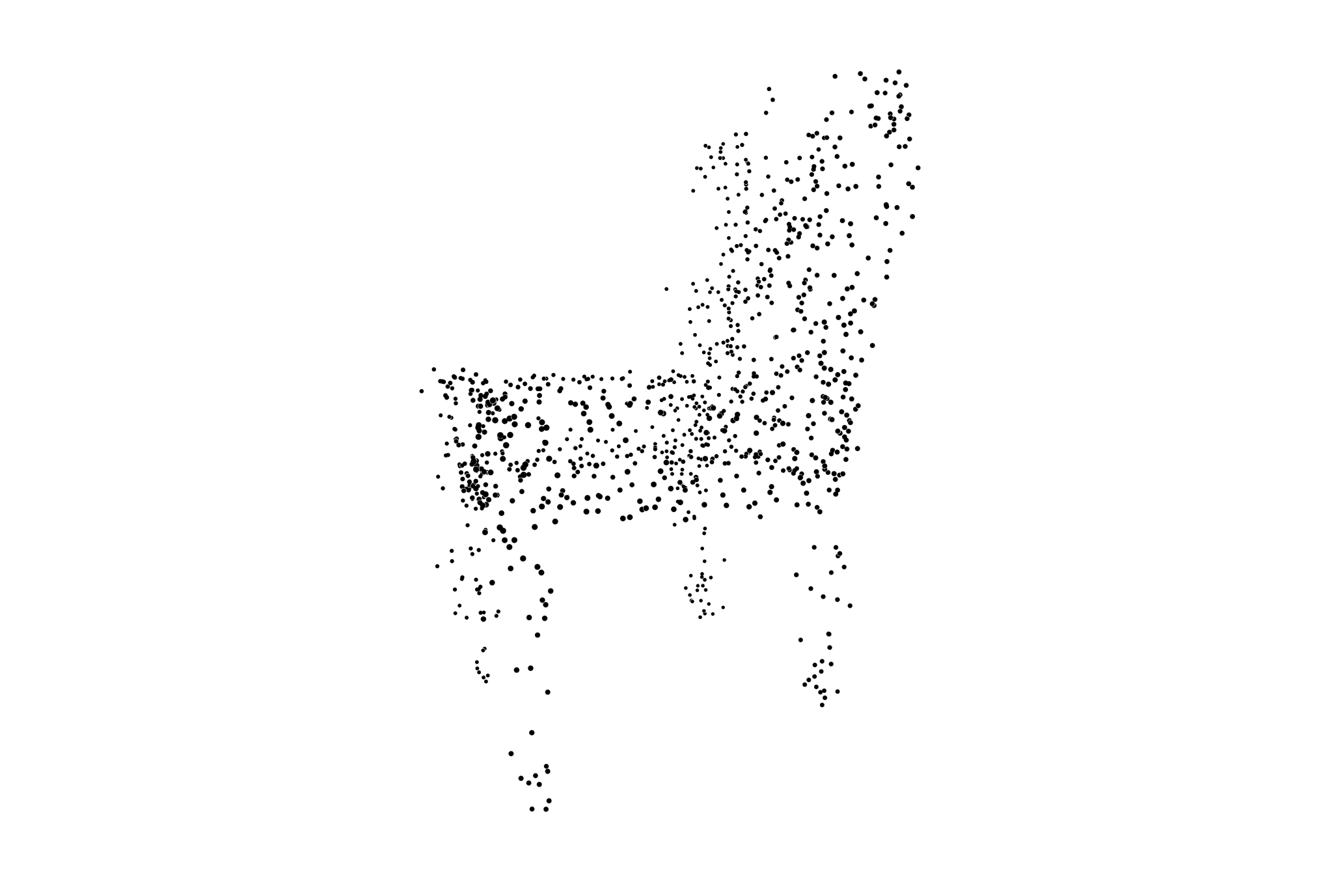} &
\figt[\sizeS]{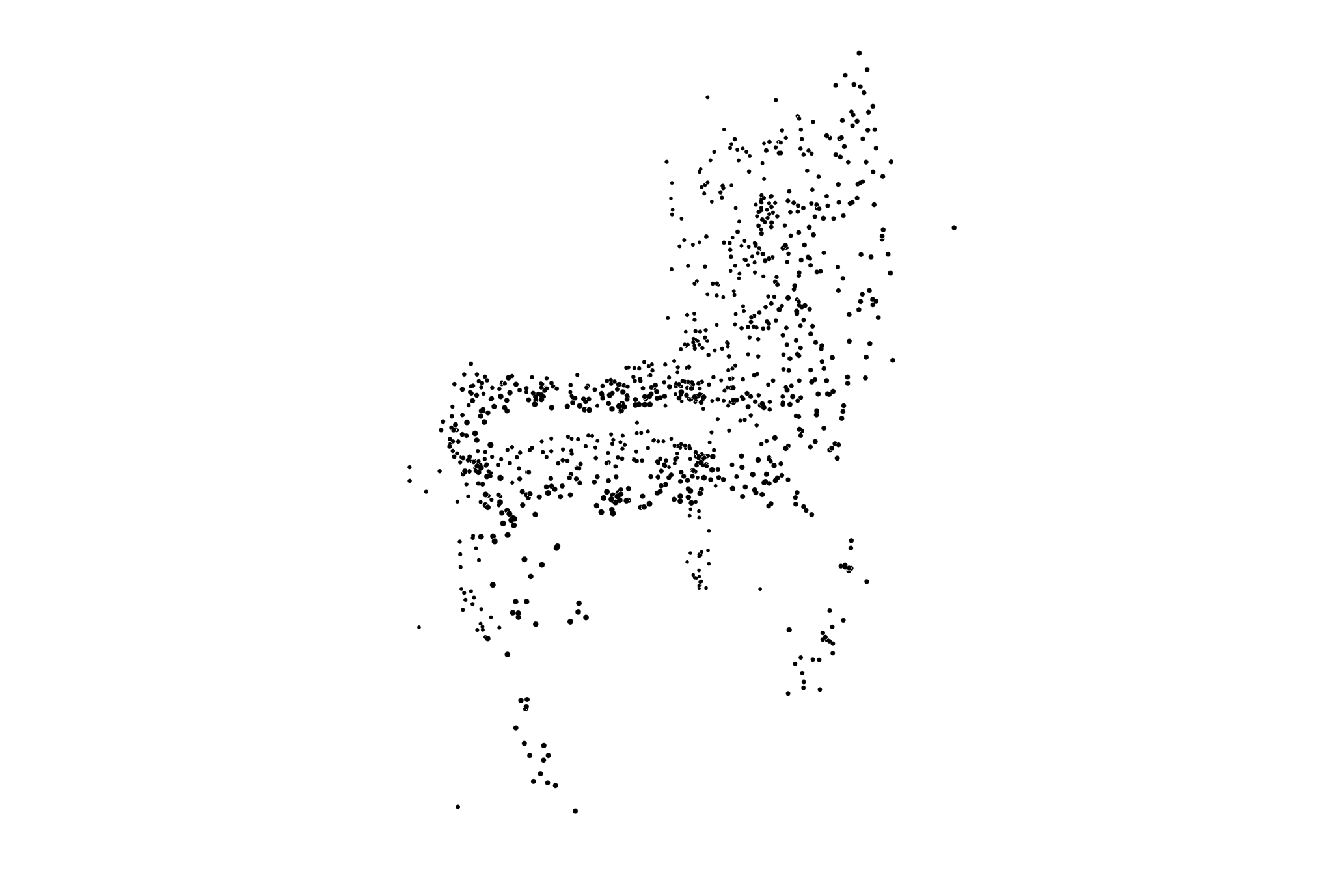} &
\figt[\sizeS]{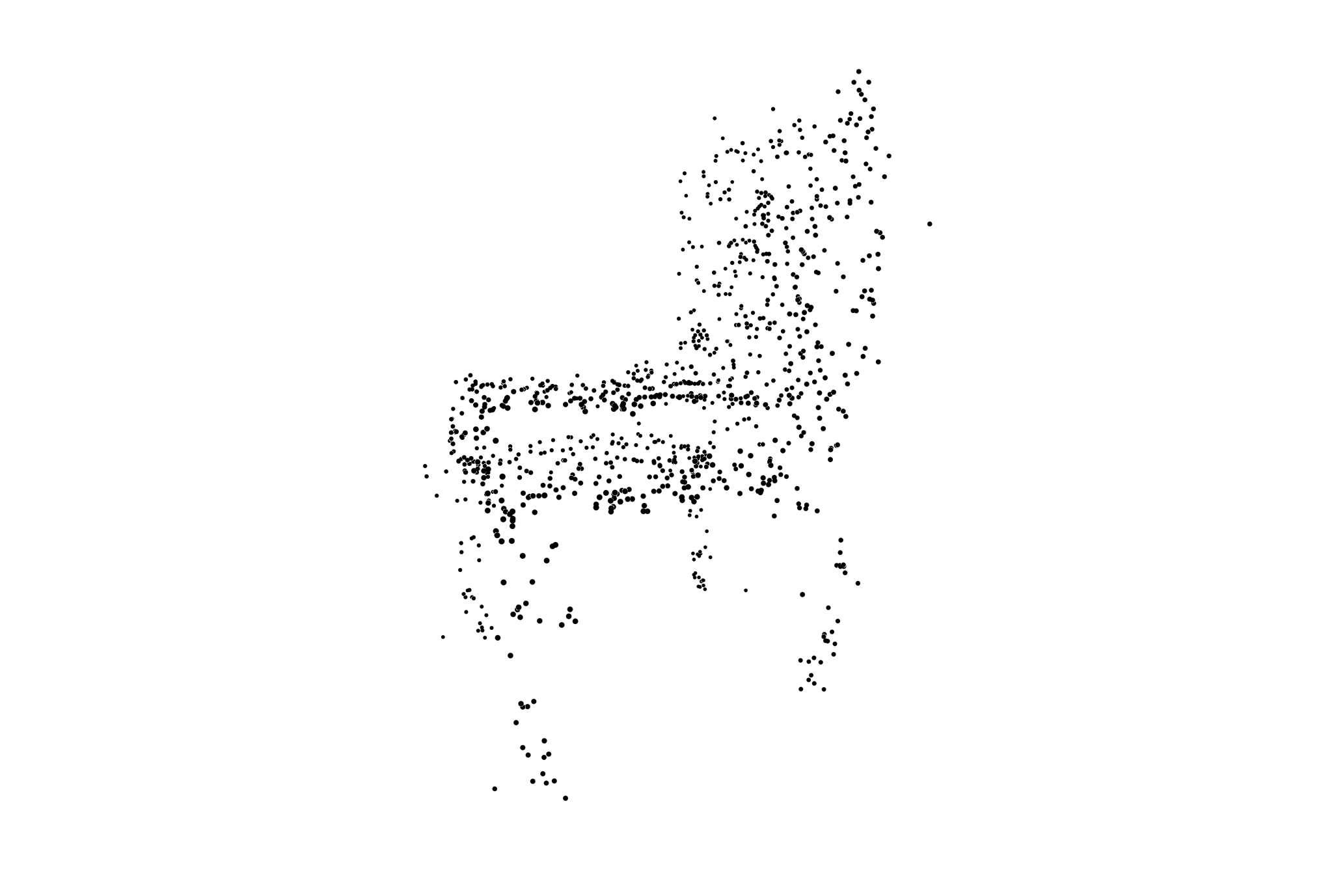} \\
(e) \blue{\cmark chair} &\red{\xmark bench} &\red{\xmark bench} &\red{\xmark bed} &\red{\xmark bed} \\

\end{tabular}
\caption{Visualization for adversarial point clouds produced by baseline methods on DGCNN.}
\label{fig:vis-3d-dgcnn-supp}
\end{figure*}

\begin{figure*}[hptb]
\newcommand{\sizeS}{.20}
\centering
\setlength{\tabcolsep}{0pt}
\begin{tabular}{ccccc}
\textbf{Clean} &\simba &\simbapp &\siadv & \projabbrv\\
\figt[\sizeS]{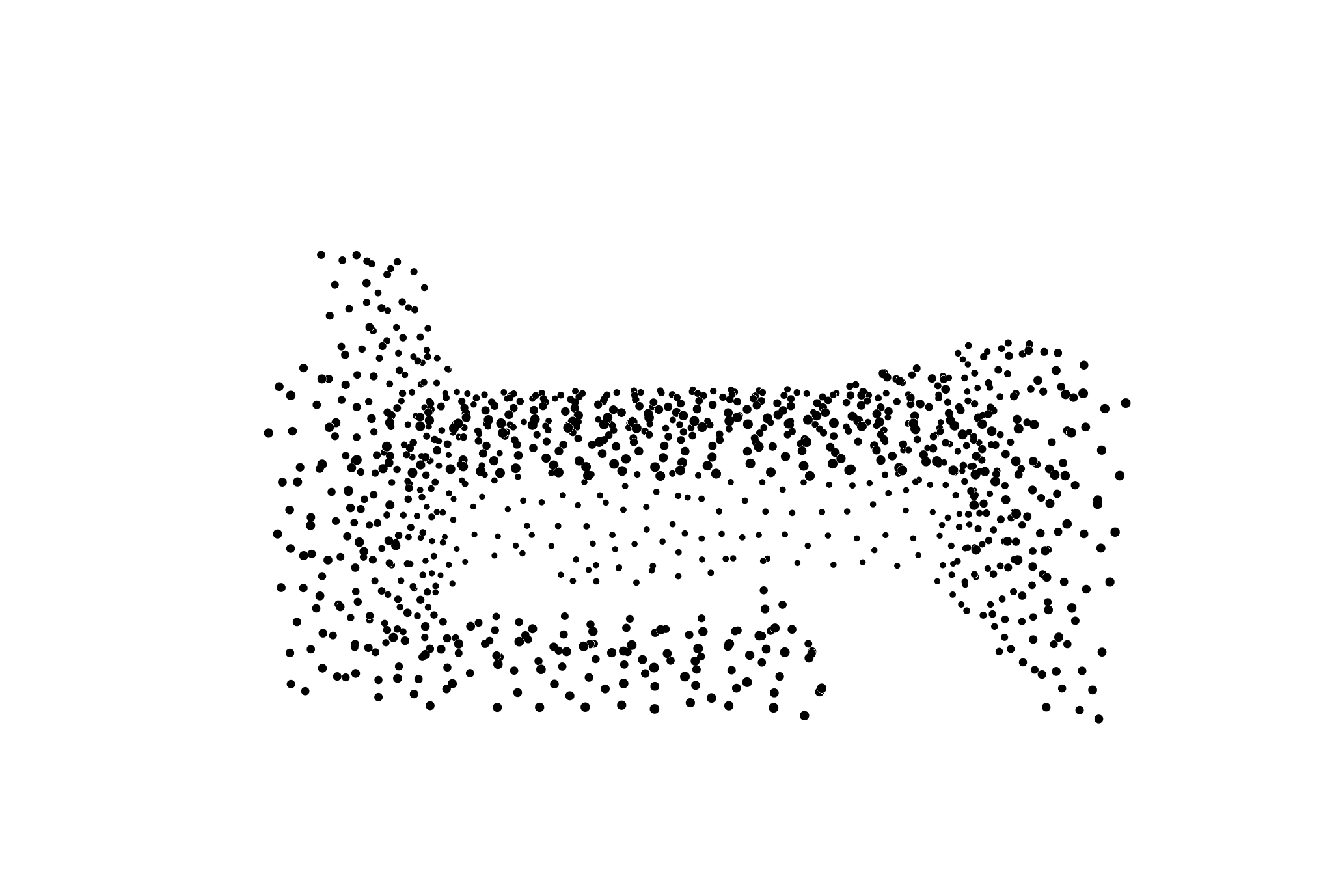} &
\figt[\sizeS]{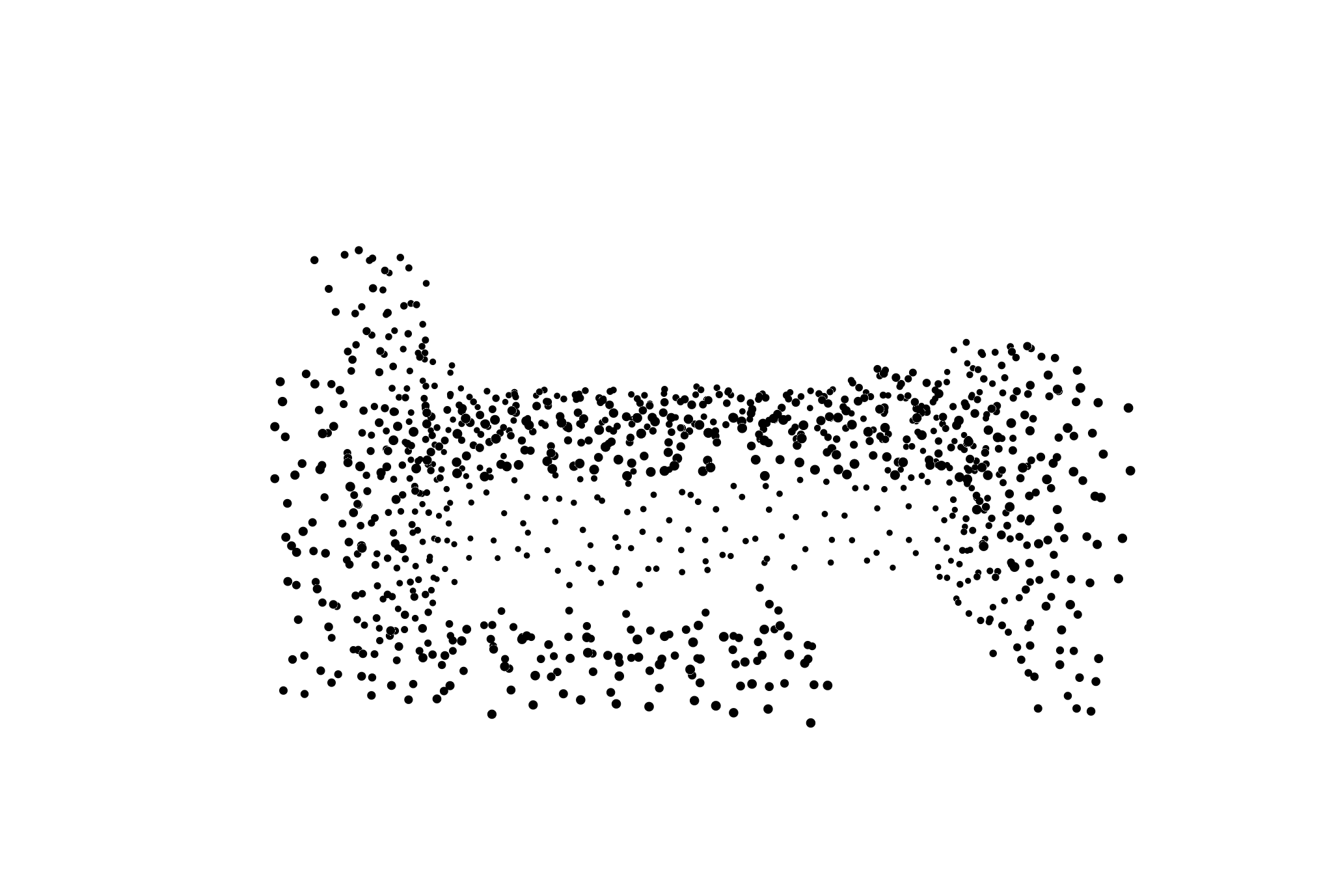} &
\figt[\sizeS]{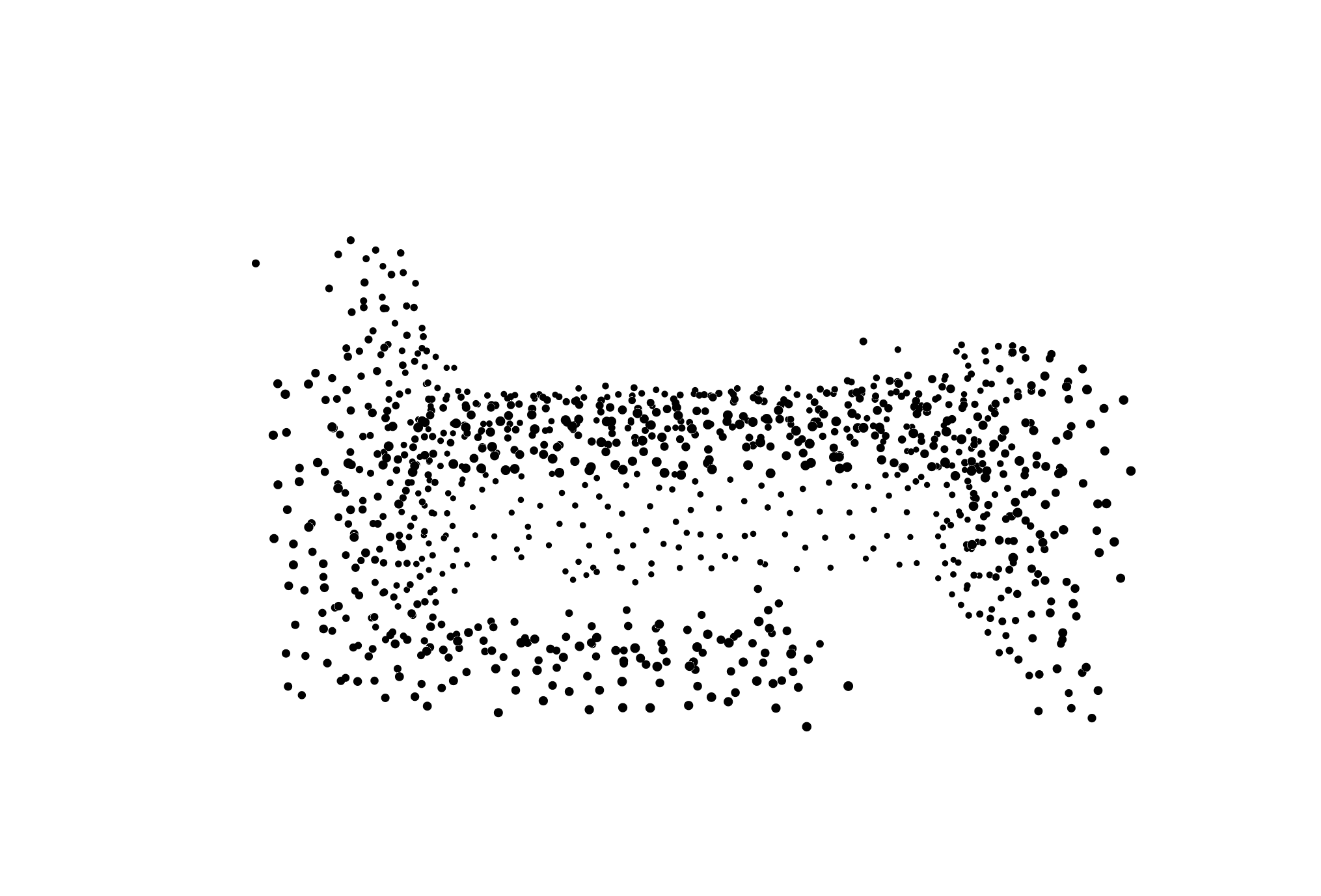} &
\figt[\sizeS]{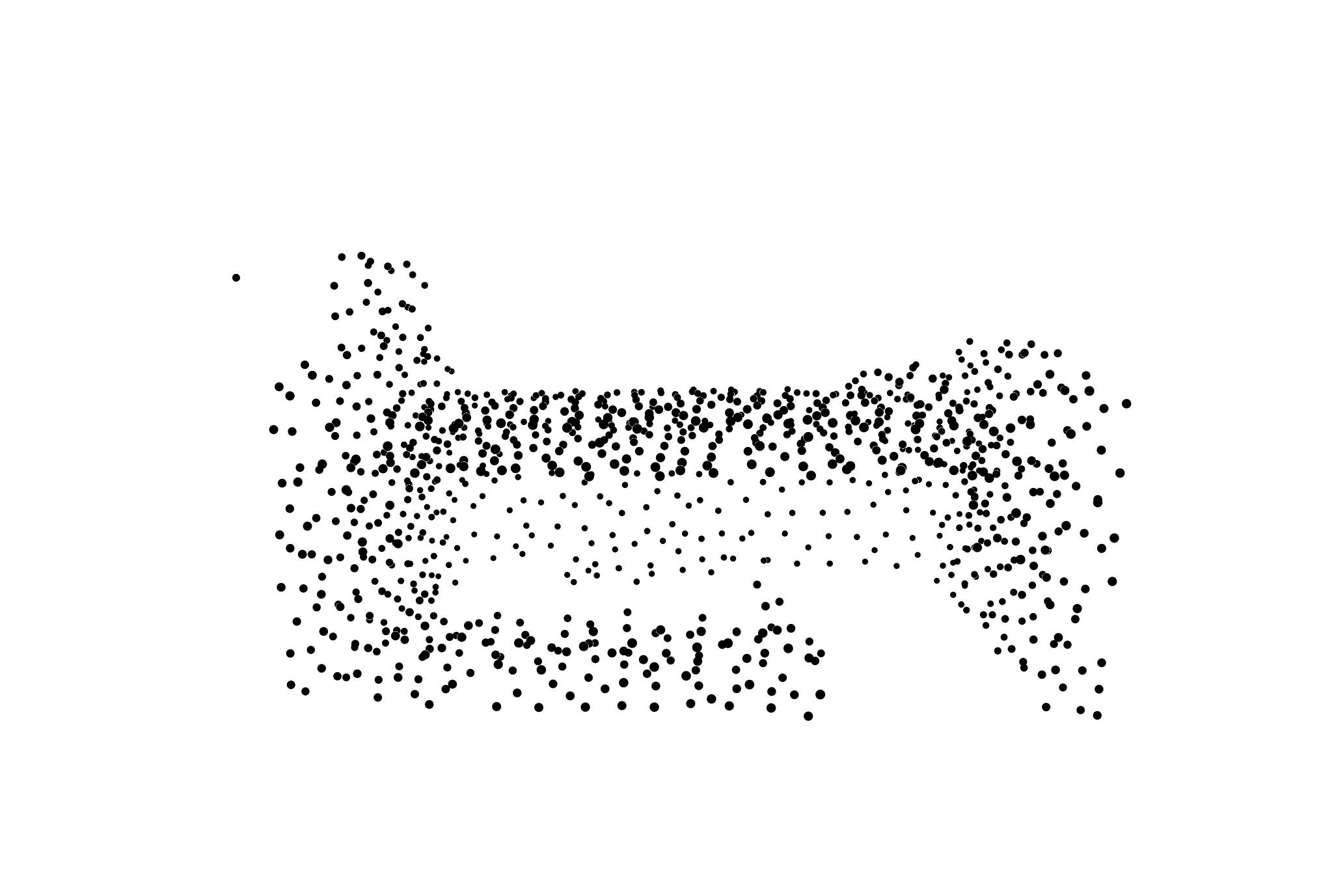} &
\figt[\sizeS]{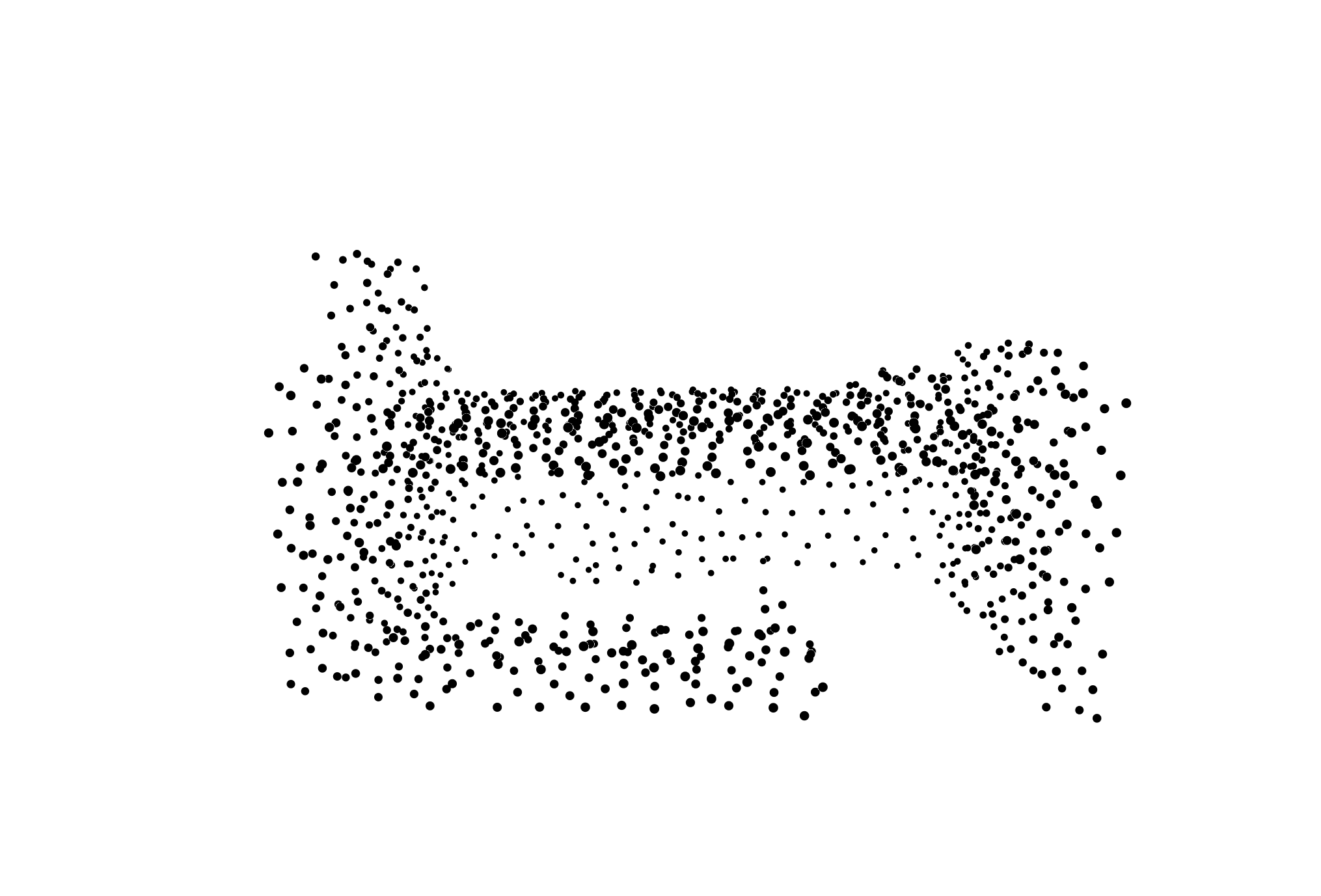} \\
(a) \blue{\cmark bed} &\red{\xmark desk} &\red{\xmark desk} &\red{\xmark desk} &\red{\xmark desk} \\

\figt[\sizeS]{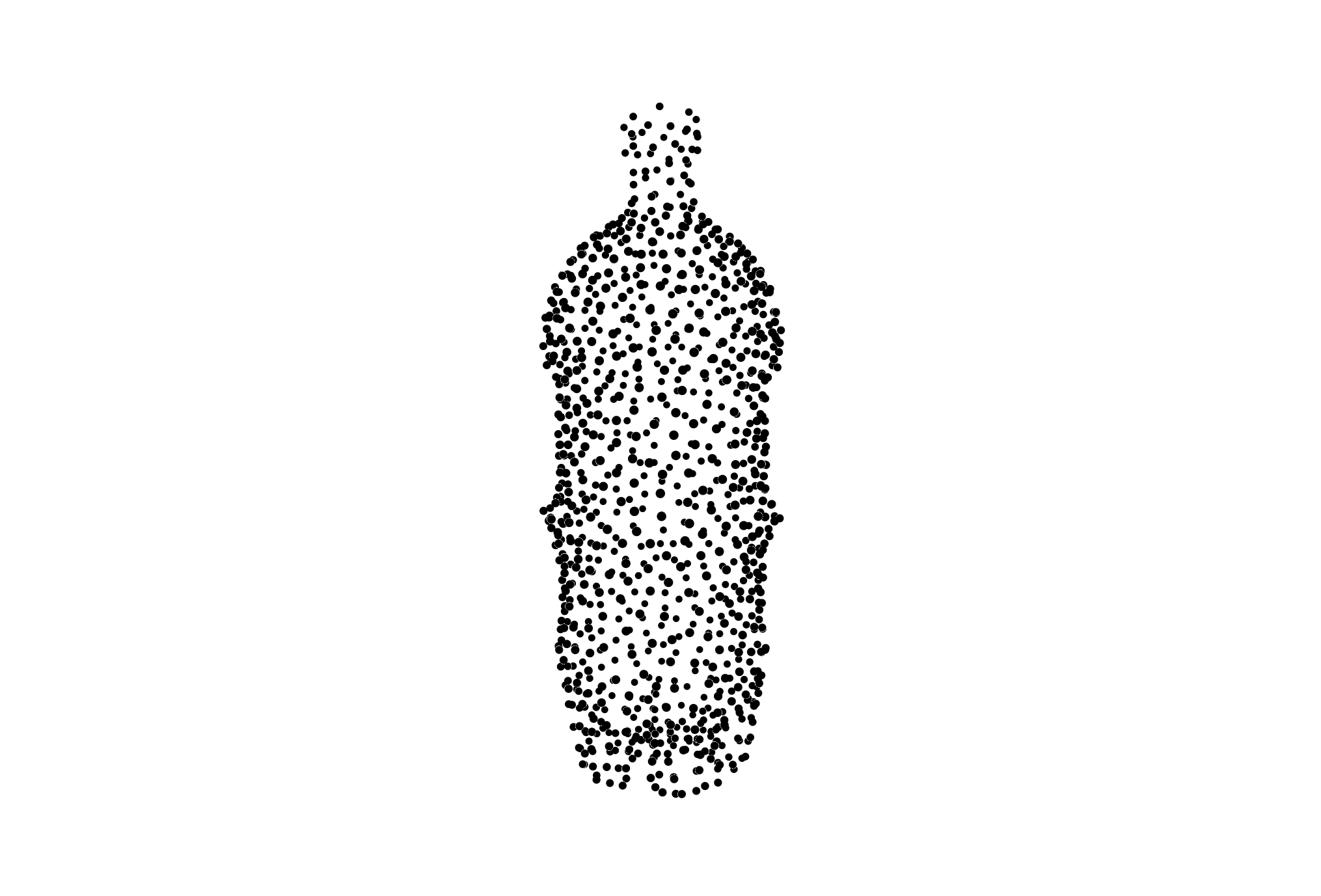} &
\figt[\sizeS]{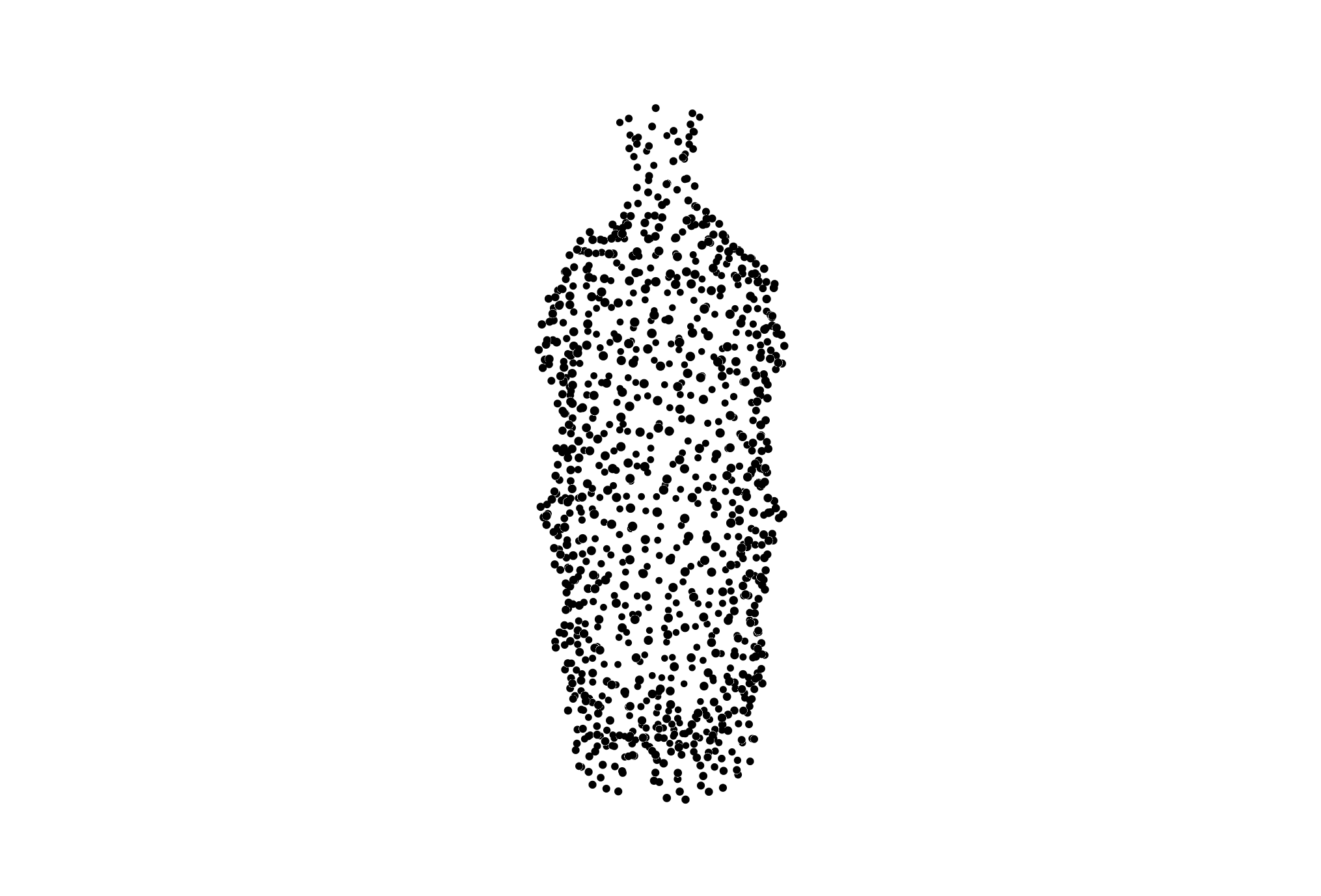} &
\figt[\sizeS]{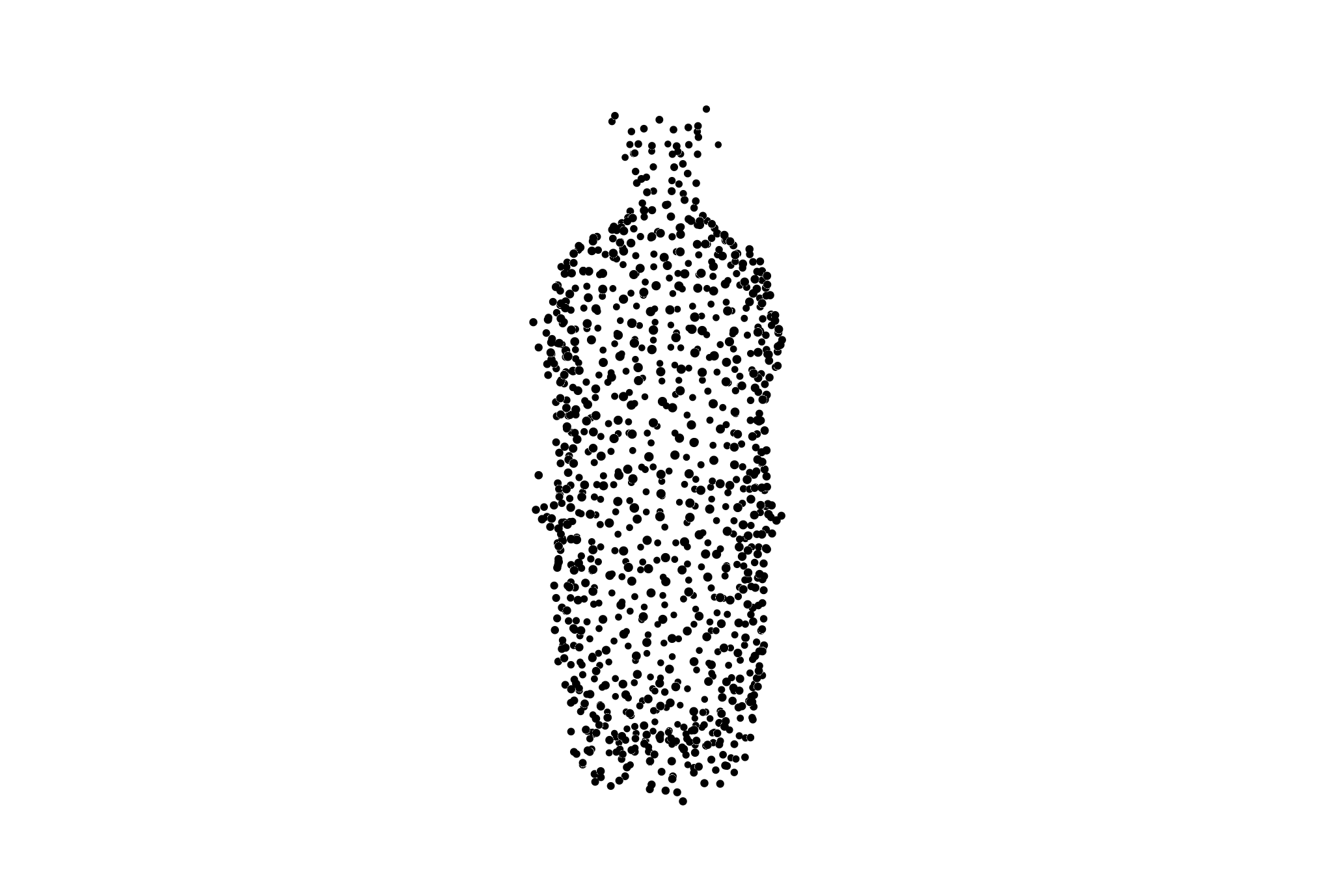} &
\figt[\sizeS]{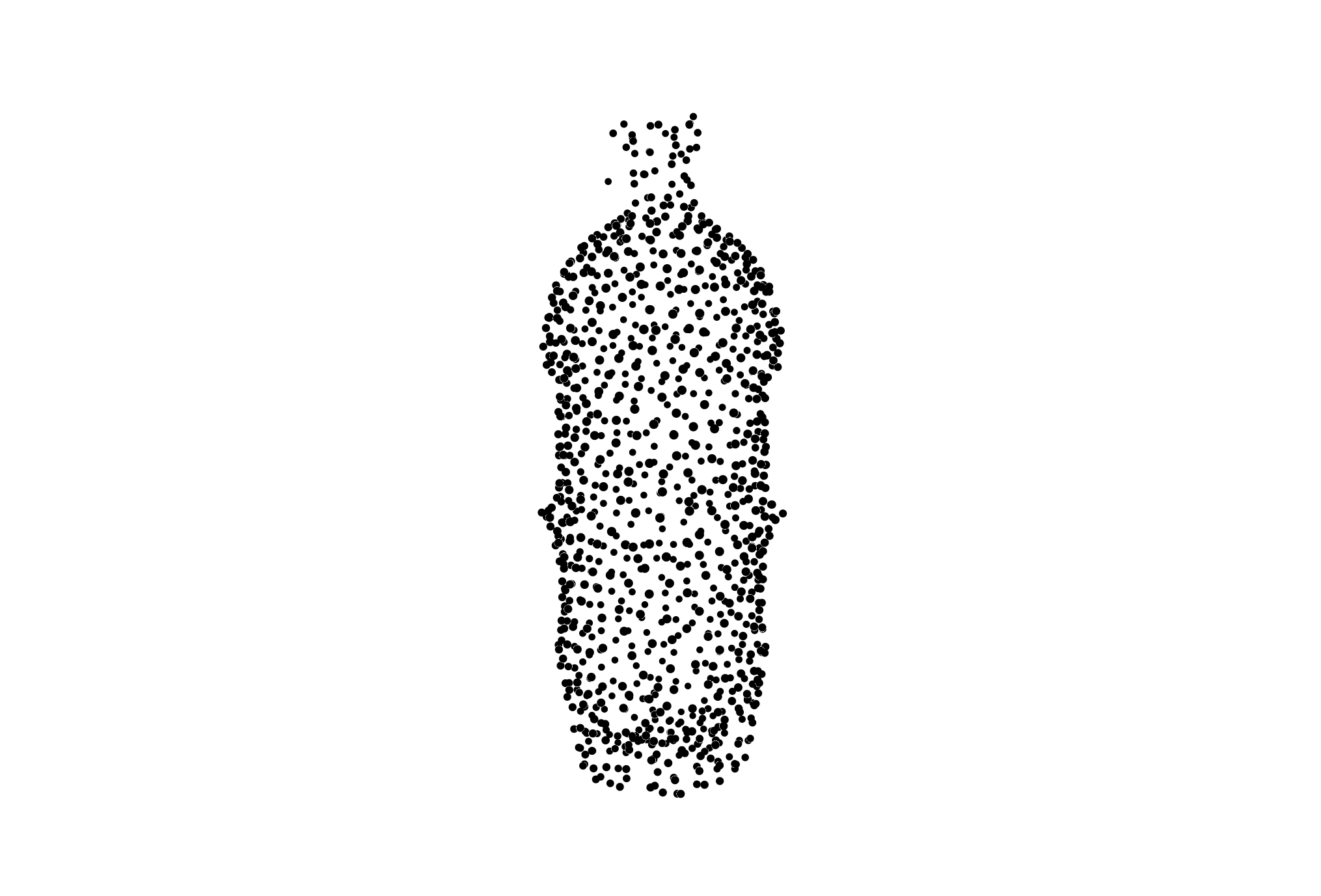} &
\figt[\sizeS]{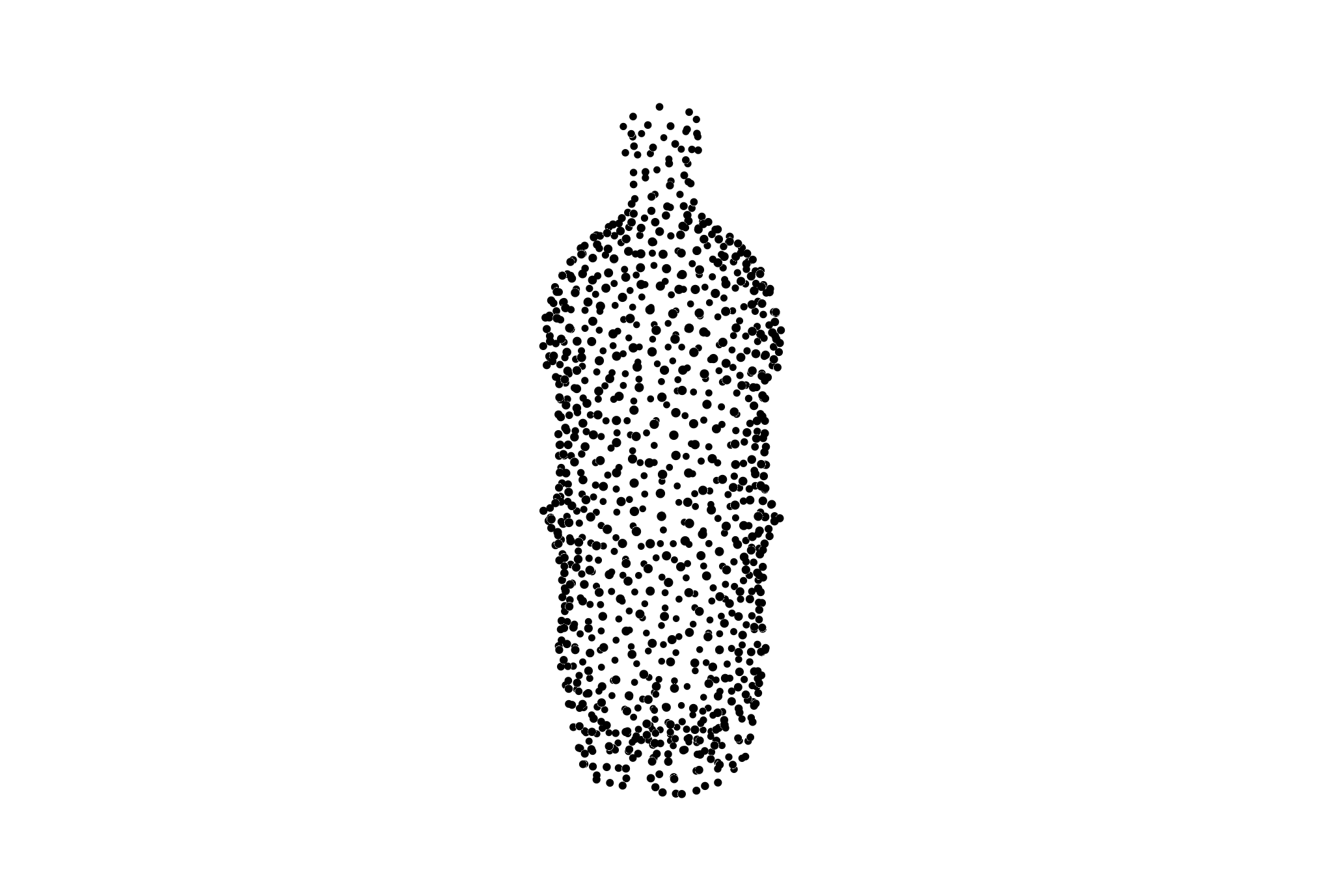} \\
(b) \blue{\cmark bottle} &\red{\xmark vase} &\red{\xmark vase} &\red{\xmark vase} &\red{\xmark vase} \\

\figt[\sizeS]{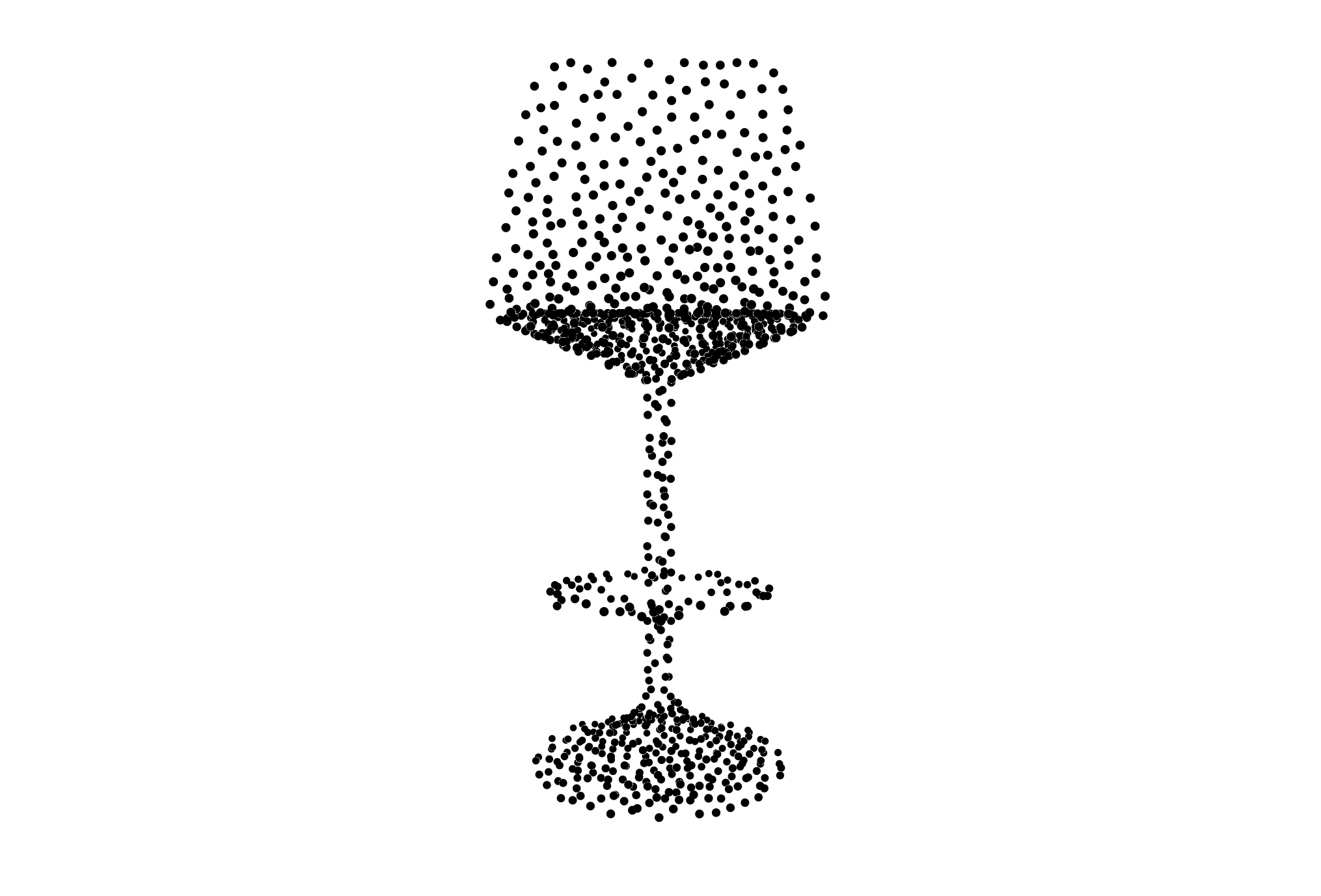} &
\figt[\sizeS]{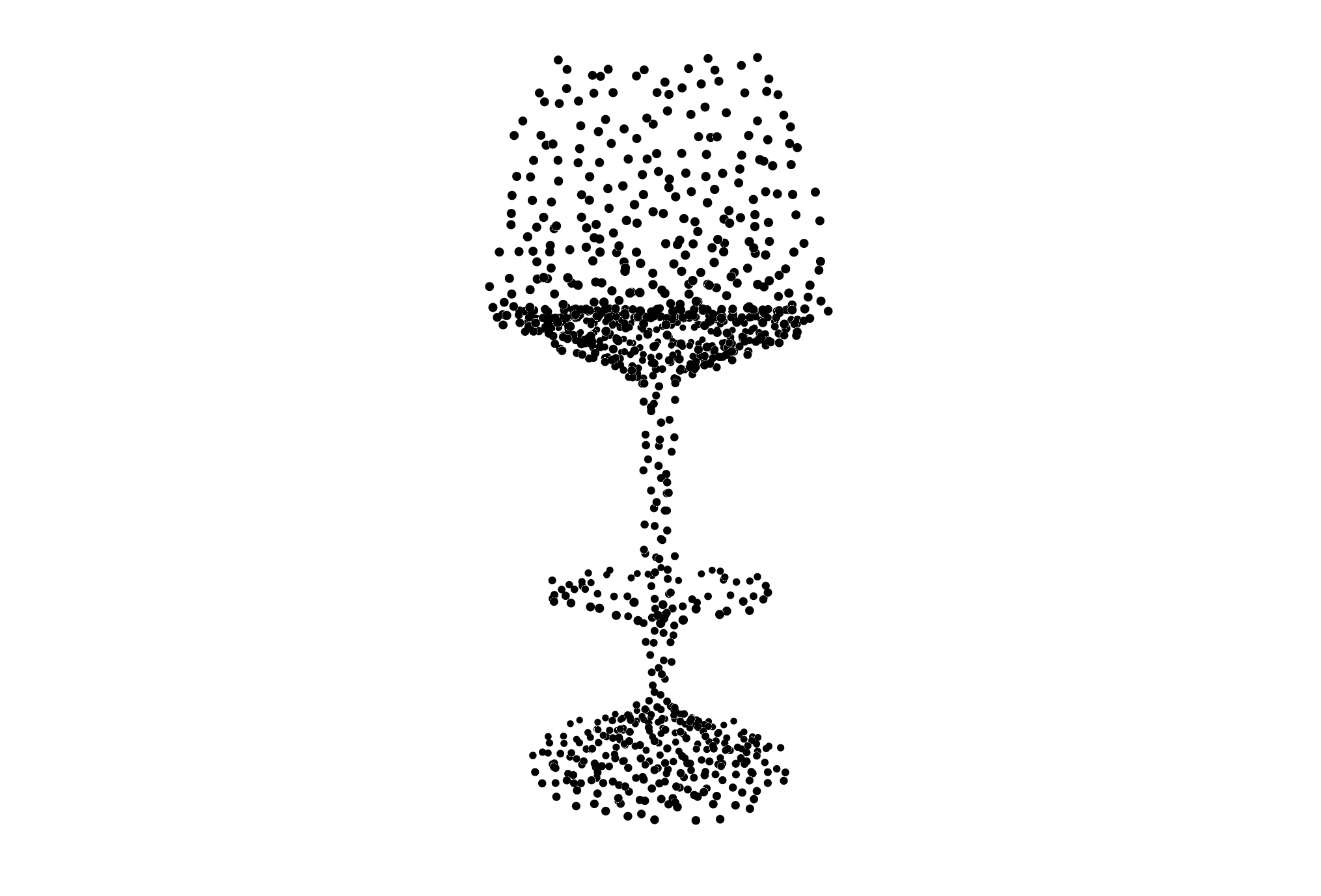} &
\figt[\sizeS]{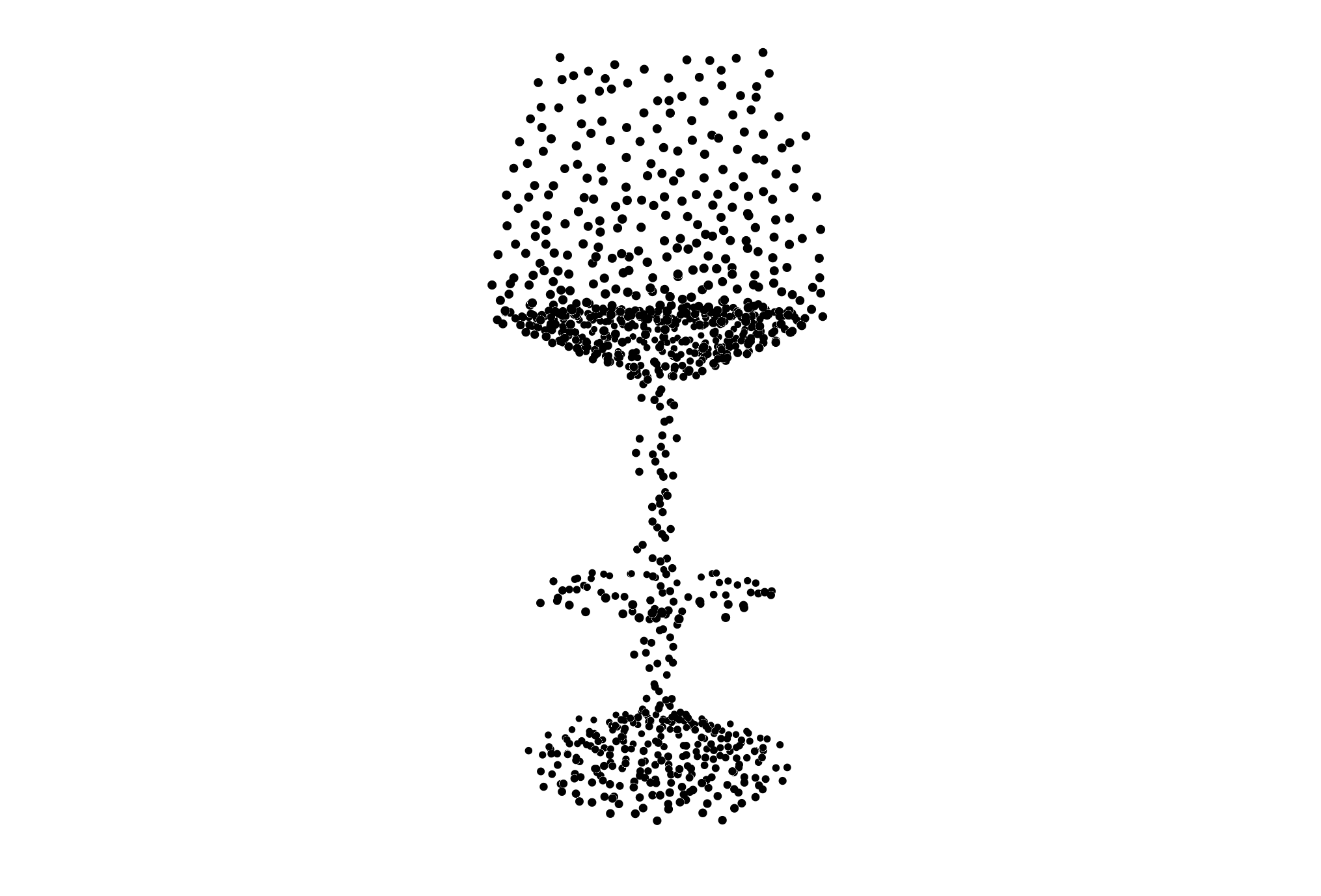} &
\figt[\sizeS]{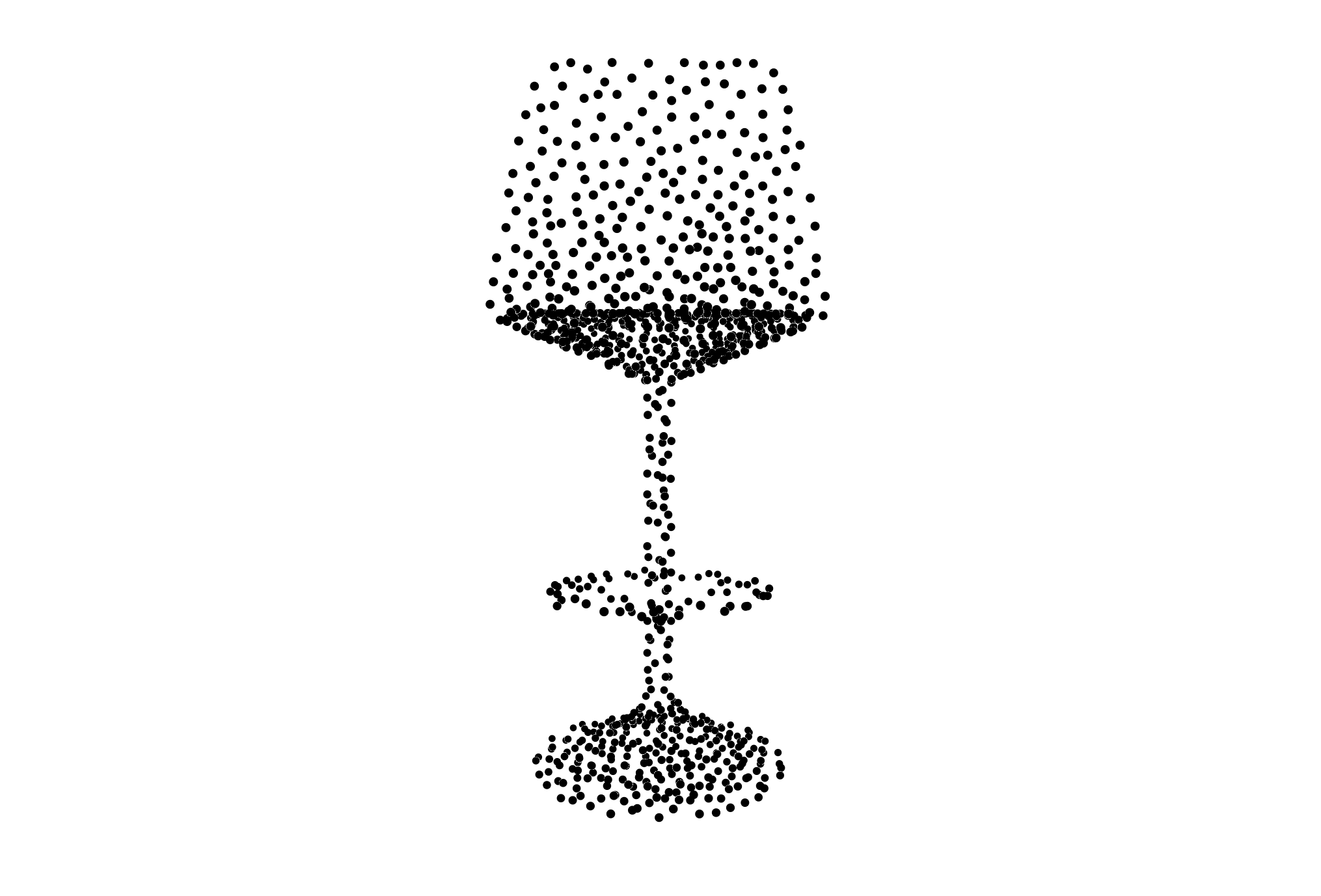} &
\figt[\sizeS]{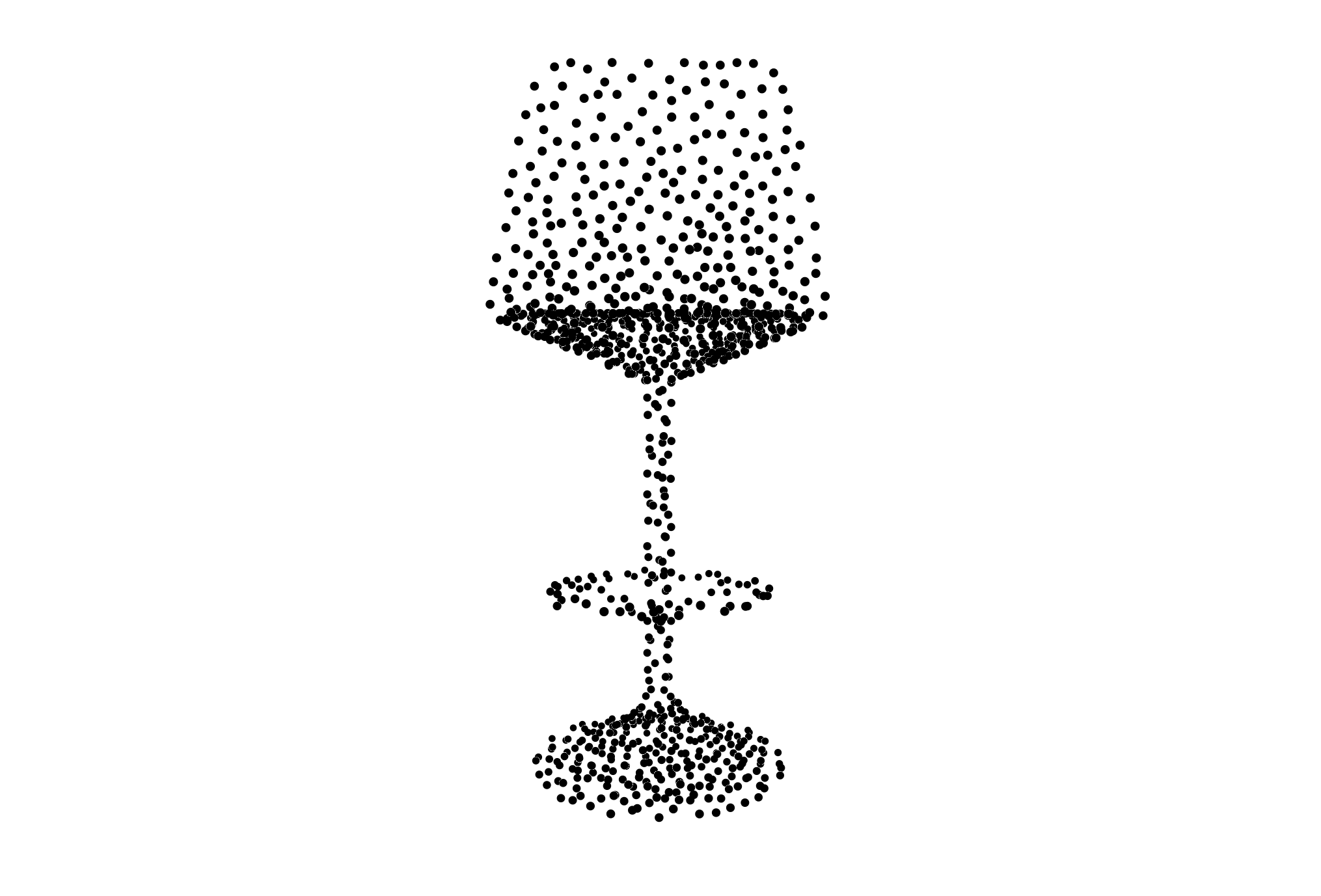} \\
(c) \blue{\cmark chair} &\red{\xmark stool} &\red{\xmark stool} &\red{\xmark stool} &\red{\xmark stool} \\

\figt[\sizeS]{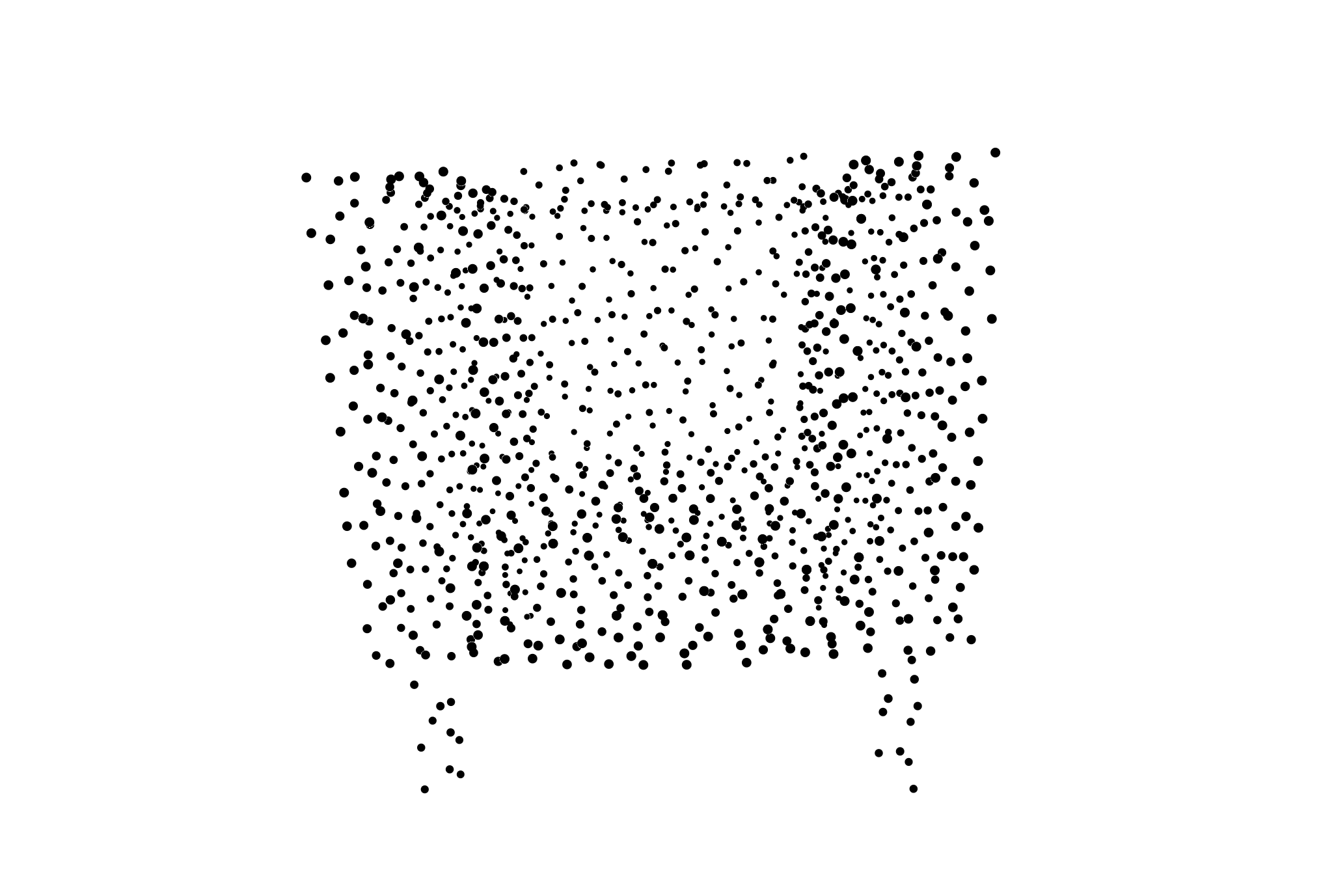} &
\figt[\sizeS]{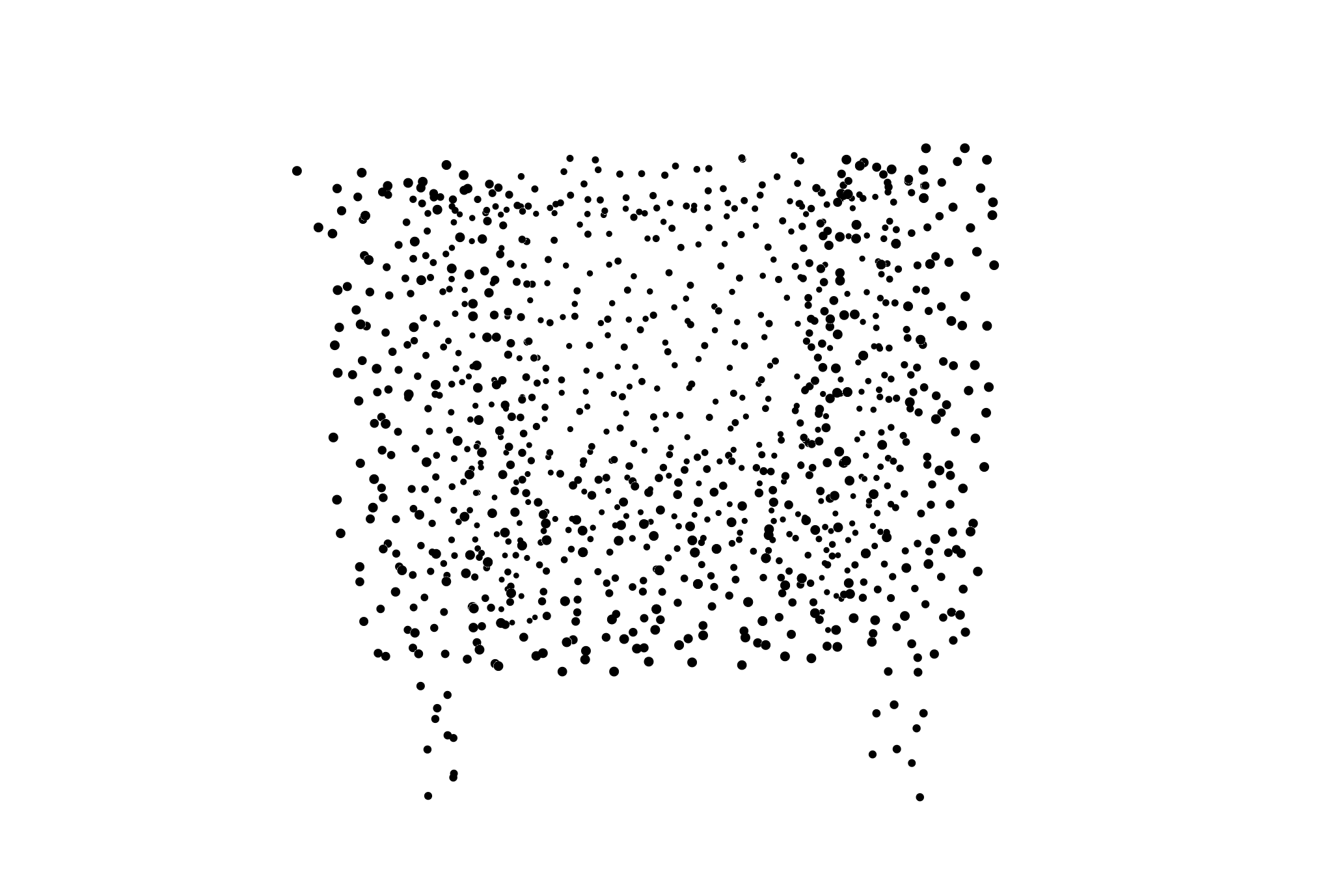} &
\figt[\sizeS]{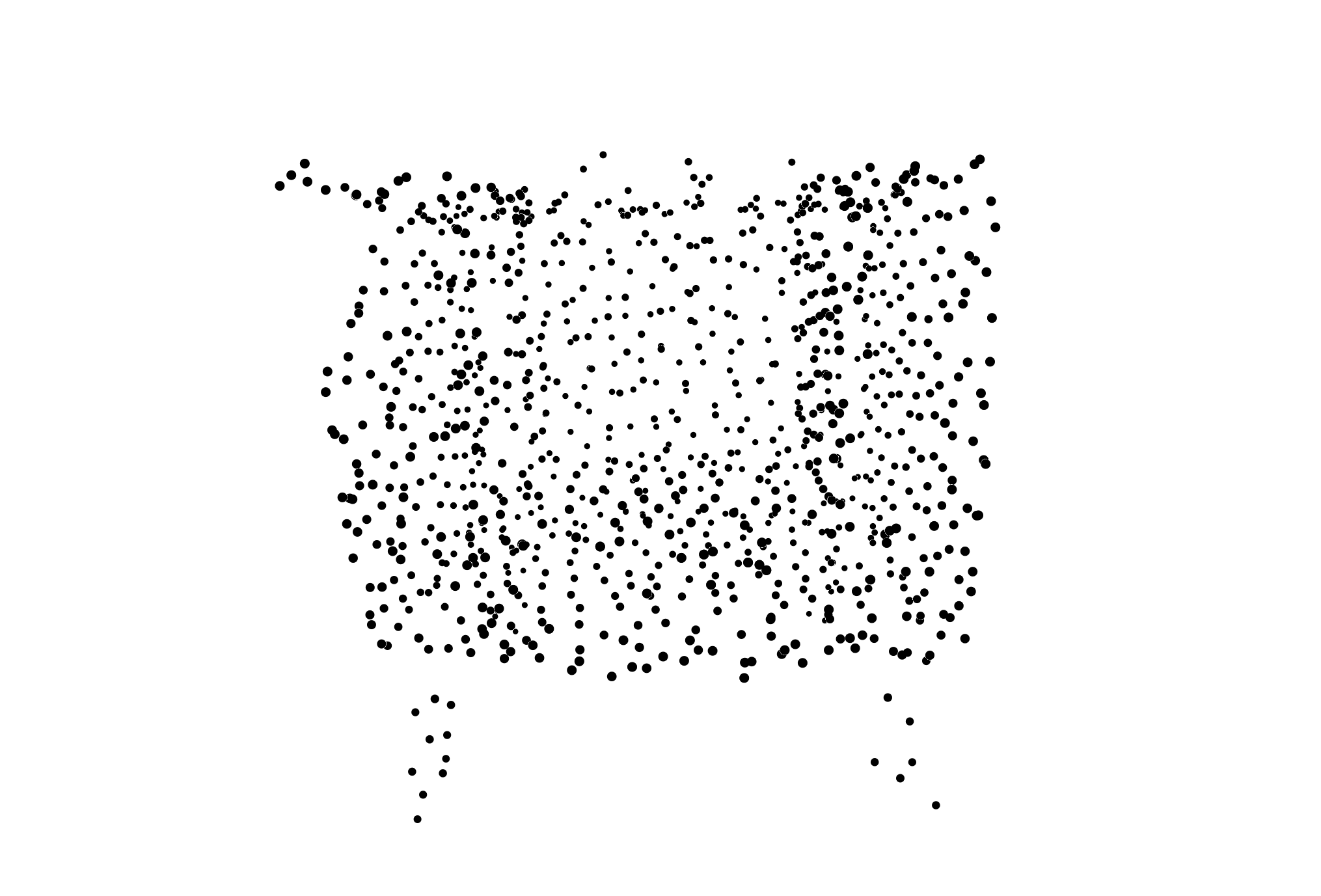} &
\figt[\sizeS]{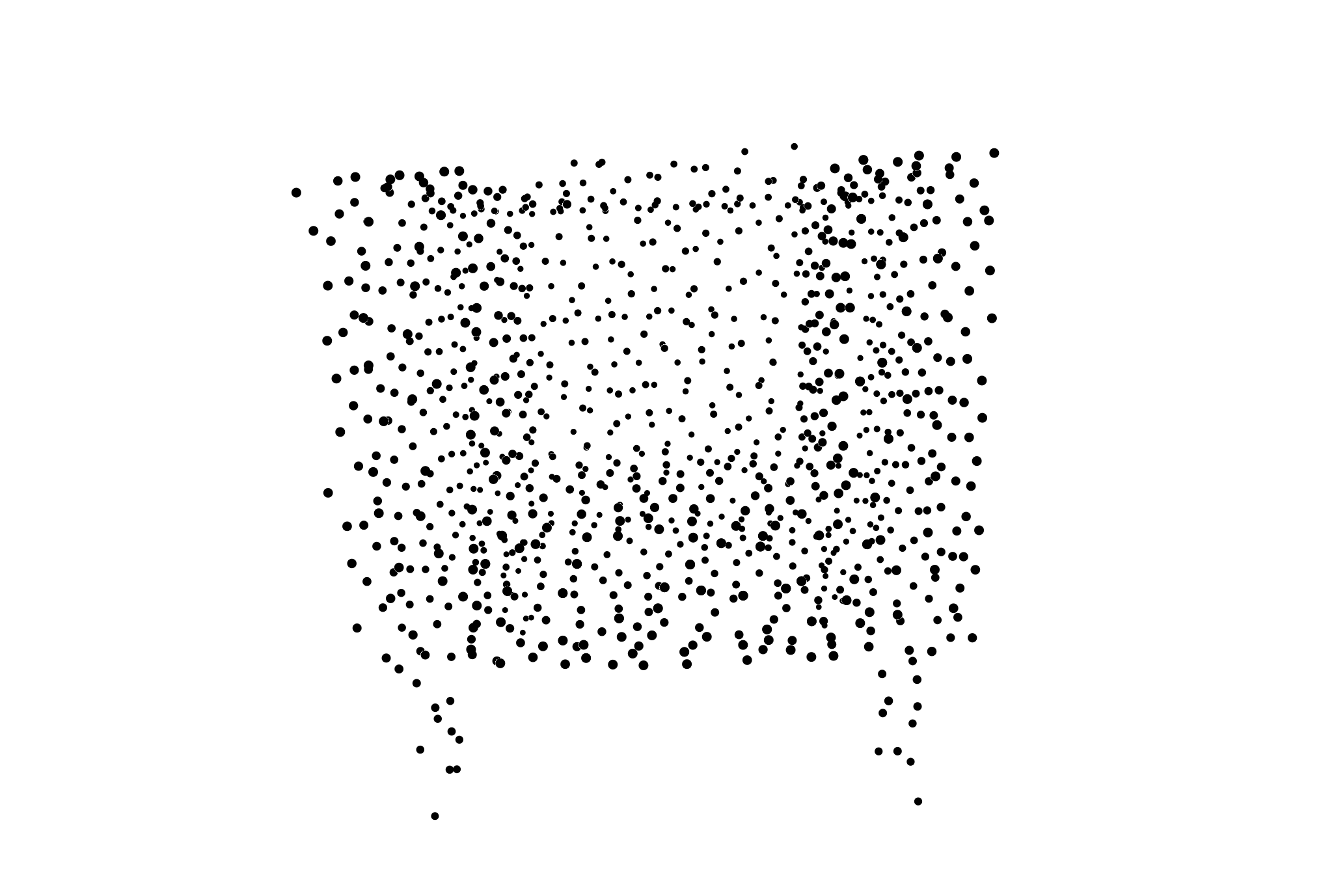} &
\figt[\sizeS]{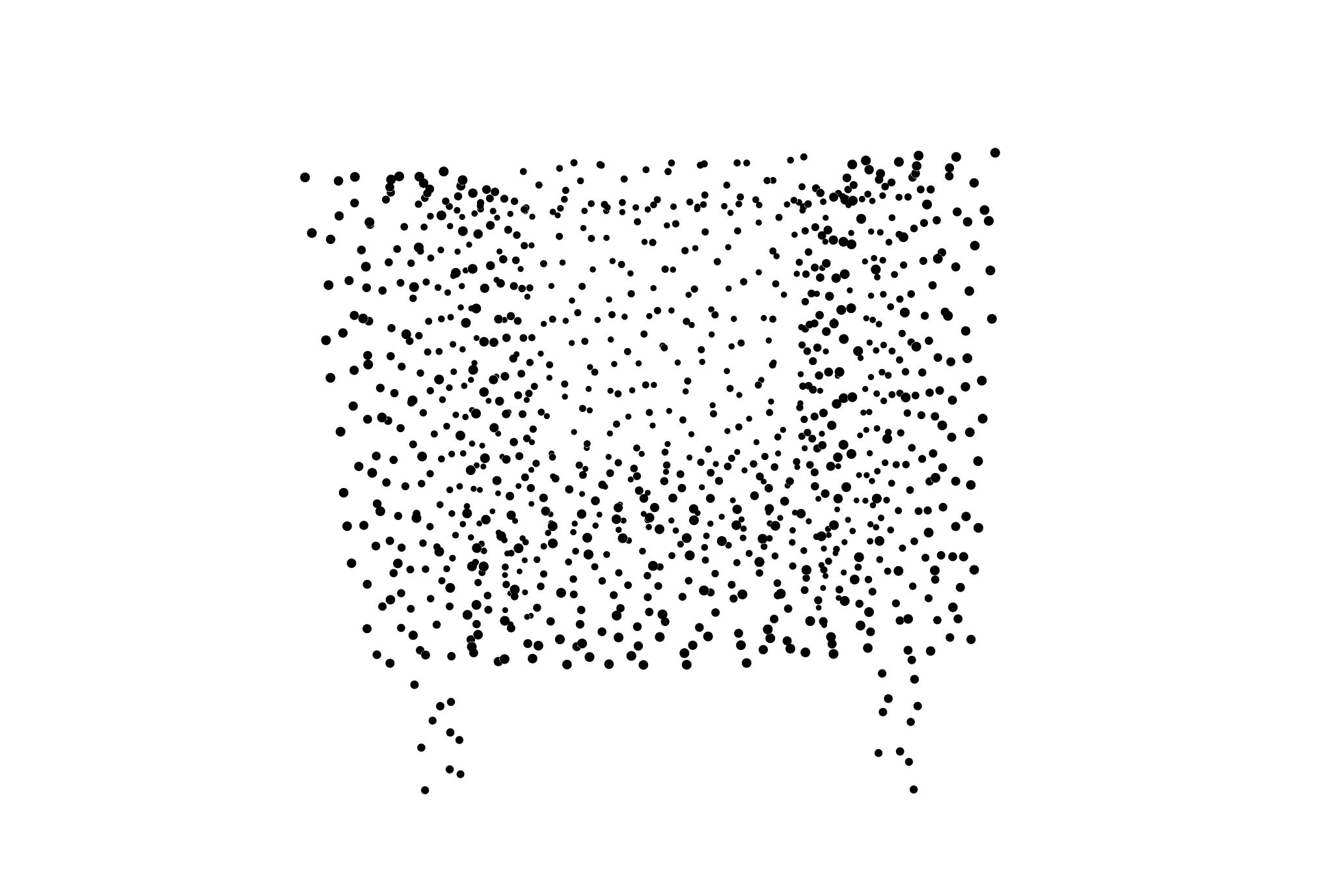} \\
(d) \blue{\cmark sofa} &\red{\xmark desk} &\red{\xmark desk} &\red{\xmark desk} &\red{\xmark desk} \\

\figt[\sizeS]{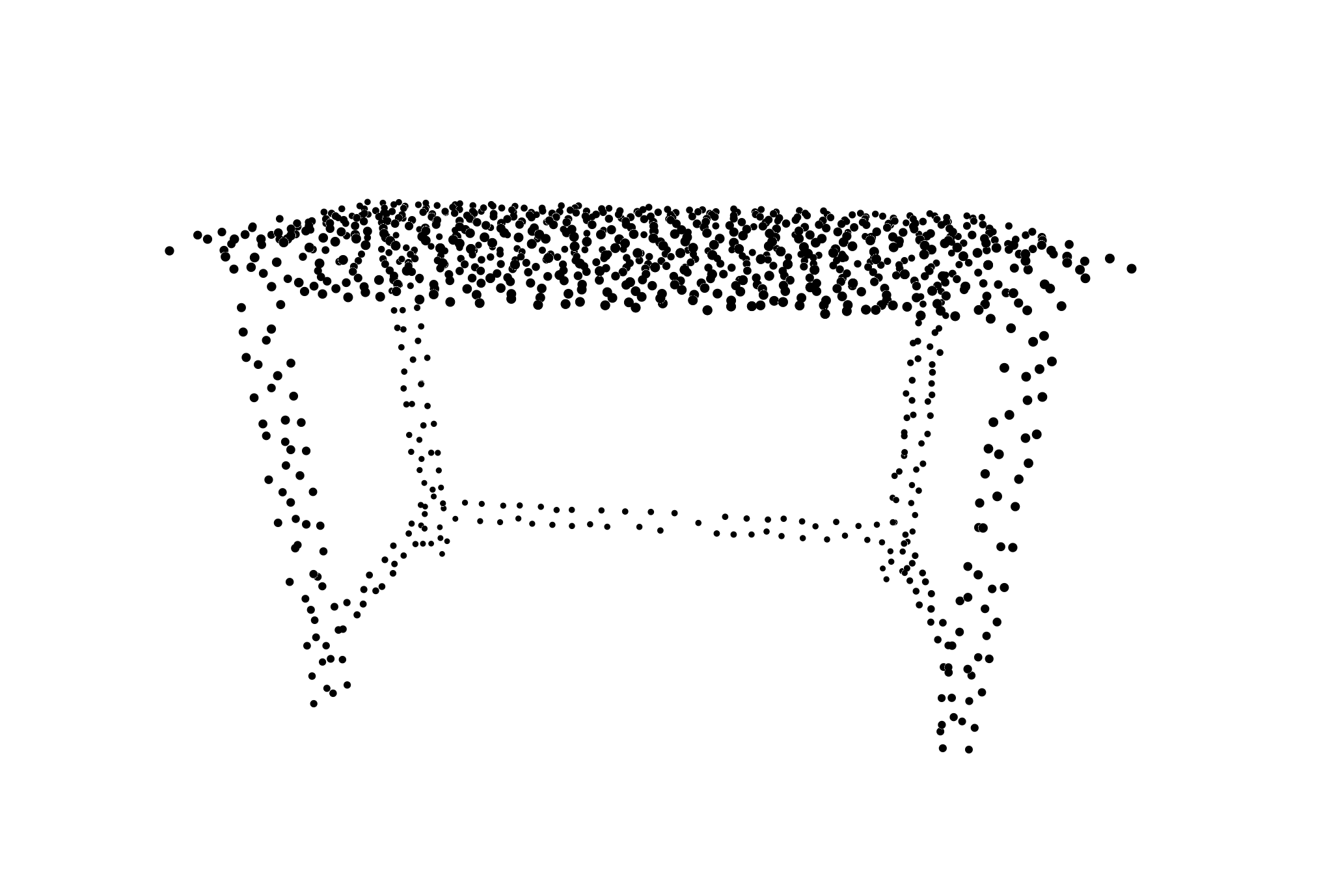} &
\figt[\sizeS]{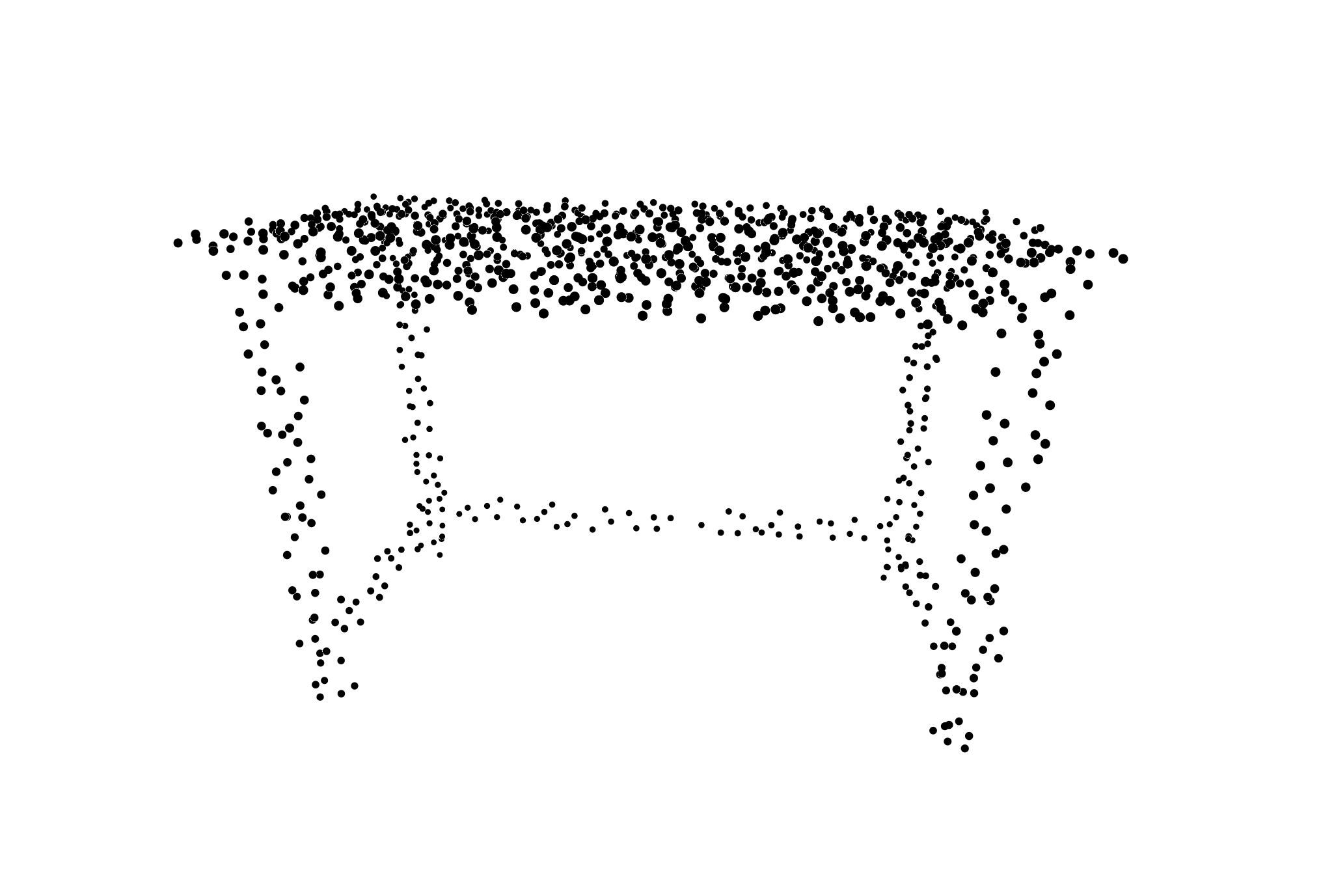} &
\figt[\sizeS]{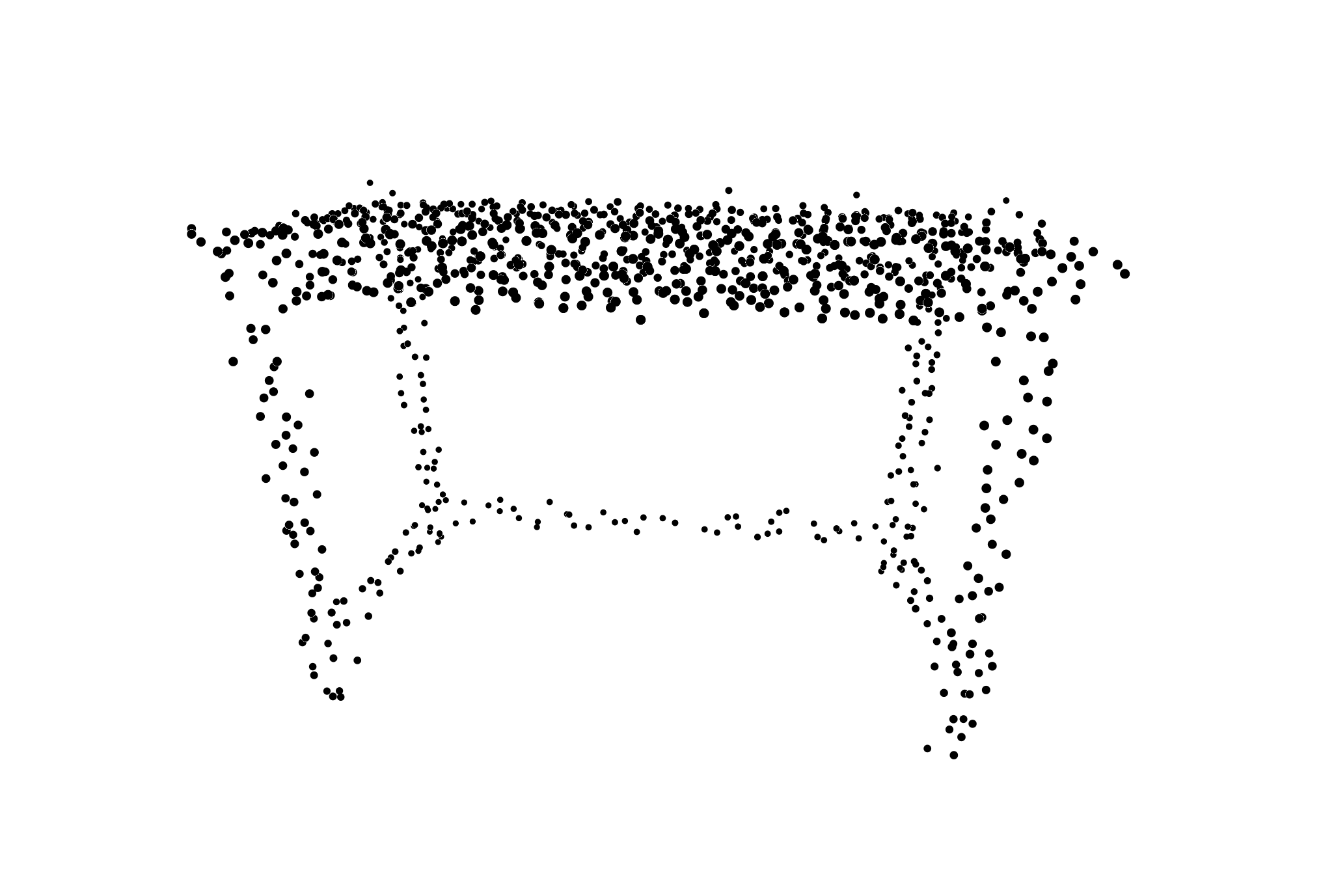} &
\figt[\sizeS]{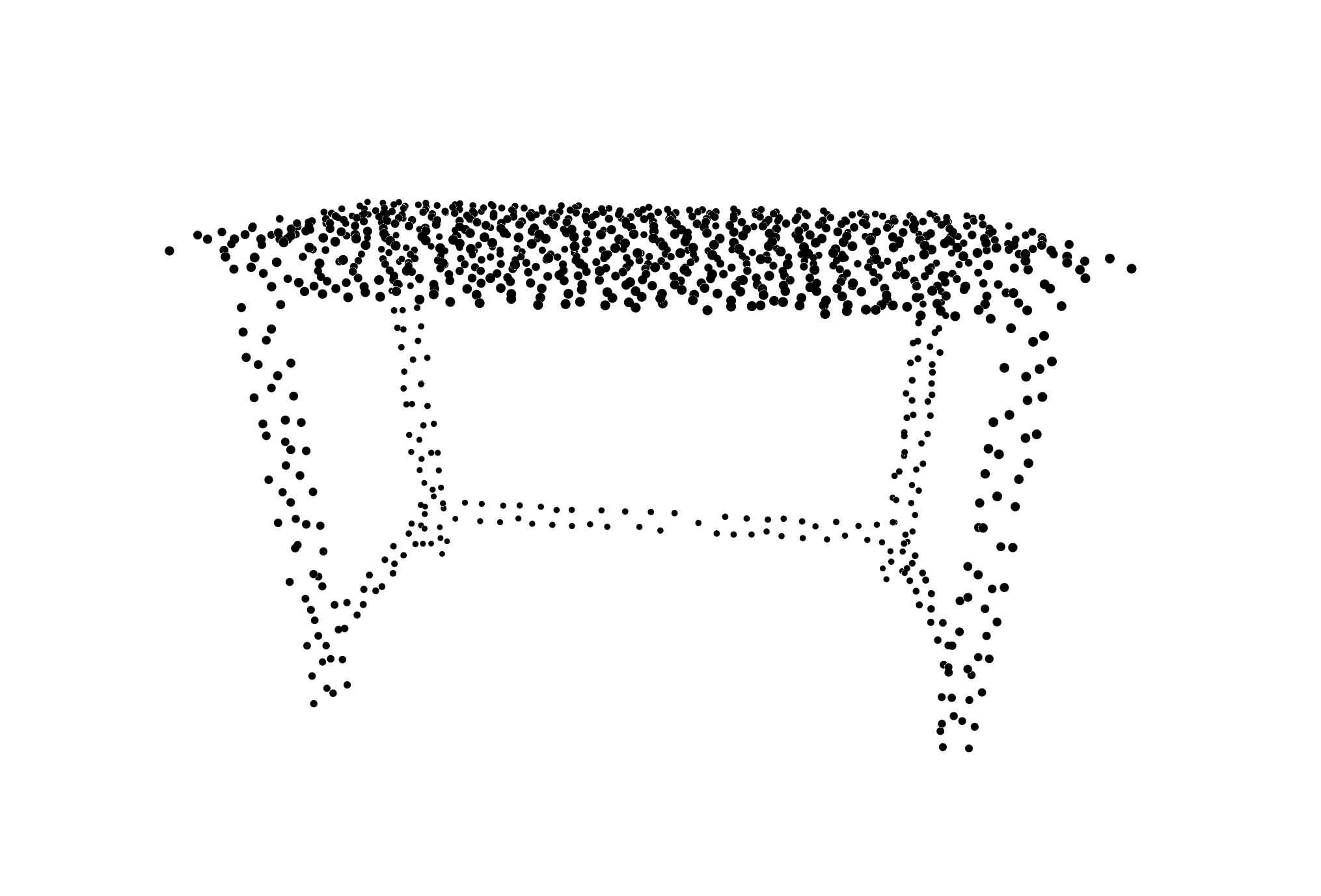} &
\figt[\sizeS]{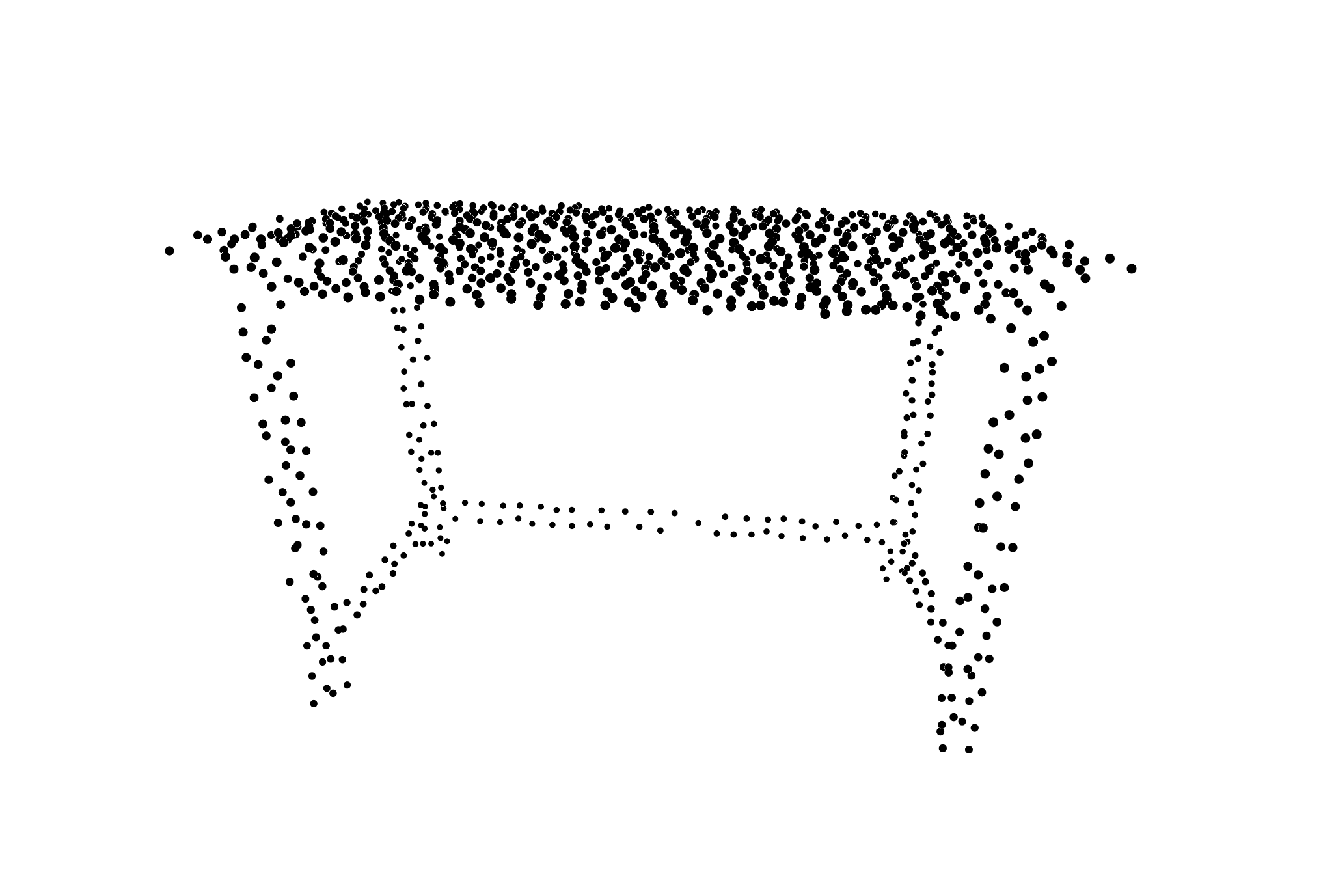} \\
(e) \blue{\cmark table} &\red{\xmark desk} &\red{\xmark desk} &\red{\xmark desk} &\red{\xmark desk} \\

\end{tabular}
\caption{Visualization for adversarial point clouds produced by baseline methods on Point-Transformer.}
\label{fig:vis-3d-pointnetTrans-supp}
\end{figure*}

We provide more visualization on PointNet (Figure~\ref{fig:vis-3d-pointnet-supp}), PointNet++ (Figure~\ref{fig:vis-3d-pointnet2-supp}), DGCNN (Figure~\ref{fig:vis-3d-dgcnn-supp}) and Point-Tranformer (Figure~\ref{fig:vis-3d-pointnetTrans-supp}). We observe that adversarial examples generated by \geoa and \gsda are very visible. \siadv trend to generate outline points. DGCNN and Point-Transformer are more difficult to attack than PointNet and PointNet++.

\section{Black-box attack}

Here, we provide more details on our black-box attack. As Algorithm~\ref{alg:black-box} shown, given an original point cloud $\cX$, a true label $t$, a set of imperceptibility regularization metrics $\cD$, a black-box model $f_b$ and the surrogate model $f_s$, we calculate the adversarial example $\cY^*$ via a query-based black-box attack. We first estimate the sensitivity map via the gradient of the surrogate model (\cf \emph{Line 4}). Since the sensitivity map is calculated under the coordinate transformation, we denote the coordinate transformation as $T_{rsi}$ and the reverse coordinate transformation as $T'_{rsi}$.
The function $SM(\vg) \defn \{ \mathbf{s}_i\}$, where $\mathbf{s}_i \defn \sqrt{g_{i1}^2 + g_{i2}^2}$. 
$Sort(\cS)$ in \emph{Line 5} means sort all the sensitivity map values of each point descendingly. 
When we do not find an adversarial example for the black-box model and we do not change every point, then we keep searching. At each iteration, we pick the top-ranked value from the sensitivity map $\cS$ as depicted in \emph{Line 7}, \ie $\cS(0)$, and then drop it from $\cS$ (\cf \emph{Line 8}). Similar to our white-box version, at each query step, the algorithm searches along the gradient for misclassification estimated in the surrogate model (\cf \emph{Line 9-12}), and minimizes the distortion along its orthogonal directions while reducing the imperceptibility regularization terms (\cf \emph{Line 13-22}).

\begin{figure*}[hptb]
\newcommand{\sizeS}{.20}
\centering
\setlength{\tabcolsep}{0pt}
\begin{tabular}{ccccc}
\textbf{Clean} &\simba &\simbapp &\siadv & \projabbrv\\
\figt[\sizeS]{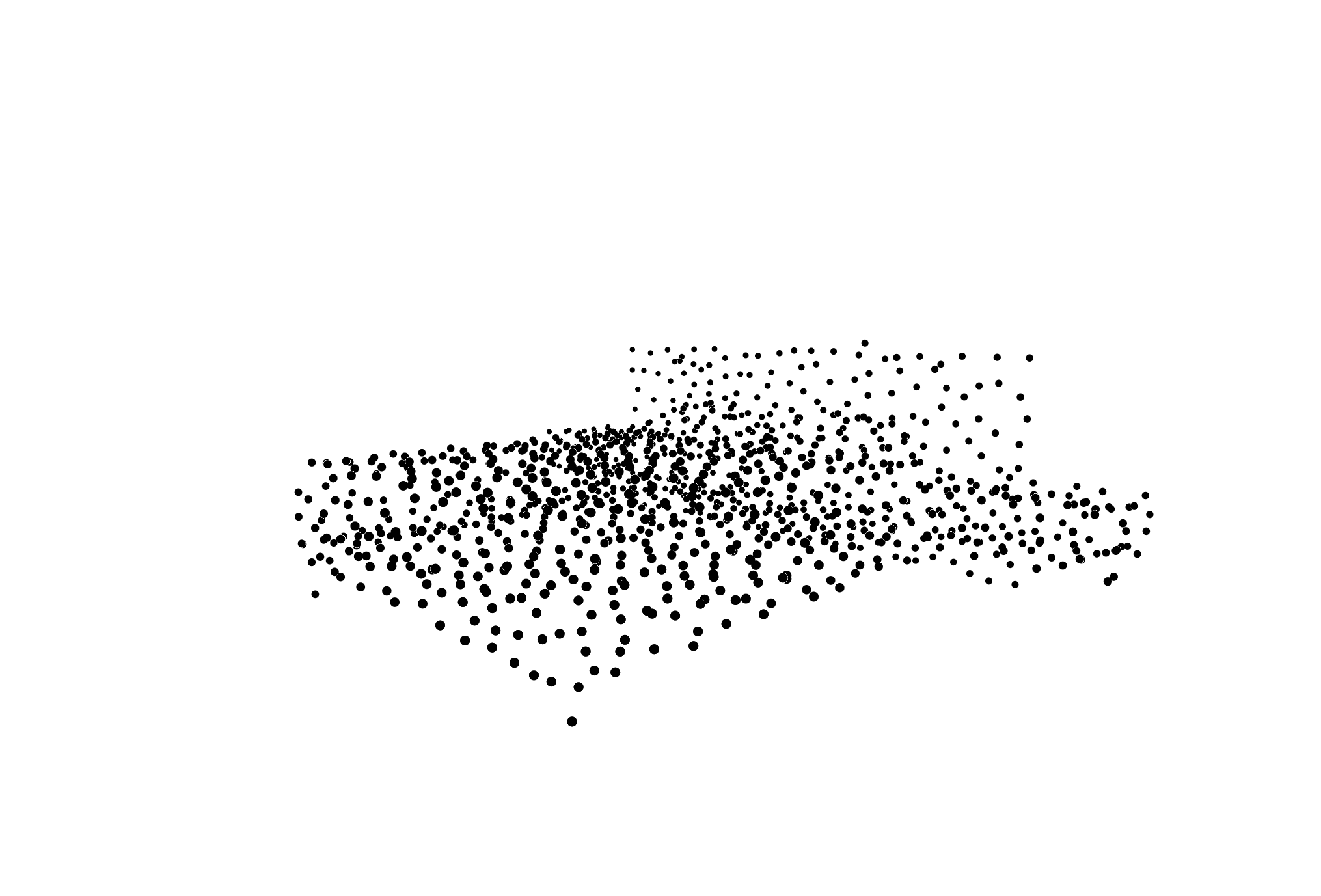} &
\figt[\sizeS]{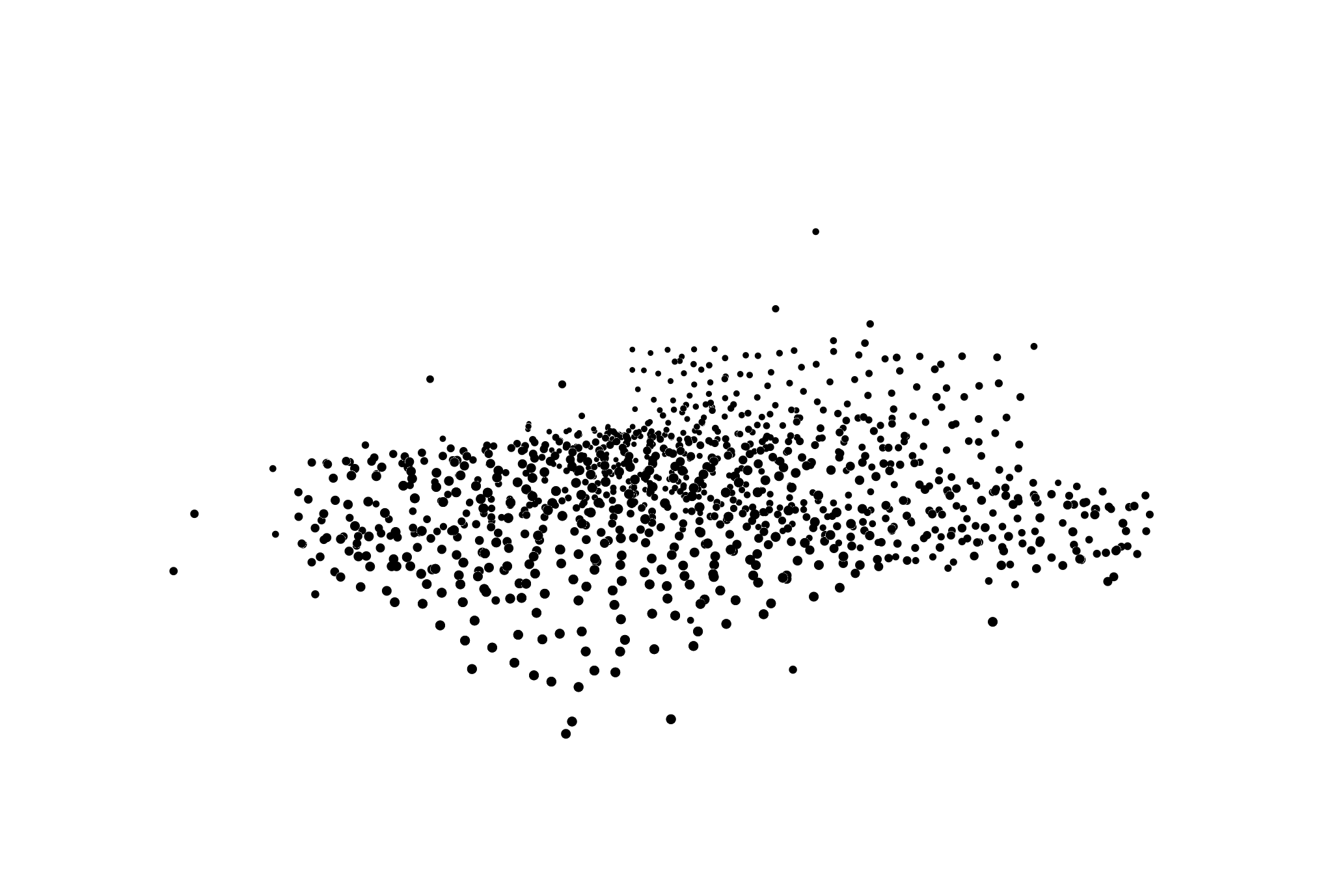} &
\figt[\sizeS]{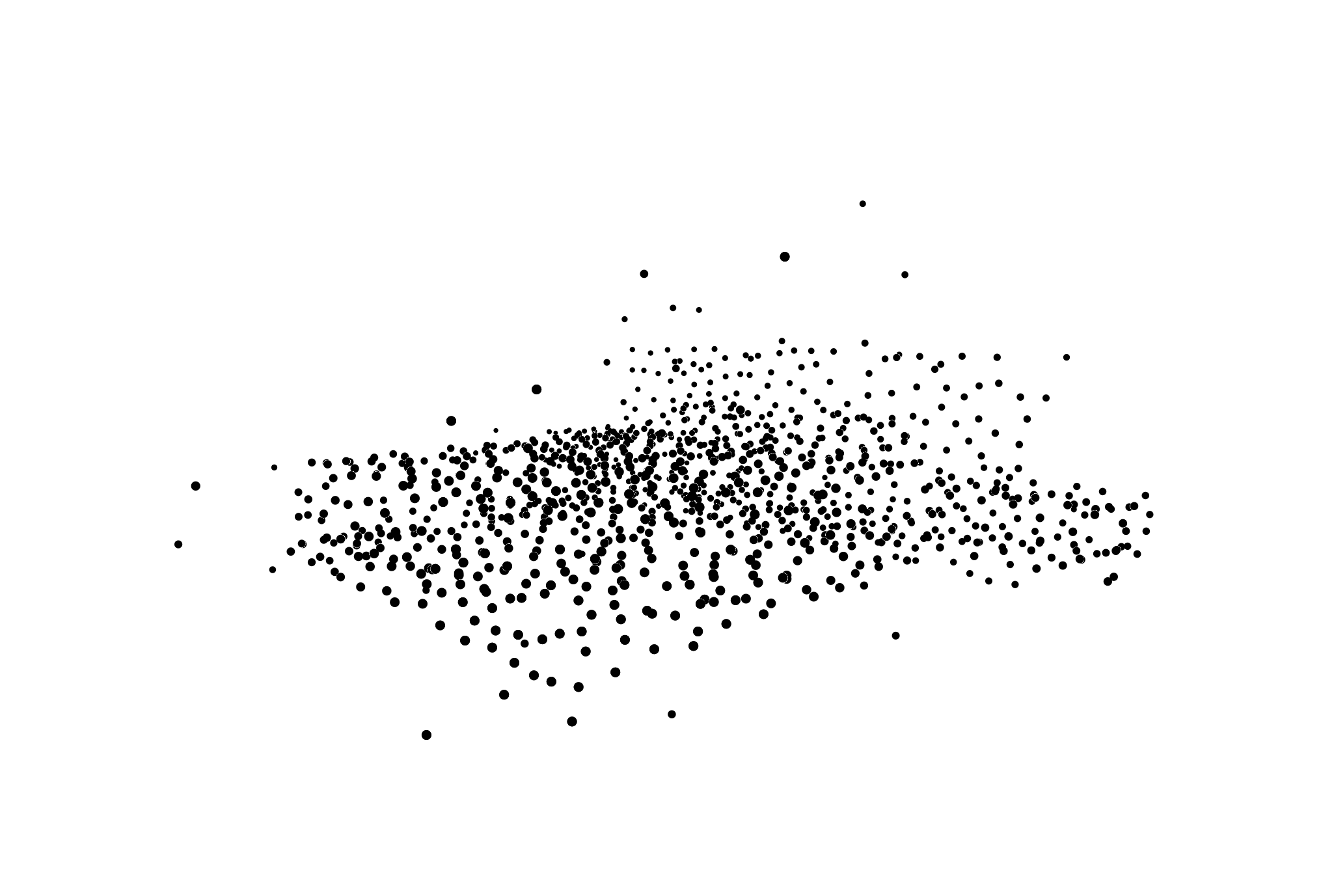} &
\figt[\sizeS]{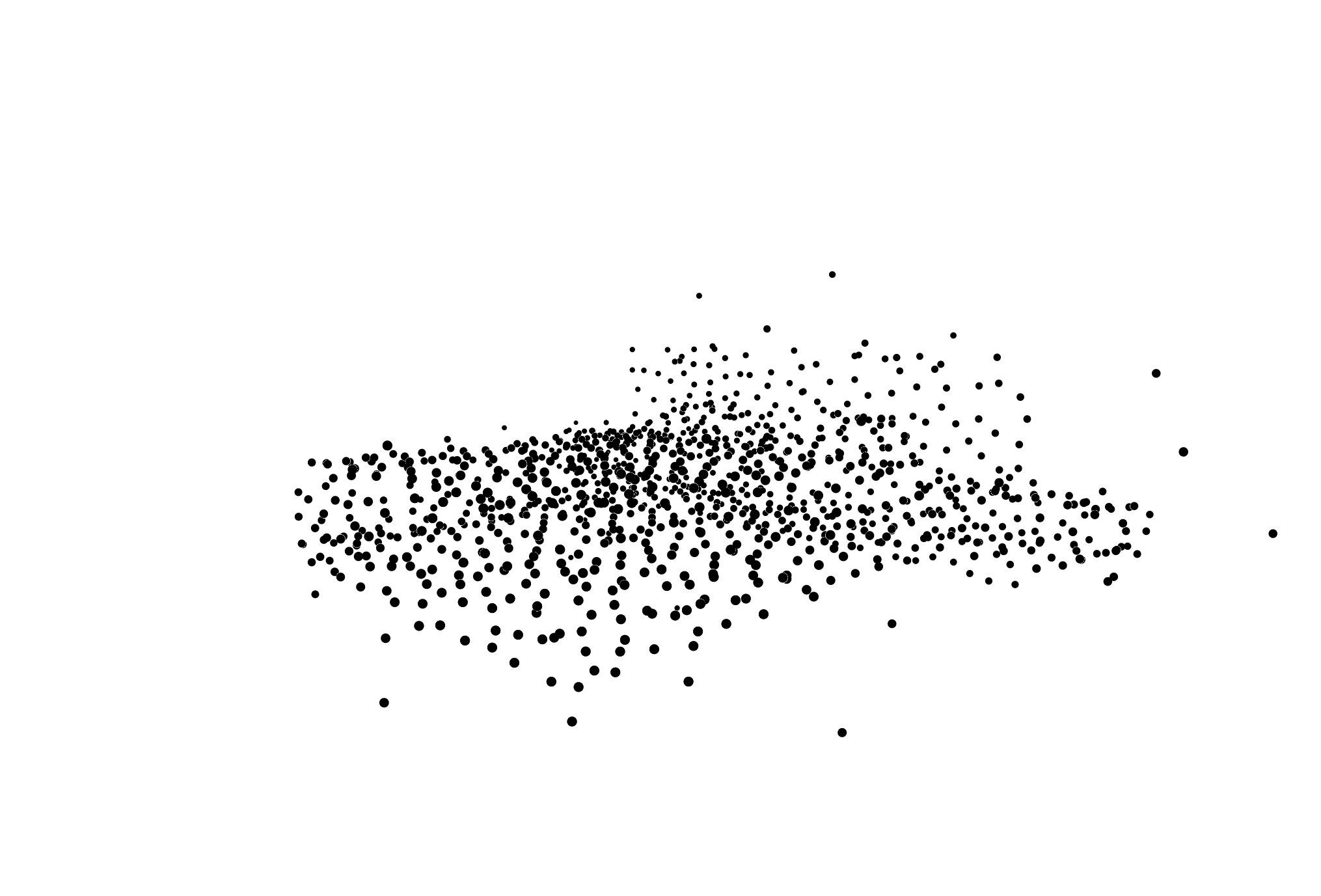} &
\figt[\sizeS]{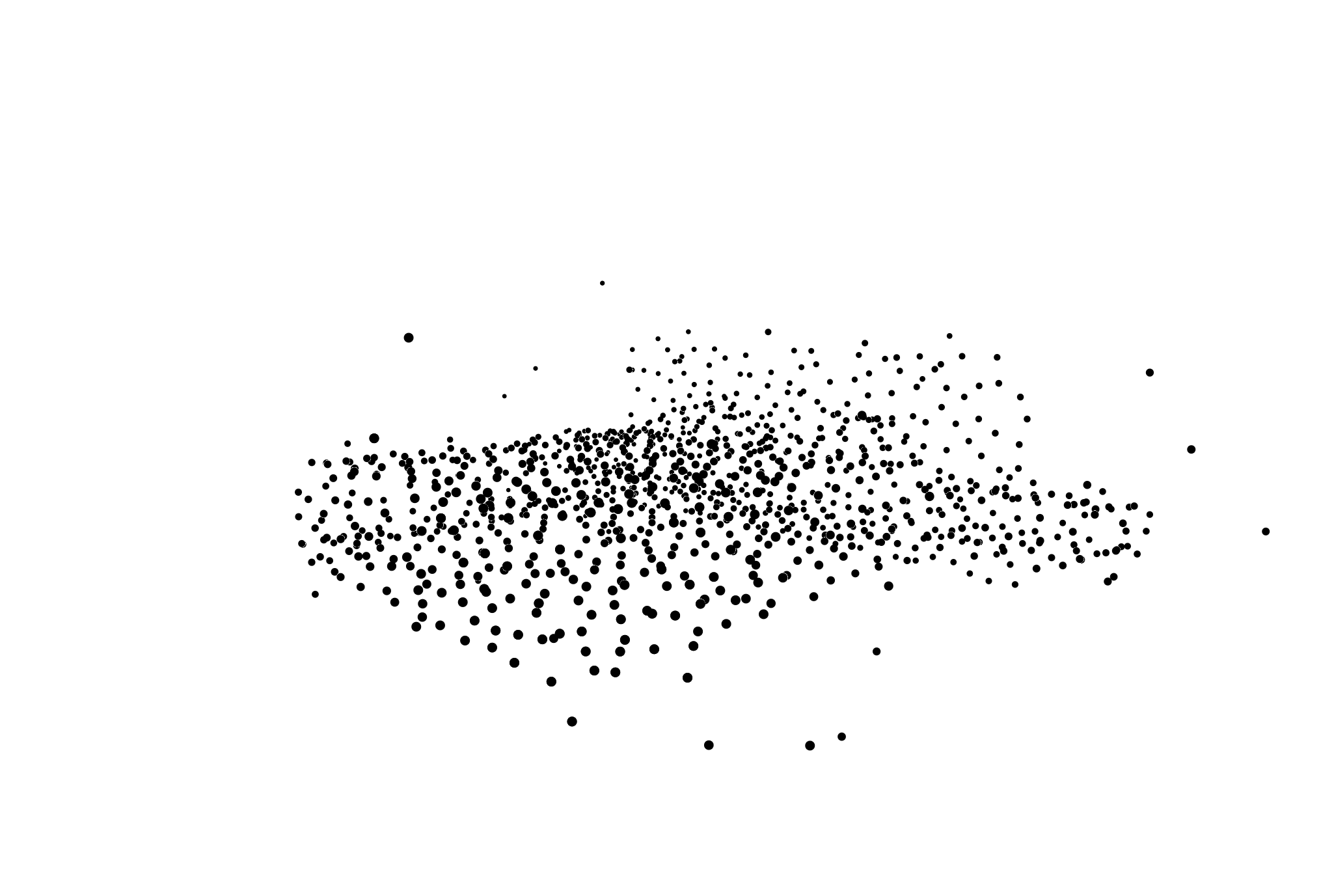} \\
(a) \blue{\cmark bed} &\red{\xmark tent} &\red{\xmark tent} &\red{\xmark plant} &\red{\xmark plant} \\

\figt[\sizeS]{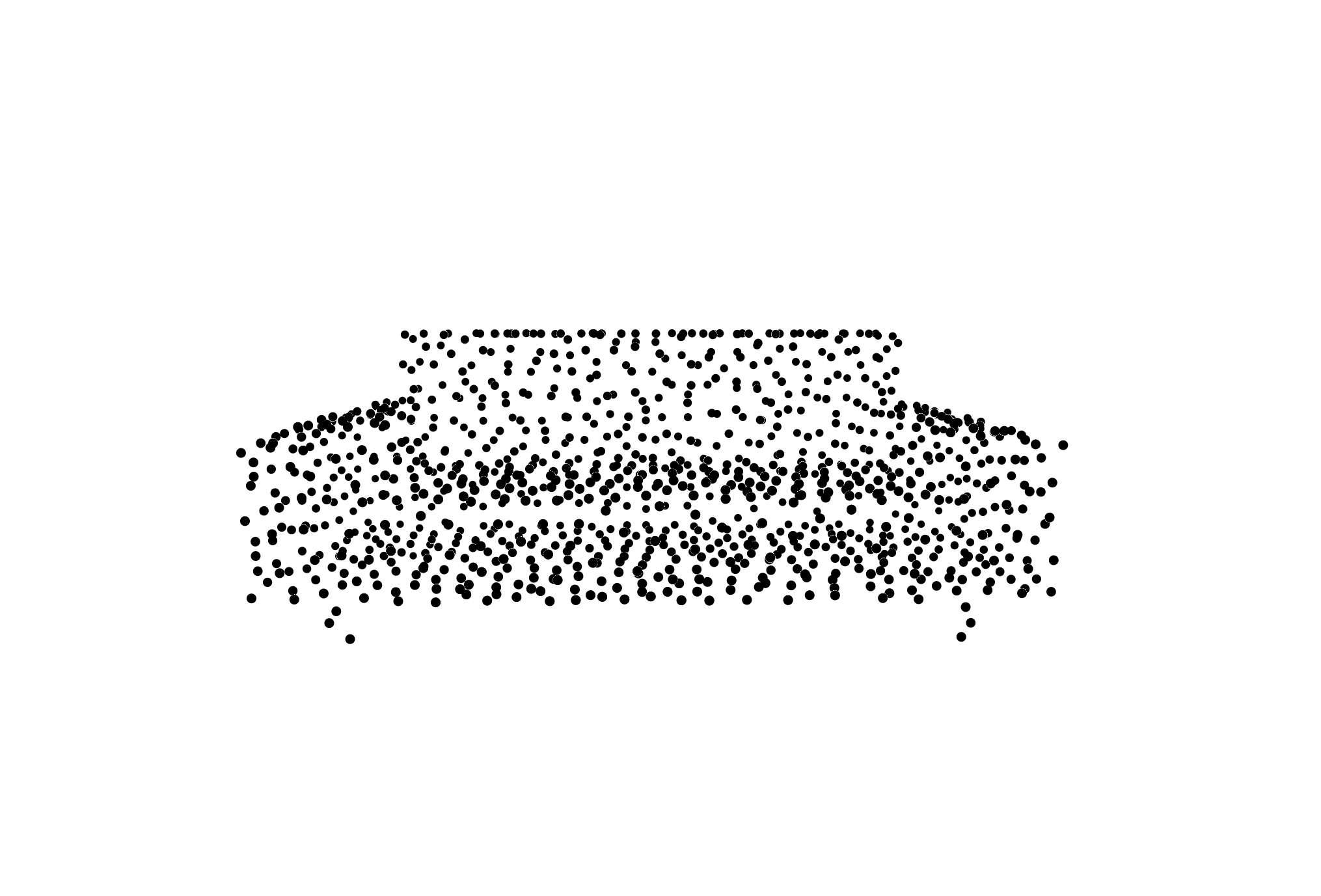} &
\figt[\sizeS]{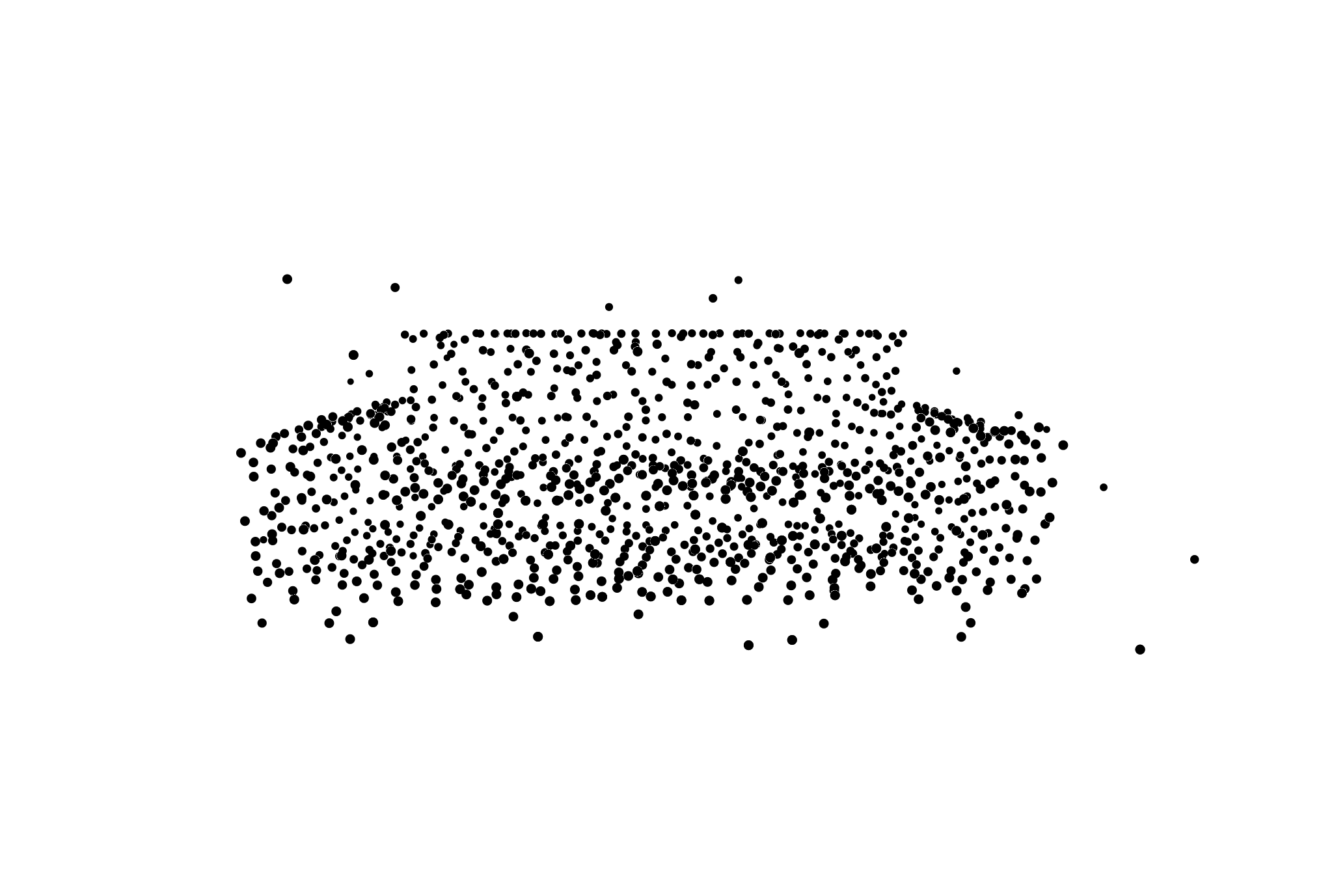} &
\figt[\sizeS]{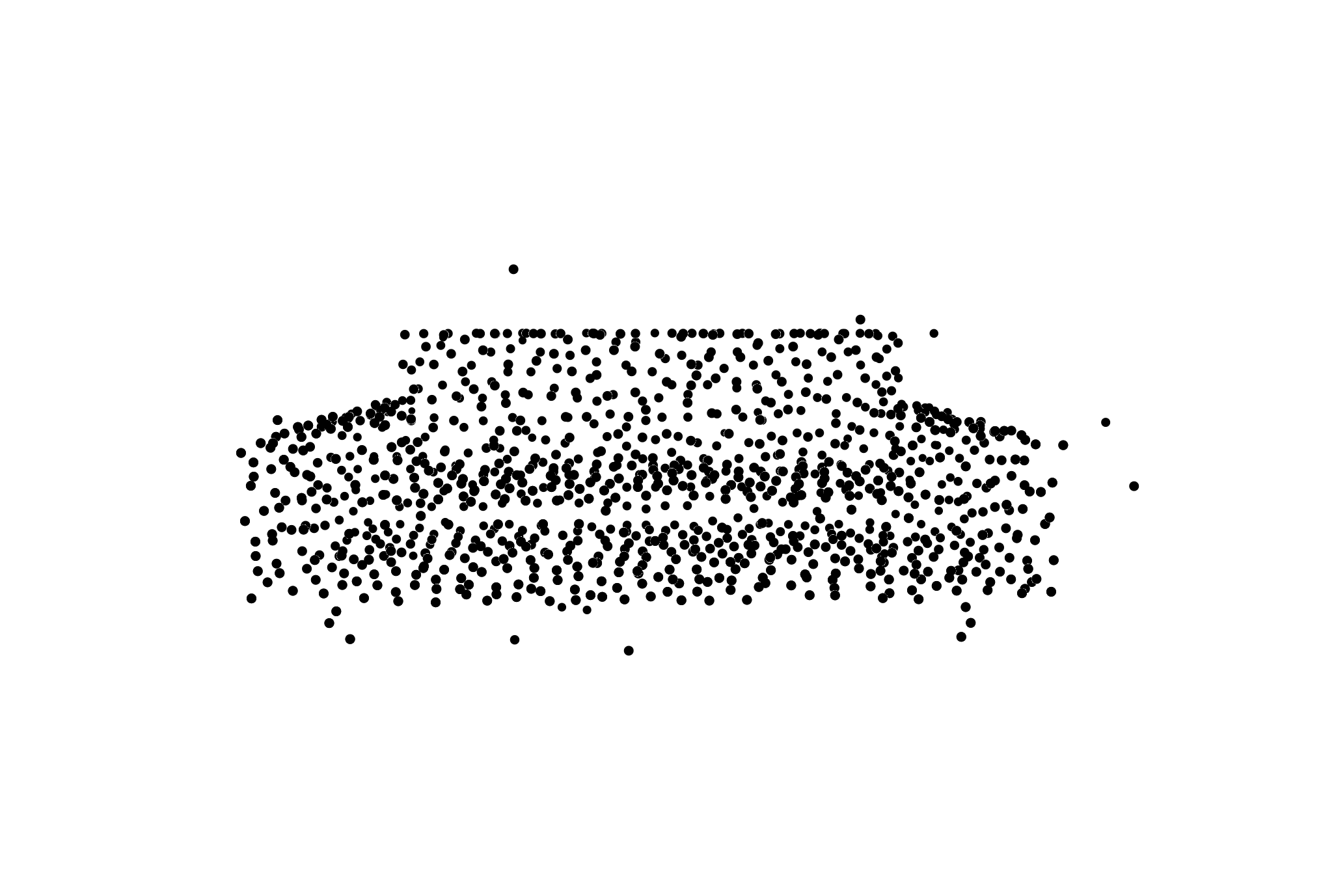} &
\figt[\sizeS]{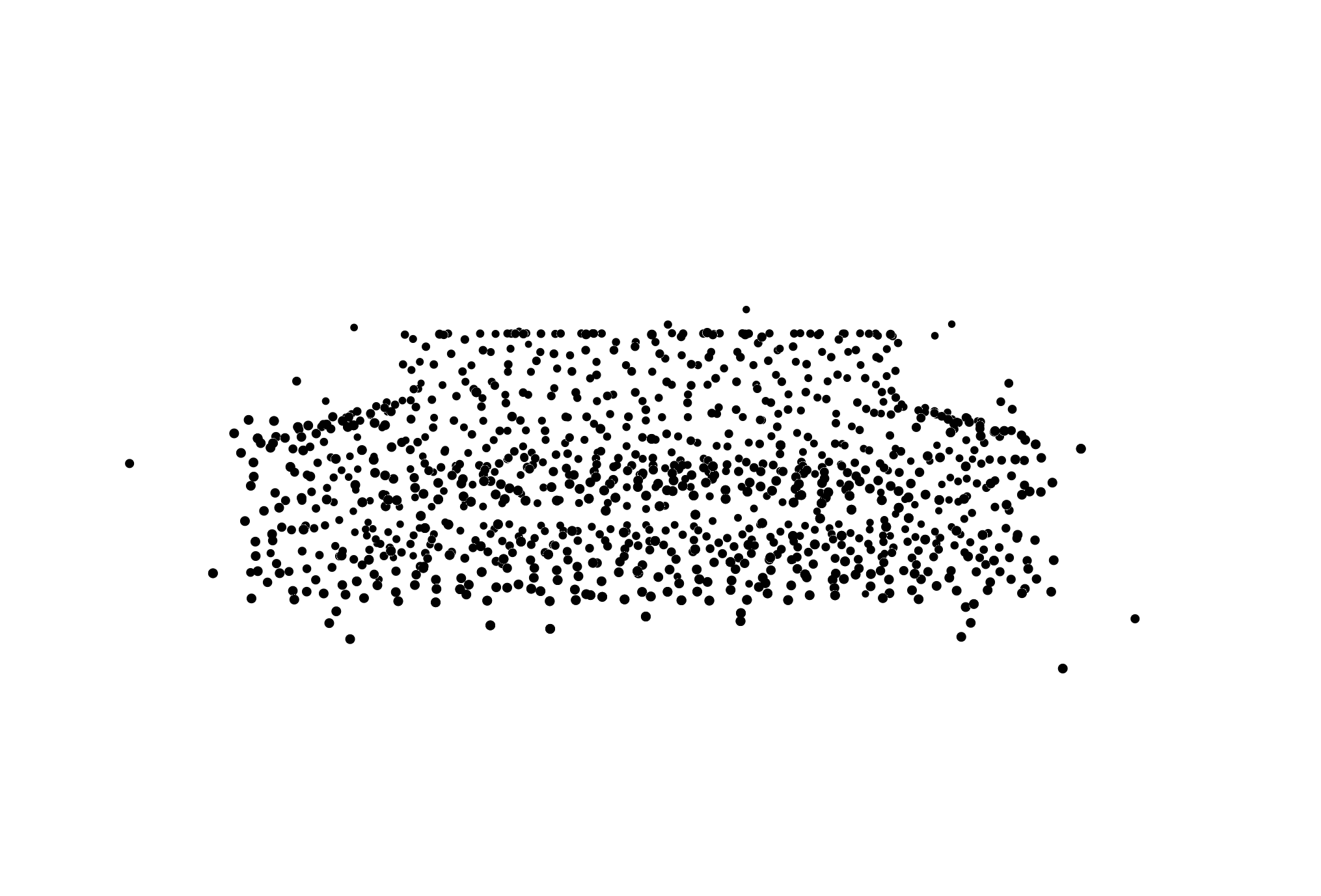} &
\figt[\sizeS]{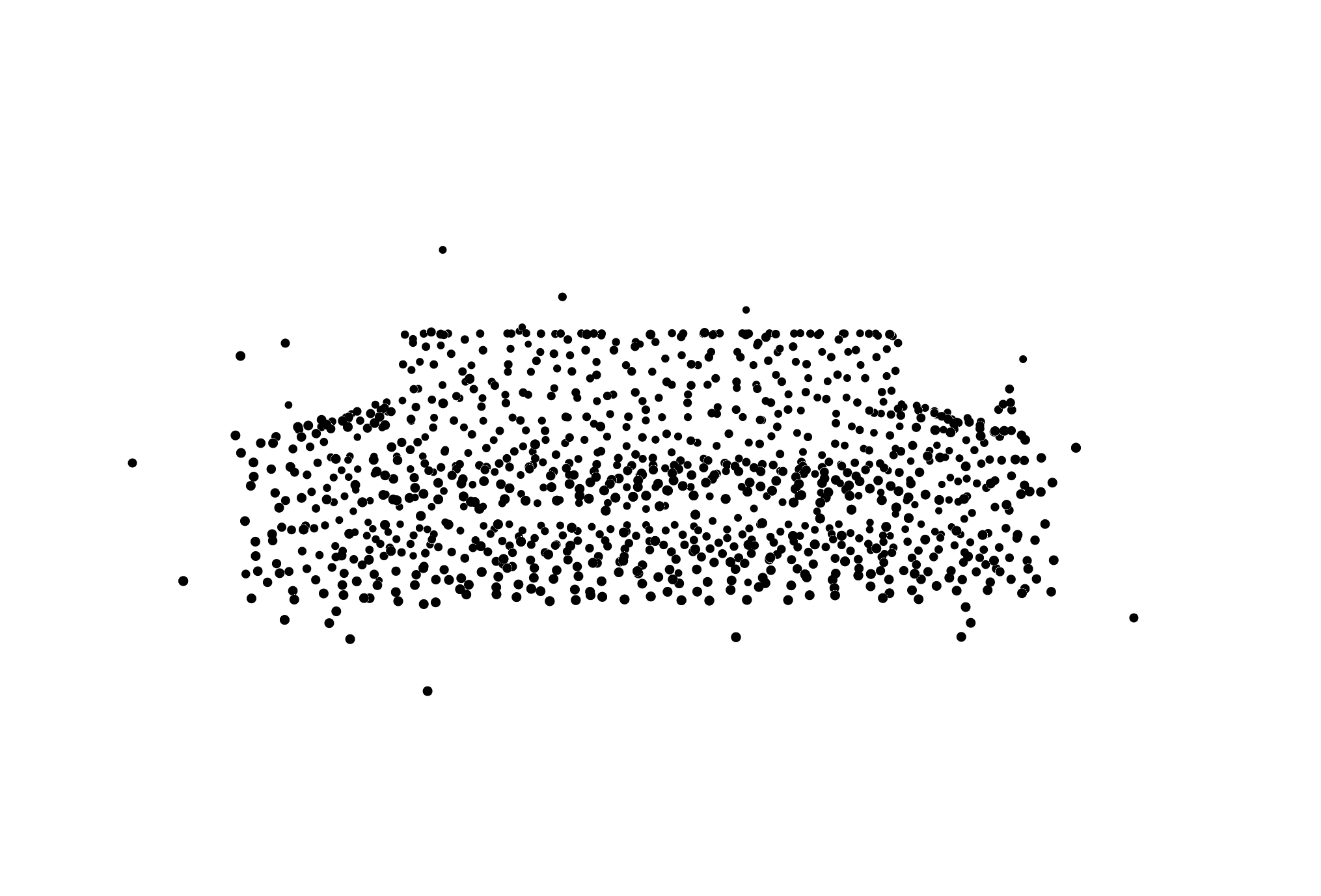} \\
(b) \blue{\cmark sofa} &\red{\xmark plant} &\red{\xmark tv stand} &\red{\xmark dresser} &\red{\xmark dresser} \\

\figt[\sizeS]{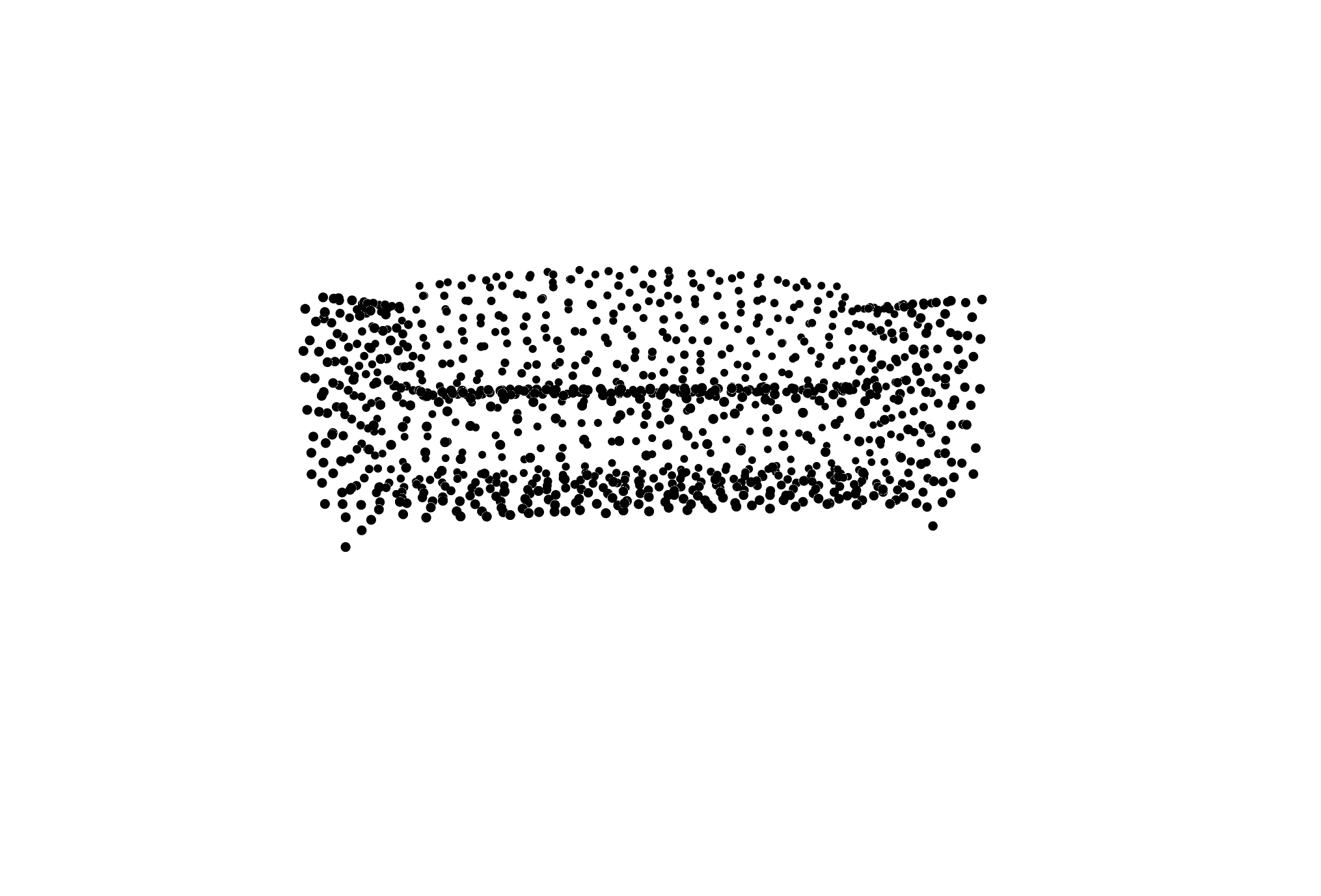} &
\figt[\sizeS]{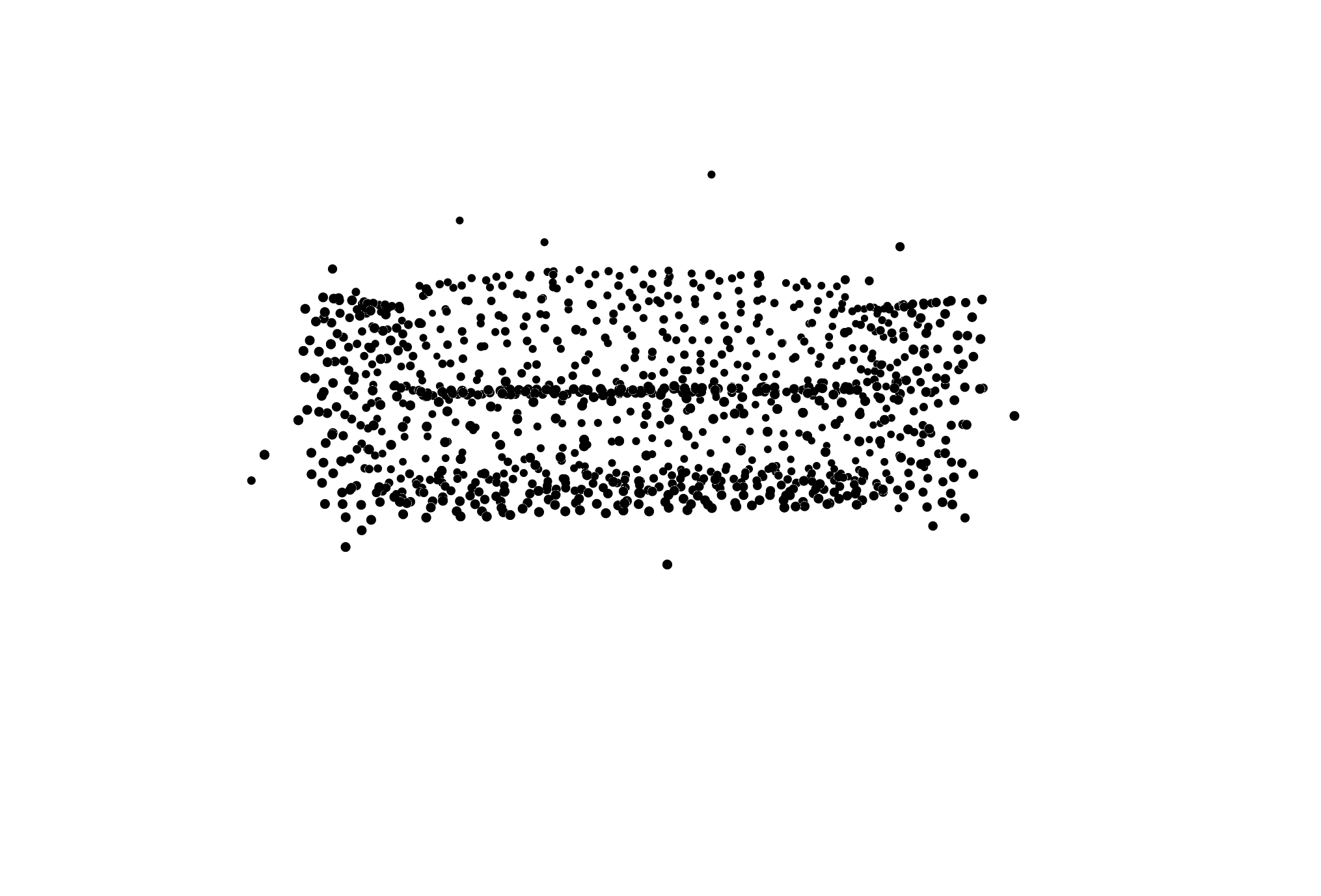} &
\figt[\sizeS]{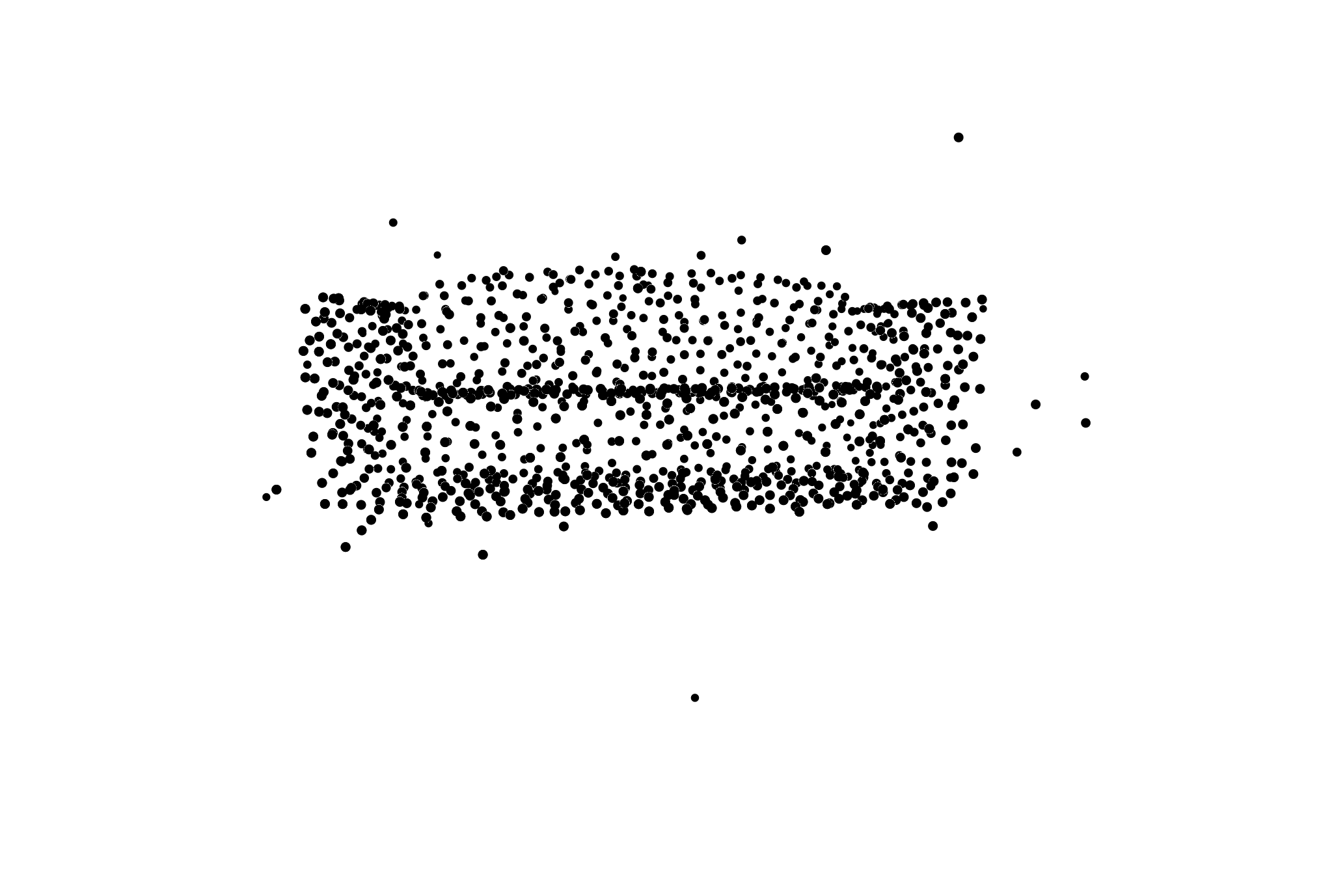} &
\figt[\sizeS]{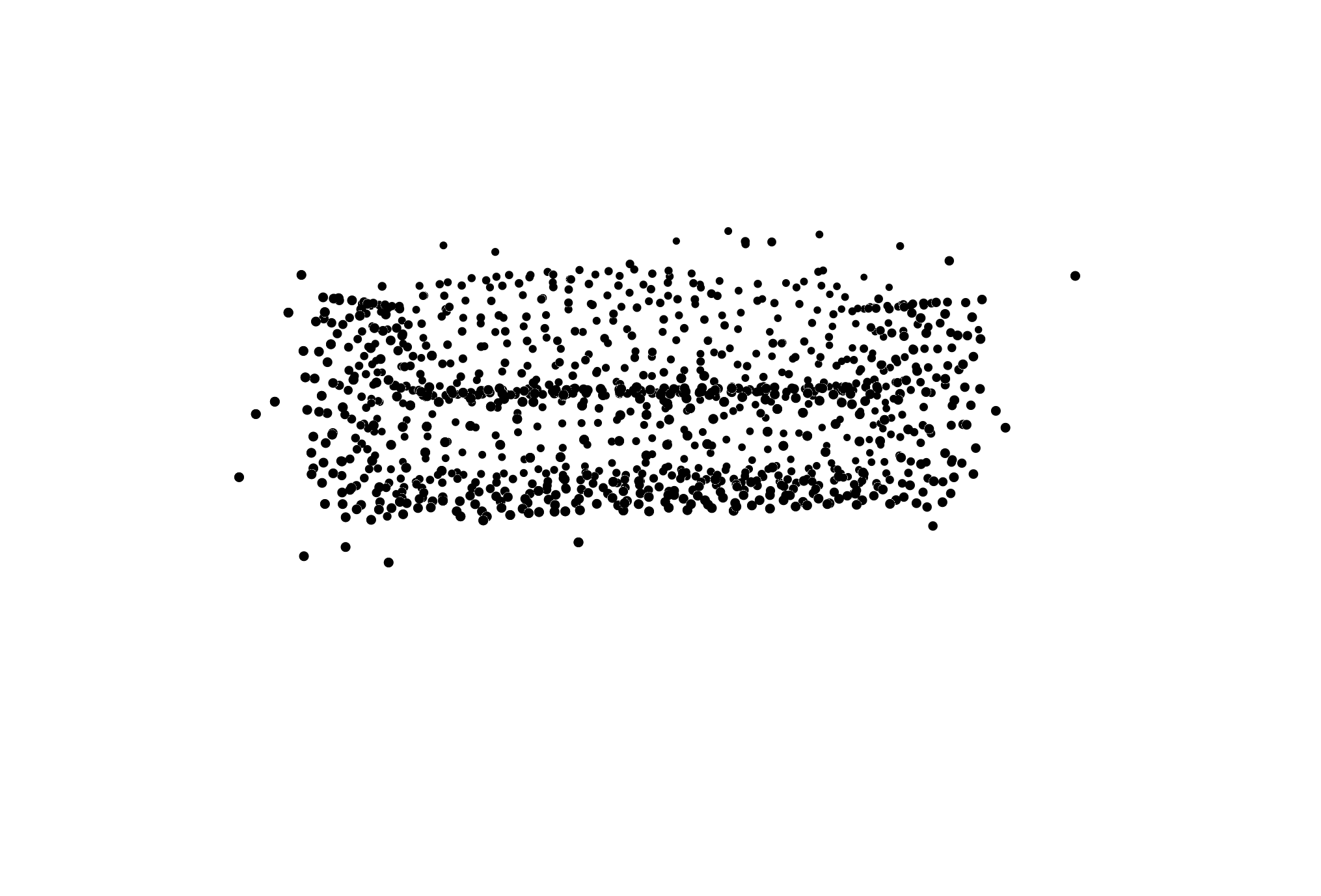} &
\figt[\sizeS]{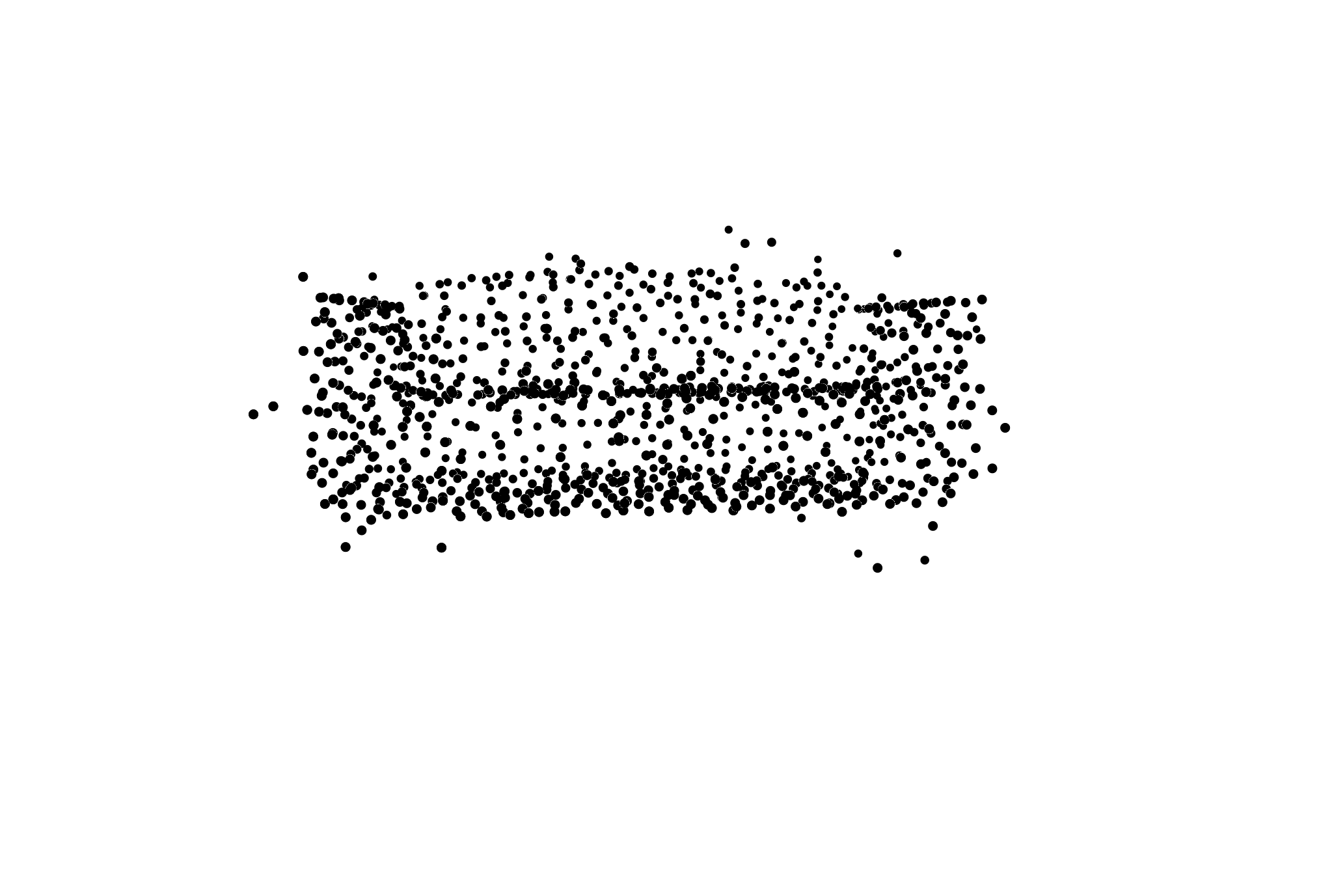} \\
(c) \blue{\cmark sofa} &\red{\xmark bathtub} &\red{\xmark tent} &\red{\xmark dresser} &\red{\xmark dresser} \\

\figt[\sizeS]{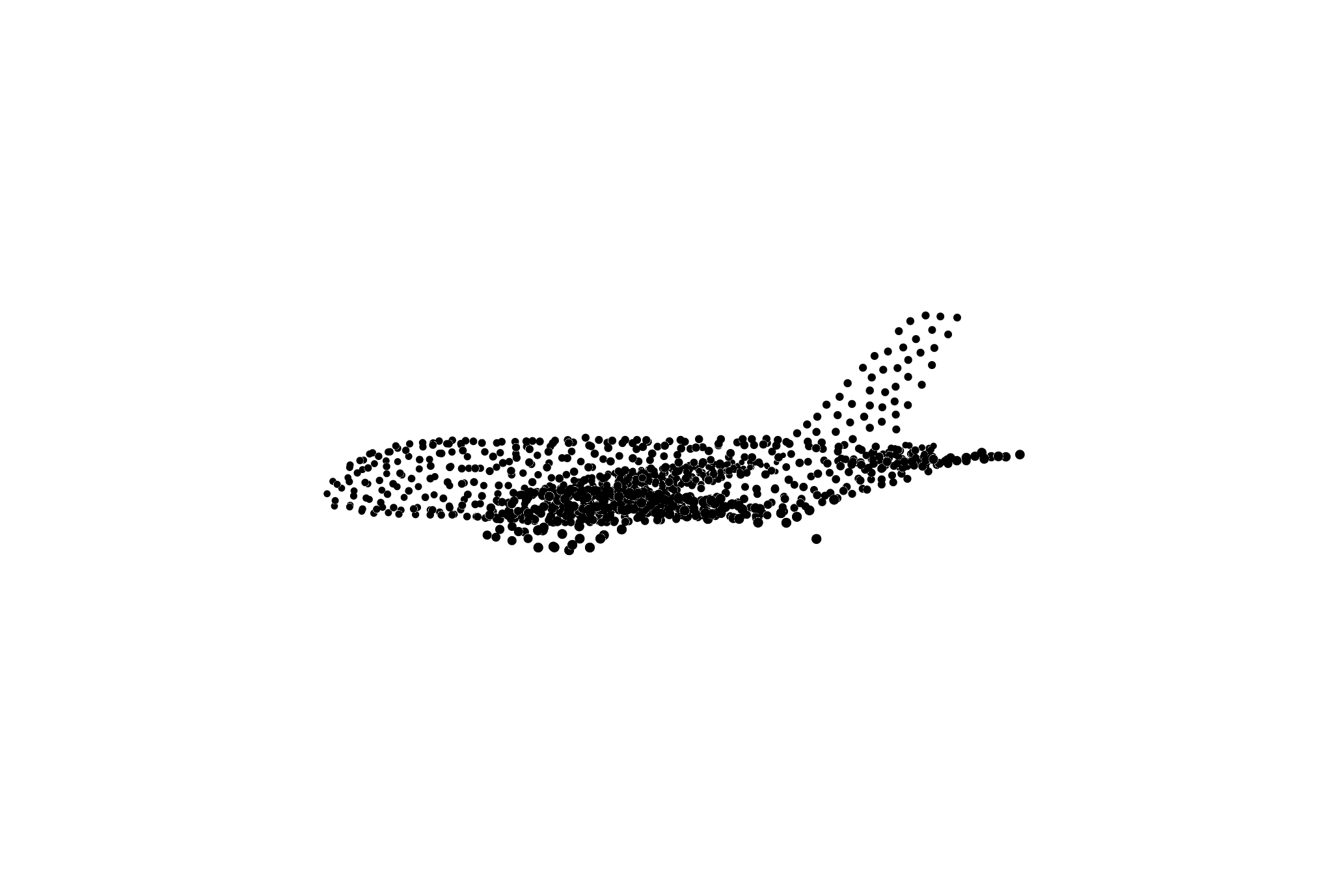} &
\figt[\sizeS]{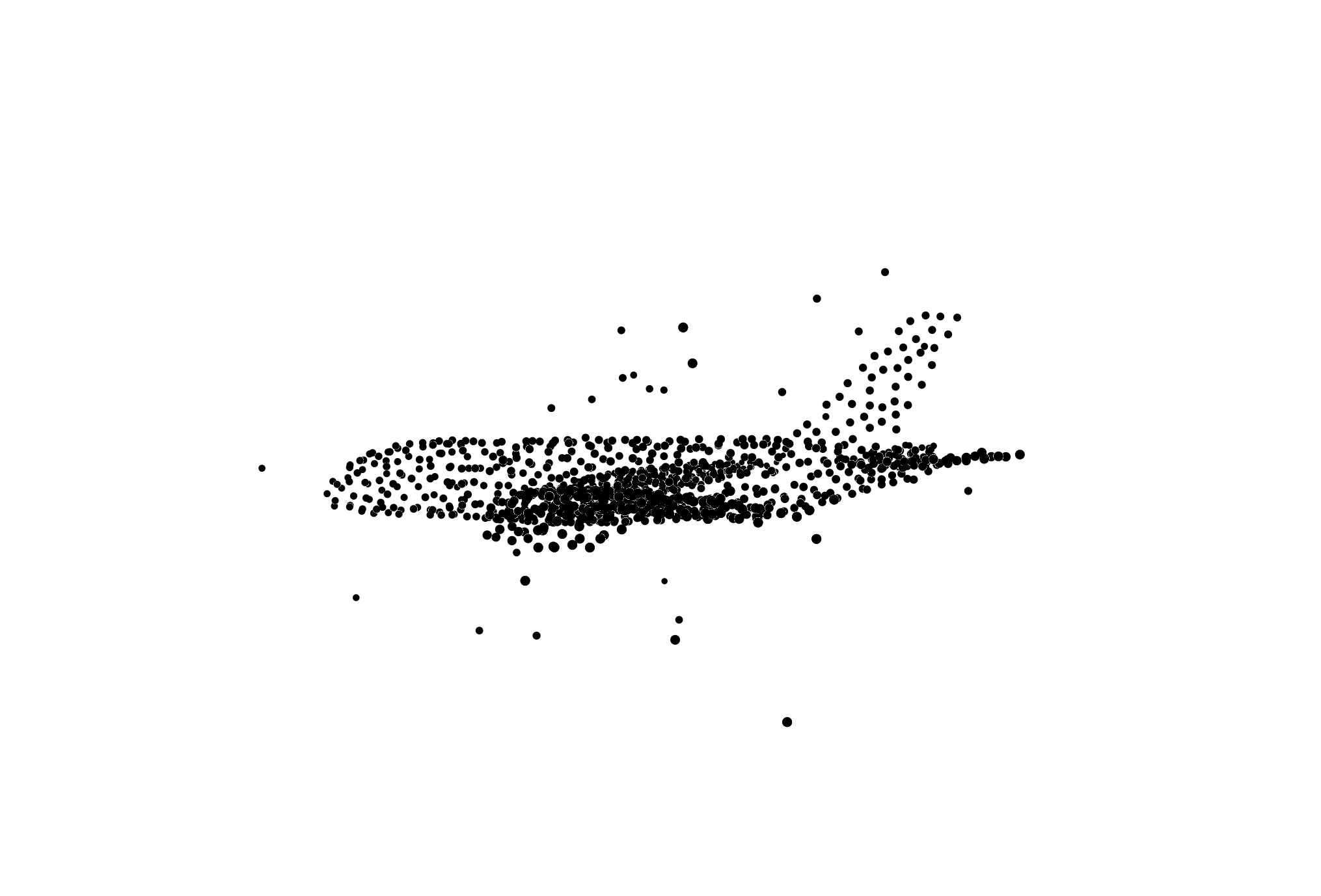} &
\figt[\sizeS]{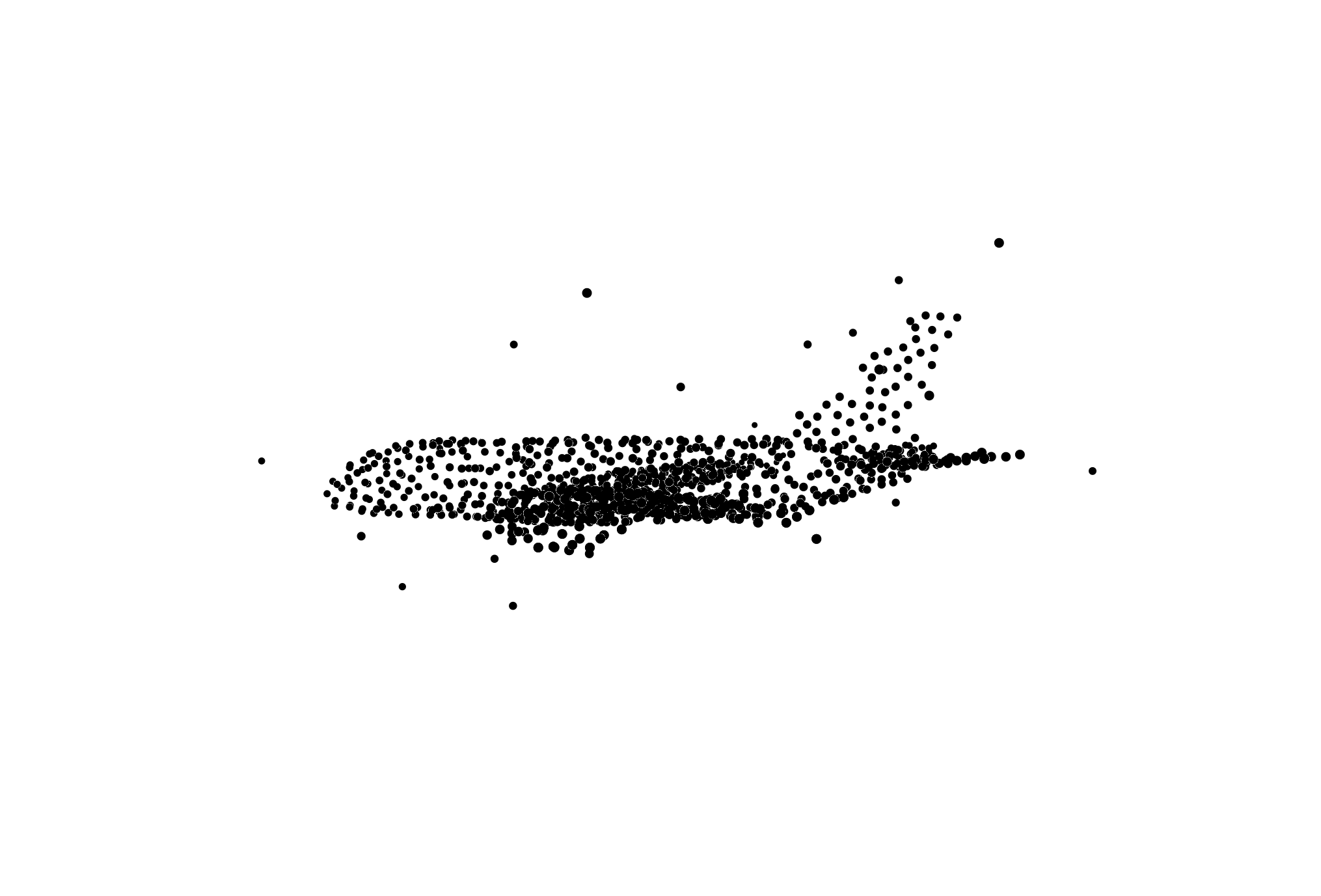} &
\figt[\sizeS]{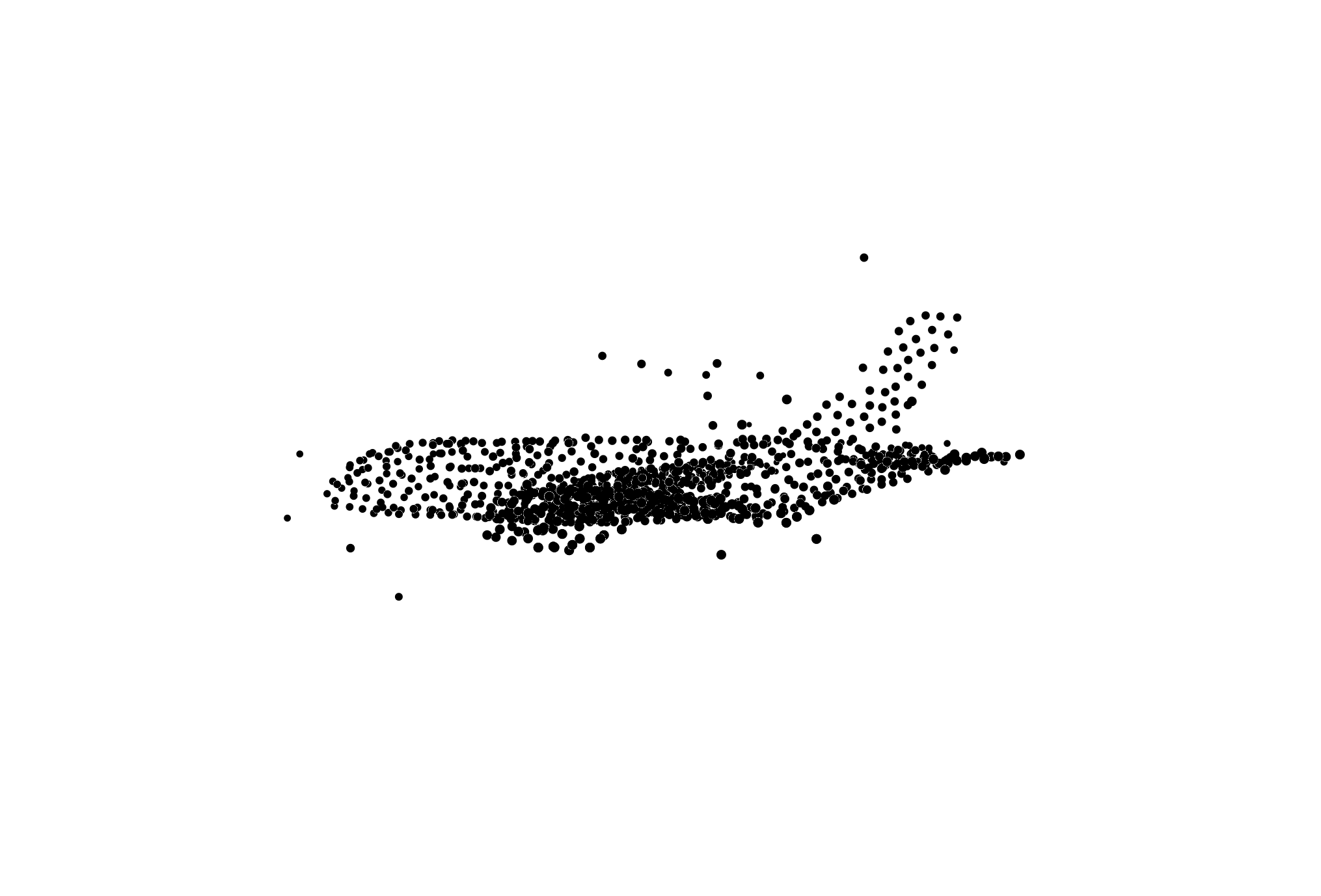} &
\figt[\sizeS]{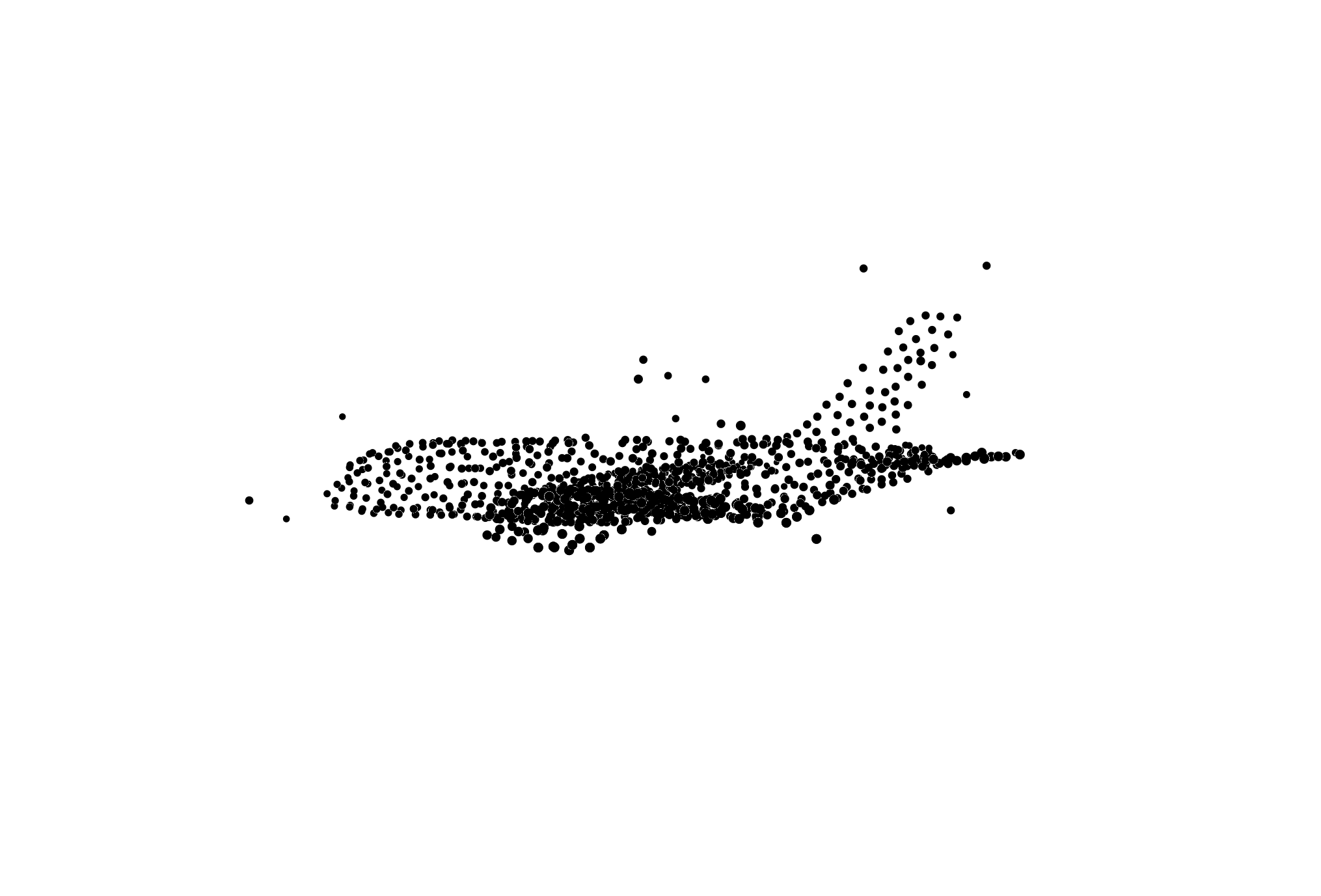} \\
(d) \blue{\cmark airplane} &\red{\xmark plant} &\red{\xmark plant} &\red{\xmark plant} &\red{\xmark plant} \\

\figt[\sizeS]{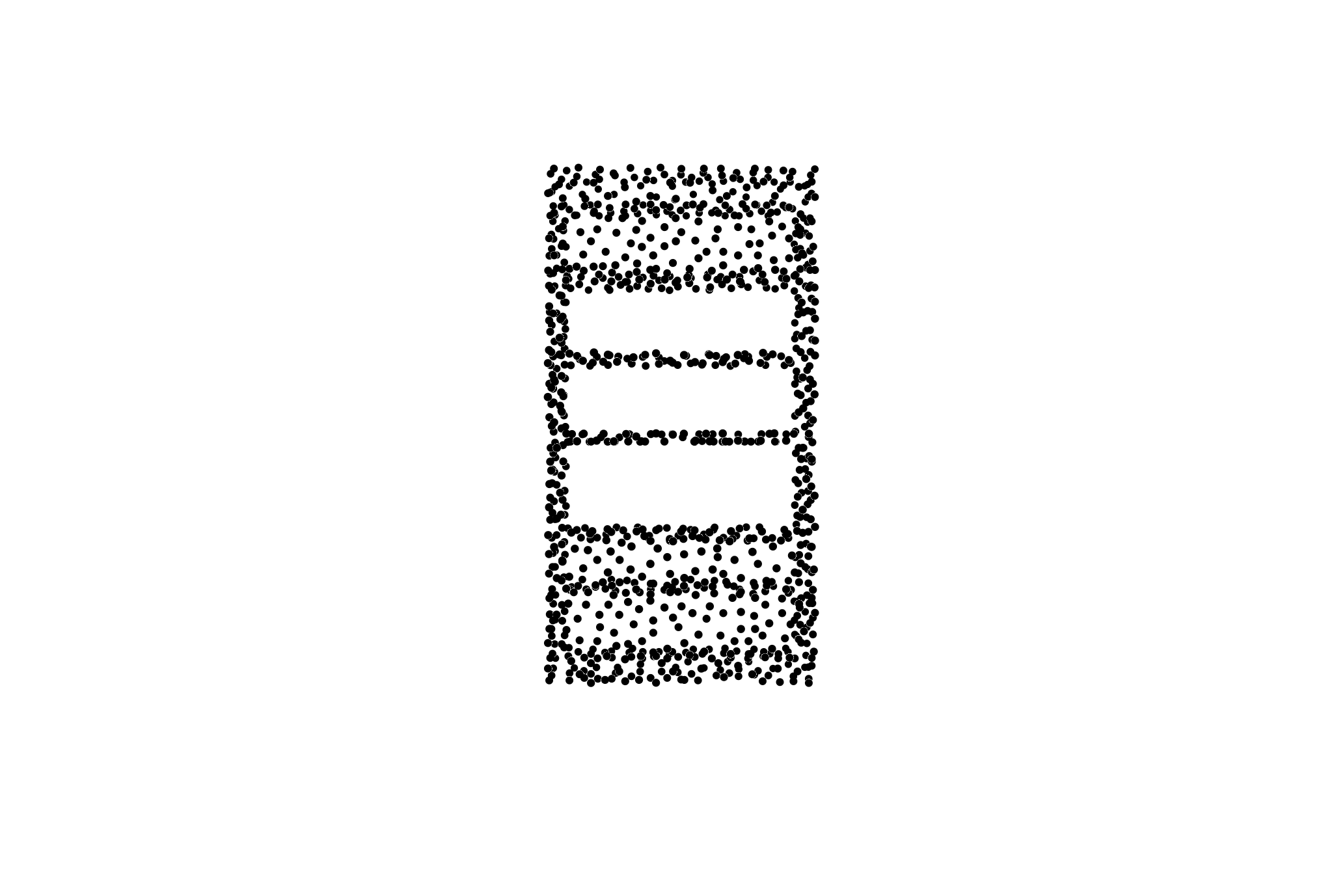} &
\figt[\sizeS]{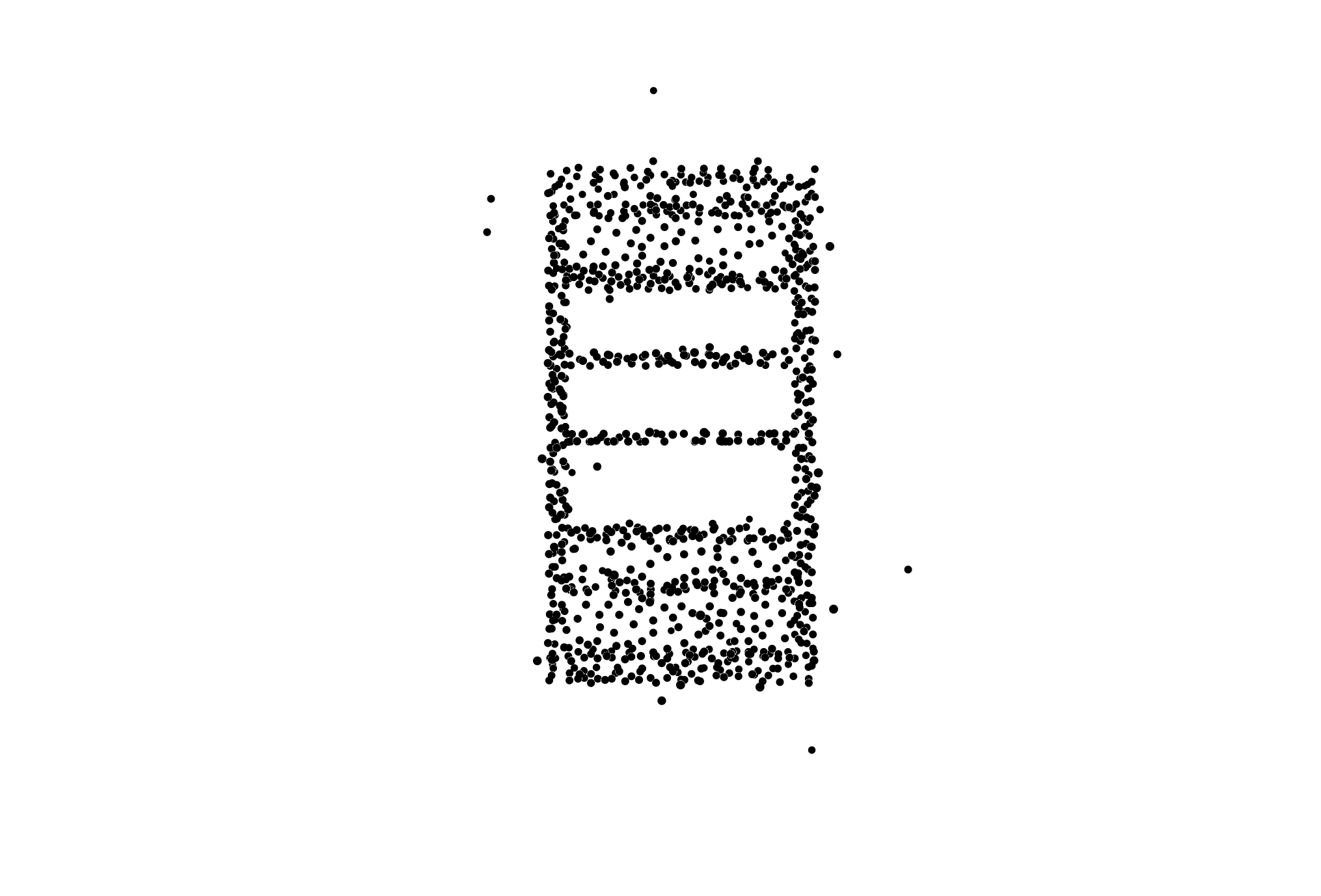} &
\figt[\sizeS]{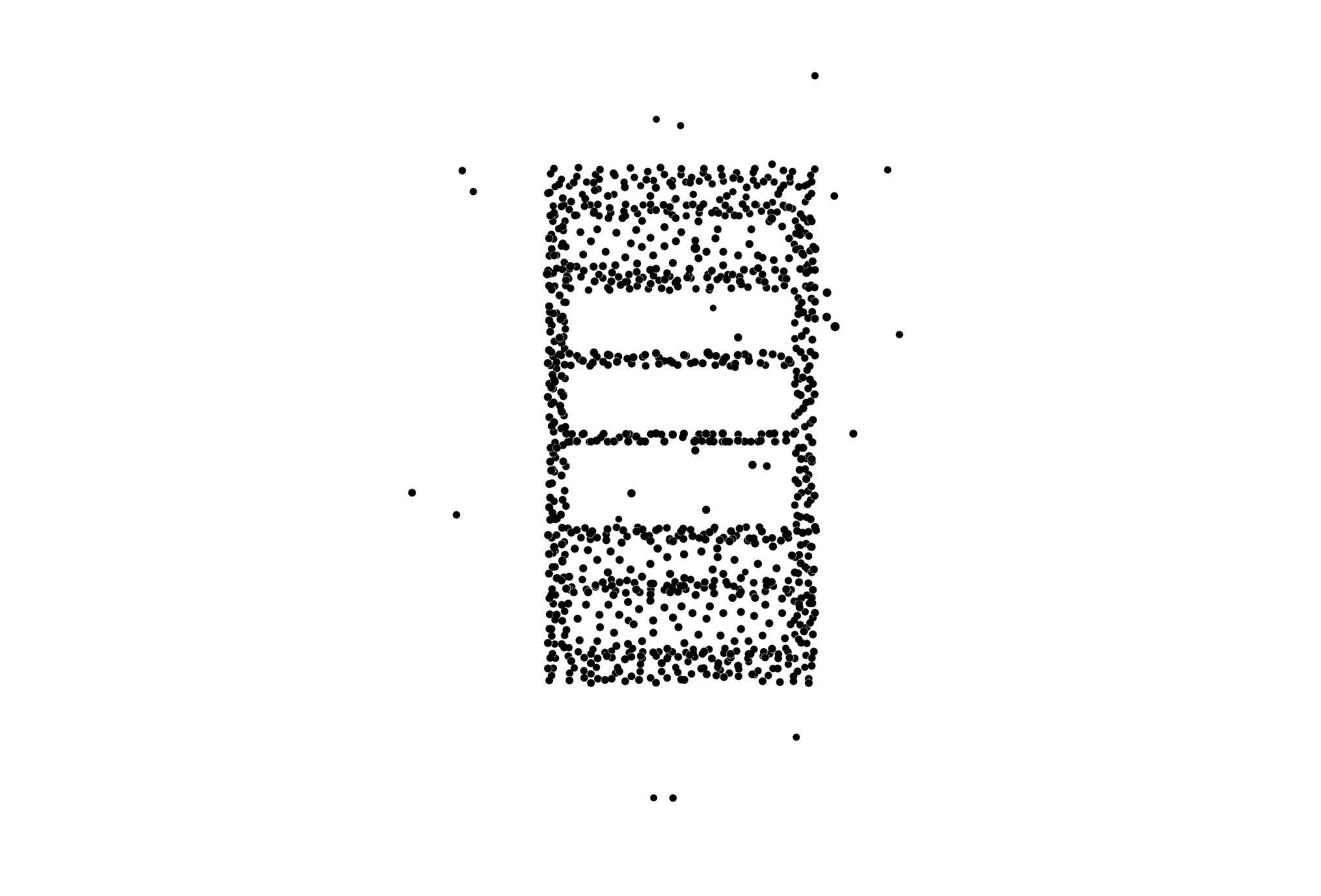} &
\figt[\sizeS]{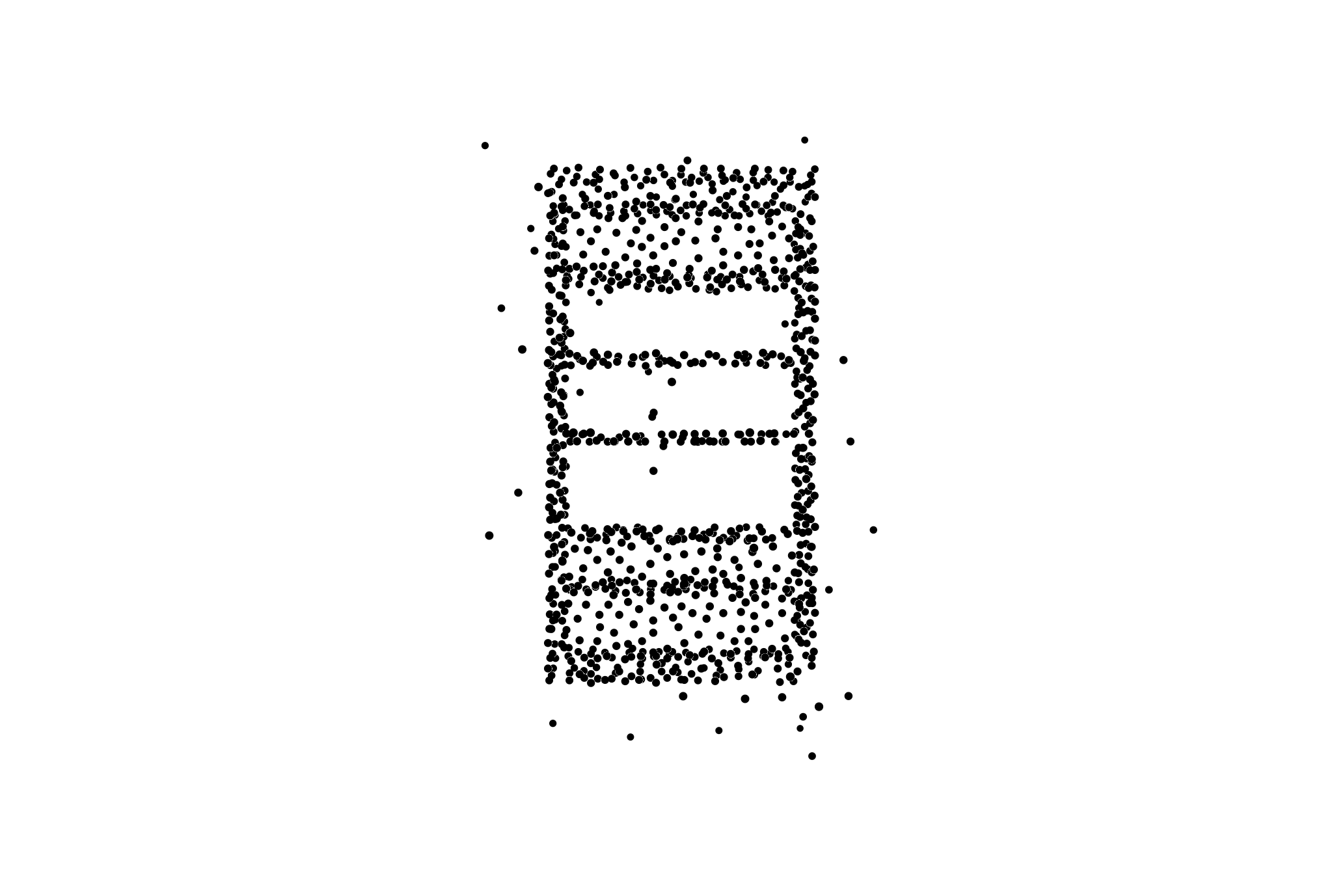} &
\figt[\sizeS]{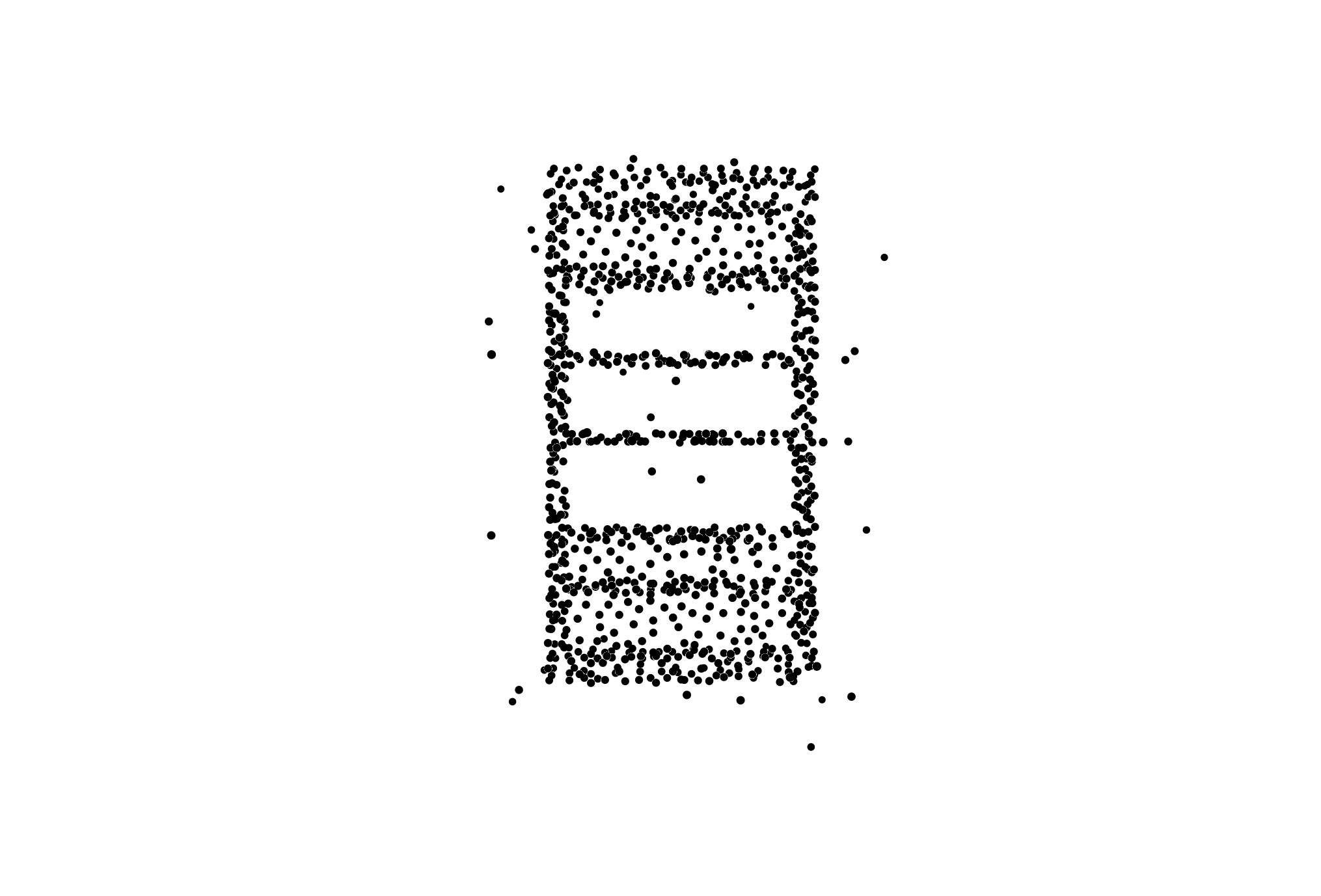} \\
(e) \blue{\cmark bookshelf} &\red{\xmark door} &\red{\xmark plant} &\red{\xmark plant} &\red{\xmark plant} \\

\end{tabular}
\caption{Visualization for adversarial point clouds produced by baseline methods under black-box setting against PACov with DGCNN .}
\label{fig:vis-3d-blackbox-supp}
\end{figure*}

In our experimentation, we select a step size of $0.32$ for searching for misclassification, \ie $\epsilon_1 = 0.32$, and $0.16$ for minimizing imperceptibility regularizations, \ie $\epsilon_2 = 0.16$, so that the ASR reaches $100\%$ to ensure a better comparison with other methods. Our method outperforms \siadv in its best parameter. It can be observed in Figure~\ref{fig:abl-black-supp}. The visualization of the black-box attacks is shown in Figure~\ref{fig:vis-3d-blackbox-supp}.


\end{document}